\documentclass[a4]{article}
\usepackage[a4paper,left=3cm,right=2cm,top=2.5cm,bottom=2.5cm]{geometry}
\usepackage{cite}
\usepackage[pdftex]{graphicx}
\usepackage[cmex10]{amsmath}
\usepackage{amssymb}

\usepackage{times}
\usepackage{array}
\usepackage[tight,footnotesize]{subfigure}
\usepackage[font=footnotesize]{subfig}
\usepackage{graphicx}
\usepackage{mathtools}
\usepackage{amsmath}
\usepackage{enumitem}

\usepackage{alltt}
\usepackage{shadethm}
\usepackage{enumitem}
\newshadetheorem{comment}{Comment}

\def\boxx{{\vcenter{\vbox{\hrule height.3pt
          \hbox{\vrule width.3pt height6pt
          \kern6pt\vrule width.3pt}\hrule height.3pt}}\;}}

\def\impos{{\;\vcenter{\hbox{\rule{5mm}{0.2mm}}} \vcenter{\hbox{\rule{1.5mm}{1.5mm}}} \;}}

\def\lrarrow{\leftrightarrow \kern-8pt \rightarrow}

\def\Unspec{A_{\rm ?}}

\def\2{\frac{1}{2}}

\def\adj{{\rm adj}}    

\def\beq{\begin{eqnarray}}
\def\eeq{\end{eqnarray}}
\def\2{\frac{1}{2}}

\newtheorem{assumption}{Assumption}

\newtheorem{example}{Example}
\newtheorem{lemma}{Lemma}
\newtheorem{theorem}{Theorem}

\newtheorem{definition}{Definition}

\newtheorem{hype}{Hypothesis}

\def\pdef{{\bf def}}

\newtheorem{corollary}{Corollary}

\def\lrarrow{\leftrightarrow \kern-8pt \rightarrow}

\def\frightarrow{\rightarrow \kern-11pt /~~}
\def\reducesto{\simeq \kern -3pt >}

\usepackage{fancyhdr}					
\begin{document}
\newcommand{\strust}[1]{\stackrel{\tau:#1}{\longrightarrow}}
\newcommand{\trust}[1]{\stackrel{#1}{{\rm\bf ~Trusts~}}}
\newcommand{\promise}[1]{\xrightarrow{#1}}
\newcommand{\revpromise}[1]{\xleftarrow{#1} }
\newcommand{\assoc}[1]{{\;\xrightharpoondown{#1}} \;}
\newcommand{\rassoc}[1]{{\xleftharpoondown{#1}} }
\newcommand{\imposition}[1]{\stackrel{#1}{\impos}}
\newcommand{\scopepromise}[2]{\xrightarrow[#2]{#1}}
\newcommand{\handshake}[1]{\xleftrightarrow{#1} \kern-8pt \xrightarrow{} }
\newcommand{\cpromise}[1]{\stackrel{#1}{\frightarrow}}
\newcommand{\policy}{\stackrel{P}{\equiv}}
\newcommand{\field}[1]{\mathbf{#1}}
\newcommand{\bundle}[1]{\stackrel{#1}{\Longrightarrow}}

\title{Spacetimes with Semantics (III)\\~\\\normalsize The Structure
  of Functional Knowledge Representation\\and Artificial
  Reasoning\footnote{This series of documents is aimed at bridging the
    world of natural science (especially physics) and computer science, with a compromise
    on the level of rigour that I have improvised with an eye on
    practical applications. It is probably too ambitious in scope and
    detail, but bridges may serve a purpose even with gaps.}}

\author{Mark Burgess}
\date{6th August 2016\\Minor improvements 1 August 2017}
\maketitle

\begin{abstract}
  Using the previously developed concepts of semantic spacetime, I
  explore the interpretation of knowledge representations, and their
  structure, as a semantic system, within the framework of promise
  theory. By assigning interpretations to phenomena, from observers to
  observed, we may approach a simple description of knowledge-based
  functional systems, with direct practical utility. The focus is
  especially on the interpretation of concepts, associative knowledge,
  and context awareness. The inference seems to be that most if not
  all of these concepts emerge from purely semantic spacetime
  properties, which opens the possibility for a more generalized
  understanding of what constitutes a learning, or even `intelligent'
  system.

  Some key principles emerge for effective knowledge representation: 1)
  separation of spacetime scales, 2) the recurrence of four
  irreducible types of association, by which intent propagates:
  aggregation, causation, cooperation, and similarity, 3) the need for
  discrimination of identities (discrete), which is assisted by
  distinguishing timeline simultaneity from sequential events, and 4)
  the ability to learn (memory).  It is at least plausible that
  emergent knowledge abstraction capabilities have their origin in basic
  spacetime structures.

  These notes present a unified view of mostly well-known 
  results; they allow us to see information models, knowledge
  representations, machine learning, and semantic networking
  (transport and information base) in a common framework.  The notion
  of `smart spaces' thus encompasses artificial
  systems as well as living systems, across many different scales,
  e.g. smart cities and organizations.

\end{abstract}

\tableofcontents


\section{Introduction} 

A spacetime is an extended network of interconnected agencies that
collectively support the notions of location, time, direction, and
dimension.  In a semantic spacetime, special meaning is attributed to
locations and times; a semantic spacetime adds interpretation, and
even function, to the experience of space by an observer.  Semantic spacetimes
may be models for everything from real structures to virtual
collections of objects, in either a solid or a gaseous state of
organization: buildings, warehouses, databases, computers, machines,
organisms. Semantic spacetimes are associated with knowledge systems;
however, we should not jump to conclusions about what this means: a
Turkish bathhouse, or a supermarket, could be representable as a
knowledge space, not only the library in Alexandria.  As such,
semantic spacetimes are intended as a universal theory of functional
interpretation within any extended system, that behaves as a natural
generalization of the physics of material scaling to more generally
immaterial worlds.

In paper I of this series\cite{spacetime1}, it was shown that, if we
take a collection of autonomous agents that promise to collaborate in
the construction of a spacetime, then it becomes possible to construct
semantics in a simple way, with clearly defined laws.  Paper
II\cite{spacetime2} showed that, by separating promises that refer to
local (scalar) properties of agents from those relating to
relationships between agents (tensors), we identify that a promise
spacetime has both scalar (material) properties and vector (adjacency,
cooperative, non-self) properties, and hence we have a description
that scales as a natural generalization of the physics of coarse
graining: we can combine agents into super-agents, or coarse grains,
without loss of information as long as we account for both reducible
and irreducible promises.  A complete definition of scaling opened the
way to a natural definition of {\em tenancy}, i.e.  the hosting of
functional agencies within a space, with properties or attributes
implemented over a region bounded by any criterion (not necessarily a
spatial boundary).  Limited range promises imply a horizon for
context, in which dominant `rules' or behavioural patterns may differ
across cells.  This is also a basis for defining and understanding
biological organisms as smart agencies. It forms a simple basis of a
spacetime interpretation of shared context, and categorization.

In this final part of the promise-spacetime manifesto, I want to
build on all these notions to explore how extended, differentiated
structures can interact with one another to encode and encode
knowledge. Such structures would find use both as repositories of
information, and as functional machinery, and as 
`smart spaces'\footnote{This includes but goes beyond the current interest in so-called
artificial intelligence, e.g. using Artificial Neural Networks (ANN) to recognize
complex patterns. Such a neural view of reasoning is too limited and
too specific to represent a comprehensive approach to engineering.}.  An
obvious application of this work is to understand the coming age of
embedded information technology: the so-called Internet of Things,
Smart Cities, etc, which are non-simply connected spacetimes of
significant scale, complexity, and inhomogeneity. The outline for the
topics is as follows:
\begin{enumerate}
\item A review of spacetime and its semantics.
\item A description of data as semantic elements, in the language of promise theory.
\item How memory can be constructed, addressed, and used, in order to equip spaces with a
capability for data representation and learning.
\item How knowledge is acquired and represented, by summarization.
\item How reasoning occurs in the shape of constrained spacetime trajectories.
\item The application of the foregoing ideas to functional spaces.
\end{enumerate}

\section{The structure and semantics of spacetime}

The artificial spaces we create are filled with meaning.  Parking
spaces are different from living spaces; a beach is different from a
cliff.  Meaning is partially engineered, by repeated trial and
error, and is partially serendipitous in its origin. To evolve
something to the point where it becomes `fit for purpose', we accept
that there must be selection criteria that prefer one functional form
over another; these criteria take on the role of goals, some of which
are engineered and some are emergent, based purely on observed success
in a functional role.

This repetitive process, which leads to an island of possible
temporary stability, is called {\em learning}, and it take the form of
an interactive relationship between and agent and its subject, whose
equilibrium outcome is what we call
knowledge\cite{certainty}\footnote{According the the UK government IT
  infrastructure library\cite{itil1}, the sequence DIKW (data,
  information, knowledge, wisdom) is the ladder of increasing meaning
that humans go through in the course of their working interactions.}.
Knowledge may decay as quickly as it has been won: it requires
maintenance, and we can talk about its stability, i.e. its usefulness
or applicability to a context. In this section, we explore how this
process occurs, with or without the intentional direction of human
involvement.

\subsection{Semantic spacetime}

\begin{figure}[ht]
\begin{center}
\includegraphics[width=8cm]{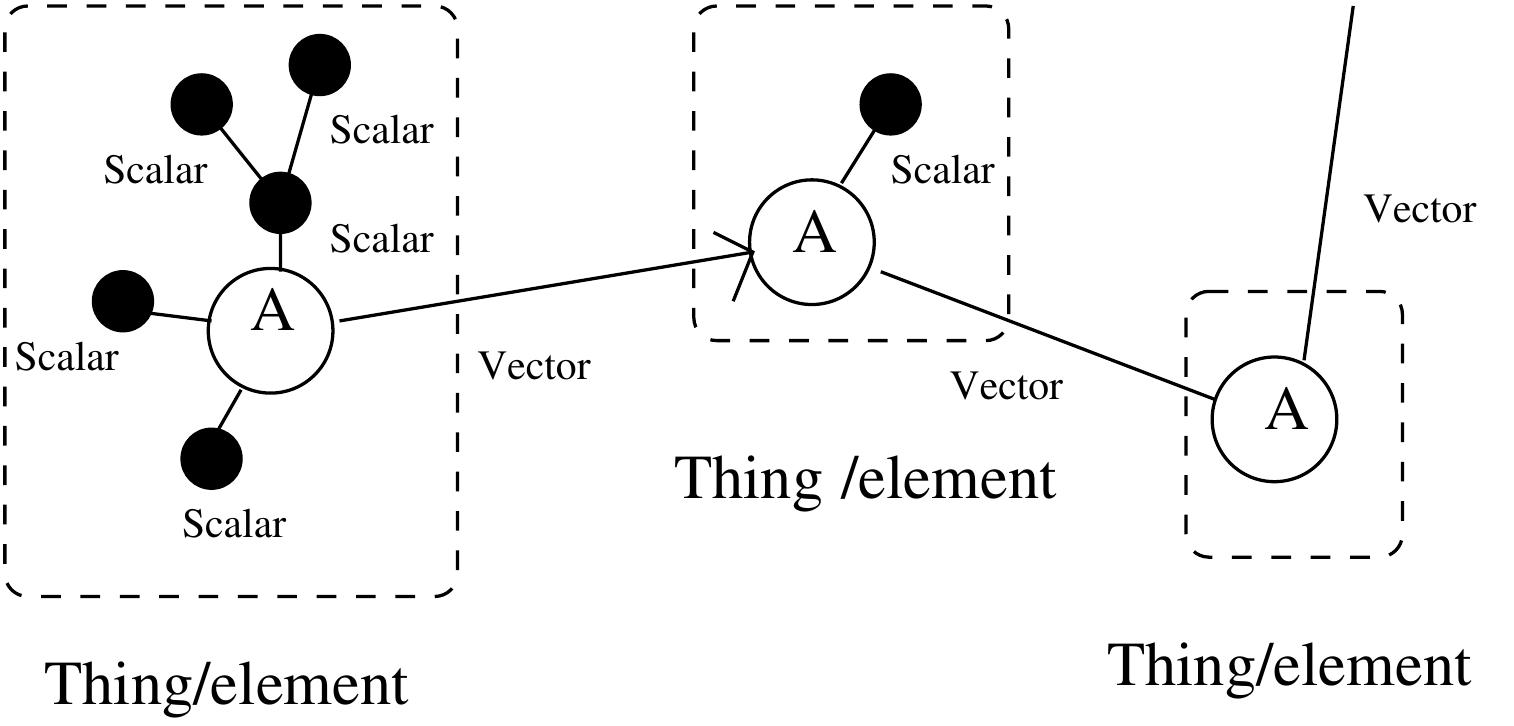}
\caption{\small Elements of semantic space may be viewed as autonomous agents
surrounded by a cloud of scalar promises, joined into a semantic web
by vector promise bindings.\label{kspace}}
\end{center}
\end{figure}

A semantic spacetime is a structured collection of semantic elements
(see section 6.6. of paper I)\cite{spacetime1}.  To make a spacetime,
with both scalar (material) attributes and vector (adjacency)
bindings, from the atomic building blocks of autonomous agents,
one defines the basic elements in a semantic space:
\begin{definition}[Semantic element]
  A semantic element is a tuple $\langle A_i, \{ \pi_{\rm
    scalar~j}, \ldots \}\rangle$ consisting of a single autonomous
  agent, and an optional number of scalar material promises.
\end{definition}
A semantic element is an autonomous agent (in the sense of promise theory),
surrounded by a halo of promises that imbue it with
semantics (see fig. \ref{kspace}). Examples include any
objects that could be relevant within a functional context: people, machinery,
and passive structures like buildings or trees.
\begin{definition}[Semantic spacetime]
A collection of semantic elements, in any phase (gas or solid)\cite{spacetime1},
for which a local change in state, promises or configuration represents
a local unit of time.
\end{definition}
There are multiple reasons to want to define semantic spacetimes. The
definition is a direct match for the day to day workings of materials,
chemistry, biology, and information technology.  More generally, the
reasons relate to the communicability of knowledge and functionality between
interacting agents:
\begin{itemize}
\item To find the requirements of a low level representation for concepts as spacetime patterns.
\item To understand arbitrary spacetimes as functional machinery, with conceptual utility.
\end{itemize}
The field of linguistics provides the essential clue that this agenda is
possible, since languages fit both of the requirements above. Indeed,
what we are looking for is a language, but a formal language that can
be used to understand the design of machinery with purpose, or discern the
relative purposes of systematic mechanisms.

Semantics are one part of successful functioning, but they are static relationships.
To utilize semantics for purpose we have to place them into a dynamically
stable context. This is where the concept of {\em knowledge} becomes important.

\subsection{The concept of knowledge as an iterated equilibrium}\label{learning}

The concept of knowledge has special status. It is defined, in these notes, by
analogy with the knowing of a person or a
friend\cite{certainty}. Knowledge is not something we acquire by hearsay,
not something acquired by reading once in a book. It is the result of an ongoing
relationship of repeated assessment, made by an agent,
over time. During repeated assessment, an agent gauges the 
relative importance of outcomes and their stability to change.
Knowledge is thus quite different from data: it is a summarized
aggregate representation of data, to which we have attached an
interpretation of significance. It has been confirmed iteratively to the point of
trust\cite{burgesstrust}, by repeated sampling. Knowledge is not `big
data' or bulk information in raw form: to sieve out knowledge from an extensive
stream of samples, a reduction over equivalences of quantitative and qualitative
measures has to take place\cite{certainty}.

The repeated sampling of semantically equivalent data by an agent is called {\em learning}.
In order for there to be learning, there must be at least a promise of offered
observability (+), and the receiver must promise to accept or receive (-)
data faithfully. The integrity of a measurement cannot exceed the overlap
of these two promises\cite{promisebook}.
\beq
\text{observed} &\promise{+\text{attributes}}&\text{observer} \\
\text{observed} &\revpromise{-\text{attributes}}&\text{observer} 
\eeq
The stability of every sampled observation made is uncertain, because
an agent does not know how fast the world is changing without
repeating its observations. Agents choose a sampling rate
$T_\text{sample}$ which (by the Nyquist theorem) should be at least
twice as fast as the rate of change of the promise it is assessing,
and hence also the rate at which it is able to form a stable assessment of what it
observes\cite{burgesstheory,certainty,cover1}. 
\begin{definition}[Learning about $\pi$]
  The sampling, equilibration, and summarization of observational
  assessments concerning a promise $\pi$ made by another agent,
  repeated over a timescale $T_\text{learn} > 2T_\text{sample}$.

Let $\alpha(\pi)$ be an assessment of promise $\pi$, made by some agent.
On each sample time $t+1$, an observer agent
applies a learning function $L$ to combine the previous average assessment $E(\alpha(\pi)_{t})$
with a new sample $\alpha(\pi)_{t+1}$ to form:
\beq
E\left(\alpha(\pi)_{t+1}\right) = L\left(\alpha(\pi)_t,E\left(\alpha(\pi)_{t}\right)\right)
\eeq
\end{definition}
Learning defines a clock that ticks at a rate $T_\text{sample}$.
\begin{definition}[Knowledge of $\pi$]
  A stable summary of the iterated assessment $\alpha(\pi)_{T_\text{know}}$,
   of one or more promises $\pi$, is formed by
  equilibration of the samples over a timescale $T_\text{know} \gg 2T_\text{sample}$.
\end{definition}
Because knowledge defines a process with a timescale, the failure to 
confirm it relative to other changes leads to its decay. Trust in knowledge
needs to be replenished. This is the main difference between knowledge and data.
\begin{lemma}[Knowledge decay]
The uncertainty of knowledge is a geometric function of the time since last
learning took place.
  If $L$ is a convex or linear function of its arguments, which weights
  old assessments less favourably than new by a factor $\ell < 1$,
  then an assessment $\alpha(\pi)_\tau$ from arbitrary time $\tau$ has
  a geometric rate of attenuation by $\ell^r$, where $r \simeq
  T_\text{learn}/T_\text{sample}$.
\end{lemma}
The proof follows from the iteration $r$ times of a convex form, such
as $L_t = (1-\ell) s_i+\ell L_{t-1}$. Let $\alpha(\pi)_\tau
\rightarrow x$, and each new sample be $s_i$, for $i=1,2,\ldots r$.
Then the contribution to $L_r$ is $\ell^r x \ll x$.  Thus knowledge
must be reassessed, and learning must continue as long as new
assessments are available, else the uncertainty of the knowledge If
sampling stops, the value of $L$ fails to track its value.  For each
sample missed, there is a cumulative error which grows like
$(1-\ell)^r$.  A corollary is that the significance of data to an
assessment decays at the same rate with its relative age.
By virtue of Nyquist's theorem\cite{burgesstheory,shannon1}, we can also say that:
\begin{lemma}[Fidelity and learning rate]
  Learning can only represent source values faithfully if the
  rate of sampling is greater than twice that of the fastest rate of
  change in the data, i.e. $2/T_\text{sample} < \partial\pi/\partial
  t$.
\end{lemma}
From the definition of knowledge:
\begin{lemma}[Knowability of information] A fact is knowable by agent
  $A$ if the promise can be assessed by an agent $A$ 
  as kept over multiple assessments.
\end{lemma}
where:
\begin{definition}[Fact]
A promise of information to be taken on trust.
\end{definition}

\subsection{The concept of knowledge spaces: spaces that can learn and represent knowledge}

\begin{figure}[ht]
\begin{center}
\includegraphics[width=12.5cm]{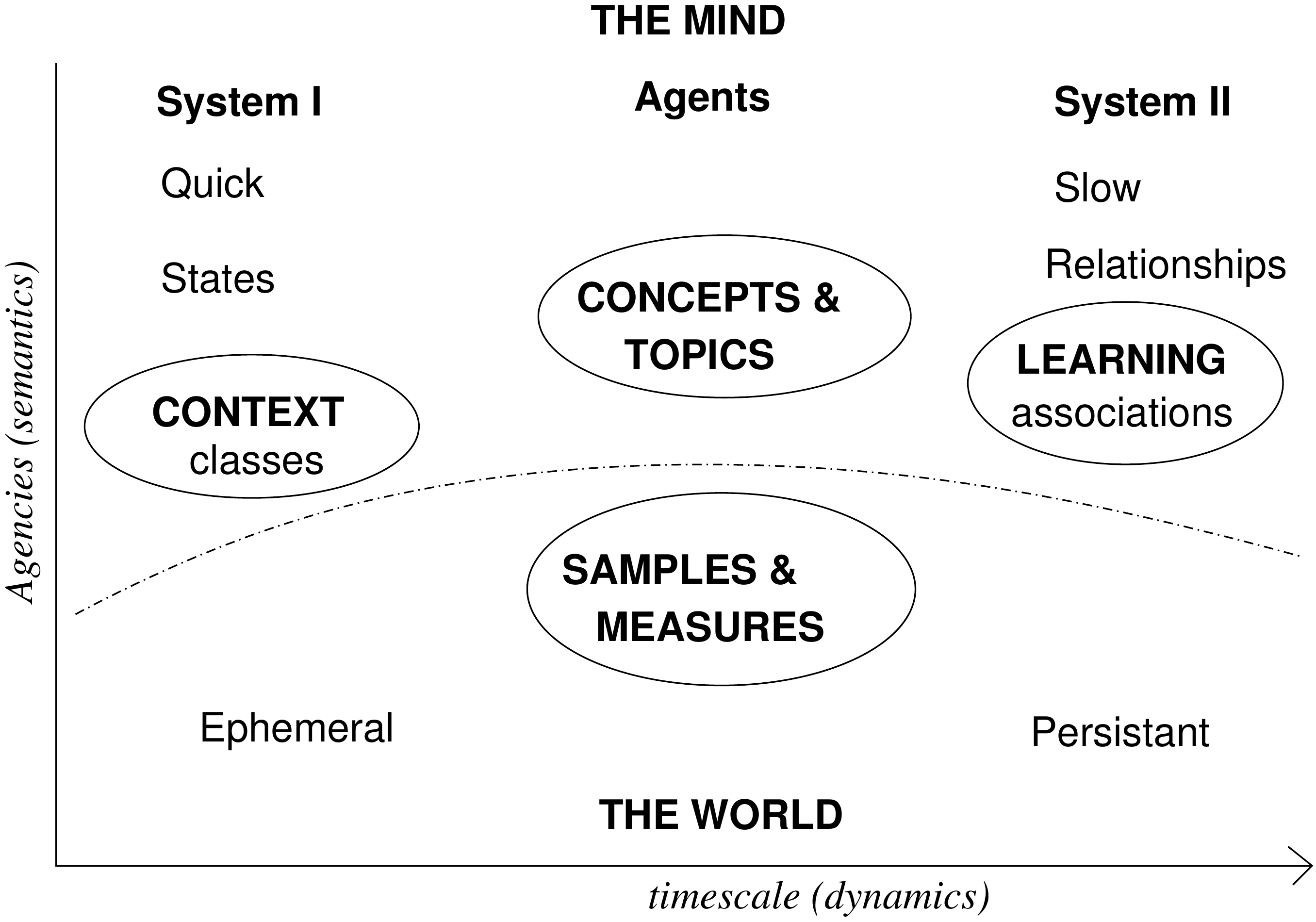}
\caption{\small To find a dynamical basis for knowledge semantics, we
  need to identify separable timescales, from which we can identify
  stable `knowable' phenomena.  Learning occurs at two timescales:
  short-lived convergent states we call `context' about `now', and
  long-lived associative bindings based on accumulated experience,
  that form the stories on which we base reasoning.\label{memorysystem}}
\end{center}
\end{figure}

What do space and time mean to us?
Let's explore the semantics of spacetime further, to better understand
how we experience it.  
The scaling characteristics of promises were investigated in paper
II\cite{spacetime2}.  One of the challenges facing us, with respect to
understanding knowledge systems as spacetimes and vice versa, is the
separation of dynamical timescales and spatial regions during
observation and measurement (see figure \ref{memorysystem}). This is a
stability criterion.  In dynamics, weakly coupled dynamical phenomena
separate along distinct timescales, and thus the semantics we build on
top of them must do the same.  A natural hypothesis is therefore that
promises, which encode information,
separate into at least two timescales: I call these {\em context} (fast) and {\em
  knowledge} (slow), such that quickly assessed context modulates more slowly
  accumulated knowledge, in order to adjust its
relevance.
From this, we see that there is a set of timescales for knowledge to emerge, including:
\begin{enumerate}
\item Sample or observation rate.
\item The time to summarize observations (form sense of context).
\item Learning rate (accumulation of experience).
\item Knowledge lifetime (decay rate in the absence of observation).
\item Reasoning about knowledge (associative narratives).
\end{enumerate}
The extraction of what we call {\em reasoning}, or stepwise
narratives, built from the semantic relationships between observations
have to be identified from the resulting network of associations. This
is the agenda for paper III.

What kind of data lead to the kind of functional knowledge about which
we can reason?  An ability to reason takes more than single samples;
it requires the learning of relationships between the things too. I'll
return to this in section \ref{learnassociation}.
For now, we may observe that there is a clear role for space and time
in data representation, and the hypothesis we shall explore in these notes
is the idea that these spacetime aspects are, in fact, the only basis for 
knowledge representation.

\subsection{Spatial semantics}

What things do we want to say about space? As
described in paper I, a space may be a collection on unrelated points
(like a gas of disconnected bubbles); but, usually, we expect points
to be interconnected allowing motion from one point to
another\cite{spacetime1}.  Adjacency is the most primitive irreducible
promise that characterizes space. In promise theory, adjacency is a
directional concept: $A$ may be adjacent to $B$ without $B$ being
adjacent to $A$. This is a deliberate feature that models functional
behaviours like semi-permeable membranes, diodes, and
authorization-access controls.

The semantics of location take on a variety of forms, representing 
four main concepts to do with space: topology, location, distance, and direction.
Many of these concepts are intertwined.
Most of our familiar conventions for space and time are based on a
history of thinking of a Cartesian theatre model, in which spacetime
is a backdrop on which interactions happen between agents,
independently of absolute location, only relative to other
agents\footnote{For a background on the concept of this philosophy, see
  for example \cite{dennet1}.}.  This separation of spacetime and
matter is rooted both in convenience, as well as in the agenda of
making science impartial.  Einstein showed, however, 
that this separation is both misplaced and inconsistent (impartial
does not mean objective)\cite{einstein1}: the presence of material agents within
spacetime is closely related, if not actually equivalent in some
sense, to the local promises made by spacetime itself. This hints that
a unified description for spacetime and matter would make more
sense\cite{spacetime1}.

In promise theory, the alphabet of symbols, in a coded structure, is composed of the atoms
of space:
\begin{definition}[Location agents]
These are irreducible sites that take up space and can emit and absorb signal agents.
They may not overlap.
\end{definition}
Agents in the role of signal agent form the alphabet of scaled inter-agent language.
\begin{definition}[Signal agents]
  They may be created and destroyed, subsequently emitted and
  absorbed, by location agents. They can occupy the same space,
since they end up and accumulate at end points.
\end{definition}

\subsubsection{Distance, similarity, and metrics}\label{shash}

The notion of distance is related to the ability to establish
a coordinate system on space.  The challenge of any coordinate system is
to make locations easy to find, and give meaning to the distance
between locations.  Two kinds of distance may be defined in a semantic space: metric and semantic.
\begin{definition}[Metric (quantitative) distance]
A measure of coordinate-similarity in position.
\end{definition}
\begin{example}[Metric distances]
Various aspects of topology and metric distance play into how we understand space.
\begin{enumerate}
\item {\bf Adjacency, connectivity, continuity, and dimension} -
  characterizing allowed transitions, doorways, the existence of
  dimensions or degrees of freedom, discrete or continuous. Which
  channels make it possible to move?  What is the smallest distance an
  agent can move? Are there disallowed transitions?  Adjacency is a
  cooperative concept (see paper I).

\item {\bf Location, position}: (nominal) identifying and promising
  position requires distinguishable coordinates.  Location does not
  specifically require numerical coordinatization of a space, as long
  as the naming of locations is unique, e.g. using signposts. The
  villages of our rural past were distinguishable in spite of a lack
  of maps. A description of relative locations depends on direction and metric
  distance.

\item {\bf Metric distance between locations}: (cooperative) given nominal positions of agents, by
  cooperation the agents can form a coordinate system with mutually
  agreed understanding of the meaning of distance. This then allows
  comparative measures.

The concept of distance is related to dimension, because a projection
of a spatial relationship onto a `screen' of lesser dimension can
distort the ability to measure or observe the distance.  What seems to
be close, in one interpretation of the observed world need not be
close in the representational facsimile world.

\item {\bf Semantics of metric distance}: What we do mean by `near',
  `far', `local',  etc (qualitative approximate discriminants).

\item {\bf Direction}: favoured adjacency amongst the possible
degrees of freedom at a location, often associated with a
coordinate representation, 
or an external relative marker that breaks a rotational symmetry, like fixed
stars (see paper I).

\item {\bf Containment}: the constraining of locations within a perimeter or other boundary.

\item {\bf State or pattern}: configuration patterns of internal degrees of freedom, whose effect may or
  may not be visible in the exterior. Hidden variables are like
  embedded dimensions projected into a subspace of the observer (e.g.
  Kaluza Klein theory).  This is familiar in physics where properties
  like charge and spin are represented as hidden, non-spatial
  dimensions degrees of freedom, generally characterized by their
  symmetries\footnote{Kaluza-Klein theory is not capable of
    representing the known symmetries as spacetime translations
    on a continuous differentiable spacetime, but
    continuum translations would not be the correct model
    in a discrete network.}.

\end{enumerate}
\end{example}

\begin{example}[Effectively fixed markers]\label{fixedstarsex}
  The fixed stars move so slowly and insignificantly relative to the
  Earth, that we can use them as effective boundary conditions, even
  though they break no symmetries.  By pegging a slow moving
  phenomenon (e.g. the perceived motions of the fixed stars) to
  something changing more quickly, one can create an illusion of
  symmetry breaking, in the form of an effectively static boundary
  condition.  The propagation itself may be complicit in the
  assessments of promises (see figure \ref{fixedstars}).

There are no cooperative promises between the fixed stars, and yet they hold
their formation in such a way that can be used.
\begin{figure}[ht]
\begin{center}
\includegraphics[width=5cm]{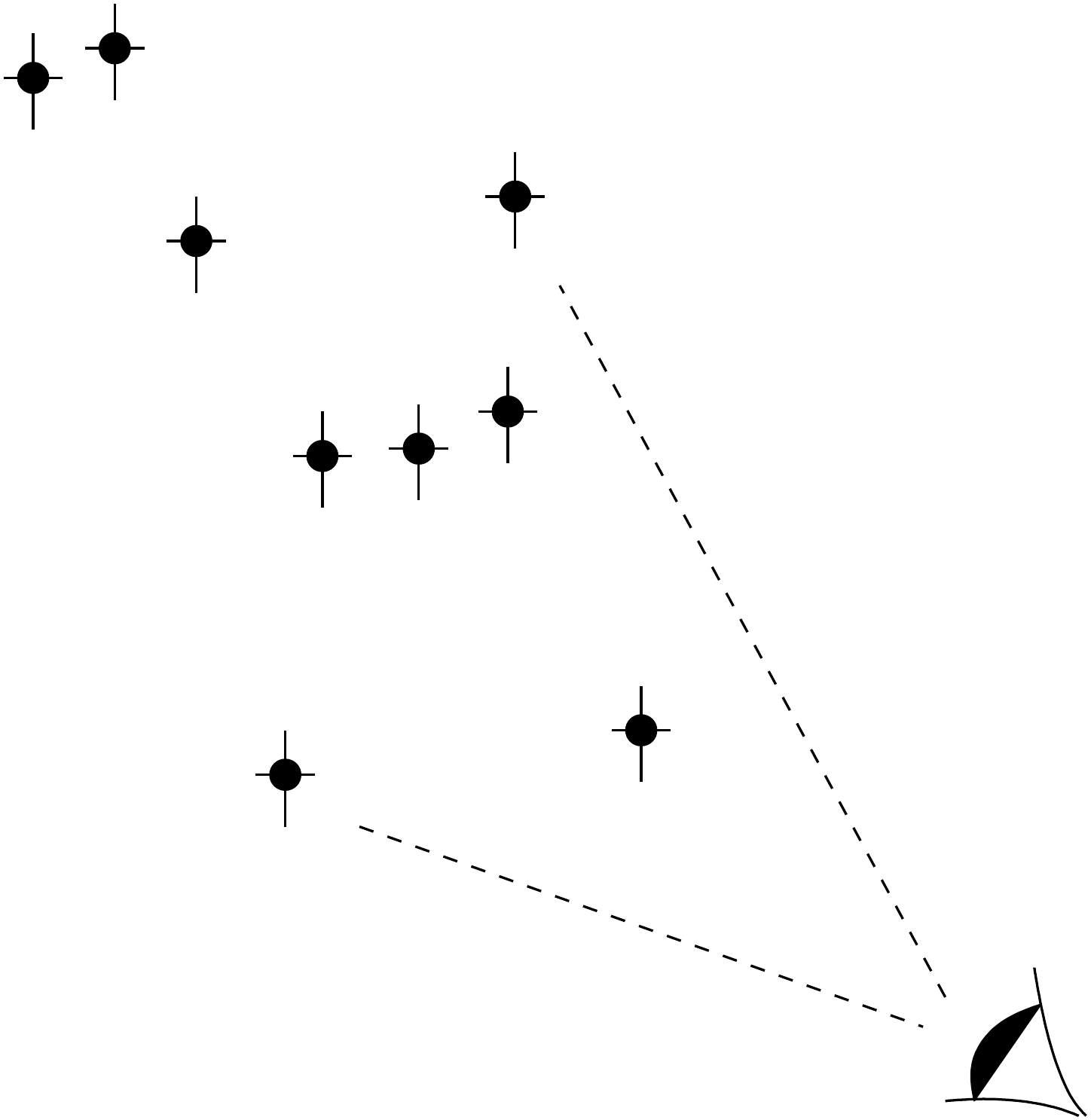}
\caption{\small Propagation of signals (latency) can effectively break symmetries too by providing
such a separation of scales as to be indistinguishable from a fixed boundary condition.\label{fixedstars}}
\end{center}
\end{figure}
Fixed stars may promise something resembling an amorphous solid, that provides a reference
without actual symmetry breaking, by virtue of an almost complete decoupling of timescales.

The fixed stars are a form of gaseous memory, joined together by our
observation on Earth.  They are (for most intents and purposes)
causally and functionally independent.  The promises they are (or
were) able to make to us are rather simple and non-interactive.  If
this situation happened on a closer scale, we would not easily trust
such unidirectionality. Science places much trust in the constancy of
the heavens\footnote{\em "I am constant as the northern star, of whose
  true-fix'd and resting quality there is no fellow in the firmament."
  Julius Caesar (III, i, 60 – 62) Just before Caesar is stabbed in the
  back for his intransigence.}.
\end{example}
When we use the term `quantitative', most of us are conditioned to thinking of distances being
real-valued continuum measurements. This may or may not be the case, depending
on the semantics of the measuring scale. 
\begin{definition}[Semantic (or qualitative) distance]
A measure of similarity in interpretation.
\end{definition}
\begin{example}[Semantic distances]
Semantic distances are based on comparison of
symbolic sequences:
\begin{itemize}
\item {\bf Hamming distance}: 

The number of symbols that differ between two sequences.

\item {\bf Hop count in an associative network}:

An obvious measure of distance in a graphical structure in the number
of hops between nodes; i.e. the adjacency count.
The concept of distance in a network of associations may simply be
used to define a notion of similarity of the concepts held at the
nodes.  

\item {\bf Semantic hashing}:

  Semantic hashing\cite{semantic hashing,hashtopics} has a specific
  meaning in the literature.  It refers to an attempt to reduce text
  in a document to a small feature space, based on importance of
  keywords and tags, in such a way that similar documents have similar
  hashes.  A hash promises a {\em semantic summary} of a superagent
  collection of individual words. The metric properties of the feature
  space are preserved by the hashing, but the choice of keywords is
  manual and ad hoc (see example \ref{ex8}).  Similarity does not promise to represent
  semantic relationships that are not derivable from the tags.

\item {\bf Sparse distributed representations}: 

Mapping similar keys to similar locations, in a sparse encoding. Similar to semantic hashes. 

\item {\bf Semantic routing}:

  The goal we seek in classifying meanings is achieve a consistent
  semantic routing of observed phenomena.  This is
  partly the goal of {\em ontology} (see section \ref{spanning}).  The
  ability to bring universal meaning to phenomena is fraught with
  misunderstandings, however\cite{burgesskm}.  Methods fail to be
  universal because they impose an ontology from the top
  down\cite{certainty}, rather than adapting to context.

\begin{example}[Nominal addressing]
By using semantic labels for locations, instead of metric coordinate, 
we can develop different approaches to locating attributes. This
idea has been exploited in Name Data Networking and Content Centric Networking\cite{ndn,ccn}.
\end{example}
\begin{example}[Semantic routing and ontology]
  Before the successes of machine learning and pattern recognition, ontologies were designed
  by hand, using logical languages like OWL\cite{burgesskm}.  This was
  unsuccessful, as ontologies are not universal; they are spanning
  sets (see paper I). Today, ontology is assumed to be discovered in
  context by machine learning methods like PCA\cite{duda1}.
\end{example}

\end{itemize}

\end{example}
Unlike metric distances, qualitative aspects are not continuously
distributed, but they can still be measured (quantitatively) using the
notion of sets and basis sets.

\subsubsection{Locatability of assets (sorting and discriminating identities)}

The ease of learning, representing, and then finding assets stored in a semantic space
(e.g. storing and retrieving concepts, from some starting key), depends on being
able to reliably route keys to the same (or at least)
similar locations in a memory every time we detect them.  If one could
further arrange for similar concepts to map to similar spatial
locations, then an approximate sensation or idea might even trigger a more
complete or specific memory, and vice versa. This would be the
beginning of what we mean by representing associative `lateral thinking'.

\begin{example}[Closeness of related concepts]
Must similar concepts be close together in a knowledge space?  This
must depend on the encoding of the space and the mechanism for
coordinatizing it. Without this mapping, there is no {\em a priori} link between
the two. 

Recent work in Nature using MRI scans of English language readings
showed how brain activations for words were hashed to broadly cohesive
cortex regions\cite{semspc}, with similarities even across different
subjects.  The authors posited that English words fell into clustered
meanings that they labelled broadly as the categories: visual,
tactile, outdoors, place, time, social, mental, person, violence,
body-part, number.  In current parlance, these could be also called
spacetime, state, self, senses, relationships, and danger. These
results are based on single words, however in Hawkins\cite{hawkins},
memory is assumed to be formed as temporal fragments, so such
individual words may not have the significance of independent semantic
atoms. Nevertheless, the results are intriguing\footnote{At the time
  of writing, some of these MRI results are uncertain due to the discovery of
a longstanding bug in the software.}.
\end{example}

\subsubsection{Physical representation of semantic differences}

Should semantic distances be proportional to metric distances when
representing knowledge as space?  Spacetime accounting principles, like the conservation
of energy, and principle of least action suggest that it would be natural
for this to be the case; however, that presumes a continuum model of space.
In network structures, small distances can be quickly amplified into
large ones by branching. If we encode processes for 
structural discrimination, which sort encoded inputs, these must lead
to bifurcations, which in turn can lead to wide (exponentially
diverging) separations at low cost across several spacetime
dimensions, so what is `close', in one measure, might actually end up
appearing far apart by another.

Finding an object may or may not imply physical visitation, so the
constraints on what `easy to find' means are relative to the agent
concerned.  In paper II, the role of directories (content indices),
for locating promises within agent boundaries, was described. A
promise directory or index is a structure that acts as a transducer
between the semantics of promises and a metric map of resources.  It
plays the role of a lookup table or hashing function, converting roles
to locations for interior agents. Directories are intrinsically linked
to dynamic and semantic transparency.

\begin{example}[Searching versus coordinatization]
  An item in a warehouse has to be retrieved by a physical visitation
  (e.g. person or robot), but retrieval of information about its
  presence may be looked up in a directory or inventory.
\end{example}
To map out a space, we need markers and/or graduations to measure distance.
Markers that select particular locations, breaking
translational symmetry, are used to locate absolute points.
Inequivalent semantics thus break translation symmetries.
To map out knowledge in spacetime, we have only spatial
degrees of freedom with which to encode persistent memory.  The
distance between locations is significant to the ability to retrieve
encoded information, as it leads to notions like {\em association} and {\em
  locatability}.  

\begin{example}[Map of spatial semantics]
How do we find a shop in a city, or an item in a
supermarket? Is there some directory or hashing function, or a funnel
that routes messengers to directed locations?
\end{example}
\begin{example}[Discriminating and sorting structures]\label{ex8}
Coin sorting machines sorts coins and eject them into different piles.
Passport queues discriminate people by their countries of origin.
Baggage handling systems discriminate luggage by origin flight number.
\end{example}

An observer examining another observer's semantic space might see a
projection of points that appears to bear no resemblance to the
relative separation of the original spatial relationships.  A metric
preserving map $M()$ exhibits isometry, i.e.: 
\beq 
D'(M(x),M(x')) = D(x,x') 
\eeq 
Conformal maps preserves angles, shear map preserve
areas, and so on.  We have to understand these spacetime concepts and
translate them into knowledge representations to appreciate the
effects they transmit to our comprehension of what we believe to be
recorded knowledge.

\subsubsection{Metric and semantic coordinate bases, spanning trees}\label{spanning}

The spanning of graphs and vector spaces with metric coordinate
systems was discussed in paper I\cite{spacetime1}. Spanning trees are
one of the main approaches to the addressability and navigation of
discrete spacetime structures, as they form natural hierarchies for
tuple-based addressing.  The same approach may be used for semantic
decompositions, with nominal rather than ordinal values.

\begin{example}[Nominal coordinates]
  Computer file systems, web documents, markup languages, or document
  data formats, like XML and JSON, all have a regular hierarchical
  structure from the root of each document.  This may be used to
  define a coordinate systems, based on the names and ordered
  numberings of identifiable objects within the document.  Locations
  are specified by URI, URN, URL path identifiers. These path locators
  provide nominal semantic coordinates based on a spanning tree formed
  from the markup elements. This is a generalization of hierarchical
  file naming, e.g. \verb+C:\parent\children\sarah+.
\end{example}
In order for similar things (semantics or meanings) to map to similar
locations, the encoding of properties that form the key must promise
to be similar too.  Thus `car' and `lorry', as words, would not be
likely to occupy a similar neighbourhood, but if both are associated
with wheels and transport, then they can be. Hierarchical
classification can thus play a role in `namespacing' concepts, as we
shall see in section \ref{concepts}. This is the basis for taxonomy
by the idea of `prototypes'\cite{langacker1}.

\begin{definition}[Taxonomy]
  A directed acyclic graph (tree) of name agents forming an ordered
  hierarchy of concepts and exemplars, with the level of
  generalization increasing towards the root of the tree, and sub-categories
diverging towards the leaves.
\end{definition}
Tree structures apply a discrimination criterion to separate and
resolve mutually exclusive branches, and separate conceptual branches
into `namespaces' (see paper II). This forms a path-like coordinate
system, as is common in computer file systems, with directories for
each nested superagent.
A generalization of taxonomy, as used in computer science, is found in
attempts to apply first order logic to relationships between named
concepts. This is called ontology. The specific meaning of ontology
has evolved since the widespread use of machine learning.
\begin{definition}[Ontology]
  A network of
  agents, representing concepts, contexts, and semantic associations,
  whose spanning tree forms a taxonomic hierarchy.
\end{definition}
It follows from basic promise theory axioms that each agent is free to
decompose the world, according to its perception, into (super)agents by its own criteria and
resources\cite{promisebook,spacetime2}.  Similarly it is free to
associate meanings to their exterior promises.  As one forms spanning
trees of such named agencies, physically or conceptually, by
annotating the representation of the physical domains with names and
labels (see section \ref{annotation}).  As an approach to addressing, taxonomies may be inefficient,
if they become too deep because they must be parsed serially and
recursively.  Fast hashing aims to reduce the dimensionality of a
space for approximate match.  Shallow, broad trees, or balanced trees
are often preferred as data storage structures for their
predictability and cost during parsing\cite{watts1,berge1}.

\subsubsection{Symmetry breaking and deterministic descriptions}

Purely relative, differential characterizations of a system (such as the
propagation rules we refer to as laws of motion), are essentially empty
of content, in the sense that they make vacuous predictions based on known
symmetries and constraints. Without fixed boundary information, 
e.g. about initial or
final states, no predictions can be made, because there is no content,
only relative constraints. A propagation rule can
thus be made to predict almost anything; only sources and sinks of
information are non-invariant, symmetry breaking, singular
states and transitions\cite{jan60} that can anchor these predictions
to some fixed information.  Boundary markers inject information into a spacetime
description (see section \ref{significance}).

The kind of deterministic prediction, which follows from a propagation of boundary conditions,
tempts the idea of a rigid universe, with mechanical predictability.
The classical notion of Newtonian spacetime is effectively such a 
solid-state lattice of points, with rigid adjacencies. It brings a mathematical
predictability to spacetime paths: if we traverse a link in a known direction, we 
can count where we are in this predictable world.
We are so used to
thinking in this way that it is tempting to believe that we cannot represent or
encode much information without such a rigid spacetime. However, this is not the case.
Even agents in a fluid phase may imitate the information in a solid phase, if there is a
sufficient separation of timescales. Example \ref{fixedstarsex}, about the fixed stars,
indicates how this comes about in a well known astronomical case. Other forms of
dynamical symmetry breaking are found in biology (where membranes and differentiated
regions of cells combine from differentiated scale interactions), and in weather
patterns (where clouds, cells, and fronts lead to the formation of spatial cells
of high and low pressure or moisture).

The separation of timescales is how effective local boundary conditions emerge
in a dynamical system, allowing symmetry to be effectively locally
(dynamically) broken for all intents and purposes, at one timescale, while actually
remaining globally invariant in the greater reckoning. However, there is a conundrum here
when we think of systems as discrete information: how can we understand timescales,
when the very clocks that define them are made of the very stuff they measure.

\subsection{Temporal semantics}

Time is the rate at which an observer experiences change, i.e. the rate at which
configurations in space are perceived to change. 
We don't need to know why spacetime elements change, but when they do, an
observer's clock ticks.

How an observer perceives time may depend on what context or state of
mind the observer is in (e.g.  awake, asleep, busy, idle, and
equivalent non-human states).  This contextual aspect to temporal
interpretation affects an observer's ability to assess and
characterize existing promises and changes; it affects
 the granularity of time's agency, and its sequential ordering.

Time cannot be defined without a semantic space (i.e. some collection of agents
that can represent state), because time does not happen unless there
is a clock to measure it.  Thus, without spacetime semantics, there is no
natural justification for a relationship between space and time, and time
could not exist in the way we understand it.
To define the smallest clock we need
sufficient number degrees of freedom to represent uniquely
distinguishable states.  Intervals of time are often defined {\em ad
  hoc}, based on a standard scale for speed, in the mathematical
tradition of imagining time to be an independent theatre, but Einstein
showed us that this is not a consistent viewpoint: we must understand
that clocks are made of space, and thus space (agency) and time are inseparable.
\begin{example}[Stories, narratives, world line knowledge]
  Spacetime structures that relate to knowledge modelling include
  processes, concepts, associations, contexts, etc. These form
  storylines or narratives, which in turn describe an experience of
  `proper time' for an observer\footnote{The concept of proper time is
    well known in general theory of relativity as the time experienced
    by an observer along its path\cite{weinberg1972gravitation}.}.
For example: a person comes into the entrance,
  goes through a changing room that promises lockers for storage, then
  the bathing rooms, and so on. 
\end{example}

\begin{example}[Temporal semantics]
The many interpretational semantics of time include the following examples. 
\begin{enumerate}
\item {\bf Local metric/sequential time}: (e.g. a stopwatch counting). In practice, all clocks
  have finite number of configurations to use for memory and thus can
  count only for a limited interval After this, time must repeat,
  though not necessarily in the same order on each covering.

  e.g. we may construct a precise equational specification of
metric time ${\cal T}_C$ as a simple counting mechanism that satisfies the algebra
  an image $Alg(\Sigma_\Phi,E_\Phi)$ of a clock $C$. This is a simple clock that wraps
around to zero when it reaches a maximum counting value. The value `1' represents a tick
of the clock. Let's give a detailed explication of this one case to illustrate.
\beq
  {\rm\bf signature} ~ \Sigma_C  &\;&\nonumber\\
  ~~~~ {\rm\bf sorts} &:& {\cal T}_C \in \mathbb(Z) \nonumber\\
  ~~~~ {\rm\bf constants} \nonumber\\
  0 &:& \rightarrow {\cal T}_C \nonumber\\
  1 &:& \rightarrow {\cal T}_C \nonumber\\
  T_\text{max} > 1 &:& \rightarrow {\cal T}_C \nonumber\\
  ~~~~ {\rm\bf operations}&:& \nonumber\\
  \cdot &:&  {\cal T}_C \times {\cal T}_C \rightarrow {\cal T}_C.\nonumber\\
  {\rm\bf equations} ~ \Phi_C\nonumber\\
  (\forall T \in {\cal T}_C) ~~~~~~~~\;T + 0 &=& T ,\nonumber\\
 0 +T &=& T ,\nonumber\\
  T+T_\text{max} &=& T \nonumber\\
  T_\text{max}+T &=& T \nonumber\\
  T_1+T_2 &=& T_2 + T_1,\nonumber\\
  T_1+(T_2+T_3) &=& (T_1+T_2)+T_3,\nonumber\\
  (\forall T, \exists -T \in {\cal T}_C) ~~~~~~~~\;T + (-T) &=& 0\nonumber\\
(-T) + T &=& 0\nonumber\\
  {\rm\bf end} 
\eeq
This is a discrete representation of $\mathbb{Z}_N$ where $N={T_\text{max}}$.
Quasi-continuous versions of a clock can be constructed by increasing the size of ${\cal T}_\text{max}$. We can extend this approach to specify formal semantics for any measure
on a collection of agents.

\item {\bf Clock/calendar time}: 

A tuple of metric time clocks that
  accumulate state in a cylindrical coordinate system e.g. (seconds,
  minutes, hours, days, $\ldots$), or `12:06:22'.  Each dimension is a
  different granular decompositions of accumulated state,
  representable as a metric clock. As the second hand of the clock
  wraps around, the next counter advances, in a cylindrical wrapping.

\begin{example}[Coordination by fixed clock]
The temporal analogy to measuring positions relative to the fixed stars, or fixed markers,
is measuring time relative to a common clock. Agents can coordinate their activities
relative to a fixed clock, or (if they have sufficient internal time-keeping
capabilities) they can synchronize their
own internal clocks by extensive peer to peer coordination. 
\end{example}

\item {\bf Time as a state and as an agent}: 

A map $T_\text{state}$ of string names $S_N$ to 
clock or calendar times ${\cal T}_C$ that characterize all of the times
on a calendar or clock times, e.g. `now', `Monday morning', `Friday afternoon', `then', `recently', `last week',...
It could be distance travelled from an origin, or amount of sand fallen.

\item {\bf Global metric/sequential time}: 

In which an entire system
  is its own clock, i.e. any change represents a tick of a global
  clock. This behaves exactly like a transaction log or software version control system,
  in which a tick corresponds to an ordered transactional `commit' to
  a database i.e. each tick is the record of a change between adjacent
  spacelike hypersurface.

\item {\bf Rate of change}: 

Distinguishability of temporal state, at fixed location, relative to an observer's separate clock.

\item {\bf Time as a resource}: 

Do you have time?

\item {\bf Trajectories}: 

Open spacetime patterns: non-repeating spacetime path.

\item {\bf Orbits}: 

Closed spacetime patterns: circling, waiting, eating, sleeping.
\end{enumerate}
\end{example}

Absolute and relative notions of time do exist (see \cite{jan60} for a discussion in the framework of
equational specifications and state machines).
A discrete aggregate concept of time is essential to creating spacetime associations, because if
time had infinite resolution, it would be infinitely improbable that coordinate pairs
would identify the intended points. 
\begin{definition}[Concurrent interval]
  A collection of events, signals, or ticks, arriving either serially or in parallel,
  that are considered to occur `simultaneously', during a single tick of a
  clock.
\end{definition}
Clocks cannot have infinite resolution, as this would lead to the need for infinite information.
\begin{example}[Concurrent data]
Any discrete aggregation process leads to a concurrent time approximation.
Each time we bake a cake, we assemble inputs into a unit output. There is
no discernable order to the mixture at the granularity of the cake. Each time
a population census is taken, the counting result stands as a single moment
in time until the next one it performed, and there is no memory of the
process by which the counting took place. The entire process is a concurrent
interval, coarse grained into a single output. The uncertainty about associating
concurrent events grows with the coarseness of the concurrent interval.

  Concurrent intervals include discretizations of streams: network
  packets, single promises, packages, enveloped messages, containers,
  words, etc. Any natural, discrete unit of aggregation may be coarse
  grained into an effective unit of time.
\end{example}
The concept of concurrent or simultaneous events is only possible in a
discrete spacetime.  The discretization of spacetime, whether one
chooses to think of it as fundamental or a limitation of an observer's
faculties, is essential to the ability to limit the resolution of
measurement information, and generate repeatable sensory experiences
that can become knowledge. In this sense, approximation is the key to be
able to represent stable knowledge.

\subsection{The semantics of spacetimes and dynamical systems}

For dynamical and functional interaction, space and time interact in
inseparable ways, so it is convenient to simply talk about spacetime.
The temporal aspect of `world lines', or paths that trace the linear
experience of an observer, play a significant role in staging
experiences into storylines that we associate with meaning.
The selection of a preferred path is a form of directional 
symmetry breaking in spacetime. Direction is one dimensional, thus time is one-dimensional.

In any extended structure, which links agents
together in chains by adjacency, sequential partial ordering is a necessary
phenomenon. Thus, timelines, world lines, and storylines are a
necessary outcome of understanding interaction trajectories for
localized observers\footnote{Even if the goal is to manufacture an
  artificial representation of a problem, such as the layout of a
  department store, such a structure has to be able to learn and adapt
  by evolving in relation to interactions with the agents that parse
  its structure.}.
This unidirectional sequencing manifests in a number of ways:
\begin{itemize}
\item Bottom up representation of communicated language, as symbolic streams.
\item Top down reduction of patterns to find atomic symbol alphabet (reductionism).
\end{itemize}
Processes that are bounded at a spacetime location lead also to a directionality:
\begin{enumerate}
\item Converging towards a fixed end state.
\item Diverging from a fixed start state.
\end{enumerate}

\subsubsection{Assigning coordinates}

Ordering of agents and signals is a cooperative spacetime phenomenon
(see paper I) that requires there to be symmetry breaking, i.e.  the
definition of a preferred direction in the degrees of freedom of a
collection of agents.  How one assigns ordered labels to a collection
of agents goes beyond the scope of this work, but has been considered
elsewhere\footnote{Unique addresses can be assigned to a gas of agents
  too in a self-organizing manner in ad hoc
  networks\cite{treecast,doss1}.  If addressing comes with
  crystallization of a addressable lattice, e.g. a collection of cells
  that self-organize into a regular pattern...then we need a notion of
  di-polarity or orientation for the joining of independent fragments
  to bring about monotonicity in the numbering.  In an already solid
  phase, the problem is simpler, and the symmetry is broken by
  choosing an end point, e.g see the self-organizing spanning
  trees\cite{borrilltree}.}.
The use of a central coordinator or observer, connected directly like a routing hub, with
the ability to discriminate paths has many advantages, because
it can be the discriminator of order as well as the agent by which
messages are routed. This obviates the need for the end node agents to
promise a specific identity.  A hub agent would naturally double as a
router for passing signals\footnote{This is true in the design of IPv6, for example.}.

\begin{example}[Cooperative waves]
Waves that pass through agents by local or global cooperation are spacetime
relative phenomena based on symmetry breaking. For example, green waves that
move through traffic lights to keep traffic moving are timed relative
to a fixed wall clock, or controlled by a single controller. All traffic lights
promise to turn green at a coordinated time relative to a fixed marker (like the fixed
stars/fixed wall clock examples).
Similarly, fan waves at sports games propagate by observation of nearest neighbours.
\end{example}

\subsubsection{Time series}

Time series are a form of data that expresses a trend or a pattern over an interval of
sequential times. The meaning of time series depends on the semantics of the time-base.
Time series may represent open ended trajectories or closed circuit periodograms to
respect and model the underlying freedoms of a spacetime trajectory\cite{box1,usenix9315,burgessDSOM2002}.
\begin{example}[Periodic time-series]
Behavioural data form time series with distinct patterns
(see figure \ref{timeseries}). In the figure, we can imagine measuring
resource consumption in a workplace, or the rate of information flow
in a computer network.  The time base covers a single working week
from Monday to Sunday. When we reach the end of one week of samples, we overlay
new data at the corresponding time of weekday by modulo arithmetic.
The aggregations occur with the same temporal semantics for each interval,
assuming that there is a repeated pattern of behaviour each working week. 
Unbiased data measurements reveal strongly auto-correlated patterns
that confirm this pattern in many systems driven by human behaviour.
By representing the pattern in the semantics of the time-base, we can effectively
learn the temporal pattern efficiently, and compare only semantically equivalent
times. This temporal cyclicity traces back to the rotation of the planets.
\begin{figure}[ht]
\begin{center}
\includegraphics[width=10.5cm]{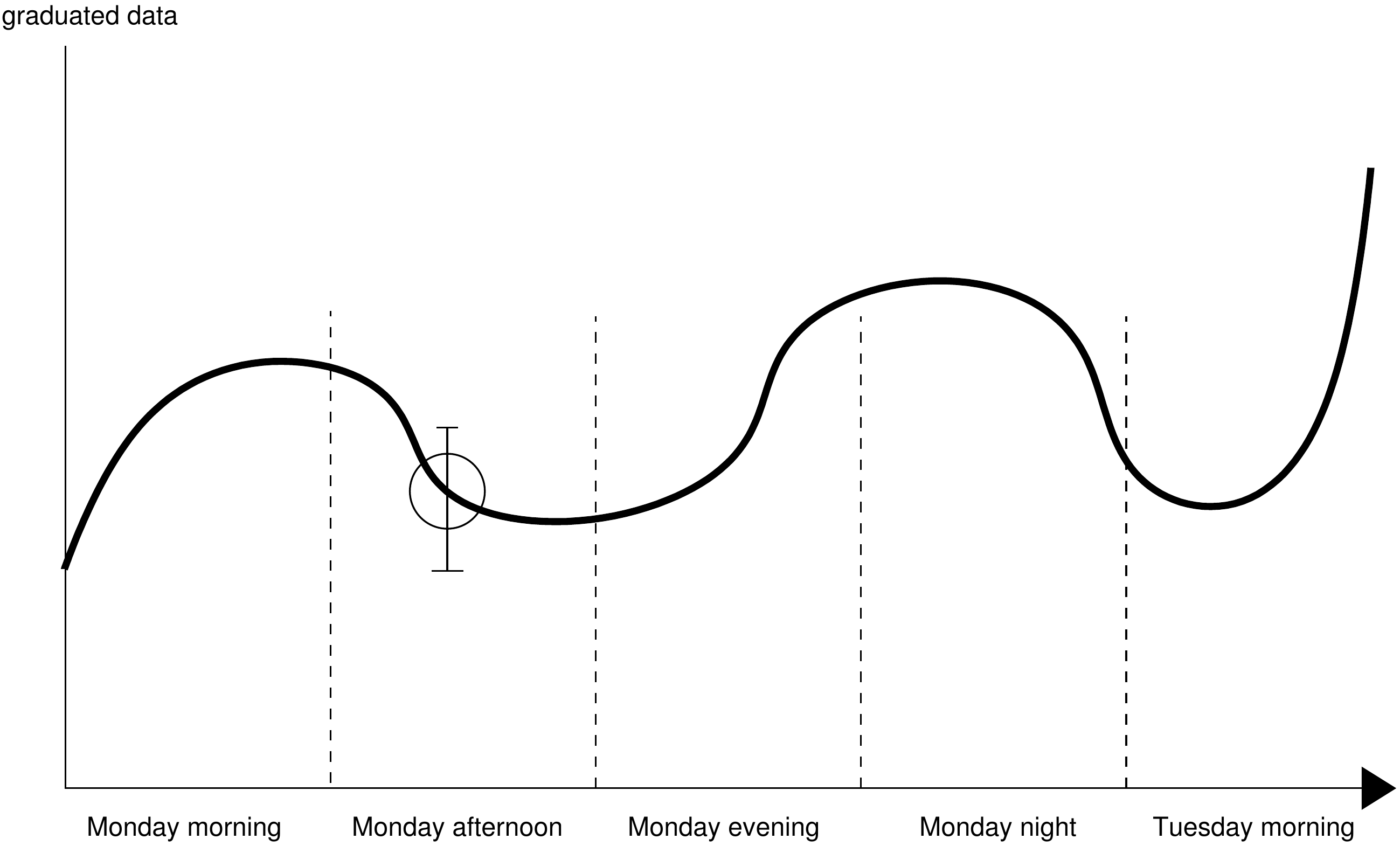}
\caption{\small A time series measures sampled data against a semantically
  significant ordinal timescale. The ordering might be open,
  bifurcating or periodic. The vertical scale resolution is the
  self-calibrated perception of sensors, and the horizontal scale
  resolution is based on sampling frequency relative to some larger
  model that is identified as meaningful in this context (e.g. the
  working week).\label{timeseries}}
\end{center}
\end{figure}
\end{example}
\begin{example}[Linear time-series]
Discrete time series are found in symbolic strings, whose time series
form syntax.  In computational and linguistic literature, a grammar is
an alphabet with a number of generative rules, leading to computable
patterns. In bio-informatics, grammars may be seen in
DNA\cite{jerne1}.  Grammars are classified by the kind of computer
needed to parse the pattern faithfully\cite{lewis1,wood1}.  In promise
theory we phrase it as a voluntary cooperation construction: a grammar
is a collection of agents and superagents that make promises, and
whose top level adjacency has the form of a one dimensional
chain\footnote{Human language is linear too, but thinking is not: we
  experience flashes of insight, in which it is not necessary to
  recount a story in our minds, step by step (though we may do so
  later to rationalize it). This is evidence that thinking is not
  performed in the manner of a Turing machine.}.  Languages are
related to algorithmic state machines, according to Chomsky's
theorem\cite{chomsky}: they are spacetime phenomena, with concurrent
words/symbols/statements and temporal progression.
\end{example}
\begin{lemma}[Inseparable space and discrete time (including time pattern)]
The representation of a timeseries, as a concurrent phenomenon,
involves spatial pattern representation within a concurrent time slice.
\end{lemma}
Any repeated pattern marks out a discrete interval whose semantics represent a scaled
concurrent phenomenon.

\section{The structure and semantics of data patterns and representations}

Underlying all communication is the ability to attach meaning to
patterns. It is the basis of information transfer in communications
theory, linguistics, and computation. Grammars are scaling patterns of the
agencies by which linguistic symbols are carried, and projected into
sequential (one dimensional) streams. Sequential streams are, by their
nature, temporal when communicated, but may be stored in a spatial
memory.  Chomsky showed that grammars are associated with transition
systems operating on discrete linear patterns\cite{chomsky}; from this
we have a theory of signalling and data encoding\cite{lewis1} that
goes beyond Shannon's basic information theory\cite{shannon1}.

The encoding of information
signals may take a variety of representations, of which natural
languages are the most complex forms.  In information
technology, it has become the norm to speak simply of `data', as a stream of symbolism 
for encoding by a plethora of underlying binary formats (ASCII, EBCDIC,
UNICODE, least significant byte first, binary-coded decimal,
big-endian, little-endian, etc) is taken for granted, precisely because
it is of fundamental importance.

\subsection{Information content of space}\label{infosec}

Consider a set of agents $A_i$, forming a semantic space. At each point, an agent
may have an alphabet of promisable characteristics 
$C_i$, with index $\alpha_i=1,2,\ldots C_i$.
We can define the distribution of promises by
\beq
p_{\alpha_i} = 
\begin{array}{cl}
p & \text{if $\alpha_i$ is promised, i.e.} A_i \promise{+\alpha_i} \Unspec\\
0 & \text{otherwise}
\end{array}
\eeq
and
\beq
\sum_{\alpha_i=1}^{C_i} p_{\alpha_i} =1.
\eeq
Then the intrinsic information at point $A_i$ is the Shannon entropy\cite{shannon1}
\beq
S_i = - \sum_{\alpha_i=1}^{C_i} p_{\alpha_i} \log p_{\alpha_i}.
\eeq
We note that the information at each point is potentially based on a different
alphabet $C_i\not= C_j$, and is therefore {\em a priori} uncalibrated. In order to compare the information
over a path or region of spacetime, a common alphabet is needed (see also the discussion
in paper II, section 2).

In order to information to propagate to neighbouring agents, e.g. to act as a
fixed marker signpost, agents need to have a non-empty overlap in their
promisable states, i.e. a common sub-alphabet
\beq
A_i &\promise{+C}& A_j\\
A_j &\promise{-C}& A_i.\label{propcrit}
\eeq
Thus, over an extended region or path, spacetime agents must have a minimum
common alphabet of symbolic promises (see also paper II, section 2). This issue
is related to the notion of schemas in data.

\begin{definition}[Translation invariance of information alphabet]
A region of space may be said to be translationally invariant over its
information alphabet if the alphabet of each agent is everywhere constant, i.e.
\beq
d_+ C \Big |_i = C_{i+}-C_i = 0.
\eeq
This is a non-local coordination, or long range ordering. See paper I about derivative $d_+$.
\end{definition}
The injection of new information, at a particular location $i$, is an insertion
of information that can be promised to agents capable of recognizing the same alphabet.
\begin{definition}[Non-local informational entropy]
The distribution information of alphabetic type $\alpha$ is defined by:
\beq
S_{\cal R}(\alpha) = - \sum_{i\in \cal R} p_{\alpha_i} \log p_{\alpha_i}.
\eeq
This represents the intrinsic information in the variation of property $\alpha$
over a region of space. 
\end{definition}
\begin{lemma}[Breaking translation invariance reduces information]
If the space is translationally invariant for all $\alpha$,
then it has maximum entropy $\log |R|$. If the $\alpha$ invariance is broken, then
$S_{\cal R}(\alpha) < \log |R|$, for any or all $\alpha$.
\end{lemma}
This is proof is self-explanatory. It refers to changes in the promises
at points within a region, for fixed alphabet.
A change in alphabet can also break
translational invariance, as observed by an independent observer
who can recognize the alphabets of all the
agents, however this change is not generally recognizable to neighbouring agents whose
alphabets are not equivalent, because they may only be able to recognize the
alphabetic properties they have in common.

\subsection{Data and data types}

Any encoded pattern, in any medium, represents data, and may be measured for its
information content.
We should distinguish data as a promise from the promise concept of 
data containers or agency.  Data
are represented by promised values, while the containers of the
representation are agents.  Thus, even non-scalar data type may be
represented by scalar promises (scalar refers to two different things here).
\begin{definition}[Data]
A stable pattern promised by any kind of agent or superagent structure.
\end{definition}
The significance of the pattern depends on the overlap of promiser and promisee's
interpretation\cite{promisebook}.  In promise theory, agents
are the only medium in which patterns can be represented, by design.
Data are thus the pattern values promised by agents.
Data values fall into two major interpretations:
\begin{definition}[Nominal data values]
The mapping of patterns indicating into a set of symbols with distinct identity.
\end{definition}

\begin{definition}[Numerical (ordinal/cardinal) data values]
  Quantitative representations indicating relative magnitude on some
  tuple of generally real-valued scales.
\end{definition}
\begin{example}[Tuple spanning of values]
The representation of numbers in the Arabic system uses tuples
based on powers of 10, using symbols 0,1,2,3,4,5,6,7,8,9. For instance,
one thousand and twenty three is written (1,0,2,3) or simply 1023, where the
coordinate tuples map powers of ten $(10^3,10^2,10^1,10^0)$.
\end{example}
\begin{example}[Data containers]
  In computer programming languages, the agents that promise data are
  called variables, constants, buffers, etc; in communications
  protocols, and documents, the agents are called fields, etc. 
\end{example}
When communicating data, a sender $S$ agent might copy data from one
representation, in its internal memory, into another, within signal
agent, which is then emitted and absorbed by the receiver $R$.

\begin{definition}[Scalar data]
The promise of scalar data, e.g. $\tt X = 8$ has the form
\beq
A_X \promise{+8} R
\eeq
\end{definition}
In general data must be `typed', i.e. must promise an intended interpretation
in order to make sense to a recipient:
\begin{definition}[Typed scalar data]
A typed promise of scalar data, is a body doublet including an intended type and a value, e.g.
\beq
A_X &\promise{+({\tt integer},3)}& R\\
A_Y &\promise{+({\tt string},3)}& R
\eeq
\end{definition}
Aggregate data are promised in the same way.
\begin{definition}[Aggregate typed data]
The promise of non-scalar typed data, e.g. \\$\tt X = ({\tt integer} 1, {\tt integer}\;3, {\tt string} \;`nine')$ has the form
\beq
A_X \promise{+({\tt integer} 1,{\tt integer}\;3,{\tt string} \;`nine')} R
\eeq
\end{definition}

\begin{lemma}[Scalar representation]
All data promised by a single agent may be represented by scalar promises.
\end{lemma}
The proof of this is trivial: promise body can contain any bundle of promises:
\beq
A \promise{+b} R
\eeq
where $b$ can represent any tuple: a scalar value, an array, a bundle of values,
and even a number of internal agents.
In other words, scalar promises are not the same as scalar values.  Scalar
data refer to single named values.  A scalar promise refers to an internal
property of a single agent\cite{spacetime1}.
We may also remark that agents that promise data form a semantic space.
\begin{lemma}[Data form a semantic space]
Agents promising data are the elements of a semantic space.
\end{lemma}
The proof is straightforward: a semantic space is a collection of
agents that make scalar promises.  They make vector promises to
coordinate in groups. Data values are spacetime scalars, whether
single valued or aggregate types\footnote{The term scalar is also used about
  single value data types, but that is a different meaning. Here we
  mean the tensor meaning of scalar.}.  Objects promise non-ordered
adjacency (superagency) in clusters of non-similar agents.  Arrays
promise ordered adjacency in clusters of similar agents.  The phase of
the space (i.e. its adjacency structure) need not be defined.

\subsection{Schemas, types, and the chemistry of composition}

If combinations of atomic data are received on a regular basis, it
makes sense to bind them into homogeneous `molecules' of fixed types, which can be
recognized and stacked efficiently in space.  Aggregated data thus form
superagents that promise multiple internal data items under a single
identity. The members inside form a `schema' which is given by the
superagent directory\cite{spacetime2}. This leads to a simple form of
hierarchical containment.  Heuristically, schemas are molecular scale
superagents, formed from atomic agents.
\begin{definition}[Schema, data class, or compound type]
  A schema $S$ is a bundle of sub-agents and concurrent promises within a
  superagent.  A schema is thus closely linked with the superagent
  directory, revealing its specific internal structure.
\end{definition}
A schema acts as a declaration of constraint on the form of a
  superagent's composition, without fixing the values of any instance
  of the pattern.
\begin{example}[Data schema]
An object $X$ may contain internal members, whose directory is a set of promises about
structure, but not value.
\begin{verbatim}
X : {
    integer subagent1;
    real    subagent2;
    string subagent3; 
    }
\end{verbatim}
The schema promises three sub-agents, each of which makes scalar promises
about their type, from an alphabet, which contains at least integer, real, and string.
\end{example}
\begin{definition}[Schema instance]
  An manifest superagent whose directory is assessed to
  be an instance of schema $S$.
\end{definition}
Schemas are well known in database theory, where tabular structures are 
ubiquitous\cite{date1}.
\begin{lemma}[A schema has constant semantics]
A schema, and all instances of a schema imply constant semantics
for each distinct promise.
\end{lemma}
As key-value pairs, the semantics of keys are constant, but the values of data are not.
The purpose of a schema is to promise constant semantics across a collection of data,
whose values may vary. This promises semantic calibration, ensuring that data can be
compared meaningfully.
\begin{example}[Schemaless data]
Data that do not have a homogeneous spatial template are schemaless.
Their representation does not have a bounded alphabet.
  A JSON file used as a specification with an ad hoc schema (sometimes
  called schemaless) has only a single data point, so we have no
  statistically significant basis for building a relationship with it,
  unless it is re-iterated intentionally to underline its authority.
  This kind of data may crave greater trust, based on an external authority,
  to reliably interpret the schema.
\end{example}

\begin{example}[Fixed or homogeneous schemas]
  A repeated schema, such as sensor data in the same format, but which
  never repeat the same value are uncertain in value, but one has
  increasing confidence in the meaning of the schema. 
\end{example}
The formation of a schema is a form of spacetime symmetry reduction:
agents take on a restricted pattern, instead of one exploring all
possible freedoms, which may be then repeated like a template.  This is an alphabet
reduction in the informational sense of section \ref{infosec}.

In a space whose agents promise a 
homogeneous schema, locations may not be equivalent in their
data values, but locations are semantically equivalent in their alphabets. 
There is still residual symmetry between agents
with the same schema. If agents are schemaless, there is complete
symmetry breaking, with neither a common alphabet nor
a common value.  The breaking of data symmetry establishes identifiable
anchors or signposts by which data can be compared and aligned, but the ability
to use these as anchors assumes that neighbouring agents can recognize
the inhomogeneity, and position themselves
relative to it.

\begin{lemma}[Schema translation]
If a collection of agents has different schemas, they may interact
with external agents via any agent whose schema alphabet overlaps with all
of the different schemas.
\end{lemma}
The proof of this follows from the propagation criterion in (\ref{propcrit}).
An intermediate agent can select and expose only those parts of a schema that are common,
and present a calibrated interface to the irregular schemas.

A schema alphabet is a simple grammar. The information contained
in its labels may be used to separate data into different contexts, as
it is parsed sequentially.  If a schema changes over time, that too
could indicate a change of context in the data, but it could be a
challenge for the semantics of the data.  In learning, we normally
wish to compare to similar data at several times, with a calibrated
schema, so that we may assume the data have the same interpretation on
each sampling (this is the cornerstone of the scientific method).  If
a schema changes, there is additional uncertainty about data
interpretation, which cannot be easily quantified.  The need for a
stable schema thus depends on the extent to which one expects to form
a stable iterative knowledge relationship with a data source.
\begin{center}
\begin{tabular}{|c|c|}
\hline
Rare events & Regular events\\
\hline
Ad hoc search/detection & Specialized sensor/cached assumptions\\
No representative & Represented by an average\\
Variable schema & Fixed schema\\
\hline
\end{tabular}
\end{center}

The usefulness of cooperation over schema calibration may change the
criterion for spacetime organization.  Different types of agent (i.e.
agents with different schemas) lead to compound that express a
functional chemistry with different functional properties.
\begin{example}[Components and compounds]
  Electronic components fall into schemas we could call resistors,
  capacitors, inductors, transistors, etc. For the purpose of
  searching it is natural to store instances of the same schema close
  together in a semantic space, e.g. in a warehouse.  When exploiting the chemistry of
  interaction, e.g. to build a television, the chief concern becomes
  to mix these components efficiently, so that their unique functions
  are near to where they are needed. Then keeping all resistors
  together is no longer supports the intended semantics.
\end{example}

\subsection{Type calibration, bases, and matroids (recap)}

\begin{figure}[ht]
\begin{center}
\includegraphics[width=7.5cm]{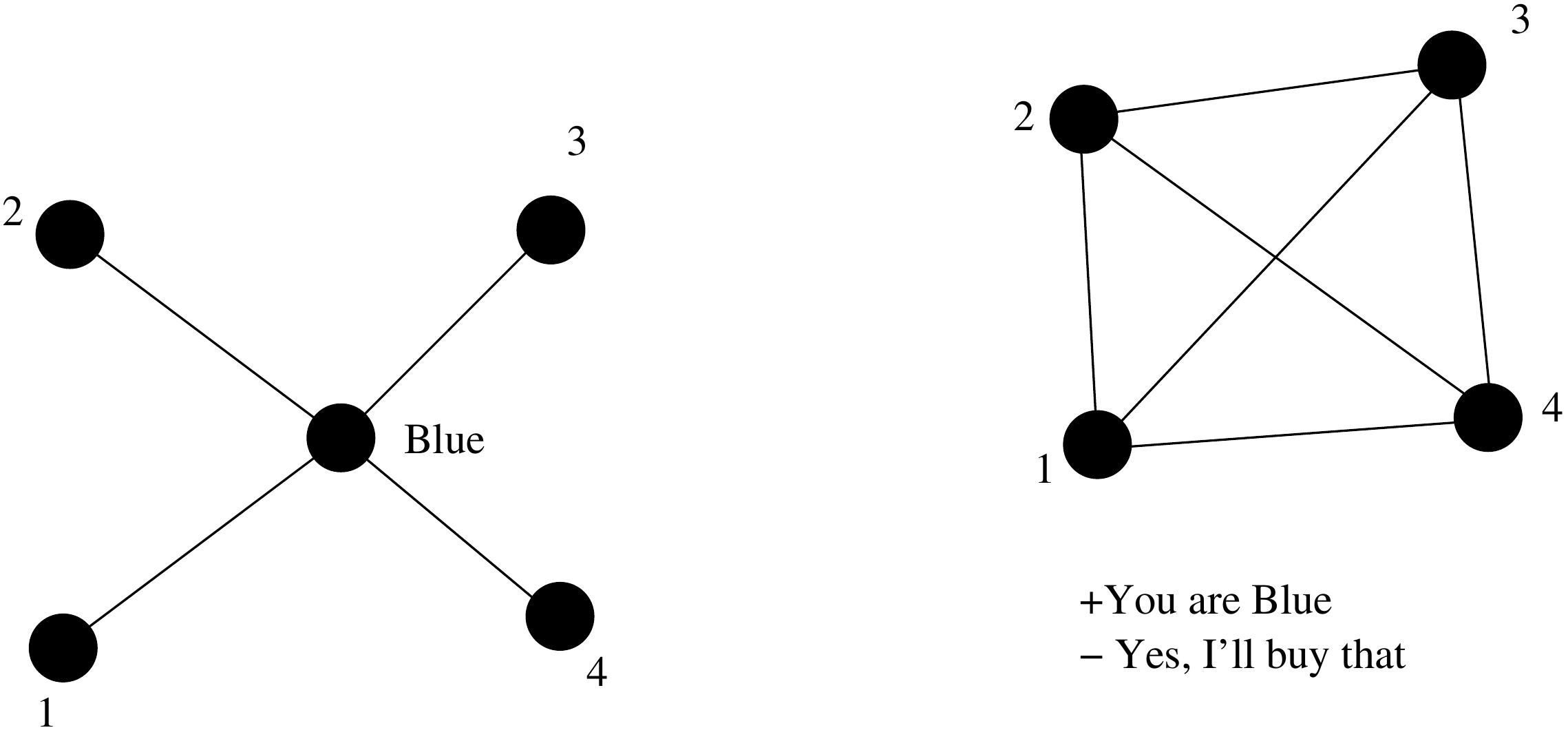}~~~~~~
\includegraphics[width=5.5cm]{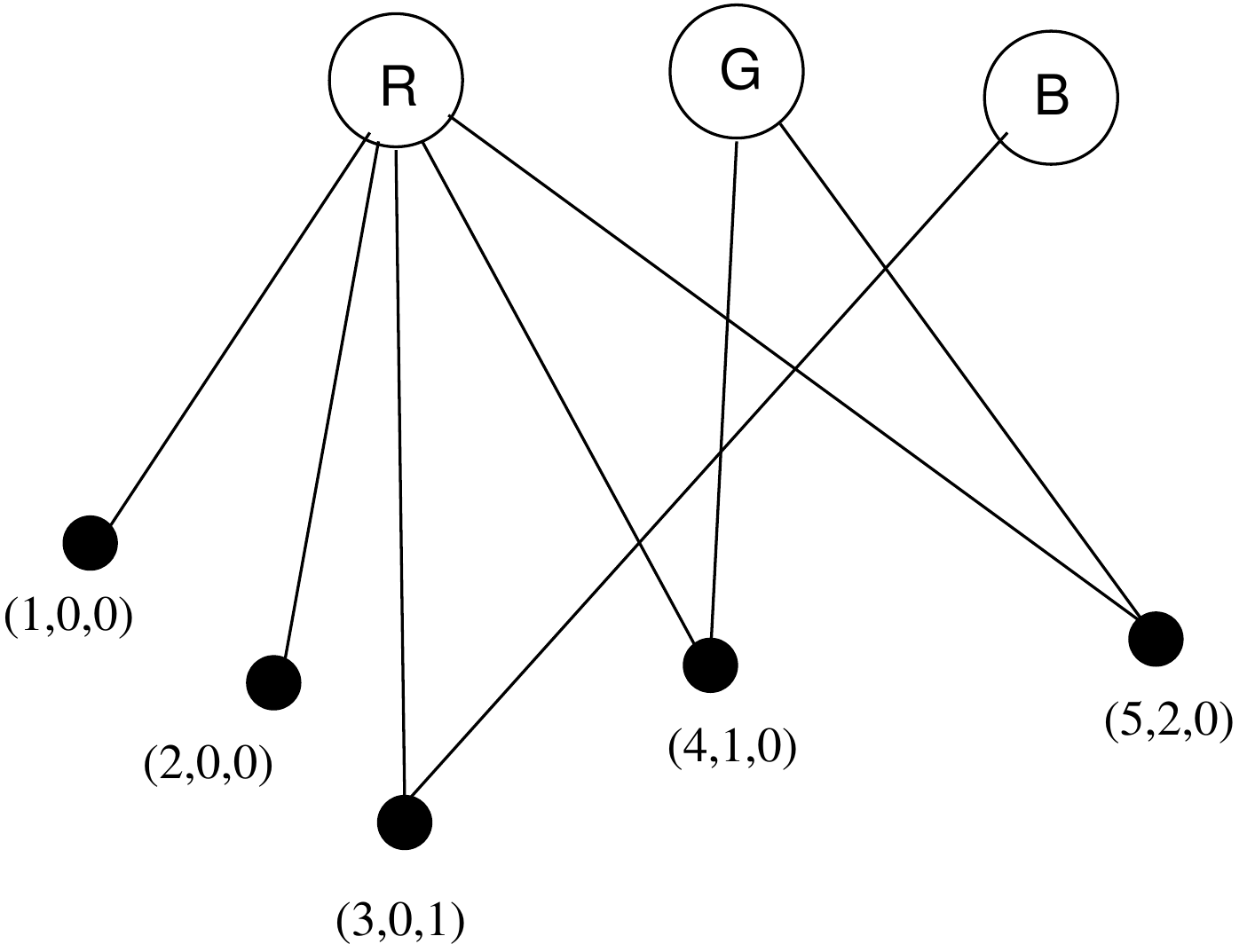}
\caption{\small Calibrating a type property, either by equilibrating with
a single reference source (singularity), or by normalizing a clique.\label{calibration}.
Right, an explicit matroid basis for global properties with three property hubs
adding three components to the coordinate tuples.\label{coordkm}}
\end{center}
\end{figure}

In a promise graph, scalar data type promises may be represented as
vector promises, connecting to an authoritative calibration agent
(matroid basis), which acts as a `standard candle'. Or, they may be
calibrated by mutual equilibrium over a collection of agents that
claim similar promises (see figure \ref{calibration}, and the
discussion in paper I).  For the former case, each such singularity
represents one unique scalar property (see figure \ref{coordkm}): this
is how one forms a matroid or spanning set for nominal properties (see
\cite{spacetime1}). Thus conceptual linkage becomes associated with
spacetime singular agents, or anchors, that ensure mutual calibration,
and semantic stability. The alternative of mutual linkage into a
complete graph involves labelling not just a single agent, but pairs
of links between all the agents, with $O(N^2)$.

The desire for uniqueness and stability often leads us to construct
the single source model for property/type attribution. This makes
concepts rigid, localized, and standardized rather than plastic,
distributed and subject to shifting equilibria. Equilibration with a
central source has a cost advantage ($O(N)$ versus $O(N^2)$),
and turns the scalar property into its own dimensional coordinate
index, as there is now a direct route from this to every instance of
it (see figure \ref{coordkm}).  This choice of basis is related to the
centralization or decentralization of knowledge representations.

\begin{definition}[Symmetry breaking]
  The loss of locational equivalence under arbitrary spatial
  translations.  Spatially distributed promises take on only a
  restricted set of values or patterns, compared to what is otherwise
  possible at only a subset of particular locations.
\end{definition}
In this sense, symmetry breaking is involved in the crystallization of
a gas with arbirary positions to a solid state, as well as any
introduction of edges, boundaries or markers, whether by equilibration
or singular agent representation.

\subsection{Trust and certainty in data calibration}

Accepting an authoritative matroid basis-agent as a calibrated standard for a
spanning property assumes a level of trust. Trust is acquired by a learning process,
i.e. it is based on extended relationships\cite{burgesstrust}.  A promise theoretical subtlety here is
that the property represented by a matroid basis agent (e.g. blue in
figure \ref{calibration}) is promised by the basis agent, and is
accepted by the associated clients.  Any agent could set itself up as
the authoritative source for `blue', and stand in competition, e.g.
namespaces and scoped blocks limit the reach of definitions.  This
becomes particular relevant as we consider representations of
authoritative identity in humans spaces , such as homes, communities,
and cities.
\begin{example}[Types in natural science]
  What agency is the authoritative source for the scalar property of
  being an electron?
\end{example}

\subsection{Data dimensionality: aggregates}

Data are processed serially (each transaction generating the notion of
system time); however, concurrent promises or parallelism may be used
to represent data in parallel threads, and nested superagency may be
used to bundle orthogonal degrees of freedom as concurrent promises.
Concurrent promises are thus the basis for aggregate data
representations.

We can use the scaling arguments of paper
II\cite{spacetime2} to express data representations in terms of agents and
superagents.  By wrapping data inside layers of containers (agents
within agents), we combine schemas, promise independent names, and form namespaces
to limit scope. Internal structure is promised by a superagent
directory or index \cite{spacetime2}.

Two kinds of aggregation may be formed from scalar values:
\begin{definition}[Data structure or data object]
A uniquely named superagent $S$, formed from uniquely named subagents $A_i$, each
of which promises values.
Let $S = \{ A_1, \ldots, A_n \}$,
where each subagent promises typed scalar values
\beq
S &\promise{+\text{name}_S}& R\\
A_i &\promise{+\text{name}_i}& S\\
A_i &\promise{+({\tt type}_i,{\tt value}_i)}& R
\eeq
\end{definition}
Some structures may be transparent, while others maybe opaque to external agents.
\begin{definition}[Transparent, typed structure or object]
An object $S_\tau$, of aggregate type $\tau$ is transparent, iff it promises directory of members
to $R$:
\beq
S_\tau  \promise{+\text{directory}_\tau} R.
\eeq
The directory of type $\tau$ contains the types and data member names of the superagent $S$.
See section 3.7 of paper II \cite{spacetime2} on superagency and directories.
\end{definition}
See paper II for a description of directories.
\begin{example}[Data object]

An object $X$ may contain internal members:
\begin{verbatim}
X : {
    subagent1 : 3,
    subagent2 : 4,
    subagent3 : 5 
    }
\end{verbatim}
If the object is opaque, it promises only its exterior name $X$, and the interior structure is
not promised, i.e. it is unknown to outside agents. We might then call $X$ a `blob'.
If it is transparent, there is a directory of its internal agents and their promises:
\begin{verbatim}
  X.subagent1 = 3
  X.subagent2 = 4
  X.subagent3 = 5
\end{verbatim}
\end{example}
Objects are inhomogeneous collections of subagents that promise any kind or
type. Arrays are totally ordered collections of agents 
promising symmetrical types (see section 5.4 of paper I \cite{spacetime1} concerning ordering).
\begin{definition}[Array]
  A totally ordered collection of semantically equivalent, symmetrical objects, under a
  single common name. Each member of an array makes all the same type promises,
  as the others, but may have different values.
A uniquely named superagent $S$, formed from uniquely ordered subagents $A_i$, each
of which promises different values of the same type.
Let $S = \{ A_1, \ldots, A_n \}$,
where each subagent promises typed scalar values
\beq
S_\tau &\promise{+(+\tau,\text{name}_S)}& R\\
A_i &\promise{+(S_{\tau},\text{name}_i)}& R
\eeq
The type and value may themselves be uniformly compound objects.
\end{definition}
Each element in an array thus promises:
\begin{itemize}
\item Membership to the compound ordered structure.
\item A value promise, indicating its primary function.
\item An index number or name promise (elements may be associative,
  i.e. nominal rather than ordinal in general), indicating its
  intended position within the collective.
\end{itemize}
\begin{example}[Data array]

An array $X$ may contain internal members of any type, including composite objects, as long as they all have the same schema. Suppose the type $\tau$ is a doublet object (integer, real)
\begin{verbatim}
X[0] = (integer, 3) (real, 2.72)
X[1] = (integer, 4) (real, 3.14)
X[2] = (integer, 5) (real, 1.0)
\end{verbatim}
\end{example}
The dimension of an object class is the number of scalar degrees of freedom,
promised by it.  The dimension of a representation is the
number of promised scalar values, or occupied slots.  In an
instantiated object, all degrees of freedom are filled, thus the
dimension of an instance could be considered zero, but it is more
useful to refer to the dimension of its schema.

\subsection{The role of agent phase in data representation}

If we follow the axiomatic
approach, and represent each independently variable property of data by a
separate agent, perhaps within the collected boundary of a superagent,
then the three levels of data aggregation are represented by the
ordered and disordered material phases of semantic space:
\begin{itemize}
\item {\bf Scalars} are atomic, ad hoc agents with only external promises.
\item {\bf Objects} are composite scalars, i.e. connected molecular superagents, with both internal and external promises, and inhomogeneous composition.
\item {\bf Arrays} are homogeneous lattices of scalar promises, i.e. agents in crystalline lattice, with homogeneous long range order.
Arrays offer limited translational symmetry, with a homogeneous schema,
by chaining together agents into a regular ordered structure.
\end{itemize}

\begin{example}[Promise graph vs JSON]\label{ejson}
Consider a typed data stream (e.g. JSON encoded data). We can extract values
and model their agency in representing functional behaviour in a system.
\begin{verbatim}
{
"kind": "Service",

"metadata": {
            "name": "myagent"
            },

"spec":     {
            "ports": [ 8765, 9376 ],
            "selector": {
                        "agent": "example"
                        },
            "type": "LoadBalancer"
            }
}
\end{verbatim}
This stream is composed of a signal agent named `kind', and two signal superagents called `metadata' and `spec'. 
The signal agents represent concurrent promise bundles within their interiors.
\begin{figure}[ht]
\begin{center}
\includegraphics[width=8.5cm]{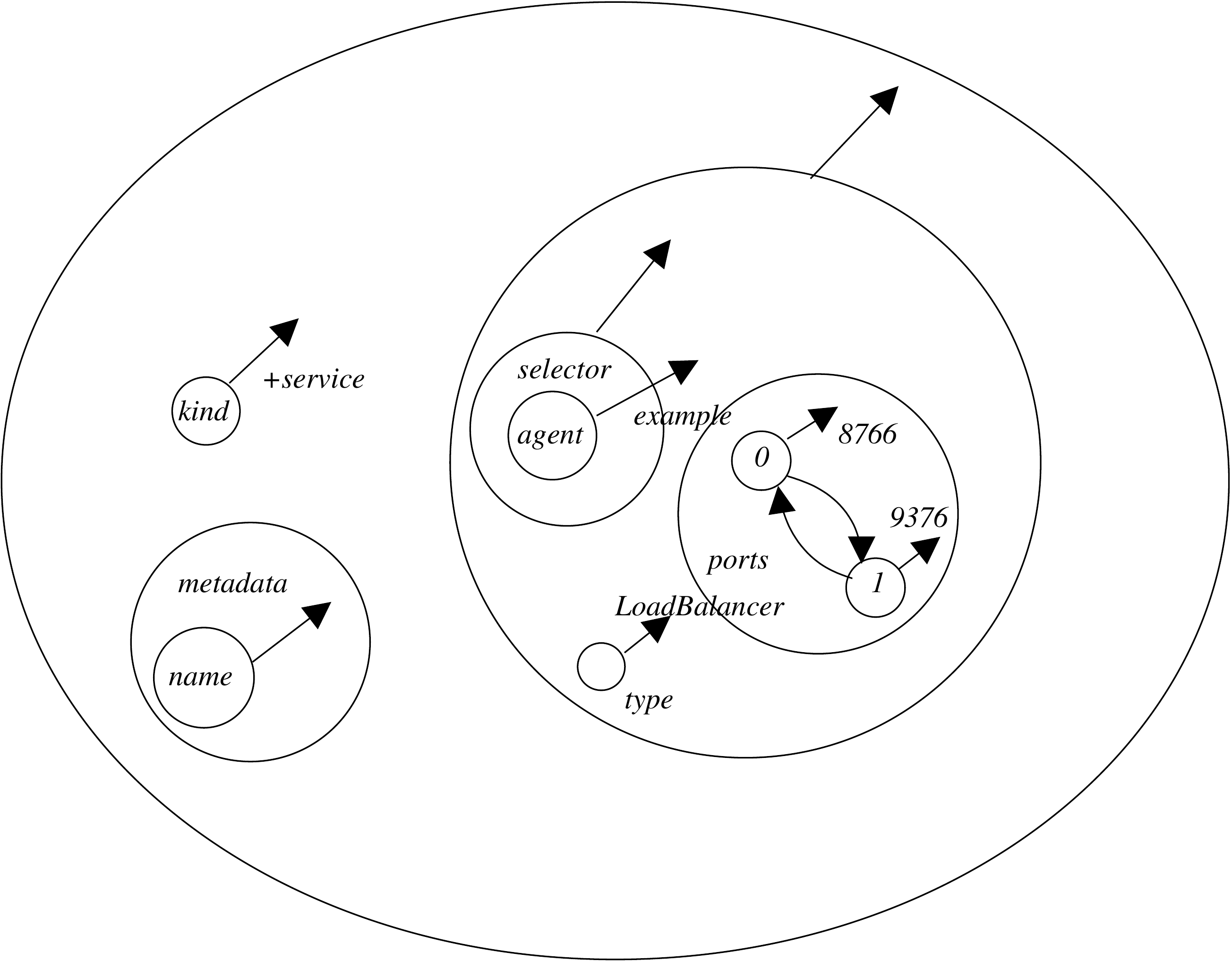}
\caption{\small Promise graph for the JSON data representation in the example \ref{ejson}. Scalar promises
are simple aliases for the agent names, while the aggregate superagent ports vector requires
an additional solid-state vector ordering agreement between the agents.\label{dataagent}}
\end{center}
\end{figure}

\end{example}

\subsection{Data encoding relative to fixed markers (protocols)}

Data may be transmitted as an array of raw values, or pre-annotated
with labels and significant symbols, in the form of a protocol. Smart
sensory apparatus may perform preprocessing that annotates data values
in a predictable protocol, while raw sensors pass only simple signal
sequences.  Decoding the significance of data adds to its semantic
value and assists in building knowledge representations, by reducing
the amount of recognition that is needed relative to the behaviour of
the sensors.  How we utilize markers (signposts) to ground and punctuate patterns
is therefore of the utmost importance to understanding and deriving
value from the structure of data. This encoding and decoding is the
realm of formal grammars, and grammatical protocol is the basis for
language and communication. Markers and delimiters break the
equivalence of array positions for raw data, establishing anchor
points for aligning expectations between a sender and a receiver, i.e.
a promiser and a promisee.

\begin{definition}[Encoding]
  The establishment of boundary conditions in a semantic space that
  break free symmetries for the purpose of anchoring an interpretation
of a collection of data promises.
\end{definition}
In promise theory, scalar promises at fixed agent locations
represent boundaries and boundary values (see paper I).  Data encoding
involves an alphabet $\Sigma$. Some symbols are reserved with special
meanings, e.g.
\begin{itemize}
\item Parentheses signify containment. Spatially, this plays the role of an embedded dimension.
\item Arrays signify collections of agency that are equivalent under the membership, i.e.
they have the same semantics.
\end{itemize}

\begin{definition}[Data stream or sequence]
A data stream is a totally ordered set of signal agents emitted from a source
agent, and absorbed by a receiver agent.
\end{definition}
A sequence transmitted from one agent to another involves a number of
promises. The transmitted signal is composed of signal agents that may
or may not promise a total ordering. These promises are made both to
the sender and the receiver, else the sequence order cannot be
transmitted faithfully. All symbolic language (including phonetic
language) depends on this construction.
\begin{figure}[ht]
\begin{center}
\includegraphics[width=7cm]{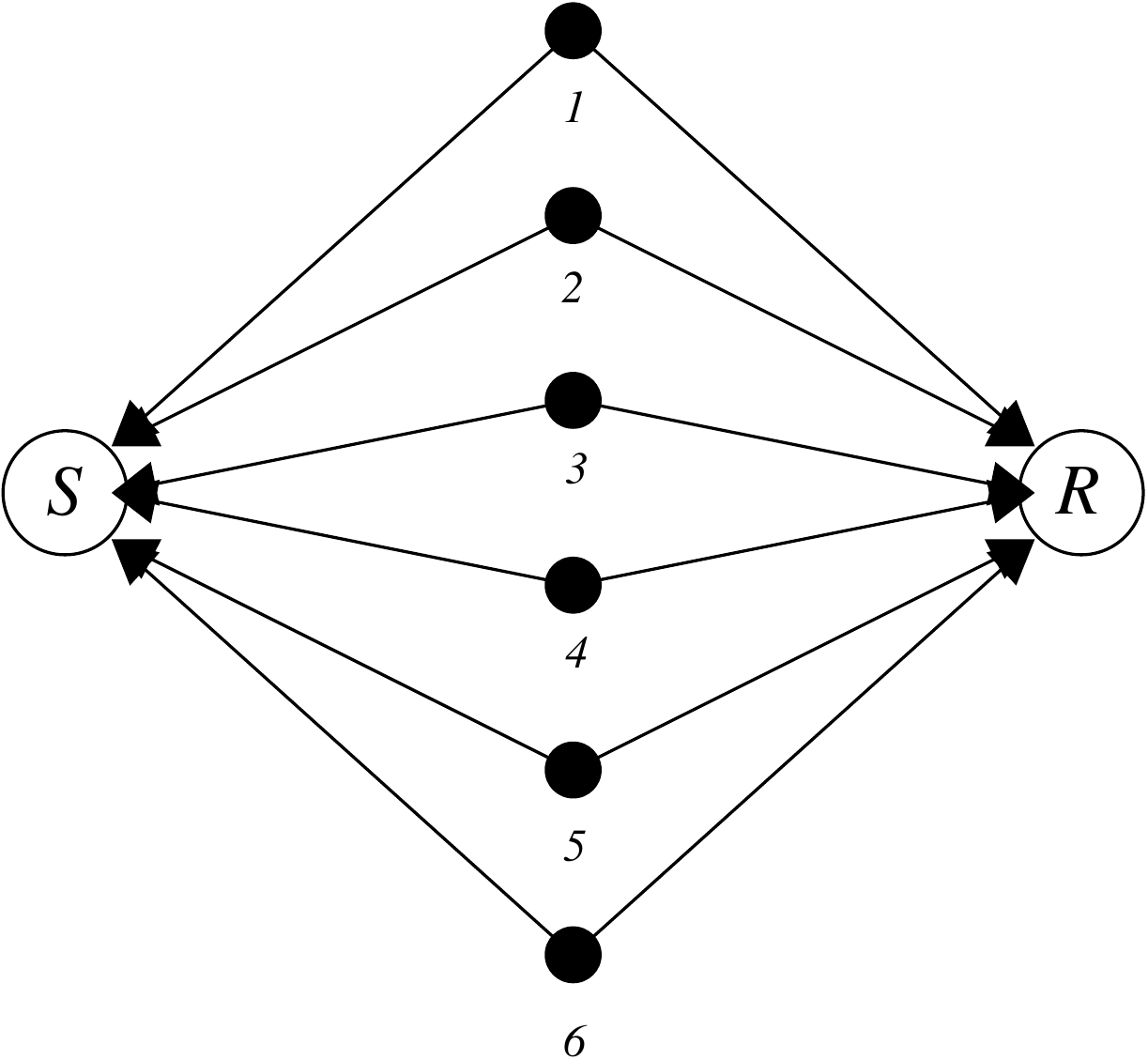}
\caption{\small A sequence is a spacetime coordinated promises.
this is identical to figure \ref{seq2} except that
the source of the data encoding is absent. \label{sequence}}
\end{center}
\end{figure}
\begin{example}[Parsing]
In the traversal or parsing of streams, an agent\footnote{Such an
  agent is called a parser or lexer.} assesses a stream of samples to
identify agents within the stream that have assumed meanings. These
are called tokens in parsing. It searches a partially ordered list of
patterns (normally sorted by largest pattern first) and guesses the
interpretation of these tokens.  Context free languages may have
specific token orders, but do not attach different semantics to
different spacetime locations.  Context specific languages (e.g.
markup languages) may attach different semantics to different
contexts, or embedded regions within a stream.
\end{example}
Encoding may be serial (timelike) or parallel (spacelike):

\begin{definition}[Serial, temporal encoding]
  In a temporal encoding there is an alphabet of symbols, some of
  which are special symbols to mark out space and time: parentheses
  for spatial digressions. Item separators, like whitespace and
  punctuation, indicate the boundaries of agency. This is a stream
  or a protocol.
\end{definition}.
\begin{example}[Non atomic symbols]
  Not all alphabets are atomic. In written Chinese, the symbols are
  formed from a deeper level of atomic motifs, combined into a
  collections of overlapping symbols. Words too are symbols that
  function collectively as concurrent data representations.
\end{example}

\begin{definition}[Parallel, spatial encoding]
  In a parallel encoding, spatial dimensions are encoded either by
  direct mapping in representation space, or by indexing with a tuple
  basis set.  This is a memory array.
\end{definition}.
\begin{example}[Parallel distributed data encoding]
  Agents may cooperate with each other to represent structures in a variety
  of ways (see figure \ref{seq2}). Parallel encoding of concurrent data, rather
  than sequential streams, are common in disk striping, RAID
  configurations for disks, etc.
\begin{figure}[ht]
\begin{center}
\includegraphics[width=3cm]{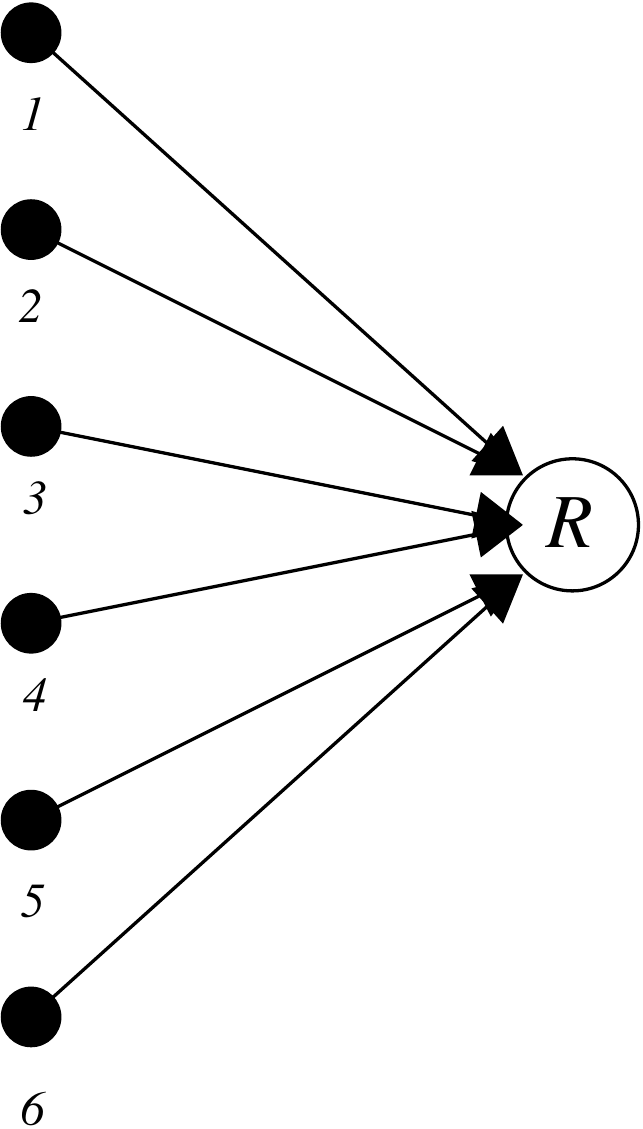}
\caption{\small A possibly ordered collection of agents may represent aggregate data in a distributed manner.
The interpretation of the collection is entirely up to the receiving agent $R$.\label{seq2}}
\end{center}
\end{figure}
\end{example}

\begin{definition}[Boundary markers and anchors (temporal punctuation)]
Special points along a stream trajectory may imbue locations with a
particular significance.  Such anchor points include boundaries,
signal delimiters, parentheses, initial and final points, etc.
\end{definition}
\begin{example}[Anchoring by agent type (spatial signposts)]
  Any exterior promise (name, charge, spin, service type, identity,
  etc) may be used to label an interstitial location within a
  homogeneous space.  This is an explicit symmetry breaking that
  injects information about location, by a reduction in non-local information.
\end{example}

\begin{example}[Recursion markers]
Parentheses may or may not be labelled to indicate different semantics.
Inside parentheses, special context dependent meanings may also apply, e.g.
separators:
\begin{verbatim}
( , , ) [ , , ] {  ; ; }  (* ; ;  *)  ($ ; ; $) 
\end{verbatim}
\end{example}

\subsection{Coordinates: spanning trees and tuples over data}

\begin{figure}[ht]
\begin{center}
\includegraphics[width=8.5cm]{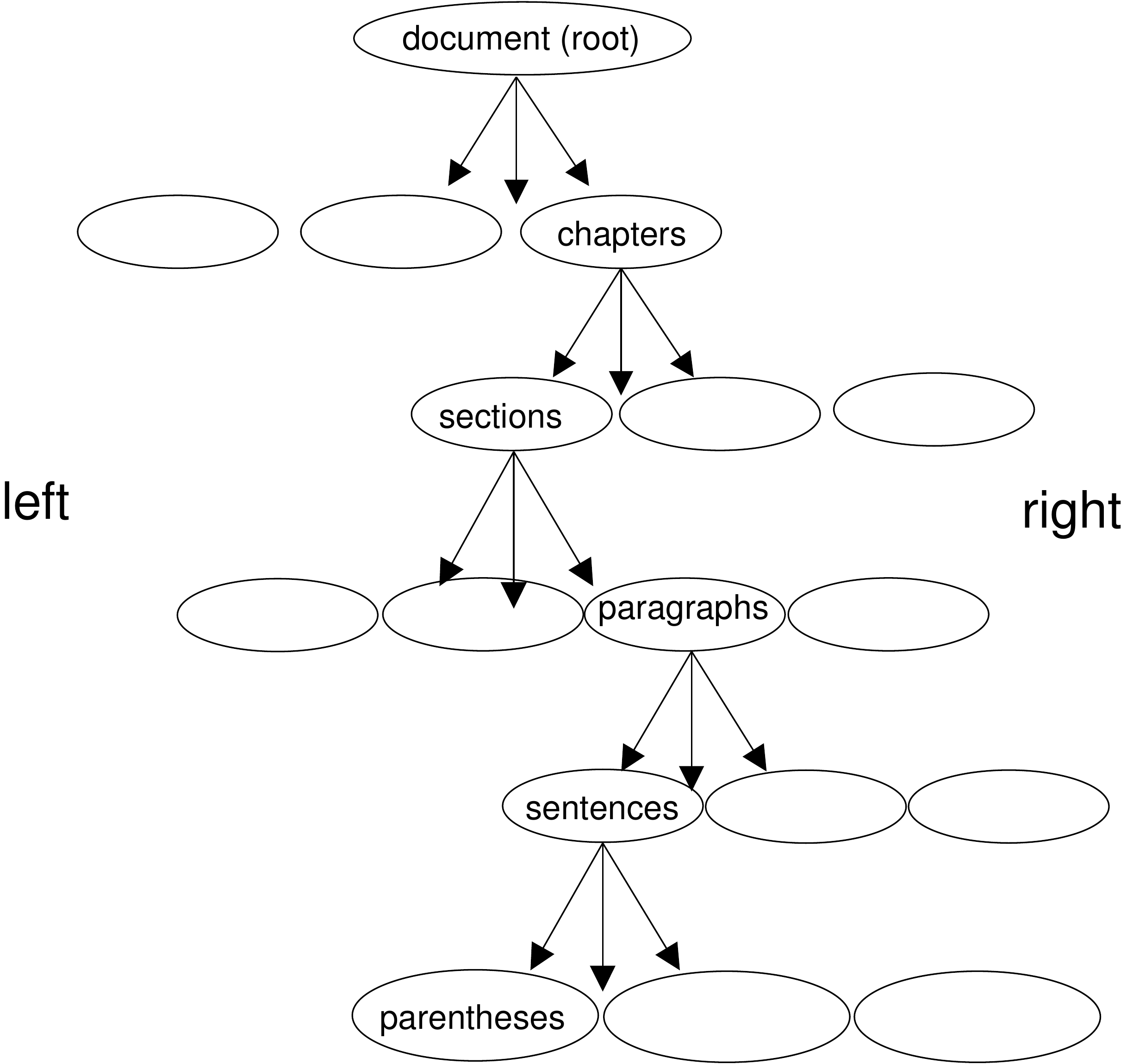}
\caption{\small Spanning tree and data hierarchy. Ordering may be preserved by
promising a policy of depth-first, left-first recursion.\label{subdivisions}}
\end{center}
\end{figure}

Data may be coordinatized in two principal ways: by spanning tree, as a graph using
nominal tuples, and as a dimensional lattice or grid, using tuples (nominal or ordinal). 
Nominal tuples might look like this:
\begin{verbatim}
(grandmother,mother,child) -> scalar value
(grandfather,mother,child) -> scalar value
\end{verbatim}
This may or may not assume ordering.
Ordinal tuples have to assume an ordering:
\begin{verbatim}
(0,0,0) -> scalar value
(1,0,1) -> scalar values
\end{verbatim}
An association with hierarchy is common, especially when the dimensionality is used to reduce
spaces into subdivisions (see figure \ref{subdivisions}).

Spanning trees may be recursive by reference only, e.g. in linked type structures,
objects can point to an object of the same type, i.e. with symmetrical semantics:
\begin{verbatim}
type data_chain
{
integer value;
data_chain  *pointer_to_next;
}
\end{verbatim}

\begin{example}[JSON and YAML data formats]
  The JSON data format is intended as a generic interchange format.
  JSON data structures are composed of three main kinds of value:
\begin{itemize}
\item Named values (scalars), represented as assignments \verb+lval : rval+.
\item Named lists of values (arrays), delimited by \verb+[+ and \verb+]+, with comma separator.
\item Lists of named values and named lists of values (objects or structs), delimited by \verb+{+ and \verb+}+.
\end{itemize}
Order is mainly important inside arrays, where order is the only semantic distinction as the
elements are homogeneous.
These correspond to the agent structures above, identifying data as agents that promise both
identities (names) and values., e.g.
\begin{verbatim}
{
"kind": "Service",

"metadata": {
            "name": "myagent"
            },

"spec":     {
            "ports": [ 8765, 9376 ],
            "selector": {
                        "agent": "example"
                        },
            "type": "LoadBalancer"
            }
}
\end{verbatim}
This structure can be coordinatized in a number of ways: e.g.
\begin{verbatim}
/kind -> Service
/metadata/name -> myagent 
/spec/ports[0] -> 8765
/spec/ports[1] -> 9376
/spec/selector/agent -> example
/spec/type -> LoadBalancer
\end{verbatim}
In this format, we understand the relational semantics as a spanning
tree of absolute paths from a root node, and the order is preserved as
a `first in first out' queue structure.  

A less opinionated coordinatization may be assigned by ordinals. Here we
cannot easily see the intended naming of regions:
\small
\begin{verbatim}
( 1,  1{,  0,  0, )[0] -> "kind")
( 1,  1{,  0,  0, )[1] -> :
( 1,  1{,  0,  0, )[2] -> "Service"
( 1,  1{,  0,  0, )[3] -> "metadata"
( 1,  1{,  0,  0, )[4] -> :

( 1,  2{,  1{,  0, )[0] -> "name"
( 1,  2{,  1{,  0, )[1] -> :
( 1,  2{,  1{,  0, )[2] -> "myagent"

( 1,  3{,  0,  0, )[0] -> "spec"
( 1,  3{,  0,  0, )[1] -> :

( 1,  4{,  1{,  0, )[0] -> "ports"
( 1,  4{,  1{,  0, )[1] -> :

( 1,  4{,  2{,  1[, )[0] -> 8765
( 1,  4{,  2{,  1[, )[1] -> 9376

( 1,  4{,  3{,  0, )[0] -> "selector"
( 1,  4{,  3{,  0, )[1] -> :

( 1,  4{,  4{,  1{, )[0] -> "agent"
( 1,  4{,  4{,  1{, )[1] -> :
( 1,  4{,  4{,  1{, )[2] -> "example"

( 1,  4{,  5{,  0, )[0] -> "type"
( 1,  4{,  5{,  0, )[1] -> :
( 1,  4{,  5{,  0, )[2] -> "LoadBalancer"

( 1,  5{,  0,  0, )  -> 
( 2,  0,  0,  0, )  -> 

\end{verbatim}
\normalsize
See appendix \ref{appendix}.
In YAML format, different
delimiters are used:
\begin{verbatim}
kind: Service
metadata:
  name: myagent
spec:
  ports:
    - 8765
    - 9376
  selector:
    agent: example
  type: LoadBalancer
\end{verbatim}
YAML emphasizes temporal structure and relative state, Now the
structure can only be understood in relative terms, as parentheses are
based on indentation level, i.e. ordinal coordinates within a two
dimensional embedding of the data within a page. Section markers are denoted by `- ' and
association is denoted by `: '.  These assume a state machine rather
than a stack for parentheses. A JSON or YAML structure is a
Directed Acyclic Graph (DAG).  The coordinatization of a JSON
structure is by name and container path (a spanning tree).
\end{example}

\begin{example}[CSV data formats]
  Comma Separated Values (CSV) are an array format. The names of data
  are represented by their position in a regular spacial array. Names
  are not provided as table headers, but are assumed to have been
  agreed by the implicit schema.  This is a tuple coordinatization.
\end{example}

\subsection{Streams, state, and memory}

In any system that observes input through some sensory stream,
assessments are converted into state, and memory is involved in
storing that state.  The timescales of real world events and
occurrences span all possible scales.  Events might be ephemeral,
relative to an observer, in which case they could only contribute to
the observer's context, or they might persist for repeated sampling as
statistical ensemble averages.

\subsubsection{Assessment and timescales}

\begin{figure}[ht]
\begin{center}
\includegraphics[width=10.5cm]{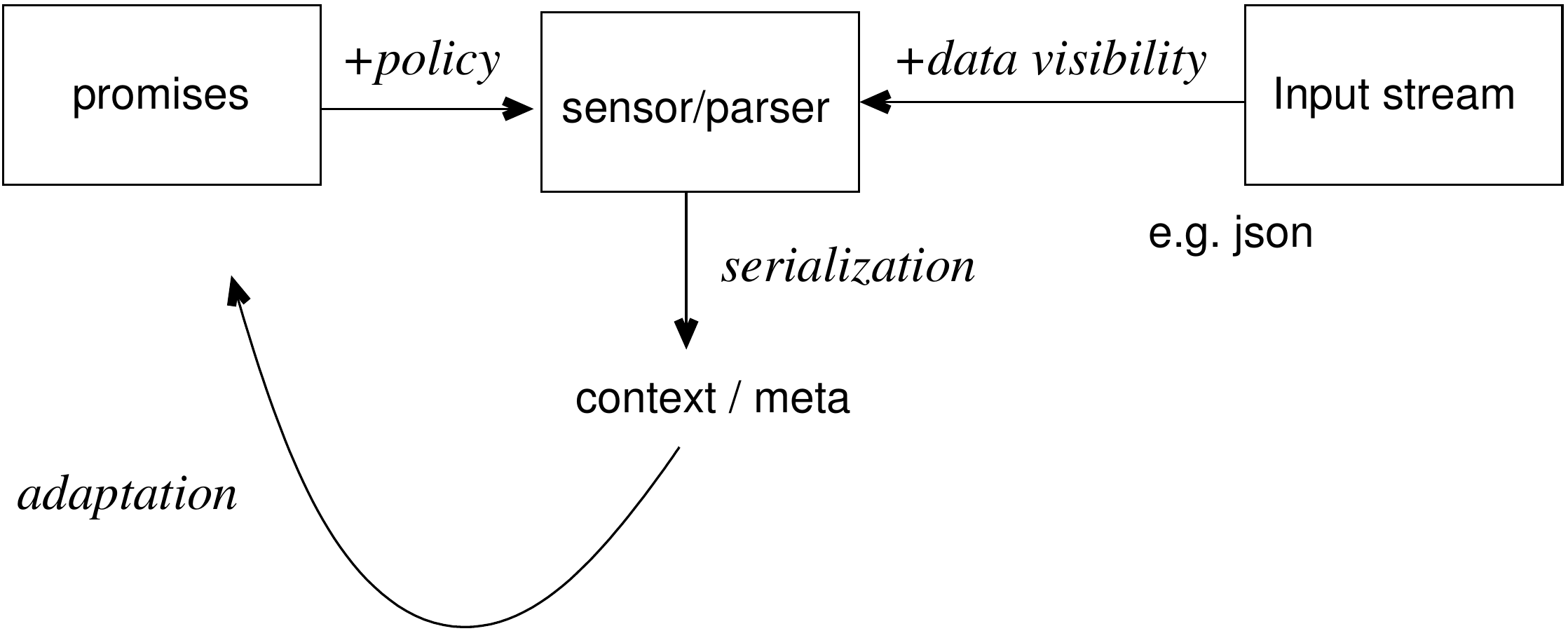}
\caption{\small Input stream is combined with a context according to promised policy, which is based
on what has previously been learned. The initial promise to learn and adapt
are the way we bootstrap a system without pre-supervised training.\label{sensorloop}}
\end{center}
\end{figure}

There are three stages to assessing data, which are explored as a simple
model for sensory knowledge in the
remainder of these notes:
\begin{enumerate}
  \item Data sampling, with timescale $T_\text{sample}$.
  \item Symbolization/tokenization/digitization, with timescale $T_\text{token}$.
  \item Contextualization (combination of multiple sources), with timescale $T_\text{context}$.
\end{enumerate}
Contextualization uses a short term memory to accumulate state, and
keeps it as an short term assessment of the state of the world. It is
the `awareness' of the agent's situation.
\begin{example}[Short term memory, modulation, and timescales]
  In later sections, we shall see how the short term memory modulates
  the long term memory, allowing large amounts of stable information
  (slow) to be switched on or off, by small amounts of context
  information (fast) (see figure \ref{sensorloop}).  The modulation of
  long term learning by situation awareness is a form of conditional
  switching.
\end{example}
Context contributes to defining the boundaries of applicability for
memories, by allowing more permanent memories to be switched on and
off in order to adapt to situations.

\begin{example}[Feeding back learned patterns]
  The working week is a pattern that humans can learn, to bring
  clarity to many processes that we sample and observe. However, this
  model may not be applicable to a train journey, or a light switch,
  or a satellite's environment. The major patterns of influence,
  experienced by an agent, need to be learned over time, and
  calibrated for each kind of environment individually.  Feeding back
  these patterns to condition the way assessments are made leads to a
  form of anomaly detection, or immediate abnormality
  alert\cite{burgessDSOM2002}.
\end{example}
Patterns are crucial to the identification and learning of semantics.
If a stable pattern is already known, it can be assumed, and learnt
permanently by hardcoding to save the cost of repeated recognition.
This is what happens in the separation between DNA and protein
networks, which are stable over very different timescales.

\subsubsection{Mapping input state to concurrent context}

A sensor is an input channel that is monitored by an agent.
Sensors are designed to select specific signals from what is promised by the
exterior environmental source.
\begin{definition}[Sensor]
A promise by an agent $R$ to accept specific data $d$ through a filter $F$, from a source $S$:
\beq
S &\promise{+d}& R\\
R &\promise{-F|d}& S,
\eeq
combined with the promise to relate an instantaneous assessment
to a subsequent agent $A$:
\beq
R \promise{\alpha\left(R \promise{-d} S\right)} A.
\eeq
Assessments form a data stream, which obey the Markov property\cite{grimmett1} in the absence
of memory, i.e. in the absence of memory,
every sample is assessed independently of what came before.
\end{definition}
Sensors contribute to the state of the agent $A$, i.e. to the
 snapshot of an agent's promisable characteristics. Memory may be added to this later to enhance discrimination.
\begin{definition}[State]
A collection of concurrent assessments over an agent's acceptance (-) promises.
\end{definition}
The ability to assess concurrently assumes memory sufficient to promise the
persistence of the assessments over the interval, and perhaps beyond.
The process by which assessments are made is familiar from information theory.
\begin{definition}[Sampling]
The projection of a sensor's assessment into the alphabet of states of the receiver system.
\end{definition}
As an autonomous agent, a sensor may filter input as it likes.
\begin{lemma}[Sensor resolution]
  The information resolution of the sensor is less than or equal to the resolution
  of the sample.
\end{lemma}
This is assumed obvious, and is stated without proof.
\begin{definition}[Parsing]
The repeated sampling of a sequential stream to decode
its concurrent symbols.
\end{definition}
Closely related to state is the notion of context:
\begin{definition}[Context (version 1)]
  A cumulative assessment of the state of 
  promises, given and received by an agent, over a concurrent interval of time.
\end{definition}
An equivalent restatement of this definition is used in section \ref{zzz}.

Context is a short term model of a system; it is
a self-assessment of an agent's place within its surroundings.
Because it needs to change quickly, in step with its environment, 
its complexity has to be limited relative to what could potentially be
stored in a long term knowledge model.  
The status of context is thus a bit like `hearsay', as contrasted with
confirmed evidential knowledge.
\begin{definition}[Situation awareness]
The assessment of {\em context}, from a set of interior and exterior
sensory inputs, relative to an agent, in comparison to an on-going longer term path of intent.
\end{definition}
To maintain context in an operational system, there has to be a
process within an agent that can work fast enough to make assessments.
This defines a timescale, which we call $T_\text{sample}$.  The
process of digital sampling (symbolization) and contextualization must
be of a similar order of magnitude in their timescales, and could well
be closely related. We may assume that these timescales are of the
same order of magnitude, since sampling and symbolization are
interdependent in any representation of assessment\footnote{Even if we
  imagine that the resolution of a sensor is real valued, with
  infinite resolution, it must be represented by internal states that
  can be read in a finite amount of time, and hence must be finite.
  Thus all observations have finite resolution.}.

\begin{itemize}
\item The process of parsing structural information has state itself (Chomsky hierarchy). This
can be referenced in the interpretation of the data.

\item The particular path taken in reading sensory input, i.e. a story or temporal structure
can form a basis for an evolving context.

\item The state may be persistent or ephemeral.
\end{itemize}

\subsubsection{Transmuting temporal association into spatial representation}

A system's spacetime state is its assessment of the present.  It
provides a notion of `here and now', which, in turn, provides a
primary original foundation for associative linkage. Each frame of
`here and now' is like a film reel that overlays new experiences and
previously experienced concepts.  It can be characterized only by the
combination of exterior sampled states with what has previously been
learned and replayed from interior knowledge representations. Interior
representations can, moreover, only be based on interior spatial
degrees of freedom, i.e.  what memory agents can promise.  Labelled
states could be assigned semantics directly, by linking context tokens
immediately into concepts, and they may take on semantics emergently
over many iterations.  Both processes must be possible for non-trivial
knowledge representation.

\begin{example}[From sensor input to concept]
Suppose we have sensors that sample and assess
the exterior environment in an aircraft.  How could we reach assessments
that have the semantics of high level concepts, e.g.
\begin{center}
\begin{tabular}{c|c}
Agent & Context State\\
\hline
 Plane & flying, crashing, standing still in a hangar.\\
 Wheel & revolving, stationary, braking.
\end{tabular}
\end{center}
from the sensory measurements of speed, temperature, acceleration, etc.
\end{example}

\begin{figure}[ht]
\begin{center}
\includegraphics[width=9cm]{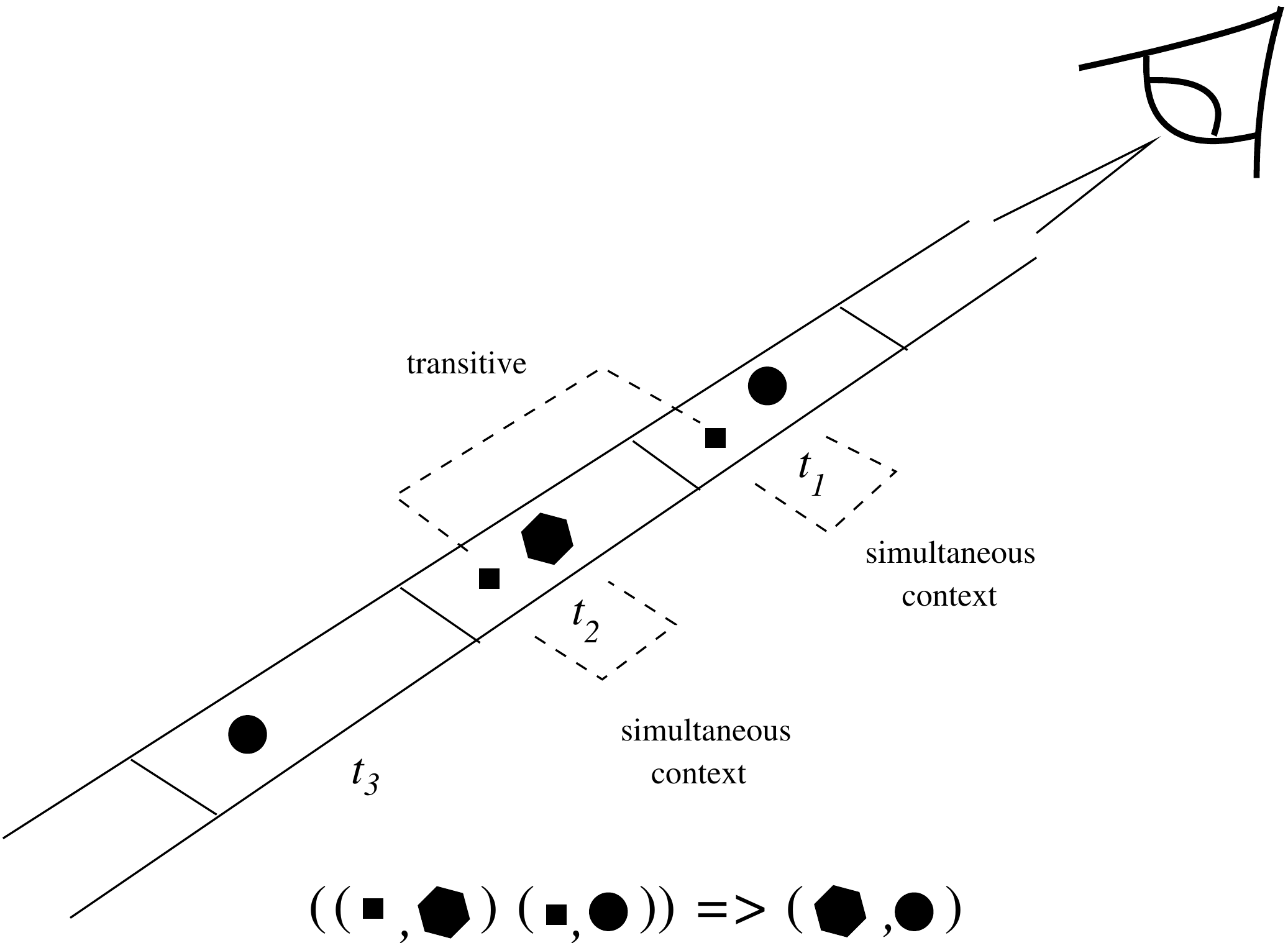}
\caption{\small As concurrent units of data are received by an
  observer, associations happen in each interval by the co-stimulation or co-activation of conceptual agents.
  This becomes transitive, by mutual association with intermediary. at a timescale
  greater than the simultaneous/coincident signal
  agents.\label{coactivation}}
\end{center}
\end{figure}

A stream of assessments, by an observer agent, leads to a spacetime
structure in which assessments become connected.  Shared semantics in
a cluster could be used to define concepts from these.
\begin{definition}[Temporally associated concepts (simultaneous activation)]
 Concepts may be called temporally associated if they occur
  simultaneously within a concurrent interval, such as a context interval, or within the body
of a single promise.
\end{definition}
The perceived simultaneity of context is where the association of long
term concepts must begin (see figure \ref{coactivation}). This is a
strategy used not only in cognition, but also in sensory systems like
the immune response where lymphocytes require co-stimulation from antigen
presenting cells to provide contextual
confirmation of pathogen recognition. Initially, during the
bootstrapping of a knowledge system, there is no longer term memory of
concepts or their associations to rely on for self-calibration of
context, but sensory context is continuously available.

\begin{figure}[ht]
\begin{center}
\includegraphics[width=10.5cm]{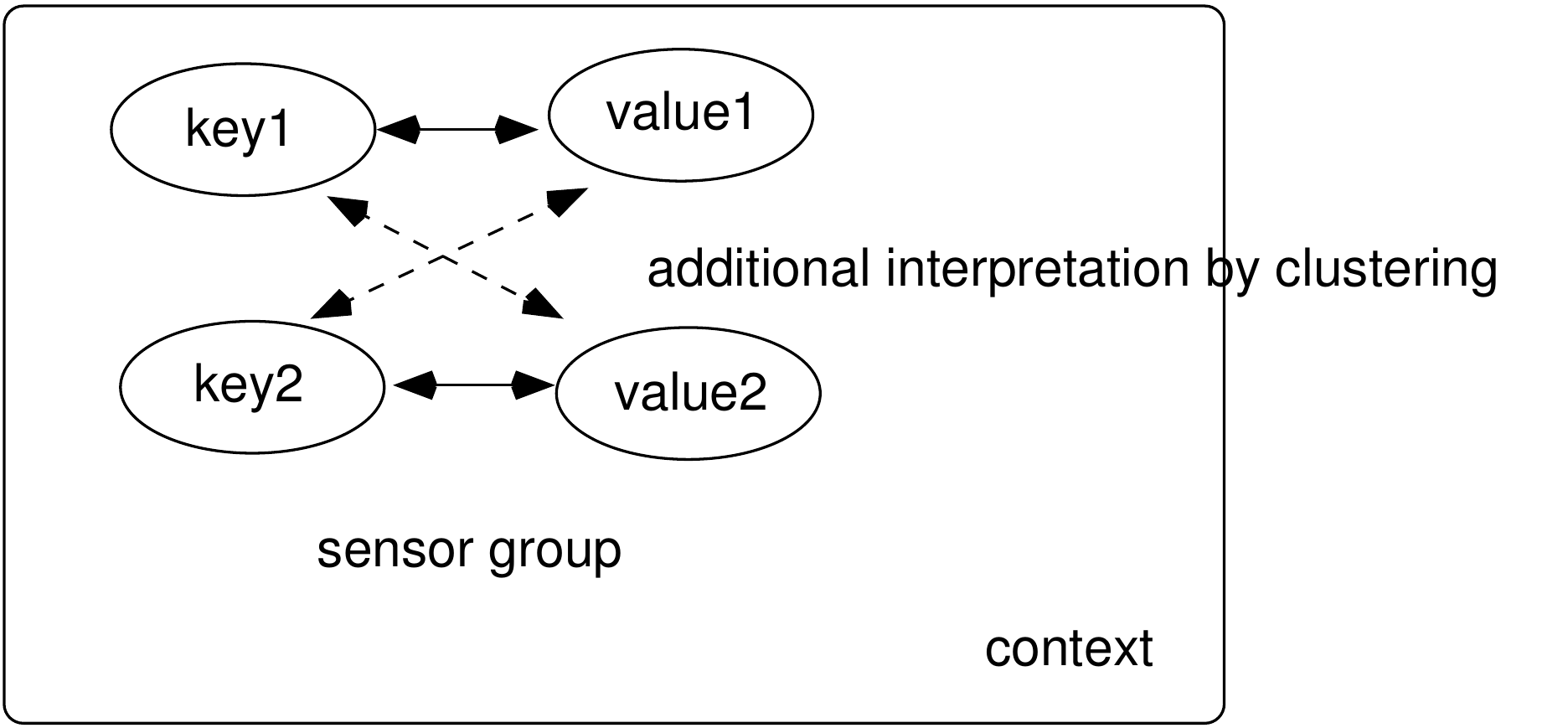}
\caption{\small An observer can annotate what is observed,
in context, by adding associations between tokens (names and
values) within a context. Depending on the memory of the observer,
the annotation may include short relative times, or be limited
to simultaneity: e.g. key 1 preceded key 2, or when key1 was
used value 2 was also in play, where key1 and key2 represent different but related aspects
of a common sensor interpretation, in (partial) pattern.\label{context}}
\end{center}
\end{figure}

Context is cumulative, over successive concurrent intervals, hence it is a
spatiotemporal state. It can thus represent regions and paths.  A
context may represent spacetime locale. It might be what just
happened, or where we are currently located.

\begin{example}[Parsing a sentence (linguistics as cognition)]\label{parsing}
The parsing of a sentence involves short term state memory (identified by 
the classifications of Chomsky theorem\cite{chomsky}, and stochastic generalizations). 
An agent parsing a stream must be able to hold all the input tokens that belong within a
concurrent interval, and are assigned semantics collectively, in order to represent
the grammar faithfully. This suggests that the transmutation, from individually sensed
input words to semantics, divides into three stages: sensor recognition (words), short term context (clauses, sentences, etc), and long term memory (meaning):
\beq
\underbrace{\text{Word}}_{sensor} \promise{\text{input}} \underbrace{\text{State} \rightarrow \text{Context}}_{\text{awareness}}
\stackrel{\revpromise{\text{feedback}}}{\promise{\text{feed forward}}} \underbrace{\text{Concepts} \rightarrow \text{Associations}}_{\text{meaning}}.
\eeq
As words accumulate in context, the scaled `simultaneity' of the complete sentence fragment
triggers associations and concepts in long term memory. 
The long term memory may conceivably feed back into short term context to further 
discriminate based on learning.
  From this perspective, language representations are
  indistinguishable from sensor arrays, in which words represent
  different classes of virtual sensor stimuli.  Next consider the
  sentence: ``I'm walking the dog.''. Taken at units, each word has
  meaning, and the sentence also has meaning. Additionally, in the
  city of Newcastle Upon Tyne, in the UK, the phrase `walking the dog'
  is a code `euphemism' for `having a beer at the pub'; thus, the
  sentence fragment has meaning at a larger scale. Meaning may be
  perceived at any scale, and patterns may trigger quite different
  associations in different contexts.
\end{example}

\begin{example}[Parsing a document]
  The parsing of a document is a scaled version of the parsing of a
  sentence.  In order to maintain longer term context, there needs to
  be continuity through short term memory. Thus, if an agent puts down
  a book and picks it up later, it needs to reconstruct some context
  from long term memory before continuing.  The process of parsing a
  linear document, or other sequence of events, leads to an evolving
  historical timeline which is cached as context.  Context must expire
  and be replaced by a new context at approximately the same timescale
  as sensing takes place.  Sometimes new context might replace, other
  times it might accumulate.
  The running context is built on spatial structure (titles and
  section containers), as well as associations and concepts triggered
  from memory.

  The low level semantic units are parsable (i.e. observable) senses,
  based on words or sounds, depending on what sensory input channel is
  active. Then clusters of words attain independent meaning. Thus,
  spacetime considerations alone suggest that, externally, linguistic
  words and phrases are just sensory fragments (as imagined in generative
linguistics\cite{generativesemantics}), while internally they
  are transduced into concepts (as imagined in cognitive linguistics\cite{langacker1}).
\end{example}

The examples above illustrate how context can be both a container for
perceived short term correlation or a dependency for switching on and
off long-term promises. Thus context is about scaling of agent roles,
by recursion and by association.

\subsection{Data classification and summarization}\label{classify}

Looping back memory along side a fresh input stream, iteratively, in
order to calibrate samples against expectations, is how a knowledge
equilibrium can be formed from assessments.  Repeated measurement is a
basic element of experimental scientific method\cite{certainty}, and
it is easy to see how this leads to a subjective notion of certainty
in the observer.  Convergence to an attractor is one way to achieve
stability; pre-existing memory acts as a seed for such an attractor.

The reliance on repetition is a robust strategy, but it is also data
intensive.  It would be impractical and inefficient to remember every
individual assessment (i.e. every data point acquired) forever, so one
needs to be able to summarize data into representative
characterizations (tokens) that are cheaper to store in
memory\footnote{There is some empirical evidence for single neurons in
  the brain being able to represent complex tokenized concepts, in
  so-called `invariant representations'\cite{invariantrep1}.}. Several
forms of summarization of data may be used to form the basis of
knowledge representations over different timescales:
\begin{itemize}
\item Context awareness (short term passive summarization): $T_\text{sample},T_\text{concurrent}$.
\item Learning (long term reinforced or stabilized passive summarization): $T_\text{know}\simeq T_\text{learn},T_\text{token},T_\text{pattern}$.
\item Adaptation (short or long term feedback based on active summarization): $T_\text{adaptation}, T_\text{attractor}$.
\end{itemize}
The principal distinctions in this list are about timescales; indeed, this
is the spacetime lesson that allows stability to converge. The
cyclic repetition of the learning process, over a noisy inputs, supplies a
constant stream of such perturbations, so persistence is a form of noise reduction
that allows stability of semantics. Stable semantics reduce the cost of
reevaluating meaning at every turn.

\subsubsection{Stable summarizations: attractors}

In order for random inputs to converge into clusters, at a stable
location, there must be a mapping whose
properties are both flexible enough to adapt to unknown inputs and
robust enough to recognize the same patterns repeatedly.  Small
variations in a pattern would preferably map to locations that were
associated with {\em attractors}, else every variation of input would
lead to an unstable and random new conceptualization. The implication
is that even associative memory would benefit from a metric quality
(see section \ref{shash}).

Iterative learning avoids data explosion by integrating 
samples into a form that is more stable
than raw data. Aggregate data measures, with a
metric nature offer such a {\em representative value} for a
collection of samples. In promise theoretic terms, a data aggregation is formally 
an external promise made by the
superagent over all semantically equivalent
samples\footnote{So-called `big data' (unstructured data) do 
not constitute knowledge, but patterns revealed within them might.}.  
Summarization of data leads to concepts like averages. Dynamical or quantitative 
averages are well known from statistics.
A semantic average, on the other hand is a
qualitative value that represents the common promises of a cluster of
agents.

\begin{definition}[Dynamical summary]
  Any single representative measure that assesses an ensemble of observations: $T_\text{know}\simeq T_\text{learn}$.
\end{definition}
\begin{definition}[Semantic summary]
  The formation of a semantic average over a bundle of related
  promises, from any number of agents, leading to i) a superagent, and
  ii) and exterior promise for the superagent ensemble: $T_\text{know}\simeq T_\text{learn}$.
\end{definition}
Both dynamical and semantic summaries are formed using the
available characteristics of data, described in section \ref{classify}.
For numerical data, the available
characteristics are the modalities and moments of data distributions, while for 
symbolic or qualitative interpretations, the averaging takes the
form of semantic generalizations and collective interpretations.
\begin{definition}[Metric (dynamical) average]
A standard ordinal scalar value that promises to
represents a body of dissimilar ordinal scalar values: $T_\text{know}\simeq T_\text{learn}$.
\end{definition}
\begin{definition}[Semantic average]
A single nominal interpretative assessment that promises to
represents a body of dissimilar promises or assessments: $T_\text{know}\simeq T_\text{learn}$.
\end{definition}
\begin{example}[Averages]
  Averages are aggregations yielding a representative result. Dynamic
  or metric averaging takes a number of quantitative scalar values and
  combines them to result in a value that promises a scalar result to
  represent the bulk data: $T_\text{know}\simeq T_\text{learn}$.
\end{example}
\begin{figure}[ht]
\begin{center}
\includegraphics[width=5.5cm]{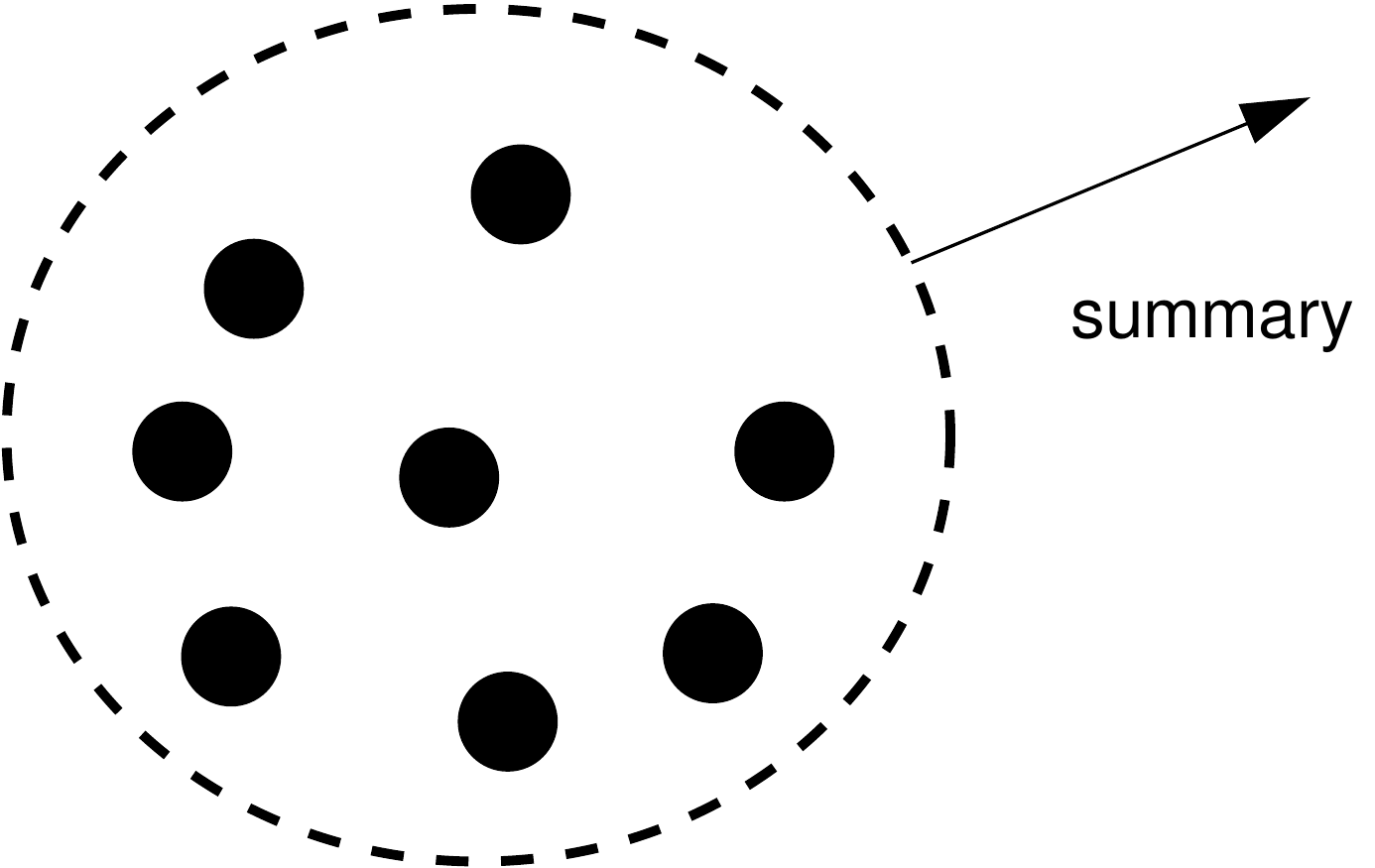}
\caption{\small Averages and representative values are exterior
  promises belonging to a superagent. Averaging, or summarization,
is thus a form of spacetime coarse graining, i.e. a scale transformation.\label{avassess}}
\end{center}
\end{figure}
\begin{example}[Semantic averages as scaling generalizations]
Semantic averages take a number of dissimilar promises and represent
them by a common standard, e.g. extracting what is meant by `a vehicle'
from `a car', `a bus', `a train', etc.
\end{example}
A trivial lemma may be noted, relating to these:
\begin{lemma}[Averages express exterior promises for an ensemble superagent]
  Any collection of agents making promises or assessments may
  collectively promise a scaled average to represent the samples; such
  a promise is exterior to the superagent formed by the ensemble (see figure \ref{avassess}).
\end{lemma}
The proof is simple (see paper II):
let each assessment by observer agent $A$ be represented by a signal
agent $a_i$. Let each agent $a_i$ promise to be members of
an ensemble (sample set)so that they form a superagent. Now,
at super-signal-agent scale there must be new external promises. 
An average is a external promise of the ensemble superagent.

\subsubsection{Repeated sampling, significance of data and attached meaning}\label{significance}

We attribute meaning to distinguishable discrete measures, i.e.
symbols, that stand out against background noise.  How we transform signal
information into meaning is thus related to a signal's entropy.
\begin{example}[Significance and entropy]
Culturally significant markers, signposts, and directional beacons, 
are towers, statues, monuments, obelisks in a landscape that is relatively
flat. Low entropy, low information structures stand out symbolically,
and we attribute them with special semantics.
\end{example}

\begin{hype}[Significance is inverse information]
At some information scale $\sigma$, we can identify a significance, of the general form:
\beq
\Big| \text{\rm Significance}(\sigma) \Big| \propto \text{MaxEntropy} - \Big| \text{\rm Entropy of representation} \Big|.
\eeq
\end{hype}
Significance gives an anchor point to which we attach
meaning. For graphs, network centrality can be used as the source of the entropy
distribution\cite{archipelago}.

The more times data are collected in semantically equivalent contexts
the more robust an average summarization of the data can be to new
variation.  By repeating a measurement, we can make one interpretation
stand out against all the others. This brings not only increasing
certainty, but increasing significance.  This is the idea behind
histograms.  Regardless of the data classification alphabet used for the sampling
procedure, repeated sampling will lead to a histogram whose
scales of variation, measured subject to Nyquist's sampling limit, that represent the
true alphabet of variation in data.

\begin{assumption}[Knowledge is an equilibrium state]
  Learning happens by iteration, and knowledge is the summary of
  iterative learning, over a timescale $T_\text{learn}$. Without continually revisiting and re-testing
  observations, both the certainty and significance of the data lose
  meaning. Stable knowledge is therefore the result of an equilibrium
  cycle, fed by observation, pattern discrimination, and
  repeated resampling.
\end{assumption}
The remainder of this work rests on this assumption.

\subsection{Distributions, patterns, and their information content}

From the summarization scaling argument, we conclude that
knowledge begins effectively with stable histograms.  Memory of multiple
episodes builds up a summary over data episodes, from which we
derive information, and thence contextualized knowledge. Summary
representation takes the form of a frequency distribution, relative to
a set of sampling classes, which, in turn, may
be tokenized into an alphabet of effective recognizable tokens or symbols.
These may be found within the spatial characteristics of the frequency distributions:
\begin{itemize}
\item Modality (how many peaks does the distribution have?)
\item Moments of the distribution (is it localized, symmetrical, skew, long tailed, etc).
\item Smooth, discrete (bar chart histogram or smooth bell curve, etc).
\item Wavy, frequency amplitude (what is its Fourier decomposition?)
\item Continuous, broken, discontinuous (is it a differentiable total function, or a partial function?).
\end{itemize}
To transmute statistical weights into symbolic representations, 
one may re-digitize these different characterizations into manageable, discrete characteristics:
\begin{definition}[Tokenization]
The process of mapping a set of assessed patterns and characteristics $P$ into names $N$:
\beq
{\cal C}: P \rightarrow N
\eeq
over a timescale $T_\text{token}$.
\end{definition}
In so doing, we must be clear that the mapping of a symbolic context
$\cal C$, based on observed data, is a function of the promisable
capabilities of the agent's sensory apparatus. Thus, there is no way
to separate concept or interpretation from cognition.  The ability to
identify patterns is critical to forming certain kinds of associations
(see figure \ref{association1} (b),(c),(d)).

\subsubsection{Ontological structure}

In a world model, there may be both observed phenomena and imagined
phenomena. We have names of things, and relationships between them.
Both of these can be imagined or real.  Thus a knowledge model can
grow to be richer than the sensory model, as it feeds back on itself.

\begin{figure}[ht]
\begin{center}
\includegraphics[width=10.5cm]{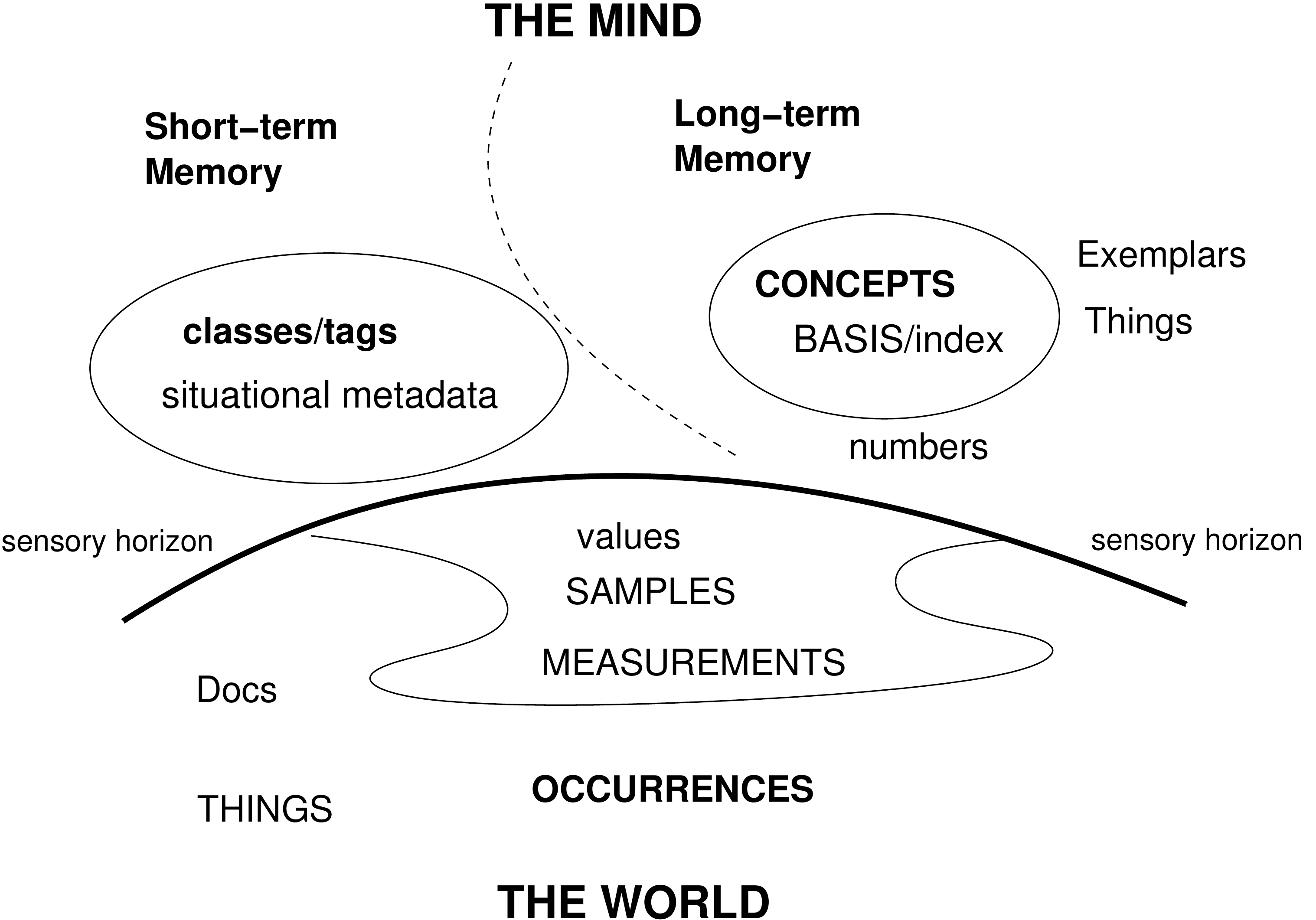}
\caption{\small Occurrences/exemplars live in their own world, with their own
adjacencies. Conceptual models form aggregate 
representations that are smaller than the sets of exemplars,
and act as indices (coordinate bases) for them, with aggregate meaning.\label{occurrences}}
\end{center}
\end{figure}
To learn an ontology, rather than assert one\cite{burgesskm}, we can
assume that samples promise consistent semantics up
front\cite{ontologies}.  It is then possible to learn structural
relationships, i.e. clustering by divining a definition of distance,
using some arbitrary dimensional spanning set, e.g. using clustering methods like Principal
Component Analysis (PCA)\cite{duda1}, assuming that the spanning set
is known.  There is a bootstrap issue, by which observational data
have to stabilize into `knowledge'.

\begin{hype}[Stability hypothesis]
Knowledge consists only of those phenomena that are dynamically
stable, hence from which we may assign consistent semantics.
\end{hype}

Before the widespread use of dynamical pattern recognition, it was a
long standing assumption that a model of knowledge had to begin with
an ontology, or a pre-imposed arrangement of `things', subjects,
issues, and relationships into a tree. That approach failed to explain
how understandings of relationships could come about by learning, or
be adapted to a changing world. Emergent models of an environment, on
the other hand, with identifiably separable clusters could be the
basis of a time-varying emergent ontology. The concept of strongly
connected clusters (SCC) from graph theory\cite{graphpaper} may serve
as an emergent definition of ontological structure; one could imagine
using different semantic filters to select directed subgraphs to form
overlapping ontological islands\footnote{Indeed, this observation was the
beginning of successful web search with PageRank\cite{pagerank}.}.

The prediction of promise theory would be that concepts form from
as a result of two processes, based on observer independence:
\begin{itemize}
\item Intrinsic properties posited by a source (+ promises).
\item Perceived and learned properties perceived by a receiver (- promises).
\end{itemize}
This basic promise theoretic theme follows from the $\pm$ 
propagation law, and runs throughout these notes, and recurs
on in many scenarios to explain subtleties of information propagation.

\subsubsection{Pattern acquisition: generalization and discrimination}

Concepts and their associations emerge through encodings that cluster related concurrent things.
This can be immediate, like a Markov process:
\begin{itemize}
\item Classification of concepts by senses alone.
\item Simultaneous activation of concepts across parallel inputs.
\end{itemize}
and it can be enhanced by long term memory, in a Non-Markov process:
\begin{itemize}
\item Approximately simultaneous activation (coarse persistence in concurrent
  chunks).
\item Feedback from previous learning into sensor stream.
\end{itemize}
We search for the most significant patterns in order to attribute
meaning to them, by:
\begin{itemize}
\item {\bf Identification by aggregation}: The precise accumulation of
  attributes by pattern activation. If every pattern can only be
  activated once without context dependency, this leads to a limited
  number of traits, which are stable by aggregation.

\item {\bf Identification by division}: A precise branching of attributes,
  leads to the discrimination of patterns into modules, which
  are fragile by branching. Within any
  branch, any sub-branchings can bring further discrimination.
  Branching is not an idempotent operation.  
\end{itemize}
The formation of a branching tree of concepts is called taxonomy or
ontology span.  Since decomposition is arbitrary (we can branch or group arbitrarily and as many times as we
like), the implication is that there cannot be a single taxonomy or
ontology based on a universal set of mutually exclusive properties.

\subsection{Summary}

A few main themes are behind the construction of knowledge from data:
\begin{enumerate}
\item The separation of scales, according to a principle of weak
  coupling.  Each scale has its own semantics and cohesion,
  structurally connected, but independently identifiable, with independent agency.

\item The identification of types, schemas, from the separation of
  spatial scales, i.e. the formation of regularized superagents, and the
  identification of context from the separation of time into
  concurrent chunks (simultaneous and sequential).

\item The breaking of `translational' symmetries is the key to
  encoding data reliably using anchors and signposts to align
  semantics relative to distinguishable markers (fixed stars to steer by).

\item Four mechanisms occur, by which intent is passed along from one agent to another:
\begin{itemize}
\item Nested containment (scaling).
\item Causal history (dependency).
\item Composition, or cooperative refinement.
\item Similarity (proximity).
\end{itemize}
\end{enumerate}
Data are encoded in the form of patterns, which are subjectively
identifiable as discrete symbols. The patterns are transmitted, or
written, in any semantic medium, by variations in the scalar promises
of spatial agents over spatial trajectories.  Symbols may be combined
into larger patterns, e.g.  words and sequences, by tracing out
spacetime trajectories in the representation medium. Data are parsed
in scaled, sequential streams of quasi-concurrent symbols. The order
of the stream has meaning, but may not be deterministically related to
the interpretation. To alter and extend the interpretation of
sequential orderings, one may define language semantics for the
encoded patterns, that are partially ordered at a larger scale
(superagency).  The representation of data as scalar promises is a
clue about how we may form memory out of any semantic space.


\section{Memory, addressing, and spacetime patterns}\label{learnassociation}

A central part of what we consider knowledge and `smartness' to be
about is the ability to learn and recall concepts and patterns. This
requires memory, which, in turn, is naturally built on spacetime concepts.
By choosing a language of semantic spaces for memory, we acknowledge
that almost any agent medium may be used to store information: whether it
be atoms, molecules, transistors, brains, or cities.

This section concerns how semantic spacetime can be used as a representation
of addressable memory, one of the key requirements for
learning and knowledge representation.

\begin{definition}[Memory agent]\label{memoryagent}
An agent $M$ may be said to promise the role of memory (like a service) if it
promises to forward data $d$ accepted from input source $S$, previously,
to another agent $R$, upon receipt of an appropriate request $r$:
\beq
S &\imposition{d}& M\\
M &\promise{-d}& S\\
R &\imposition{r}& M\\
M &\promise{+d|r}& R\\
R &\promise{-d}& M
\eeq
(see figure \ref{transwitch}).
\begin{figure}[ht]
\begin{center}
\includegraphics[width=7.5cm]{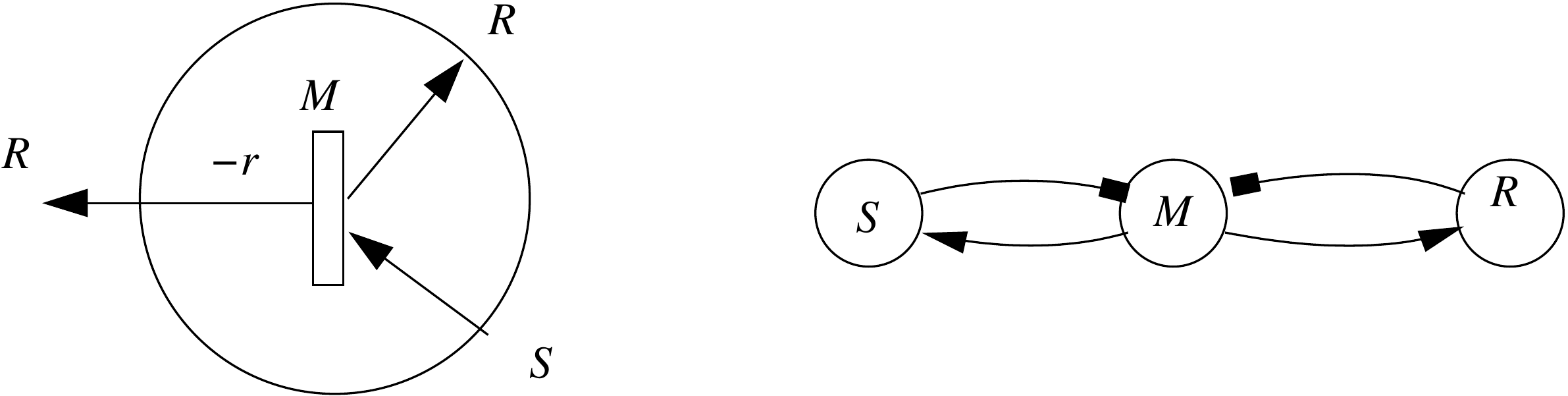}
\caption{\small Memory acts as a switch, with the promises here likened to a transistor. Data are pushed to memory, and accepted on trust. Requests for data are pushed to
memory and a conditional response is returned, as a service transaction.\label{transwitch}}
\end{center}
\end{figure}
The replacement of impositions with promises, in these interactions,
is also acceptable for voluntary cooperation lookup, such as in 
scheduled access. In that case,
the request $r\rightarrow\emptyset$ may be empty,
for spontaneous delivery, like a repeating beacon.
\end{definition}
A memory agent is thus a kind of delayed forwarding device
for information. This is explored in more detail below.

\subsection{Relationship between switching and memory}

In order to write to memory, by changing the patterns represented by a spacetime
configuration, agents must be capable of keeping conditional promises,
which are the representation of switching interactions\cite{promisebook}.

\subsubsection{Conditional promises}

Conditional promises are the basic atomic representation for a
variable memory representation, because properties are promised subject
to conditional biases, e.g.  an agent may promise to record sensor data subject to
prevailing conditions, or may promise to return a data value, subject to
providing the agent address.
\begin{definition}[Switch]
  An agent, which promises a distinct outcome, for each distinct
  assessment it makes of a promised input.
\beq
\text{\rm switch} \promise{\text{output(condition)}|\text{condition}} A
\eeq
\end{definition}
In other words, the output is a function of the condition. The
condition modulates the promised role.
\begin{lemma}[Dependencies are switches]
  Any functional dependency represents the modulation of a promise,
and hence acts as a memory element.
\end{lemma}
The proof follows from the structure of a dependency (see figure \ref{transwitch}):
\beq
S \promise{+b|\text{dependency}} R,
\eeq
where $b$ is effectively a trivial function of the dependency, as the promise
body is $+b$ if the dependency is given, and $\emptyset$ is not.
\begin{figure}[ht]
\begin{center}
\includegraphics[width=5.5cm]{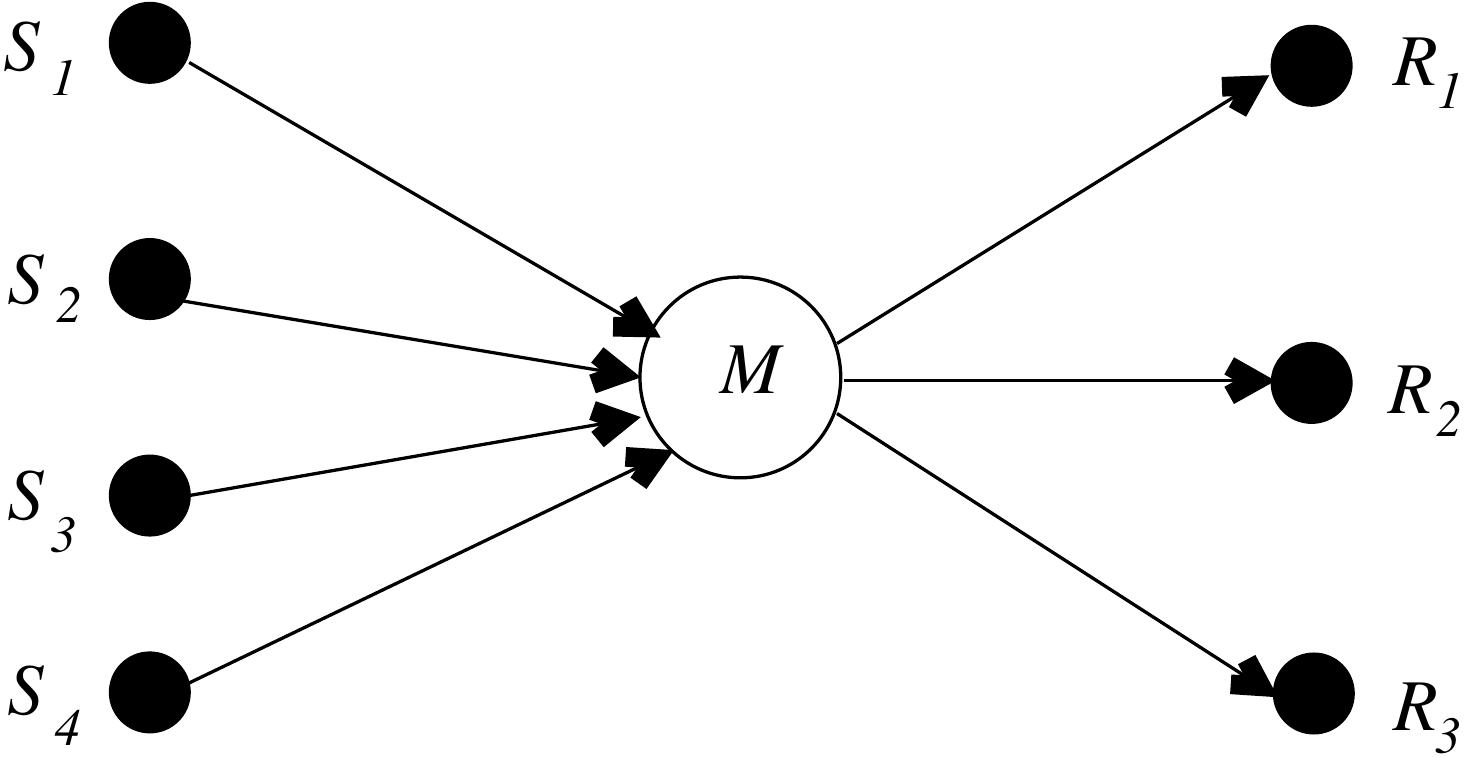}
\caption{\small The flow of data through discriminator switches: with
  this generic form, we have the basis for predictable addressing. The
  ability to uniquely address locations in space, and to promising
  routing of data to that single destination is what allows spacetime
  to encode data reliably, and for us humans to attach significance to
  the values.\label{switching}}
\end{center}
\end{figure}
Switches may be more complex than a single input, and may give up their stored
data by different channels (see figure \ref{switching}). More complex
switching structures can be used for forwarding or routing of information that
can perform sorting functionality, which is important for addressing specific elements.
\begin{definition}[Routing (multi)switch slave]
An agent with several imposed input channels $S_i$ and several output channels 
$R_j$ (see figure \ref{switching}). For example, a simple configuration might impose:
\beq
S_i &\imposition{d_i}& M_{ij}\\
M_{ij} &\promise{-d_i}& S_i\\
R_j &\imposition{r_j}& M_{ij}\\
M_{ij} &\promise{+d|r_j}& R_j
\eeq
The switching agent $M$ is now a matrix whose promise to forward depends on 
the request issued by a receiver. Again, the imposed request may be null to
represent a pull request for data by the receiver.
\end{definition}
By extending the conditionals to a matrix, a policy for forwarding can be
hardwired into the intermediate agent, as an explicitly programmable relay.
\begin{definition}[Forwarding or relay agent]
An agent $F$ which promises to accept data from an input channel $S_i$
and relay it to an output channel $R_j$ unaltered, subject to a promised condition
$r_{ij}$:
\beq
S_i &\imposition{d_i}& F\\
F &\promise{-d_i}& S_i\\
F &\promise{+d_i|r_{ij}}& R_j\\
\Unspec &\promise{+r_{ij}} & F
\eeq
The condition $r_{ij}$ is called the routing or forwarding table and may be promised
by any agent $\Unspec$, including $F$ itself.
\end{definition}
If we allow the intermediate agent to promise with full generality,
the implication of this matrix of promises is that the 
router acts as memory in the form of a routing table.
\begin{lemma}[A routing switch can act as memory agents]
$M_{ij}$ is equivalent to a collection of memory agents with $\dim j$ retrievable values.
\end{lemma}
The proof is trivial from definition \ref{memoryagent} by identifying
$M \rightarrow M_{ij}$, 
$S\rightarrow S_i$ and $R \rightarrow R_j$, with data $d\rightarrow
d_j$ for requests $r\rightarrow r_j$. Similarly:

\begin{lemma}[A routing switch can act as a data forwarder]
The multi-switch $M_{ij}$ contains a forwarding agent $F$.
\end{lemma}
The proof follows by identification $M_{ij} \rightarrow F$, $R_j \rightarrow \Unspec$.

\begin{example}[Network routers and forwarders]
It might seem that a routing agent does not store data for
very long, but this is only a design choice of most routing switches.
The definition above states that data values can be retrieved from
memory at some later time, which is a sufficient condition
to act as memory. It does not specify how many times values can be
retrieved, or even how quickly. Routing switches
could, in principle, keep all the data they receive permanently, as
long as they have sufficient internal capacity, and a sufficient
number of addressable channels, but that is not usually of interest.
Some firewalls retain more detailed information for forensic analysis.
\end{example}
The configurations and examples above illustrate how
memory retrieval by address is directly analogous to the routing of information by
request, and that the requests may be pre-determined in the form of
a memory policy. We can apply these structures to the problem of learning
systems, and knowledge representation, by thinking of memory addresses
or data requests as being conditional biases which come via a sensor
or input channel, arriving as a queue of changing impulses. What this
shows is that almost any spatial structure capable of being modulated
through its adjacencies to neighbours can store and retrieve state under
the right circumstances.

\begin{example}[Artificial Neural Networks]
  Artificial Neural Networks (ANN) are sequential fabrics made of
  non-linear aggregating switches, whose agent connections
  promise weighted threshold values to other agents. By training, i.e.
  adjusting, the weights of the whole layered fabric to match a
  desired state output, for given input, they approximate something
  like $N:1$ attractors. ANNs are effective on complex patterns, e.g.
  facial recognition. They are used for sensor pattern recognition,
in the language of this work.
\end{example}

\subsubsection{Timescales for deterministic switching}\label{tscond}

To be able to construct stable knowledge systems from switching agents, their promises
need to be kept over timescales that support predictability
even with non-deterministic inputs.
The timescales for change and operation of a conditional promise tell
us about the dynamic and semantic stability of switching.
We may define the stability of this promise. Let us denote the timescale
for change of promise components by $T(\cdot)$.
\begin{definition}[Semantic stability of a switch or conditional promise]
Consider a promise
\beq
S \promise{+b|c} R
\eeq
with timescale $T(\Delta b)$ meaning the timescale over which
one expects a change in the promise body, and $T(\Delta c)$
meaning the timescale over which one expects a change in the
dependency modulation. The stability of the promise means
\beq
T(\Delta b) \gg T(\Delta c),
\eeq
i.e. the body of the promise $b$ is constant over the switching on and off
by modulation. If $b,c$ are continuous functions, then we may express this
as
\beq
\frac{\partial b}{\partial t} \ll \frac{\partial c}{\partial t}.
\eeq
\end{definition}
If these stability criteria are not met, subsequent switching operations will lead to
inconsistent results, i.e. the promised behaviour will not be a total function of $c$.
\begin{definition}[Dynamic stability of a switch or conditional promise]
Consider a promise
\beq
S \promise{+b|c} R
\eeq
with timescale $T(b)$ meaning the timescale it takes to keep
a promise $b$ from any starting state, and $T(\Delta c)$ means the 
timescale over which one expects a change in the
dependency modulation.
\beq
T(b|c) \ll T(\Delta c),\label{xcc}
\eeq
i.e. the promise $b$ must be keepable quicker than the conditionals vary
by modulation.
\end{definition}
\begin{lemma}[Non-deterministic switching]
If the dynamical stability condition in (\ref{xcc}) is not met, there must be either
a queueing instability or an overlapping and inconsistent state in the switch.
\end{lemma}
The proof is straightforward. Let the arrival rate be $\lambda \simeq 1/T(c)$
and the service rate be $\mu \simeq 1/T(b|c)$. If $T(b|c) \ll T(\Delta c)$
then $\lambda \ll \mu$, and there is no queue, else there must be a finite queue
length. If the agent $S$ does not support queueing, then its state must be
a superposition of all the queue states, i.e. $c$ must be in an indeterminate state.

\begin{example}[Memory and learning network stability]
  Memory networks, such as computer memory, Bayesian or neural networks,
  etc., are dynamically and semantically stable if their training $b$
  does not change significantly for the duration of their operational usage $c$.
\end{example}

\subsection{Spatial representation of encoded memory}

Data are encoded and represented as strings of symbols. Symbols
are represented as discrete (but possibly overlapping) patterns in a
spatial medium, i.e. they are patterns in the properties of spacetime.
Symbols are thus promise theoretic agents (or superagents).
\begin{definition}[Alphabet]
  A collection of unique symbols, represented as discrete spacetime
  patterns over any configurable spacetime property.
\end{definition}
Alphabetic symbols form serial structures called strings. Some strings may promise to 
conform to languages\cite{lewis1} (see also the discussion in paper II). 
\begin{definition}[Language]
Aggregate superagents of symbols that promise constraints $L$ on their interior 
composition, classified by Chomsky rules and their stochastic generalizations, etc.
\end{definition}
Natural languages comprise strings of words and phrases (such as in this document or in
spoken words). Formal languages, used in
control systems, typesetting, protocols, etc, make more rigid promises about internal structure
of language objects. For the purpose of this work, it is not necessary to go into more detail
than this.

Data may be transmuted through many different alphabets during the course of reading, copying, transmitting, and storing sampled data in different memory representations.
\begin{example}[Multiple language representations]
  Imagine reading aloud from a book, to an illustrator, who depicts a
  story in pictures, with captions written in another language. Each version
  of the story represents the story in a different semantic (spacetime) medium. 
\end{example}
Representations may not be fully equivalent.
\begin{example}[Binary representation of languages]
  The alphabet of symbols for contemporary computers comprises the
  binary symbols 0 and 1. These are combined in fixed size strings
  (superagents) or tuples to represent arbitrary pattern. At a higher
  level, these are grouped into 8 bit words over ASCII symbols, or 16
  bit Unicode symbols. These symbols form strings in different
human and computer languages.
\end{example}

\begin{example}[Codons]
  In biology, two alphabets of the genetic language are composed of
  the four almost identical sets of symbols for DNA/RNA: G,A,T/U,C.
  The language of proteins is encoded by codons (triplets of DNA/RNA
  symbols). These cluster into words formed of three letters, called
  codons, which represent amino acids.
\end{example}
The symbols in an alphabet represent a collection of states which can be
promised by spacetime agents.
Signal agents, promising messages in these languages, can be
emitted and absorbed in streams, as promise bodies, and act as symbols 
themselves in order to communicate between location agents\cite{spacetime2}. 
These few points about language should be enough to describe serial
communication in a general way.

\subsection{Spatial addressability of memory}\label{floodroute}

Storing information as spatial structures is, by far, the dominant
form of memory representation at every scale.  The interior nature of available
spacetime agents is central in determining the sophistication of the
patterns that can be supported by a memory representation.  Simple
agents, technological or biological, promise limited interior
capabilities but can make up for information density through numbers.
On the hand, a medium like the biomolecular language of proteins has
immeasurable complexity.

A minimum requirement for agents is that they must promise labels to be individually
addressable, in such a way that information stored may be retrieved
from the either the same location, or one that is deterministically
equivalent. For this to be true, there must be cooperation between the
agents, and each must promise a unique identity, as either a name
or a number, within their localized namespace (see paper II).
\begin{example}[Pages of a book]
A book is a collection of pages. Each page agent is numbered
allowing addressability of its contents.
\end{example}
Two approaches may be used for addressability.


\subsubsection{Address discrimination by constrained flooding or broadcast propagation (-)}

All agents promise to recognize their own address or unique name.
A sender imposes (floods) addressed information to all the agents, and every agent promises to 
only accept information labelled with its own address. 

Let $M_i$ be a set of memory agents, and $A$ be any agent that can
read of write to the memory. Each agent $M_i$ promises a value $v_i$, and a unique address
${\rm addr}_i$, sometimes referred to as its media access control address, to all other agents:
\beq
M_i \promise{+(v_i,{\rm addr}_i)} A, M_j  ~~~~~\forall i,j.
\eeq
In general, address uniqueness cannot be promised without coordination. Such coordination
could be expensive, requiring $4N^2$ promises amongst $N$ agents, or
$2N$ promises with a centralized source authority. In practice, memory addresses
are imposed on memory agents, by a single source authority, which we may call the factory $F$: 
\beq
F   &\imposition{{\rm addr}_i}& M_i\\
M_i &\promise{-{\rm addr}_i}& F.
\eeq
To write a value to a particular $M_p$, and agent $A$ imposes a data value $v$, along with 
the intended address onto all agents (this is called broadcasting or flooding=:
\beq
A \imposition{+(v,{\rm addr}_p)} \{M_i\}
\eeq
All agents promise to accept such messages, conditionally if ${\rm addr}_i = {\rm addr}_p$:
\beq
M_i \promise{-v | {\rm addr}_i = {\rm addr}_p} A, M_j, ~~~~\forall i,j.
\eeq
These promises are hardwired into the agents in many cases, however,
unreliable agents
may or may not promise not to accept messages not addressed to them.
\begin{example}[Unreliable agent addressing]
If we chose the $M_i$ at the scale of representing students in a
classroom, then we could not assume that they would all promise to
behave in such a simple mechanistic manner. Several students could
have the same name, as they originate from different `factories', and
so on. In other words, depending in the scale and nature of the
agents, we should be careful about what assumptions are made.  
\end{example}

As the number of addressable agents grows, a flooding scheme
imposes an increasing burden on the capabilities of the receivers, to
i) recognize longer and longer addresses, and ii) to respond quickly and unblock
the channel for future messages.

\begin{example}
  The biological immune system uses a flooding mechanism to signal
  cells.  It is impractical for cells to have unique addresses, but
  cells of a given type are considered redundant, and have the same
  function. Bio-addressability targets classes of cells rather than
  individual cells. Cells have antigen receptors (Major 
  Histocompatability Complices (MHC), etc.), which act as addresses
  for classes of equivalent cells.
  These sites bind only to particular messenger cells like antibodies, and thus 
  receive only the signals intended for them.
\end{example}

\begin{figure}[ht]
\begin{center}
\includegraphics[width=6.5cm]{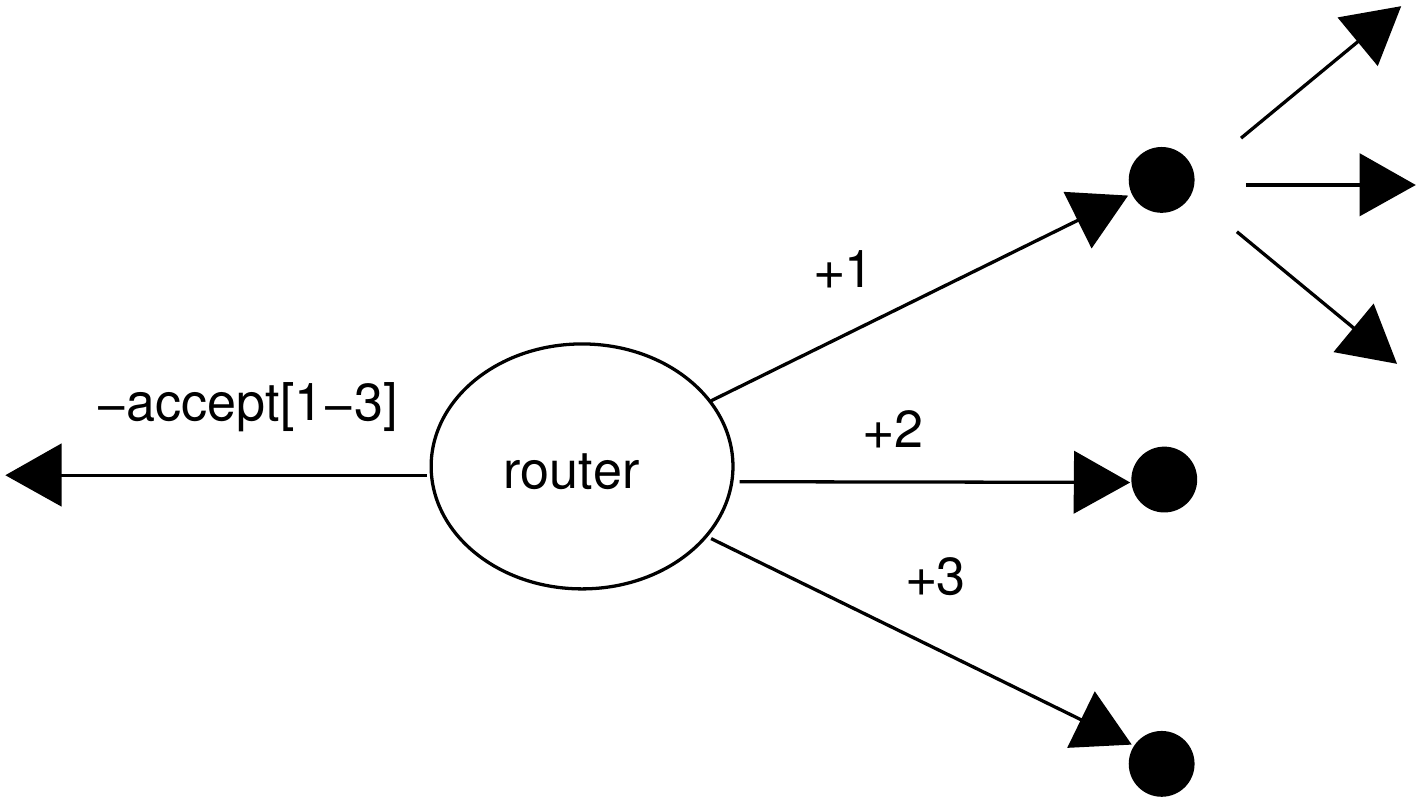}
\caption{\small Individual identity promise, and voluntary identity
  recognition is a part of both flooding and non-flooding
  addressing schemes, if one considers
  the scaling of routed addresses. A router accepts a certain range of
  addresses and routes precisely. If the paths (or ports in the
  language of \cite{graphdynamics1}) to 1,2,3 were not unique, the end
  nodes would also have to promise to know their identities and
  discriminate. Routing allows end nodes to operate with few promises.
  The principle can be applied recursively to address arbitrarily
  large numbers of agents, at the cost of an increasing number of
  trusted intermediaries.\label{acceptor}}
\end{center}
\end{figure}

\subsubsection{Address discrimination by routed propagation (+)}

A way around the scaling limitations of flooding is to hardwire
spacetime adjacencies so that there is a unique and direct path to each agent from a source
(see figure \ref{acceptor}).  An intermediate agent $R$, which we call
a router, is introduced between the sender of a message $A$ and a finite number
of receivers $M_i$, say in the range $0 <i < N$. Each leaf agent now has a unique identity
to the intermediate routing agent, by virtue of its unique adjacency relationship, and 
being isolated in its own cul de sac of the hub $R$.
The router $R$ accepts data from an agent $A$, if the address lies in its range:
\beq
A &\imposition{+(v,{\rm addr}_p)}& R\\
R &\promise{-(v,{\rm addr}_p) | 0 < p < N}& A
\eeq
Further, it promises (or imposes) to forward the message to the $M_i$, without broadcasting it to all
its children:
\beq
R \promise{+(v,{\rm addr}_p) | p=i} M_i
\eeq
and $M_i$ accepts:
\beq
M_i \promise{-(v,{\rm addr}_p)} R.
\eeq
It is no longer strictly necessary for $M_i$ to make acceptance
conditional on the address, if one assumes that the configuration of
agents is fixed and that $R$ is a trusted intermediary and delivers
the data correctly. Promise theory makes it clear that one cannot
place complete trust in intermediaries: they may be unable or
unwilling to keep their promises to relay information.

An alternative (which is more in keeping with the agent autonomy) is
that the relationship between $R$ and $M_i$ is reversed. $M_i$ can promise
its identity or leaf address to $R$, to be placed in a routing table
(and index or directory) for forwarding:
\beq
M_i &\promise{+{\rm addr}_i}& R.\\
R   &\promise{-{\rm addr}_i}& M_i.
\eeq
The difficulty with this is that the $M_i$ have no way to make their
addresses unique, without exterior coordination\footnote{In Internet
  addressing, which uses this scheme, the uniqueness of addresses must
be handled manually, out of band, with an additional network of promises.}. This conundrum is a namespace management
issue, which I won't explore further in these notes.

In summary, the intermediary $R$ has a topology that gives it a unique path to
each memory address. The router promises to keep an index overview, or directory
map, enabling it to switch on connected routes (see paper I).  It selects the path
corresponding to the promised address, and forwards information
accordingly, i.e. it acts as a switch that discriminates direction, based on the
promise of an address. 

To scale this further, addresses may be made
hierarchical, be splitting them into a routing prefix address and a
leaf address. Then router would modify its promise to accept data only
if the routing prefix were addressed to it specifically. This
principle can be applied hierarchically to provide tuple addressing
with direct memory access along dedicated connections.

\begin{example}
  Computer networks use both kinds of address discrimination. Bus
  architectures, like Ethernet or WiFi, assign a factory unique
  machine (MAC) address to each network agent. Messages containing
  serially encoded addresses, are flooded to all agents in a region
  and receivers promise to accept messages addressed to them. No
  ordering of addresses is needed: the agents are as a gas.  For
  routed architectures, like the Internet protocol (IP), addresses are
  routed by selecting the known pathway to the destination. This
  presumes a cooperative ordered arrangement of IP addresses, in a
  solid state.
\end{example}

\begin{example}
  Computer memory chips assign addresses to memory locations by using
  a discriminatory switch, attached to an address bus. The address bus
  acts as the envelope for messages on the data bus, imposing
  addresses on to the leaves.  Collaborating memory chips are arranged
  in an order with hardwired signals to tell them their address
  ranges. A hierarchical system of relative addressing, with bit
  discrimination, allows memory chips to accept messages only for the
  address ranges they represent, and single out the addressable memory
  registers uniquely.
\end{example}

\subsubsection{Criteria for addressability of data in memory}\label{addressability}

By either of the foregoing methods, absolute addressability of memory
implies the ability to uniquely signal or retrieve information from
every agent. Some basic promises enable this.

Names or addresses\footnote{Whether we call this a name or an address
  is not really important. An address is just a long name, usually
  given relative to a namespace.} must be unique for all agents within
a hierarchy of namespaces that connect senders and receivers of data,
i.e. every memory agent must promise an
identity that is uniquely distinguishable by its neighbours within its
scope.
Every agent must be distinguishable from every other locally, else it
would be impossible to intentionally retrieve data stored in a memory
agent, or for a memory agent to forward to another agent
deterministically.
\begin{lemma}[Deterministic addresses]
  Let $M_i$ be a collection of memory agents. $M_i$ may be
  deterministically addressed by any other agent $M_{j\not=i}$, iff
  $M_i$ promises a unique distinguishable identity.
\end{lemma}
The proof may be found by considering the properties of memory agents.
If $M_i$ and $M_j$ cannot be distinguished, then 
\beq
S_k &\imposition{d_k}& M_i\\
S_k &\imposition{d_k}& M_j
\eeq 
cannot be distinguished, and data
may be stored in unknowable locations, and may overwrite existing
values. Moreover, 
\beq
M_i &\promise{+d|r}& R \\
M_j &\promise{+d|r}& R
\eeq
cannot be distinguished, thus an agent would retrieve an arbitrary
value from an unknowable location.  If the agents are distinguishable,
the agents act independently and deterministically according to
definition \ref{memoryagent}.

The ability to coordinatize a space, to assign unique tuple addresses,
depends on its topology, i.e. on agent adjacency. Non-trivial
topologies may have loops and multi-valued coordinate, causing
significant difficulties. This is why most artificial structures
are based on trees (directed acyclic graphs) or partially ordered
lattice structures.  The acceptance of
address labelled data cannot be avoided, even in a direct tree routing
approach, however.  Agents need to recognize their own messages (see figure
\ref{acceptor}).

\begin{figure}[ht]
\begin{center}
\includegraphics[width=13.5cm]{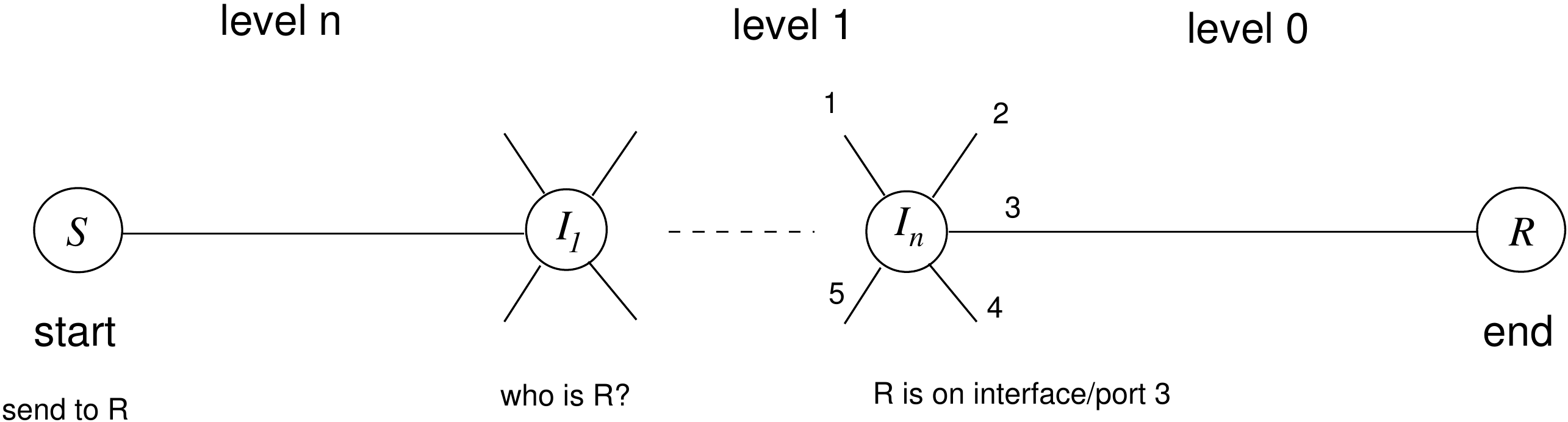}
\caption{\small Agent names or addresses have to propagate out from
  their location $R$ throughout a semantic space, before other agents
  $S$ can i) know that they exist, and ii) know how to reach them as a
  destination. Messages could be given a maximum range, relative to a
  current position, iff the geometry of the space is fully understood
  by all agents.
\label{addressable}}
\end{center}
\end{figure}

Further to this, we may say that in networks, where there is not direct adjacency with memory
locations, the existence of a complete name or address has to propagate from every
memory location to all remote agents in order to become
addressable in practice, i.e. readable or writable\footnote{Note that the ability to route
to $R$ should not be confused with the ability to locate named data objects
stored in the memory network without pointing to their
specific location, as in\cite{ndn}. The latter is an independent problem. Only the forwarders that connect
the address space need full unique addresses.}.
\begin{lemma}[Name/address propagation]
  An agent $R$ may be addressed or signalled directly by an agent $S$
  iff the full name of $R$ propagates through all intermediary agents
  $I_1\ldots I_n$ to $S$, and $I_1\ldots I_n$ commit the route by
  which the information propagated to memory (see figure \ref{addressable}).
\end{lemma}
The implications of this lemma are that deep hierarchical naming
or addressing of agents requires a routing infrastructure based itself
on memory agents, thus scaling of memory may require recursively nested memory.
To prove this, we may assume that $R$ promises its own identity:
\beq
R \equiv A_R \promise{+R}.
\eeq
In order for any other agent $A_i$ to learn this identity, we must have
\beq
A_i \promise{-R} R.
\eeq
Knowledge of $R's$ identity can propagate iff
\beq
A_i &\promise{-R}& A_j\\
A_i &\promise{+R}& A_{j\not=R}, ~~~\forall i,j\not= R.
\eeq
Since $S \in A_{i\not=R}$, $S$ can learn of $R's$ identity, and may try to promise
or signal to $R$. The name of $R$ may not be private. All agents must be in
scope of its name promise. Now, in order for $S$ to signal $R$, there must
be a connected path along intermediaries $I_1\ldots I_n$, all of which promise
to forward data to an output interface along the path. This information can 
only come from back-propagation from $R$ to $S, I_1\ldots I_n$, and these agents
remember $R$ and the route `$r_{ij}$' to select the appropriate interface direction.
The intermediaries must therefore have memory of their own.

The costs of routing or flooding a network are quite different.  In a
pure flooding approach, with no optimized routing, all agents are
blocked by a single agent's use of a shared network, locked into a single concurrent timeslice.
In a routed network, there is a setup cost of propagating addresses
to the forwarding tables, and then a latency proportional to $r^N$
for path length $N$.

\begin{lemma}[Forwarding, name/address scope, and embedded memory]
  If intermediate agents $I_1\ldots I_n$ forge a path from $S$ to $R$,
  in a hierarchical spanning tree of non-overlapping nested regions
  (`subnets') with $N$ tiers, then $N$-tuple addresses may be used
to bound (but not eliminate) the memory required for forwarding messages,
at each level of agency:
\begin{itemize}
\item $R$ needs only $R$
\item $I_n$ needs only $R, I_{m<n}$
\item $S$ needs the full path.
\end{itemize}
\end{lemma}
We can see this as follows.
In order to route at each stage of the path, the intermediaries must have
a routing table `$r_{ij}$'. Without embedded memory, this cannot be achieved.
In order to discriminate its own level, as well as pass on data to all sub-levels,
unidirectionally, this information is used. Other parts of a tuple address
can be discarded.

\begin{example}[IP addressing]
  The IP address subnet mask divides an encoded IP address into a
  prefix and a local address part: (prefix,local). This is a two level spanning
  hierarchy, where local addresses are used at the end node, and
  prefix is used for forwarding. With only two levels, the approach
is simple but not very efficient, leading to high memory usage in
forwarding agents. If a deeper hierarchy, with longer tuples were used,
the memory required to forward could be reduced to enable local uniqueness.
\end{example}
\begin{example}[Ethernet MAC addressing]
  Ethernet addresses have no namespaces (ignoring VLAN), and so
  delivery has to be by public flooding and self recognition. No
  memory is required for this, but each data transmission occupies the
  entire channel for all agents, blocking their ability to promise
  simultaneous availability.
\end{example}

The foregoing results show how
hierarchy plays a key role in scaling the local uniqueness of addresses.
By exploiting superagent structures as `fixed star' markers, 
the addresses of individual agents can be made relative to the superagent containers. This includes the addresses of sub-agents within them, and so on, recursively down the levels.

\begin{example}[Tuples, namespaces, and the Arabic numerals]
  Hierarchical namespaces are what allows us to represent arbitrary
  numbers with only the Arabic symbols 0-9. By forming tuples of
  numbers we allow multiple digits to represent different blocks of
  the final digit, e.g. 13 is address 3 in bank 1, and 23 is address 3
  in bank 2. Without such tuple addressing, ordinal values would have to
be represented by nominals, i.e. we would need a different symbol for
every number. For example, singlets, using nominal alphabet [A-Z], can encode only
the addresses [1..26], but doublets (a,b) using relative addresses [0-9][0-9]
can encode numbers from 0..99, as they two values vary independently, with 
tuple elements counting as logarithmic partitions.
\end{example}

\subsubsection{Non-local link memory}

In spatial memory, key-value networks may be approached in
two ways: by storing data locally at fixed agent locations, and non-locally by storing
data as processes that span links.  The memory models described thus far may be called
deterministic, local models of storage: they map a fixed address to a fixed
location.  A different kind of memory is possible, based on network
linkage, by observing that, between any number $N$ of agents, there
may be as many as $N^2-N$ links. Thus, for large memory spaces,
storing memory in the links rather than the locations could be very
efficient. The challenge lies in addressing and encoding the memories.
This is the idea behind Bayesian networks, and the related Neural
Networks (for a review, see for example \cite{duda1,bayes,pearl1}).

\begin{figure}[ht]
\begin{center}
\includegraphics[width=7.5cm]{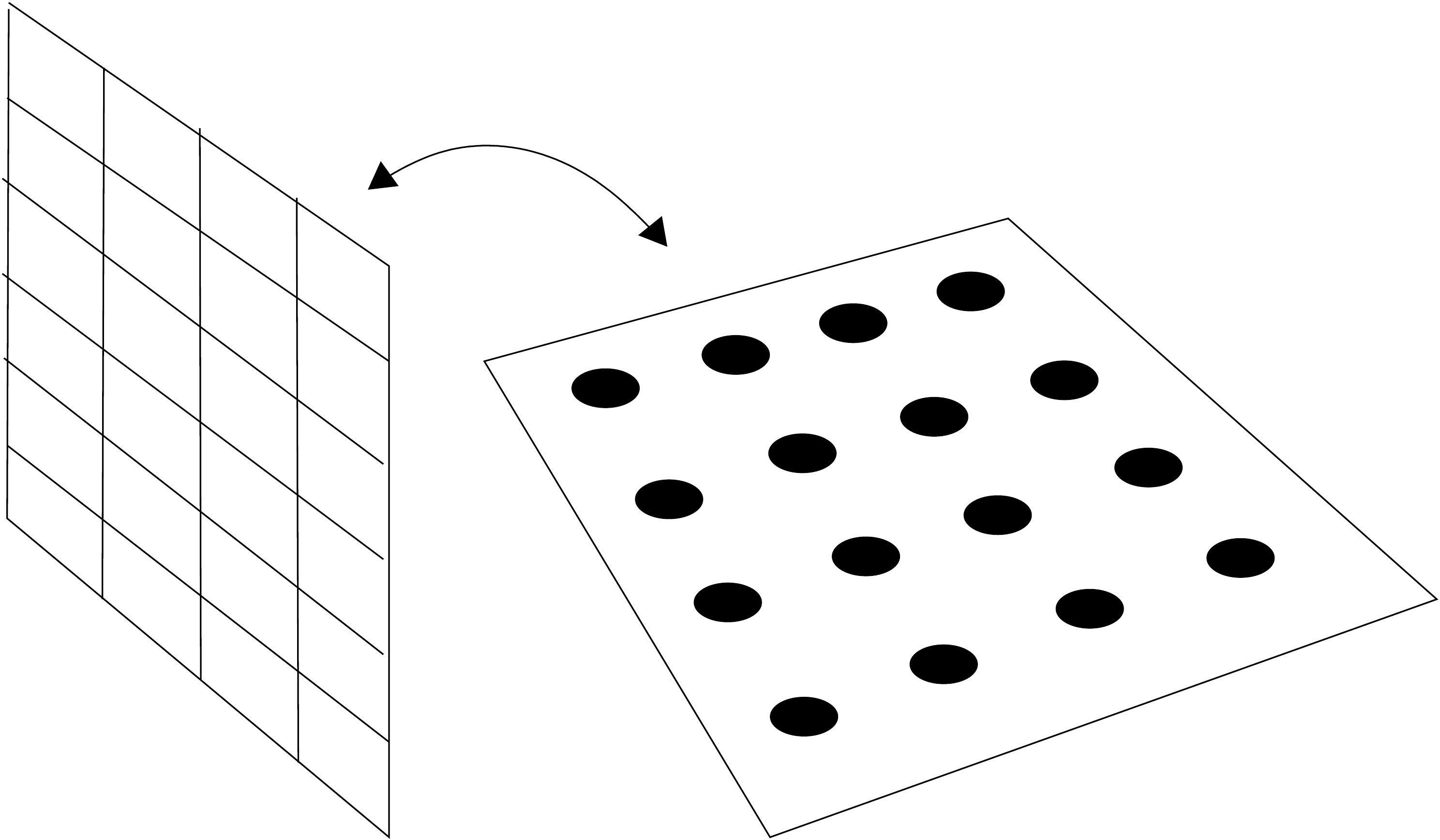}
\caption{\small Mapping an observed exterior space into an interior representation, 
through a sensory apparatus, may imply an essentially isomorphic representation.\label{spacemap}}
\end{center}
\end{figure}

\subsubsection{Teaching spacetime to tokenize patterns}\label{topofit}

Sensors are often designed, shaped, or honed, over their evolution,
specifically to register particular spacetime phenomena\footnote{This seems to be what
Ashby called homomorphic machines\cite{ashby2} in his work on cybernetics.}. 
Such sensor `designs' favour the
recognition of patterns by virtue of their structural properties, and minimize
the need for data analysis and post-processing of interpretation. For example,
the ability to judge distance might be enhanced by a sensor array with
constantly spaced pattern between locations, acting as a graduated measuring stick.
Through this specific adaptation,
sensors are made efficient as pattern discriminators, by learning
that is cached by evolutionary selection, over longer timescales.  There is
evidence that this is true in the human senses: eyes and ear do not
merely pass data to the brain but perform staged preprocessing and
limited `type' recognition, through the existence of specific cells
for the measurement of motion.  Research has further identified
fabrics of place cells, grid cells, head direction cells, border cells
for seeing navigation of spacetime, and even facial recognition
regions, and so on\cite{facecells,placecells,gridcells,gridcells2,Fox05072005}.

Software techniques, using machine-learning, can simulate this preprocessing,
by long-term sensor adaptation, to train sensors with particular semantics; this
is typically done by training artificial neural networks (ANN) to identify patterns\cite{deeplearning3}\footnote{Thus ANNs are
  more akin to the preprocessing of a visual cortex than perhaps the
  brain that receives the interpreted signals.}.  This ability to
optimize the recognition by evolutionary learning should be considered
highly significant from a spacetime viewpoint.  If one can ask a
specific question about external phenomena, with a specific answer, it
is possible to build a discriminator based on the scaling of a fabric
that uses semantic addressing to yield a direct symbol token.  This
amounts to a large compression of information being forwarded into a
knowledge representation, and suggests that tokenization is a powerful
strategy for making cognition efficient.

In sensor training, implicit associations become encoded, over
multiple link patterns, by traversing spatial memory lattices,
whose geometry is specialized to play the role of parsers that trigger a specific
tokenization of the inputs. This is somewhat analogous to the creation of
a directory or index for fast lookup. A particular symbolic token may be recalled because it is
adjacent, in the index coordinatization, for the context we
are currently exploring. With non-local link encoding, this process is
still somewhat mysterious, but it amounts to a form of semantic hashing.

This training of specialized sensors seems almost fantastic, however,
it makes considerable sense from a processing viewpoint. Without a
pre-seeded similarity of structure to lay out boundary markers, there
is no obvious way in which an unbiased associative structure could
spontaneously break a symmetry to recognize spacetime
characteristics like distance, direction, dimension, location, etc;
yet we know that they do.  What long term learning enables is the
caching of environmental boundary conditions that train discrimination
structures by selection.
\begin{example}[Head direction cells]
These are cells that perform a form of context assessment, approximately
tracking a compass direction of the head by relative motion.  Grid
cells and place cells in the brain have a regular lattice structure
that allow an image of space to be projected into an array of agents
that mimic the same kind of connectivity.
\end{example}

This partial mimicry of patterns by spatial structure goes beyond the recognition of
exterior sensory patterns.  It applies internally too, in the
representations of concepts.  In a knowledge system, we have to answer
how similar patterns may be mapped to similar regions, so that
uniqueness is assured over an appropriate region, (e.g. in a brain
network, how does the same concept avoid being senselessly repeated at different
locations, like a plague?). A notion of idempotent convergence seems
important to maintain stability (i.e.  concepts as singular desired
end-states)

\begin{example}[Deep learning applied to reasoning]\label{ann1}
  Artificial Neural Networks (ANN) or deep learning networks have
  proven to be effective at forming stable attractors in the
  recognition of discernable patterns, even though they may be
  inefficient in terms of computational resources. Foreknowledge of a
  pattern, combined with system design (where possible), can sometimes
  be used to simplify the recognition of patterns with much lower
  cost.

  In \cite{deeplearning3,deeplearning4,deeplearning5}, a multi-stage,
  highly trained series of ANNs is used as an advanced discriminator
  for sentence parsing. This approach is interesting, as it does not
  attempt to preserve grammatical structure, except in word compound
  groupings. Moreover, using multiply, individually trained stages, it
  approximates intentional promises through a mixture of emergent
  behaviours and hand-crafted training design.  The naming of concepts
  is emergent (non-intentional), buried in the internal layers of of
  the neural network. Thus intermediate concepts may be presumed to
  exist, but are only accessible from the encoding at the input, and
  the associative weights that lead to particular network activations:
  their internal names are non human-readable.  The use of specific
  associative forms, belonging to the four irreducible types features
  in the authors' solution.

  The model in this work leads to the kind of representation one
  expects for the processing of an {\em intentional} process:
  information is promised at the inputs and outputs, and training
  shapes the network weights to propagate input to output conditions,
  assuming the existence of sufficient internal degrees of freedom to
  compute stable attractors.  We can thus understand ANN training as
  an approximate, non-linear, best-fit propagator matrix for
  input/output, with hand-crafted spacetime symmetry breaking.
\end{example}

A working hypothesis, based on a semantic spacetime model, is that one
would expect a successful sensor apparatus to align its operation with
causal structure and precedence, and operate like a `prism' which
transmutes spatial patterns into separated sequential frames of input
to be parsed in chunks\footnote{Might we even show that the brain
  cannot simply be an aggregation of homogeneous autonomous agents, as
  the cortex seems to be, without some pre-seeded symmetry breaking
  from sensory arrangements?  There must be symmetry reduction in
  order to support the different tenant modules as well as to
  distinguish different datatypes?}.
This hypothesis further seems to be a necessary condition for
remembering location, and navigating in artificial memory, e.g.
chips, cities.

\subsection{Temporal addressability of memory}

Temporal memory includes the encoding of data like transaction logs,
melodies, video, and any sensory stream.  How do we encode and recall
such sequences in a reliable way?  

\subsubsection{Sequences and oriented paths}

To encode an arrow of time, i.e. a directed sequence, it's necessary
to break the symmetry of a memory space in some way, by establishing a
precedence relation. One way is to distinguish locations and order
their consecutive addresses, either by labelled addressing or by using
a hardwired discriminator structure.  These may be formed by any of
the causal, irreducible promise types:
\begin{lemma}[Partially ordered promises]

Partially ordered addressable structures are representable as partially ordered promise types:
\begin{itemize}
\item Precedence adjacency gradient (queues):
\beq
A_1 \promise{\text{precedes}} A_2 \promise{\text{precedes}}  A_3.
\eeq
\item Conditional dependence or prerequisite:
\beq
A_3 \promise{\text{later} | \text{earlier}} A_2 \promise{\text{earlier}}  A_1.
\eeq
\item Superagency and encapsulation:
\beq
A \promise{\text{inside}} S \promise{\text{inside}} S'.
\eeq
\end{itemize}
\end{lemma}
By breaking the symmetry with an arrow of time, one finds a
representation for sequential memory. Since learning is not sequential
but cyclic, long term memory must be able to reliably revisit such
sequences to reach an equilibrium along the time vector to end in a
summary state. It is plausible that each agent, as part of a sequence,
could reach some kind of independent averaged equilibrium, with multiple sequences
representing the same semantics. This begs the question: how does an
observer know whether two sequences have the same intended semantics
or not? 

\begin{example}[Sequences with constant semantics]
  When does it make sense to compare two sequences (e.g.  comparing a
  shopping list with wedding invitations does not make sense)? The
  simple answer to this question is that such comparisons must be
  baked into the immutable aspects of the sensory stream. The boundary
  conditions of sensory experience are the only source of information
  on which to base such semantics. Thus we hypothesize that this is
  where long term evolutionary adaptation of sensor recognition play a
  role. A separation of timescales allows long term adaptation to
  similar spacetime processes to feedback and shape the short term
  sensory recognition of similar patterns.
\end{example}

\subsubsection{Knowledge as equilibrium, with constant semantics}

I have proposed that knowledge is to be understood not as information
but as an the stable equilibrium of an active relationship\cite{burgesskm,certainty}.
We may state this hypothesis as a less ambitious lemma:
\begin{lemma}[Learning with immutable semantics]
  Learning by repeated observation over semantic sequences, is
  possible iff all samples promise to be piecewise graph homomorphic to one
  another.
\end{lemma}
Without a consistent promise graph, data cannot be combined or mapped
into a coordinate system reliably, with preservation of structure.
The simplest way to make measurements promise similar graphs is to use
the same sampling apparatus or sensor for all samples. This fixes the
boundary conditions, and suggests that, even if sensors share certain
infrastructure, it is natural for specific intermediary stages to be
routed to specific spacetime discriminating structures that encode
graph-isomorphic structures with the appropriate ordering and
interpretation.
\begin{example}
The eye shares a lens and optical infrastructure. The captured
light may be processed by different specialized memory cells
as part of decoding or imposing semantics on the data:
\beq
\text{Lens} \rightarrow \text{Retina} \rightarrow  \left\{
\begin{array}{l} \text{Colour}\\
  \text{Grid shape}\\
  \text{Orientation}\\
\ldots
\end{array}
\right.
\rightarrow \ldots \text{Concepts}
\eeq
Even with substitution of an element in the process, these have shared
structure, and can share semantics.
\end{example}

\subsubsection{Advanced-retarded spacetime propagation and addressability}

The promises needed for propagating addressability of space, by
comparing agent identities, are related to the concepts of so-called
{\em advanced} and {\em retarded} boundary conditions in the physics
of field propagation and quantum mechanics\cite{burgesscovariant}.
Referring to the discussion in section \ref{addressability}, we may define.
\begin{definition}[Advanced propagation]
Information originates from an end point and
expands towards possible starting points.
\beq
R \promise{-d} \{ I_1\ldots I_n, S\}.
\eeq
\end{definition}
Advanced propagation is like `back propagation', as referred to in the training of
learning networks.

\begin{definition}[Retarded propagation]
Information originates from a starting point and
expands towards possible end points.
\beq
S \promise{+d} \{ I_1\ldots I_n, R\}.
\eeq
\end{definition}
These two propagation methods are conjugates of one another,
and both processes are involved in determining a route from start to
end in a complete path.  (see figure \ref{addressable}).

Only the end points represent information sources, assuming that
intermediaries admit zero distortion\footnote{This is a common
  assumption, but also one that is wholly unjustified. The presence of
  intermediate agents is likely to distort the propagation of
  information, without a considerable infrastructure of promises (see
  the discussion of common knowledge and the end-to-end problem in
  \cite{promisebook}).}. The propagation of a message $d$ from one to
the other is possibly universal and carries no significance. However,
the joining of these defines an overlap of information transfer (what
is sometimes called a transaction).
\begin{lemma}[Path overlap of advanced and retarded propagation]
The set of oriented transitions from $S$ to $R$, or vice versa
is given by the overlap of advanced and retarded propagations for data $\pm d$
originating from $R$ and $S$ respectively.
\end{lemma}
If there exists one or more connected paths from $S$ to $R$, involving
an arbitrary number of intermediate agents, then it make be possible to
propagate information in one or  both directions to connect $S$ with $R$.
These mutual flooding processes can be joined at any intermediate location
along a viable path, provided the message types are compatible ($\pm d$ as a lock and key).

\begin{example}[Quantum mechanics as an address routing protocol]
  This construction is very like Feynman's notion of a spacetime path
  formulation of quantum mechanics\cite{feynman1}, and the Feynman
  Wheeler absorber theory\cite{feynmanwheeler}, as well as Schwinger's
  source theory\cite{schwinger1}. This is discussed in the
  Transactional Interpretation of Quantum
  Mechanics\cite{transactionalqm}. Between an interacting system and
  observer, there is an `offer wave' $\psi(x,t)$ described by the wave
  function of the system, and a `confirmation wave'
  $\psi^\dagger(x,t)$ travelling from the observer back to the system.
  The overlap between these $\psi^\dagger\psi$ is the probability of a
  connected path between them, as $\langle \psi|T|\psi\rangle$ are
  matrix elements for specific path transitions. In Feynman's
  construction is represents a sum over all possible connected paths
  that satisfy the energy conservation payment conditions (see figure
  \ref{qm}).
\begin{figure}[ht]
\begin{center}
\includegraphics[width=5.5cm]{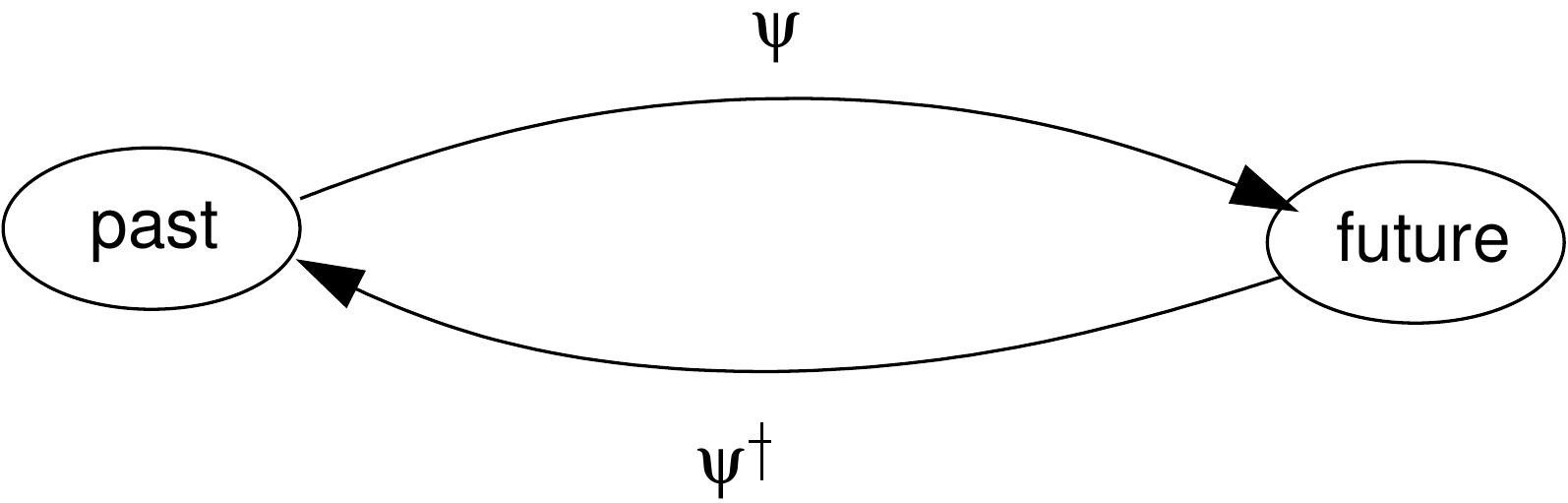}
\caption{\small A handshake between two propagating waves represents
a set of overlap path transitions connecting them. \label{qm}}
\end{center}
\end{figure}

In this view, the wavefunction of a system plays the role of a
pre-arranged address structure (like a directory, or name service) to
distinguish spacetime locations. Indistinguishable locations cannot
have different connection probabilities, else they would be
distinguishable.  The wavefunction is thus an equilibrium information
map of a system at the snapshot instant of an interaction, and the
solutions for $\psi(x,t)$ have the status of an on-going knowledge
relationship between observer and observed. Note that, although the
time $t$ appears in as an evolutionary learning parameter in
$\psi(x,t)$, this is not the same time as is represented by the
propagation from boundary points $S$ to $R$.  Moreover, the
time-reversal invariance represents path-reversal (route) invariance
rather than absolute time.
\end{example}

\subsubsection{Splitters and oriented discriminators, encoding decisions}

How is a preferred direction realized (reified) in a representational
spacetime?  There any may processes that routinely break directional
symmetries in the exterior world: the position of the sun, rivers,
etc.  On the interior of a knowledge representation, there are no such
landmarks, so associations are always relative to something in the
interior. There are two ways around this:
\begin{itemize}
\item Encode absolute boundary conditions spatially in memory as 
homomorphic `concepts', and then
  associate directions relative to those concepts.

\item Form a Cartesian theatre\cite{dennet1} facsimile of the outer
  world and use interior properties to represent directional
  processes: electrical currents (e.g.  in a brain, or an electronic
  circuit) select a direction of flow; the act of pushing a memory
  by imposition or force breaks its directional symmetry; start and
  end boundary conditions select a direction, e.g. , like a contact
  potential, or being inside the plates of a capacitor, or a
  temperature field, where some field gradient sets a bias.
\end{itemize}
\begin{figure}[ht]
\begin{center}
\includegraphics[width=9.5cm]{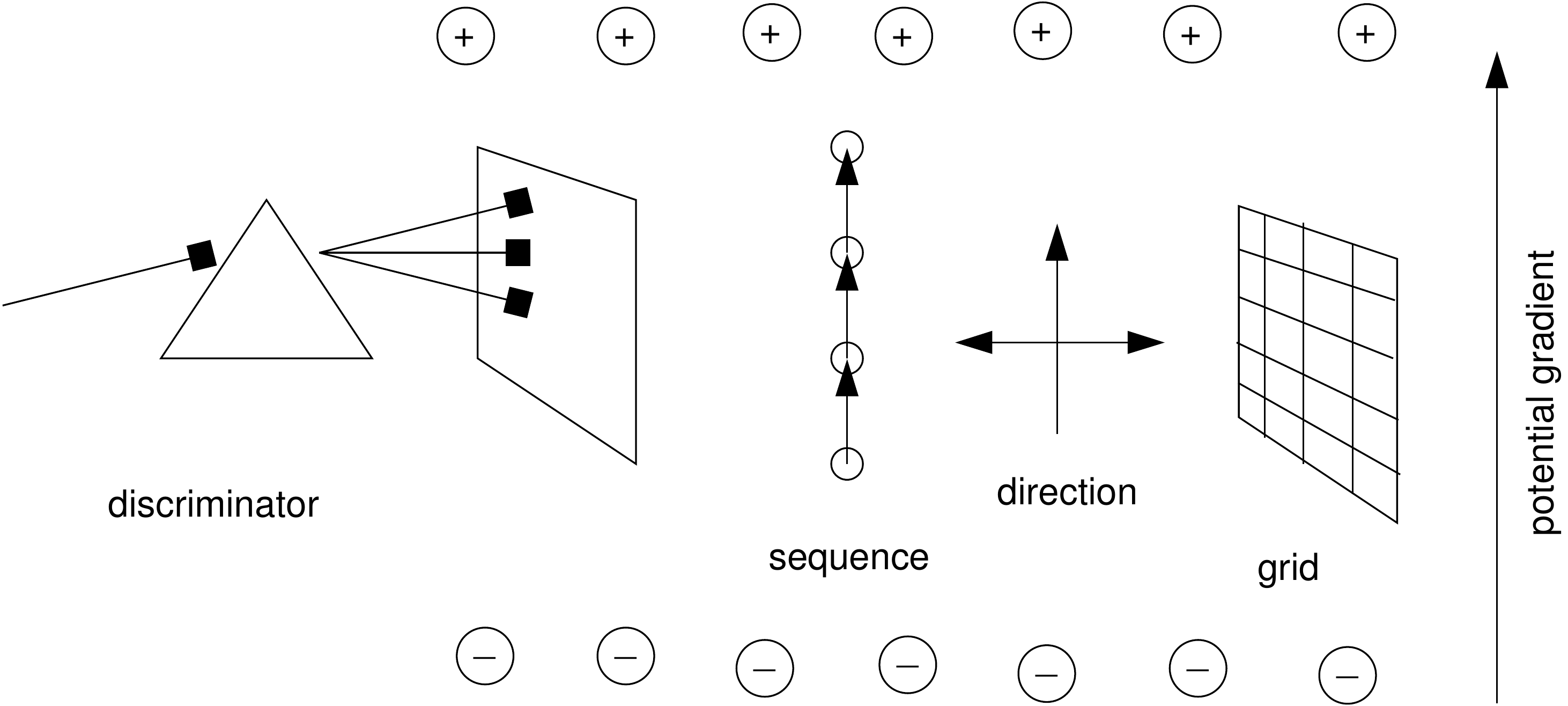}
\caption{\small Ordered spacetime structures form directed graphs,
  either by autonomous cooperation of memory locations, or by external
  calibration through discriminators, routers, prisms, etc. \label{temporalgrad}}
\end{center}
\end{figure}
Certain spacetime processes break symmetries, by virtue of their nature,
e.g. prisms, gradient fields, etc (see figure \ref{temporalgrad}). Sensors
can make use of these devices in adapting to pattern encoding.
Directional memory exhibit any axial symmetry. Branching treelike
(conical) structures are sequential and directional, with
discrimination built in. In a treelike bifurcation, semantically close
concepts could end up close or very far apart in memory
space\footnote{Hawkins has presented ideas about Hierarchical Temporal
  Memory, in which sparse distributed representations are used to
  store sequences, however too few details are in the public domain to
evaluate the claims  \cite{zhao1,hawkins2,hawkins,facecells}.}.

\begin{example}[Optical interferometry and correlation]
  The optical versions of neural networks, which have much the same
  kind of spacetime structure, are diffraction gratings, optical
  correlators, holographic systems, and other interferoemeters. These
  encode information in frequency and phase and yield fringe patterns
  that can be used for optical processing. This underlines the point
  that massively parallel `spatial computation' may be performed in
  many ways\cite{CTOptics}.
\end{example}

\subsubsection{Role of separated timescales in tokenizing}

As usual, the separation of scales informs the dynamical basis for the
semantic machinery.  Three timescales for memory are generally
accepted in human learning, and translate naturally for other learning
systems by spacetime considerations\cite{neurosciencebook}:
\begin{enumerate}
\item {\bf Immediate memory}: for data collection, i.e. the ability to
  hold ongoing experiences from fractions of a second to seconds.  The
  capacity of this register is very large, involves all modalities
  (visual, verbal, tactile, and so on), and provides the ongoing sense
  of awareness or context.

\item {\bf Short-term memory}: allows humans to hold information in
  mind for seconds to minutes. This is associated with gathering of context.

  A special sort of (procedural) short-term memory is called working
  memory, which refers to the ability to hold information in mind long
  enough to carry out sequential actions. An example is searching for
  a lost object; working memory allows the hunt to proceed
  efficiently, avoiding places already inspected.

\item {\bf Long-term memory} entails the retention of information in a
  more permanent form of storage for days, weeks, or even a lifetime.
This is the equilibrated memory of a learning process.
\end{enumerate}

\subsection{Nominally addressed memory}

Numerical addresses, based on metric coordinates are well suited to
the labelling of ordered structures\cite{spacetime1}.  Nominal
addresses (semantic addresses) may also be used to represent ordered
structures, if an intermediary supplies the partial ordering of names
by transmuting names into partial ordering, through its spacetime structure.

Address discrimination, using names and locations, is made using the
two approaches in section \ref{floodroute}, i.e. intermediary forwarding switch 
or broadcast flooding (called a bus
architecture)\footnote{If the brain were a bus architecture, memory
  would be routed by address. We know of no way in which individual
  neurons or clusters could know their own addresses in a coordinated
  way, nor of a unique routing structure that discriminates on
  conceptual lines.  It is plausible that sensory bundles from major
  sensory regions: colour cells, place cells, frequency range cells,
  etc, could provide some of the labelling needed for form an
  approximate regional (digital) address.}  .  Flooding is an
inefficient approach since it occupies an entire region of space for
the duration of a transaction.
\begin{lemma}[Address routing]
  To transport a signal or agent from one location to another, it is necessary
  and sufficient for agents to promise the ability to distinguish
  direction, and a map between a unique absolute identity and the
  desired spacetime location.
\end{lemma}
The proof is as follows. If an agent does not promise a
unique name, i.e. a unique public scalar promise, it cannot be distinguished
from other agents. If the location has a unique identifier, and other
agents promise to recognize this unique name, the location will self
identify once it has been located. It can now be searched for by
covering the entire space. Once located, sign posts or complete paths
can be left to mark routes. A signpost is a lookup table (map) that
associates a class of locations with an identity. If agents do not
promise a unique identity, we have to be able to measure metric
distance from a signpost or origin. Then we can assign numerical
coordinates based in distance and direction.  Each coordinate system
breaks translational invariance by choosing an origin.  Either the map
is fixed in the name (coordinates) or paths are encoded into spacetime
itself The topology of tuples has to match the promised directional
topology of the spacetime. If neither direction nor some encoded name
are given, unique locatability is impossible.
We can now define semantic routing as routing without a metric
coordinate system:
\begin{definition}[Semantic routing]
  The use of conditional forwarding based on persistently promised directions and
  the use of lookup tables (directories) to match names and forwarding
  directions to the next signpost. Signposts form a vector field that
  converge onto an externally stable set of destinations\cite{burgessbook2}.
\end{definition}
Example \ref{ann1} illustrates the use of multi-stage semantic routing without explicitly known
names. Another example that uses naming based on weighted paths is the following.
\begin{example}[Breadcrumbs, ants, and road-signs]
  Road signs use nominal addressing, with arrows pointing to the
  direction to follow to find the next clue about a named object.  The
  formation of a trail for nominal addressing is basically a
  stigmergic process of discovery and breadcrumbs or pheromone
  trails\cite{swarm}.
\end{example}
Names (and thus nominal addresses) need not only be based on written
language.  To our senses, the name of a door handle is given by its
shape and relative position, just as the name of a protein is an
attached protein with a shape that gives it the ability to bind to a
receptor, like a lock and key.

\begin{example}[Machine addressing]
  Memory addressing in integrated circuitry and layer 2 (e.g. Ethernet) networks
  attached to buses makes use of self identification addresses. Data
  are addressed to a destination and flooded to all agents in their
  region of space.  The owner of the intended address accepts the
  data.
\end{example}

\begin{example}[Named Data and Content Centric Networking]\label{ndnex}
  Named Data Networking\cite{ndn}, and other name-based location
  addressing schemes, like Content Centric Networking\cite{ccn},
  Content Delivery Networks\cite{cdn} use names as the destination
  addresses of locations on the Internet.
\begin{figure}[ht]
\begin{center}
\includegraphics[width=6.5cm]{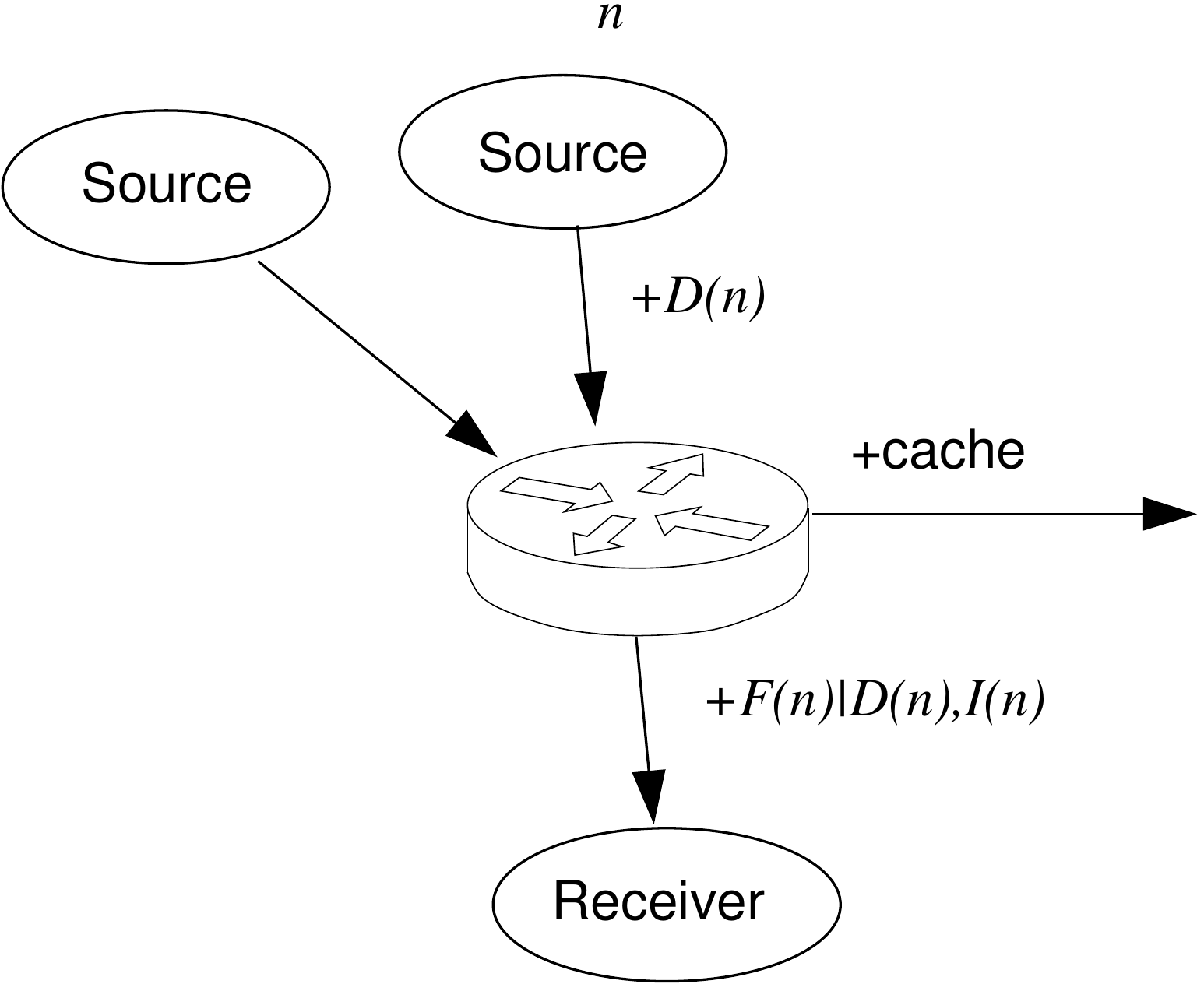}
\caption{\small In Named Data Networking (NDN), a trusted intermediary caches and passes along data
by name.\label{ndn}}
\end{center}
\end{figure}

  Unlike the impositional approaches to data transmission, NDN is
  based on the idea of voluntary cooperation\cite{burgessrpc}, and is
  a form of conditional forwarding based on
  identity\cite{aredo1,knowledgemace2007,promisebook,BorrilBCD14}.
  Sources of data promises promise their data $+D(n)$ by name $n$, with a
  signature of the originator to specify provenance (see figure \ref{ndn}). Willing receivers
  promise that they want data by name $+I(n)$ using `interest promises' also using the name.
  Name discovery requires some kind of directory.
  Forwarding agents (routers) promise to accept both kinds of promise
and forward data them by name, if they have received both voluntary
declarations, with a conditional forwarding promise $F|D,I$:
\beq
\text{Source} &\promise{+D(n)}& \text{Router}\\
\text{Router} &\promise{-D(n)}& \text{Source}\\
\text{Router} &\promise{+\text{cache}(n)}& \text{Source}\\
\text{Receiver} &\promise{-I(n)}& \text{Router}\\
\text{Router} &\promise{-I(n)}& \text{Receiver}\\
\text{Router} &\promise{+F(n)|+D(n),-I(n)}& \text{Receiver}.
\eeq
The intermediate router must be completely trustworthy to both parties,
as it has the ability to distort the data and even replace it\cite{aredo1,promisebook}. The authenticity
is presumed by virtue of signing.
Although neither the publishing of the data, nor the interest request
specify where to find the data (i.e. they need not have any concept
of location address), the router and the end points do need to understand one anothers'
locations or addresses. Identity might be encoded by direct and unique adjacency (signpost or interface direction),
if the routes are unique. There is always the possibility that the source and recipient may have
moved relative to the router's understanding of topology (e.g. in a wireless of gaseous network,
where broadcast flooding is the only approach). Thus the router has to make further promises
to give spacetime structure, but which do not involve data:
\beq
\text{Source} &\promise{\pm \text{addr}}& \text{Router}\\
\text{Router} &\promise{\pm \text{addr}}& \text{Source}\\
\text{Receiver} &\promise{\pm \text{addr}}& \text{Router}\\
\text{Router} &\promise{\pm \text{addr}}& \text{Receiver}.
\eeq
\end{example}

\begin{example}[Sparse Distributed Representations]
  Sparse Distributed Representations\cite{sdm1,sdm2} use a kind of semantic
  hashing to map similar things to similar locations. Some distance
  measure is used for this, e.g. the Hamming distance of keywords. An
  interesting property of Sparse Distributed Memory is that each
  keyword is its own memory address, i.e. data and index hash are the
  same.
\end{example}

\begin{example}[Conceptual memory]
  An intelligent agent has to be able to process multiple concepts and
  draw an imaginary boundary around them in order to deem them
  related, like in a namespace. A simple machine needs to be able to
  do this based on a mechanical principle, without some
  brain-like reasoning skills presumed\footnote{We may ask: how does
    the brain itself associate concepts? Simultaneous activation is
    one of few possibilities, as explained later.}.
\end{example}

\subsection{Long range ordered spacetimes: cellular automata}

A priori oriented structures are not the only way of breaking
spacetime symmetries to distinguish memory locations.  In discrete
spacetimes, with sufficient symmetry, edge effects and interactions
can also lead to dynamical effects that form memory representations.
Regular edge geometries encode directional information.

Many of the spacetime notions that we take for granted in the
classical formalization of dynamical systems do not apply to networks
or arbitrary graphs (see paper I).  For example, translation,
velocity, momentum, derivatives, conjugate pairs, phase space
(position and momentum), waves, Fourier transforms, etc---all these
concepts derive from the existence of a smooth manifold, with a
clearly definable set of coordinates\footnote{I don't want to get into
  a discussion of differential or algebraic geometric ideas here, so I
  am being intentionally simplistic.}. In generalized graphs, where
the topology of space changes at every hop, the idea of continuity and
conservation are simply nonsense.

If so many of the classical concepts are absent or non-trivial to define,
then how can we relate the concepts of a discrete spacetime to classical
notions?
Finite state machines are the dynamical equivalent to canonical phase
space for discrete graphs.  The correspondence is 
\beq (\vec x,\vec p)
\leftrightarrow (A_i,\pi_{ij}) 
\eeq 
where $\vec x$ is a position
coordinate, $\vec p$ is a momentum vector, and $A_i$ is an agent,
$\pi_{ij}$ is the promise graph over $A_i$\footnote{A half-hearted
  attempt to relate graphs to a Hamiltonian formulation has been posed
  in \cite{qgraphity}, leaving many questions unanswered.}. Although analogous,
these constructions are in no way comparable structures, unless there is
a basic homogeneity or long range order in the spacetime adjacencies (the
long range ordering of the scalar promises on top of the graph is a
separate issue). The study of such graphs as dynamical systems (rather than
as approximations to Euclidean or Riemannian space) has only received
attention very
recently\cite{graphdynamics1,graphdynamics2,graphdynamics3}, and turns
out to be a quite non-trivial problem. To make progress, one needs to
fix a coordinate system (naming convention) for the links emanating
from each node agent (referred to as `ports' in
\cite{graphdynamics1}).  Graphs with this property are related to
Cellular Automata (CA)\cite{barricelli,wolfram1,wolfram}.

The hypothesis explored in these notes is that any stable spacetime
phenomenon can be used as a form of memory, and be attributed
semantics by context.
The behaviours of cellular automata models are essentially those of
reaction-diffusion models, and their study goes far beyond the scope of
this work, but they are related to the propagation of intent discussed
in \cite{faults}. What is important to note here is that only special
cases of CA may be used to act as memory that does not simply leak away.
\begin{example}
  Cellular automata (CA) have been used, in combination with learning
  methods, to model the evolution of land use in Geographical
  Information Systems
  (GIS)\cite{cellaurANN1,cellaurANN2,cellaurANN3,cellaurANN4,cellaurANN5}.
  By modelling each point in space as an agent (cell), with promises
  based on on iterative dependency on nearest neighbours, one achieves
  a local CA model. The challenge with CA models is in matching
  boundary conditions around the edges, in order to break the
  symmetries of the lattice. This is how one injects information about
  actual models. To accomplish this, neural network techniques of
  back-propagation have been combined with the CA models.
\end{example}
Even with a notion of propagation, it is not trivial to define
something like a momentum, related to a graph transition function (see
section \ref{graphdyn}), and the usual formulations of Poisson
brackets and Hamiltonian or action principles make little sense
without a scaling limit to a Euclidean space.
\begin{example}[Self-healing communication trees]
  The cellular connection network infrastructure, proposed by
  Borrill\cite{borrilltree}, to self-organize a self-stabilizing
  spanning tree on top of a homogeneous trusted lattice of agents with
  a fixed number of ports, is a cellular automation whose function is
  to route messages using an optimized flooding scheme and nominal addressing
  (flooding is optimized by learning of routing tables by the memory agents).  The
  operation is similar in concept to NDN, and makes use of purely
  local agent adaptation, with only emergent organization. The aim is
  not to stabilize a rigidly addressed memory, but rather to stabilize
  an adaptive spanning network across fixed nodes, each of which can
  act as memory. This is best suited for retrieving encoded memories,
  already placed at the nodes, but does not immediately support a
  natural ordering of locations for storing distributed sequences from
  an external source.  The latter can be arranged by adding a globally
  ordered naming convention, in accordance with the reduction
  law\cite{spacetime2}.
\end{example}

\subsection{Graph dynamics, thinking, and reasoning}\label{graphdyn}

Knowledge spaces of concepts, adjacent by association, go beyond memory
systems. The fact that they can propagate intent, means that can they also {\em compute} results, in the sense of executing
algorithms. As feedback networks, the stability of concepts and outcomes is thus a matter for
serious investigation. Our understanding of the dynamics of generalized graphs is still
in its infancy, but a number of special cases submit to analysis, and challenge
us to think about spacetime concepts more carefully.
\begin{example}[Causal Graph Dynamics]
  The dynamical evolution of labelled graph systems is a theme that
  can be related to a vast number of fields of study practical
  examples, from cellular automata models of land use, to biological
  tissue, to ecological systems, and so on.  Recent work, which tries to
  address the dynamics at a formal level, builds on notions of
  dynamical systems from physics\cite{graphdynamics1,graphdynamics2,graphdynamics3}, by using
  labelled graphs, without explicitly modelled semantic constraints.
  Graph vertices (analogous to promise theory agents) make connections to
  other agents by named `ports', i.e. links that are promise labels from a fixed
  alphabet of types. The semantics of the types are unspecified, and
  agents do not have to have all kinds of ports. The type alphabet
  thus acts as a matroid basis for vector promises. The model of
  causation is based on locality of interactions, i.e.  neighbours,
  within some adjacency radius. The locality is provably related to
  causality.

Causal Graph Dynamics shows that it can be a model of computation,
thus (as a constrained subset) a fuller spacetime dynamics can also
represent computation. This ought to be obvious, since computers are
made from labelled spacetime, but it is still a source of controversy
in some fields. There is also a tangible similarity with quantum field
theory.
\end{example}
\begin{example}[Differentiated spaces]
Spaces like rooms, buildings, and cities are strongly differentiated
spaces, where the forces of morphology have already reached an
equilibrium that decouples their function from the dynamics within. However, the
origin of that morphology presumably begins from a less differentiated
state somehow. Modelling the emergence of structure by exploring boundary
conditions and symmetry breaking mechanisms is also a major industry
in science.
\end{example}
\begin{example}[Cellular automata]
  Cellular automata have been studied since the pioneering work of
  Baricelli and von Neumann and
  others\cite{barricelli,vonneumann1,vonneumann2,wolfram1,wolfram}.  A
  cellular automaton is a graph that normally forms a regular degree
  graph of neighbouring cells, forming a lattice, each of which is an
  autonomous agent. Each agent has state and that state evolves
  according to a promise dependent on its neighbouring cells.  The
  dynamics are local, but the relative times of transitions of each
  neighbour are not completely defined. A normal approach is to
  consider a complete cellular automaton as a rigid spacelike
  hypersurface that evolves according to a global clock, with a
  transition function mapping all cell states $c_i(t)$ synchronously:
  \beq 
T: c_i(t+1) \rightarrow c_i(t), ~~~~~\forall i.
\eeq 
\begin{figure}[ht]
\begin{center}
\includegraphics[width=6.5cm]{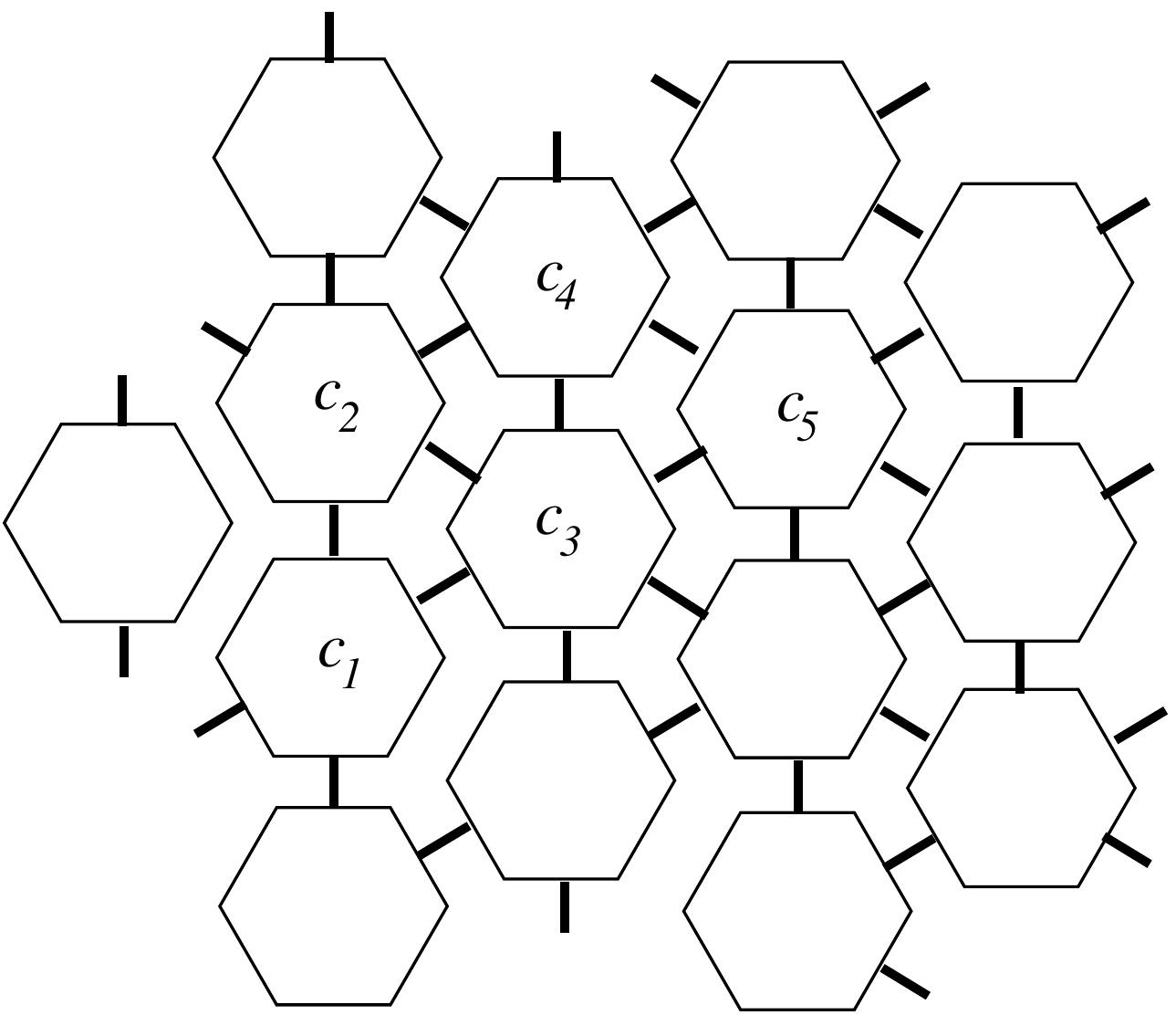}
\caption{\small Cellular automata are regular degree graphs whose cellular agents promise
to modify their state based on the states of their nearest neighbours.\label{cellaut}}
\end{center}
\end{figure}
The state at $t+1$ for all agents is thus based on the frozen snapshot
of states at time $t$, and all update synchronously.  This does not
imply infinite speed of propagation, because influence is restricted to
nearest neighbour interactions (see figure \ref{cellaut}): the velocity of influence is one hop
per time tick. However, it implies infinite speed of coordination:
i.e. long range order
in spacetime is assumed, which is phenomenonologically plausible
for small scale living systems observed by light and sound, etc.  Monte Carlo updating is also possible
to simulate quantum physical systems like the XY-spin
model\cite{burgessC9}. Due to the multi-dimensional dependency of
cellular automata state transitions, the evolutionary outcome of the system
is strongly dependent on the updating semantics of the transition function.

Cellular automata act as computers (e.g. see \cite{wolfram,certainty}), and can perform
routing functions as addressable networks (e.g. see \cite{borrilltree}). 
\end{example}

\begin{example}[Directed eigenvector centrality]
  Eigenvector centrality is a measure of a graph based on the
  observation that the graph adjacency matrix is a transition matrix,
  whose self-stable flows result in a shift of link weights onto
  nodes, through the principle eigenvector (a consequence of the
  Frobenius-Perron theorem).  This was well understood for undirected
  graphs.  A variant of this technique was used by Google search for
  its PageRank method of search ranking\cite{pagerank}.  The result
  was extended for directed graph flows in \cite{graphpaper}. For a
  directed graph, the result is somewhat different to that for a
  non-directed graph. Only absorbing sinks nodes end up with any
  weight, because transitive adjacencies funnel all the flows into
  sinks. If we apply this to a reasoning process, it would imply that
  directed learning networks would not be able to distinguish between
  reasoned outcomes by relative weight. However, a number of remedies
  were noted in \cite{graphpaper}, including the equivalent of
  back-propagation, and feedback by pumping results back into the
  sensory input end of a network. This link between search ranking and
  reasoning illustrates how search ranking acts as a simple kind of
  Bayesian network. The technique is particularly interesting as it
  leads to stable regions\cite{archipelago} that could naturally be associated with
  concepts, based purely on spacetime connectivity, and thus it
  offers insight into the way concepts may emerge without deliberate
supervised training in a true artificial intelligence.
\end{example}

Much remains to be understood about the behaviours of systems on which
we are relying for artificial knowledge representation and reasoning.
It is tempting to think that, because certain applications approximate
deterministic behaviour, when driven by a dominant directional
process, like input sensing, all behaviours must be therefore be
deterministic. That is far from the case. Indeed, even in the case of
humans, we know that when external input is switched off, the brain
enters states of hallucination and dreaming, where reason seems to
follow quite different rules.
Several questions stand out:
\begin{itemize}
\item How do graph dynamics play a role in knowledge representation?
\item Is knowledge storage and recovery stable to the dynamical states of a knowledge system?
\item Can all concepts be observed (read or written) simultaneously, or are is it possible to
excite partial eigenvectors that are not simultaneously compatible, as in the case quantum mechanical
canonical pairs\cite{weinberg2012lectures}?
\item What role do pointed graphs\cite{graphdynamics3} and spanning trees play in bounding
the determinism of reasoning?
\item How does parallelism affect the linear nature of sequential reasoning?
\end{itemize}
We return to the matter of reasoning in section \ref{reasoning}.

\subsection{Agent phase and conditions for memory, smart behaviour}

Agents may be in a free gaseous state, as atomic or molecular agents, or bound together
into a solid lattice with limited long range order (see paper I\cite{spacetime1}).
\begin{itemize}
\item In a gaseous phase, agents mix freely and encounter one another
  by random walk, with short-lived adjacencies. This is a useful way
  of mixing and discovering new information. Coordinates in a gas can
  only be through arbitrary numbering and message broadcast. Think of
  people milling around the lobby of a hotel, meeting by chance, or
  calling out ``Come in number 9, your time is up!''.  Extended
  structures like strings, stories, paths, cannot be remembered in a
  gaseous phase, because they have no stable or persistent relationships.

\item In a solid phase, information is located in a very regular, ordered
 way. Adjacency is a
lasting {\em knowable} relationship. One can form a regular coordinate system
with purely local predictability. Search is not required.
Think of visiting a specific tenant in a numbered hotel room.
\end{itemize}
The transfer of intent (including structure) requires both (+) and (-) promises
to propagate, mutual bindings or tensor promises are involved in structuring
of data.
\begin{lemma}[Sequential memory implies vector promises]
To promise stable sequences, agents must bind with vector promises.
\end{lemma}
This was discussed under temporal memory.  To the extent that a solid
implies a stable ordered sequence of vector bonded agents, without
necessarily having significant long range order, then memory requires
a kind of permanent solid state.  Given this, there may be something
to learn from cellular automata about how a smart knowledge
representation (i.e. `mind') can remain sufficiently plastic to
change.

\begin{example}[Immunology and neurology]
  Memory structures and pattern recognition discriminators are often
  constructed from solid state regular lattices, that provide regular
  calibration grids for observation. The brain cortex, and other
  related memory structures are form a largely solid state network,
  albeit irregular on a small scale.  The immune system, by contrast,
  is capable of learning and rememberings previous infections based on
  populations of antigen carrying lymphocytes (B and T cells). In
  order to accomplish this, the individual agents must be able to
  promise greater capabilities as individuals. Thus, in a gaseous
  phase, they substitute scalar promises for vector and tensor
  linkage. For a review, see \cite{perelson1,lisa98283}.
\end{example}

\subsection{Summary}

Information may be encoded as memory by using the familiar spacetime
properties of spacetime:
\begin{itemize}
\item If a spacetime is deterministically addressable, it can store patterns of
  information in a retrievable manner.

\item Separation of scales enables scale-free addressability.

\item Consistent addressability is possible if memory agents are
  distinguishable, and there is end-to-end propagation of data between
  the agents in two directions between readers/writers, and memory
  agents.

\item Four encoding methods may be matched with mechanisms for propagation of intent:
\begin{itemize}
\item Containment and aggregation.
\item Causal ordering.
\item Composition or cooperation of parts.
\item Proximity or topological nearness.
\end{itemize}
\end{itemize}
These elementary considerations show that any semantic spacetime 
can perform a learning function, and can therefore acquire information,
and store it, leading to a knowledge representation. We can now study
the structures of tokenized knowledge in detail.

\section{The structure and semantics of knowledge} \label{knowledgerep}

The hypothesis, which forms the basis for this work, is that knowledge
is the sum of an observer's tokenized sensory inputs, iteratively
confirmed, over multiple observations, and laced with a feedback that
is integrated into a running assessment of `here and now'. This
process leads to a unique world-view, for each observer, resulting
from each agent's unique world line history.  The non-linear feedback
amplifies the fact that experience is non-deterministic, but if
knowledge emerges around stable attractors, it may be cached in a
compressed, tokenized form.  This is essentially like a linguistic
representation of experience\footnote{This might help to explain why
  we experience such a close connection between thinking (in the sense
  of storytelling and reasoning) and language.}.

Constructing a plausible model for this is the subject of this
section.  Given the network complexity of associations formed during
full-sensory learning, it is unrealistic to expect tokenized concepts
and associations, committed to memory, to be complete and faithful
representations of the input.  It would be a misunderstanding of
knowledge systems to expect any such thing.  A symbolic knowledge
representation can only be a tokenized {\em approximation} of the
fruits of learning\footnote{In principle, the same level of
  approximation need not be a limitation of an artificial knowledge
  systems, with simple sensors, but the economics of learning point to
  a diverging need for resources, and hence the eventual importance of
  approximation.}.  The tokenized form is not a substitute for the
actual experience, only a summary of its interpretation.  The
implication is that we cannot pass on complete experience in a
tokenized (linguistic) form, because it enters a system through
different sensors and touches fewer concurrent sensations.

\subsection{Learning, familiarity, and linguistic tokens}

The implications of learning are profound. They suggest that deep
knowledge cannot easily be imposed onto an agent from the `top-down',
in symbolic form. In other words, one cannot induce knowledge by a
short-cut, like reading or listening to a lecture. One may try to
simulate a complex experience, in a linearized stream of
communication, just as a Turing machine can simulate any information
based process eventually, assuming perfect fidelity, but it cannot be
done efficiently or with a predictable outcome, especially under the shadow of
approximation, because the process of learning is iterative and
non-linear\footnote{Promises of knowledge, made by other agents, can
  be assessed by a receiver and either accepted as trusted facts or
  merely as data. When concepts and semantics are taken as
  authoritative standards, they change from being voluntary emergent
  promises to impositions, and hence they lose their ability to avoid
  inconsistency and be represent local circumstances.  Standards may
  be useful as scoped definitions, but their validity is fragile to
  variation in the environment.}. The implications of this are not
well understood, and yet we work hard to represent knowledge in a
variety of media, and design `artificial intelligence' that has its
own approaches to learning.

\begin{example}[Is a book knowledge?]
  Is knowledge the same as a representation or facsimile of knowledge?
  Do we possess knowledge simply by owning a book?  Is the process of
  knowing the content of a book about revisiting it to gain
  familiarity? There is a clear temporal aspect to knowledge that we
  need to address.  The taking of a photographic snapshot of a space
  (a time-slice) may capture information, but we do not `know' that
  information by simply capturing the image: knowledge comes from
  revisiting the snapshot, relating it to other familiar concepts, to
  the point of familiarity. Thus even an approximate representation
  does not become knowledge simply by being disentangled from its
  network and encoded into a form that can be passed on.
\end{example}
To understand the implications of these remarks, in as broad a range of
scenarios as possible, we need to get to grips with how knowledge is
valuable to different kinds of processes, without preconceived
notions!  Knowledge is not just that which is familiar, but also
descriptive of context.

\subsubsection{Can familiarity be passed on?}

Familiarity is a major part of our understanding of knowledge.
\begin{definition}[Familiarity and significance of knowledge]
  The attribution of a weight or score that grows with frequency of
  use or accumulated usage, in a memory network.
\end{definition}
The non-linearity of learning and memory networks leads to a simple point:
\begin{lemma}[Familiarity cannot be propagated]
  Familiarity of a learning network cannot be transmitted faithfully
  through a serial (linguistic) representation.
\end{lemma}
This observation follows from the sensitivity to boundary conditions
in non-linear systems, and the relativity of familiarity 
with each concept to neighbouring concepts, in spite of intentionally deterministic
procedures\cite{chaosbook}. No two agents can be prepared and
maintained in exactly the same state for an extended process of
simulated learning.  They would have need to have to begin with
identical initial conditions, and maintain identical external
conditions throughout the process of information acquisition.  It
could not be identical to the knowledge acquired by experience,
because a serial stream cannot be processed in the same way as a
parallel sensory experience, simultaneously, this is impossible (on
many levels).

The network aspect of knowledge is crucial.  Knowledge is not
isolated: it has lasting bonds to notions of {\em context}.  If lasting
connection were enough for represent knowledge, any solid object (a
building, a vase, a book) could be used as a representation of
knowledge. Any specific arrangement of space, persistent in time with
sufficient stability, or thing rendered timeless by isolation from
change, could encode what has been accumulated, and could be
transmuted into familiar knowledge by repeated interaction.

\subsubsection{Tokenized knowledge: topic maps and semantic indices}

Knowledge management has been studied for many years in the context of
library systems and of the semantic web.  One of the early
technologies for representing concepts formally was the idea of Topic
Maps (for a review, see \cite{topicmaps}). This emerged from the need
for document classification and indexing.  An approximate rendition of
the models is as follows:
\begin{itemize}
\item The world consists of subjects (things, people, ideas, etc) about which we can say something.

\item These subjects map to `topics' in a software data model.

\item Our linguistic understanding of subjects is glued together with
  associations we make.  These are descriptive relationships between
  linguistic (nominal) data.

\item Associations are made by individuals, making topic maps
  subjective viewpoints. They can be merged to mimic learning from
  one individual to another, or a blending of experiences.
\end{itemize}
The net result of a topic map is a graph structure whose nodes are
topics and whose links are associations. Additional structure is given
to type concepts, which are meant to model context. The information
model is simple to parse and to search, in the manner of an index.

The strength of Topic Maps lies in their autonomous, subjective
viewpoints, based on keywords. This further allows topic maps to be
merged without conflict (the language is promise oriented).  However,
their weakness stems from a model that is deliberately abstract.
Nonetheless, it supplied much of the inspiration for the present
work\footnote{Topic maps probably suffered from premature
  standardization, an artificially taxonomic approach to
  classification ontology (to fit into a data model technology based
  on typing), and closeness to a particular imagined use-case; thus
  the scaling of the model was never considered at all.}.  By
re-examining the ideas without those constraints, using an independent
framework of semantic spaces, we end up with a more detailed model
that retains the simplicity of the old at a particular scale. A
critique of topic map knowledge representation was given in
\cite{burgesskm}.

The idea that we can take a complex set of observed attributes, from a
mixture of sensory inputs and pre-learned concepts, and project them
into a simplified `symbolic' representation is profound; it is the key
to `approximation', without which it would be likely impossible to
bound the resources associated with reasoning (see section
\ref{storysec}). Tokenization solves three issues:
\begin{itemize}
\item Reduction of the dimensionality of sensory data.
\item Compression of sensory information to a singular alphabetic form.
\item Mapping variable inputs to an invariant representation.
\end{itemize}

\subsubsection{A minimal observational ontology for knowledge, based on promise theory}\label{concepts}

Starting with agents and promises as a simple axiomatic framework, we
can try to formulate a minimal set of agencies< and promises that would
be needed to represent knowledge.  In order for concepts to be
encoded, there must exist some form of semantic space. The precise
nature of the space may limit what knowledge it can learn and encode.
Based on a scaled view of spatial agency, and the axiom that knowledge
is acquired independently by each observer, based on a context of its
temporal interactions with exterior agents, we can now develop some
minimal number of structures for knowledge representation, by mapping
the parts systematically to agents and promises.
\begin{figure}[ht]
\begin{center}
\includegraphics[width=8.5cm]{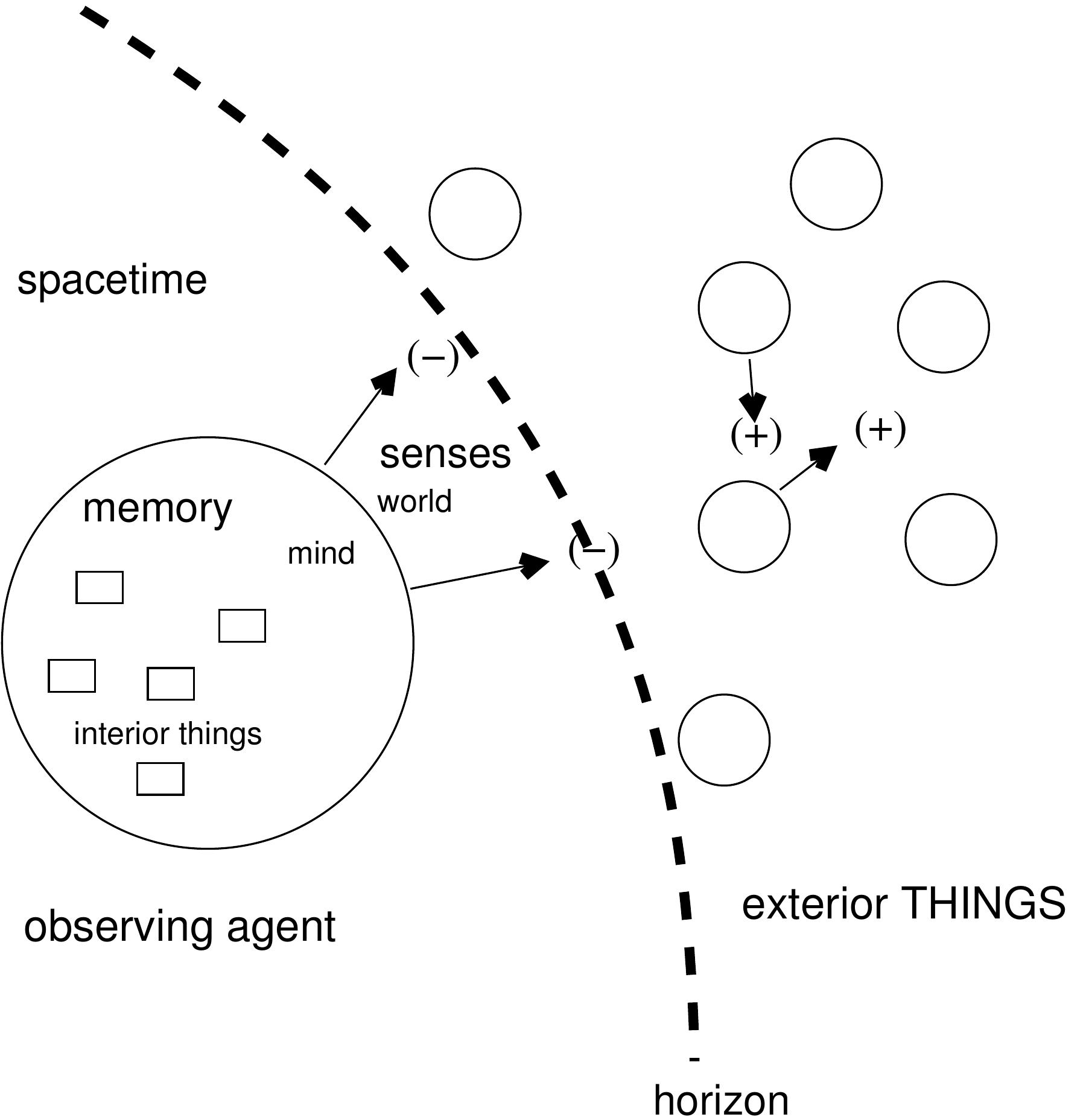}
\caption{\small Knowledge representation is based on the viewpoint of
  an observer agent, assessing the sensory promises made by the agents
  in its exterior, and encoding its understanding in a system of
  interior agents making `knowledge promises'. These knowledge
  promises map to `associations' in semantic memory.\label{mindagent}}
\end{center}
\end{figure}

\begin{example}[What happens in space stays in space]
Is every familiar solid object something that can represent
knowledge?  Certainly, why not. However, knowledge tokens are only
isolated characterizations: a kind of knowledge, but not a faithful copy of the
process that made them.  They are discretized tokens that
have been isolated from their origins, i.e. abstracted representations.
The fidelity of knowledge is not guaranteed.
\end{example}

\subsection{Labelling and naming conventions: the observer's view of the world}

With reference to figures \ref{mindagent} and the original figure
\ref{memorysystem}, we begin by with a model of a single agent
observing the exterior by assessing exterior agents through a number
of sensors. Inside the observer (within its superagent interior) are
conceptual agents, which we may refer to as its `mind', of unspecified
nature, used to equip the superagent observer with memory and
knowledge representations.  The primary distinction is thus the
sensory horizon between interior and exterior agencies.

\subsubsection{Interior and exterior spaces of an observer}

Distinguishing the exterior or physical world of an agent (that which
can be sensed) from its internal representations\footnote{Art is the
  lie that enables us to realize the truth, said Picasso.} is
the first step in discussion knowledge.
\begin{itemize}
\item {\bf Exterior} `world' agents promise varying levels of transparency
  and predictability.  The host agent's understanding of the world
  comes from interacting with and assessing these, if it can.

\item {\bf Interior} mind agents: memory agents that record a semantic
  imprint in the form of concepts. Each interior represents a single
  observer viewpoint, and an independent knowledge representation.

\item {\bf Interface sensors} promise to observe and assess that which has been
promised observable.
\end{itemize}
We need not be more specific about how the representations work until
a particular realization is needed. 

\subsubsection{Interior and exterior agents}

I use caps for exterior agents and lower case for interior representational agents, thus we may define:
\begin{definition}[Exterior {\sc thing}]
An agent that makes promises in the exterior world of a specific agent, which observes.
Whatever the nature of the {\sc thing}, it is the promises it makes that
make it observable.
\end{definition}
This maps to an observer's characterization and concept:
\begin{definition}[Interior thing]
The interior representation of something promised on the exterior
of an observing agent, i.e. a full or partial concept.
\end{definition}

The agent structures we need include the following\footnote{I hope the
following definitions suffice as plausible to most readers (without
trying to appease other nomenclature in the literature).}.

\begin{definition}[Exterior (real) agents]
Agents that can be distinguished by any observer in the exterior
world. These are not interpretations, but structural spacetime identifications.
\begin{center}
\begin{tabular}{|l||l|}
\hline
exterior & physical source agent\\
\hline
\hline
\sc Thing   & Exterior agent\\
\sc (+) Promise & Information supplied by a thing agent.\\
\sc Connection & a physical adjacency between things.\\
\sc Pattern & A geometric configuration of agents.\\
\sc Path  & A one-dimensional ordered pattern of \sc things.\\
\hline
\end{tabular}
\end{center}
\end{definition}
These basic exterior agencies require a memory representation within the observer.
The interior representations are one to one, but the representations have the 
ability to make different kinds of promises:
\begin{definition}[Interior (representational) agents]
Internal structural agencies needed to map {\sc things} in
the exterior world to interior `thing' representations.
\begin{center}
\begin{tabular}{|l||l|}
\hline
interior & mental representation receiving agent \\
\hline
\hline
 Thing   & An internal representation of a \sc thing.\\
 Concept   & A scaled aggregate `thing'.\\
 (-) Promise & Information accepted from the {\sc thing} to condition the internal thing.\\
 Connection (association) & An adjacency between interior things, with interpretation added.\\
 Pattern & A geometric configuration of agents.\\
 Path  & A one-dimensional ordered pattern of \sc things.\\
\hline
\end{tabular}
\end{center}
\end{definition}

\subsubsection{The named roles in knowledge representation}

The basic roles for categorization may be formed from the irreducible association types, by alias.
\begin{definition}[Role promises by interior agents]
  Roles are promise patterns, common to a variety of agents. These
  translate into concepts in an interior representation.  A subset of
  knowledge structural relationships promised by interior agents. Some
  examples:
\begin{center}
\begin{tabular}{|ll||ll|}
\hline
interior agent promise & mental promises\\
\hline
\hline
image / representation (default)  & internal representation of a thing\\
location & a thing that represents a spatial position\\
event & a thing that represents a spacetime position\\
concept & a generalizing label representing common promises\\
pattern & a geometric configuration of agents/promises\\
association & a promise relating subjects \\
story  & a linear collection of agents ordered by associations\\
topic / theme   & image of a Thing or Object\\
subject /theme & image of a Thing or Object\\
model   & a superagent promise of complex spacetime representation\\
categories &a generalization of several concepts\\
class  & " \\
idea   & a selection of things taken together\\
\hline
\end{tabular}
\end{center}
\end{definition}
Note that, by contrast with Topic Maps, I treat the term `subject' as an interior `mental'
agent, rather than a physical agent.
Associations between the interior concepts are explicit {\em vector} promises.
Any scalar attribute can be made into a generic vector promise by
introducing a central matroid agent which represents the named
attribute and serves as a membership hub for all agents that possess
the attribute.

These conventions follow from the basic promise theory representation of types
(see paper II, section 2.6) as agents that make specific behavioural
promises, i.e. what we call roles. Thus types are simply aliases for
roles.  A basis agent `thing' can thus play many roles in knowledge
representation without being a fundamentally different type of object:
roles, concepts, exemplars, generalizations, classes, categories,
models, ideas, etc are all roles played by thing representations
rather than different fundamental types of thing/{\sc thing}. 
There are relatively few agent roles in the exterior world, because there are fewer
ways to make feedbacks and links to form more complex representations. In the
interior (what we can refer to as `mind'), there are
many more possibilities, unconstrained by reality. Imaginations and
analogies that result from feedback and mixing of patterns provide the richer
menagerie of forms that we use to describe knowledge.

\subsubsection{The observer's clock, and sense of time}

When an agent samples its sensory stream, the formation of a context
from the samples sets a timescale associated with the input $T_\text{sensor}$.
\begin{figure}[ht]
\begin{center}
\includegraphics[width=10.5cm]{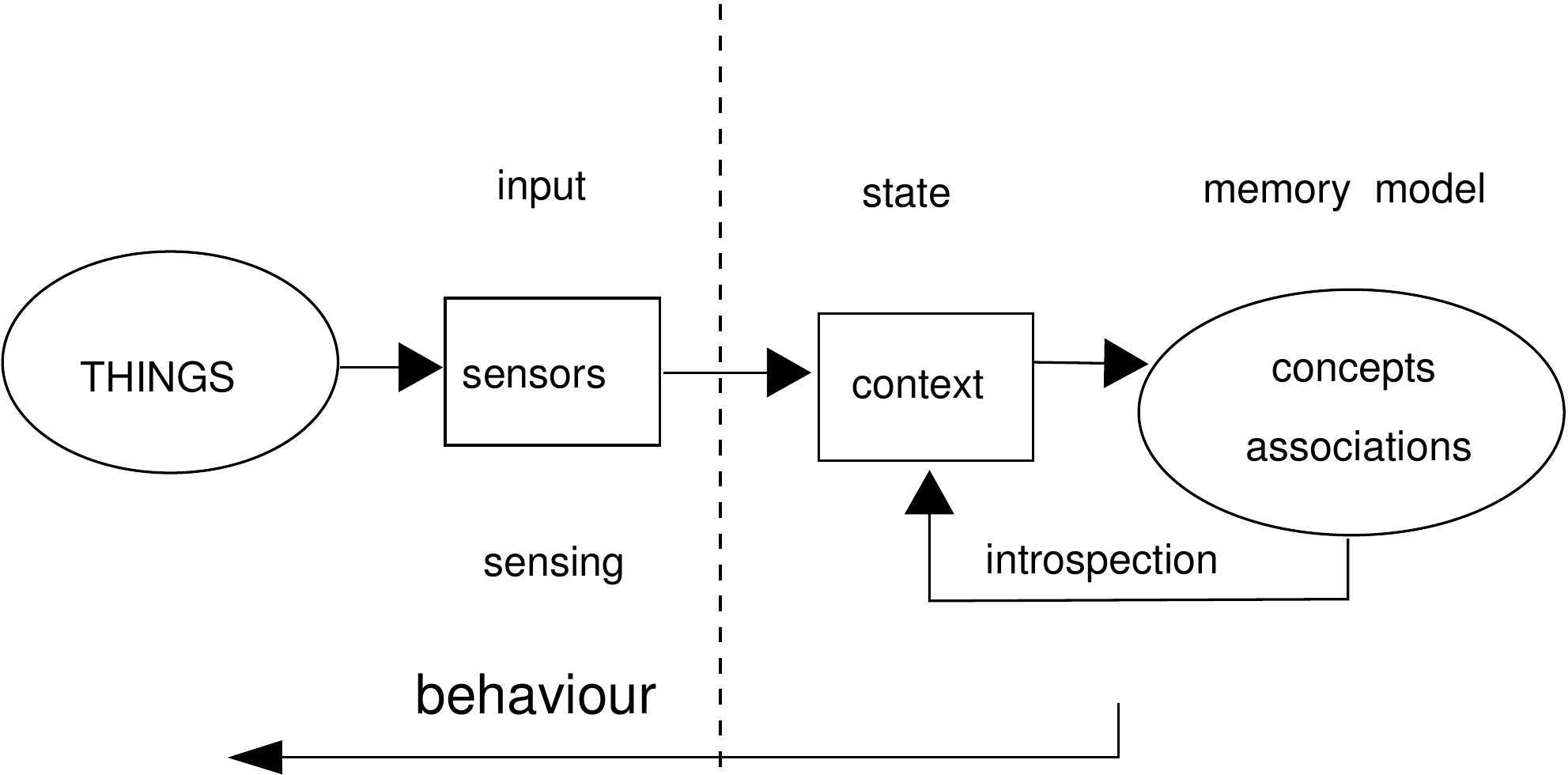}
\caption{\small The functional roles involved in continual
  transmutation of observation into state and then concepts (memory
  model), from left to right. Notice that agents' promises (not the
  agents per se), on the left, determine what later agents, on the
  right, can learn and know. Memory feedback, from the right, can also
  cycle back around, and be placed on a par with direct observation.
  We could speculatively refer to memory feedback, without new input, as
  dreaming.\label{awareness}}
\end{center}
\end{figure}
This is the clock by which observations are {\em accepted} (not
the rate at which impulses are necessarily given). It determines an agent's
ability to receive information and pattern it into a sense of `now'.
\begin{lemma}[An agent's internal observational clock completely determines its sense of time]
  Let $A$ be an agent promised a steam of data that arrive with an
  average interval of $T_+$ units, and are accepted every $T_-$ units. Thus
the expectation time to receive a sensory signal is greater than or equal to $T_-$.
\end{lemma}
The proof follows from agent autonomy: no change can occur in the
agent unless it accepts the external data or modifies its own internal
data by its own mechanisms. This is what is meant by its internal clock.

\subsection{Concepts, interior structure, and semantic superagents}

The obvious function of concepts is to aggregate observed {\sc things}
into summarizable tokens, labelled by the context in which they were
observed.  Concepts begin as representations of external spacetime phenomena, 
in the internal spacetime of an observer.
However, one there is a sufficient library of concepts, new ones 
can also be invented {\em ad hoc} from within, by mixing and distorting
existing ones.  If an agent forms a `new' idea, this can be learnt and
remembered, as a separate internal concept, with the iterative learning
replaced  by repeatedly rehearsing the idea, simply by thinking a lot about it.
\begin{definition}[Concept]
  An agent, or superagent, within the internal semantic memory space of an observer, 
  which promises a representation of some identifying characteristics,
  associated with imaginable or externally observed phenomena, and addressable by a
  locally unique name, for each context in which it arises.
\end{definition}
This slightly involved definition allows for concepts to be observed directly,
modified internally, or even invented randomly.
Concepts are distinguished from one another, by their agent boundaries.
Being discrete, they can be thought about
separately, thus they are discrete semantic units. However, the fact that they can
be thought about independently of sensory input
means they can also be invented without observation.
Their physical encoding in spacetime may overlap with other concepts,
depending on the scale and composition of their representation.  Concepts need not
be mutually exclusive: they may overlap with one another. Their contextual
(conditional) nature means that they are unlikely to be completely
reducible to unique, mutually exclusive semantic sets (like a
periodic table of atomic elements); rather, they will be like a rich 
and varied chemistry of nuanced compounds.

New concepts are created all the time in our human minds, and it makes
sense to extend this process, by analogy, to recognize the possibility
for ideation in any kind of learning system. The graphical
representation of concepts then takes the form of patchworks, i.e.
non-orthogonal networks of connections, linking to matroidal basis
members. This structure reflects how the concepts coalesce from rough,
noisy environmental interactions over a range of contexts.  Concepts
(including synonyms, aliases, etc) might well reach into a postulated
`space of all notions' pervasively, in some regions, and barely cover
at all in others. Concepts may also overlap by sharing crucial 
sensory ingredients.

\begin{figure}[ht]
\begin{center}
\includegraphics[width=8.5cm]{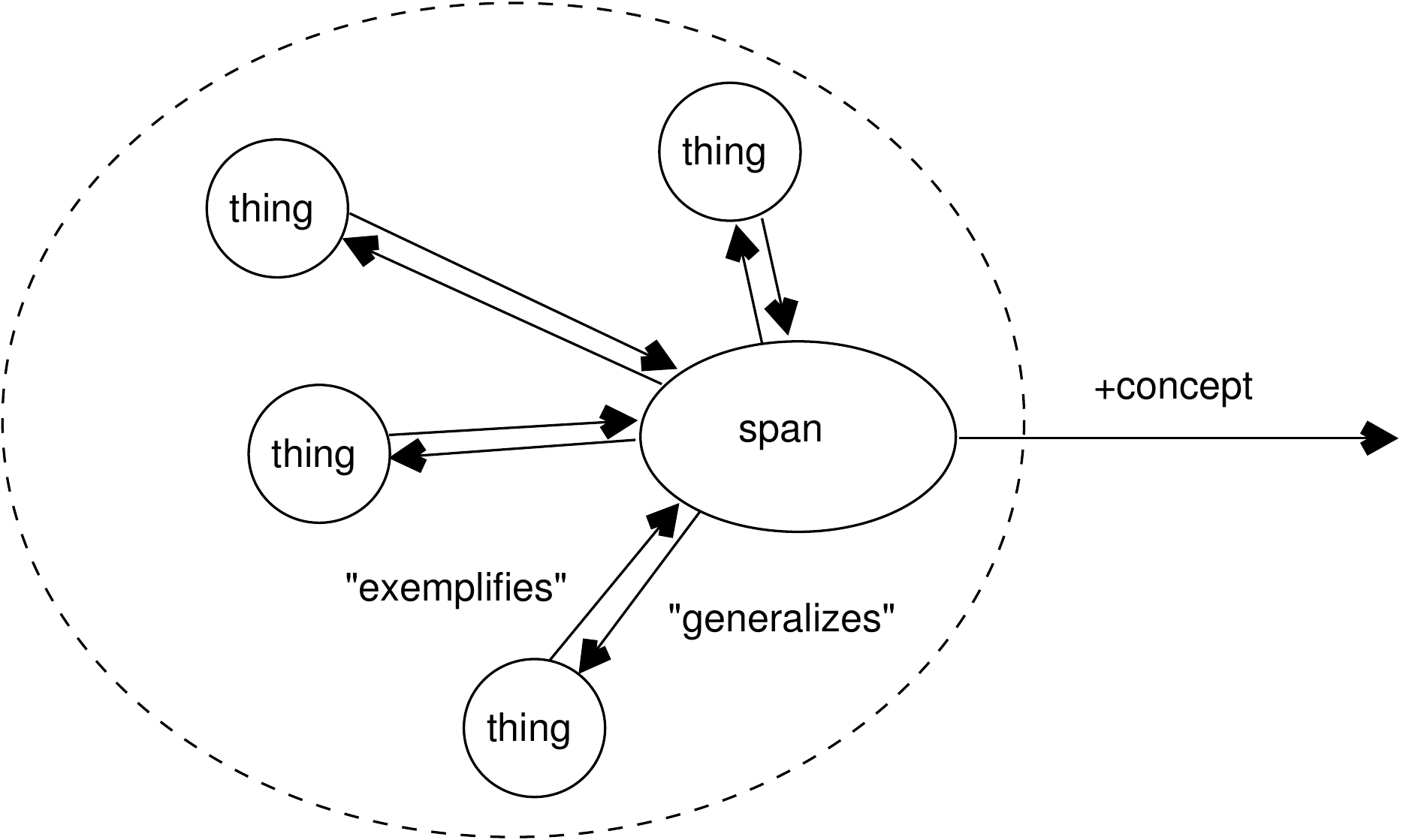}
\caption{\small Concepts promise a basic set of characteristics that generalize
the things that play the role of its exemplars. Note the spacetime structure
is a very familiar one of aggregation and scaled representation (a scaling transformation).\label{ovelap2}}
\end{center}
\end{figure}

\subsubsection{Encoding or realization of concepts (reification)}

\begin{figure}[ht]
\begin{center}
\includegraphics[width=6.5cm]{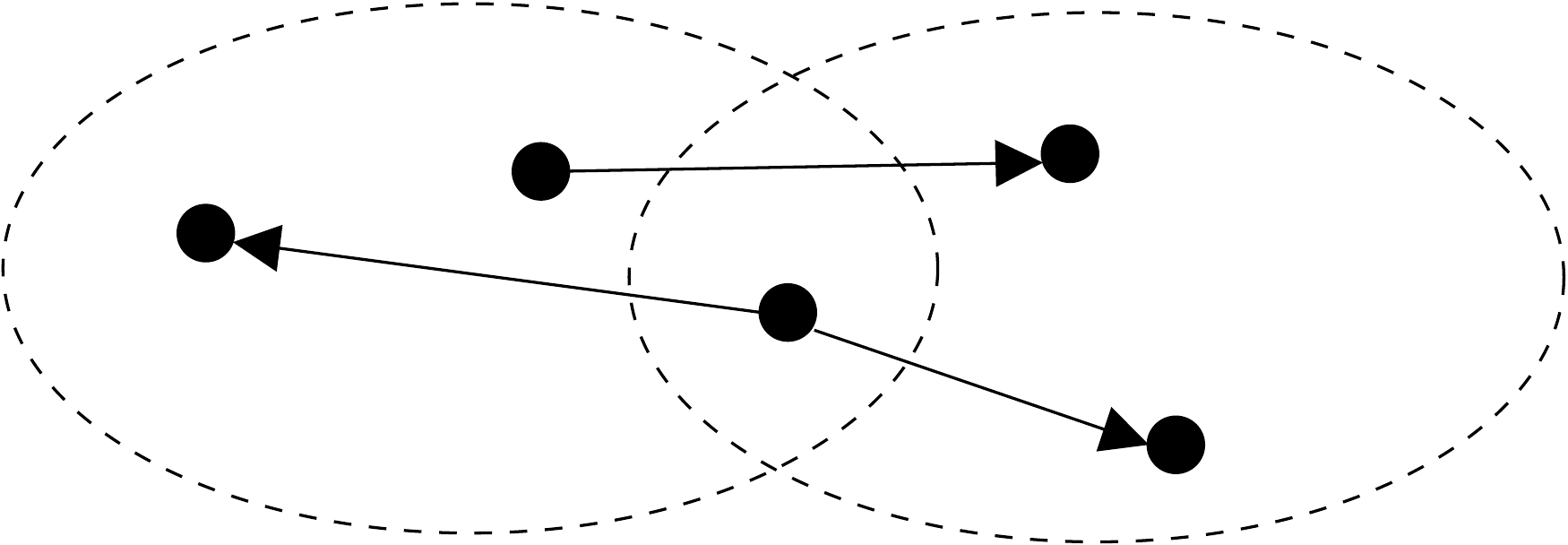}
\caption{\small The overlap of two conceptual regions might involve agents
who represent common concepts, but might only include their promises.\label{ovelap}}
\end{center}
\end{figure}

To represent, encode, or `reify' (in knowledge jargon) concepts in a
knowledge representation, we identify the available agents, belonging
to the representational medium, at the representational scale (see
figure \ref{ovelap}). It seems natural to assume that any concept may
be a superagent, formed from collaborative subagents, in the general
case. This means that concepts may overlap in two ways: through the
presence of shared interior subagents that belong to more than one
concept, and through the repetition of exterior promises made between
concepts (see figure \ref{ovelap}).
\begin{example}[Concept representation medium]
  Concepts may be attributed to memory configurations written on any 
  semantic space or medium, e.g. a crystal lattice, a collection of
  proteins, an agricultural planting pattern, the layout of a town or
  city, etc. 
\end{example}
 It may be  assumed that there is a ground state for the conceptual fabric, 
 i.e. that there exist basic building blocks for concept agents, in the form of
  discrete semantic units. All concepts would be broken down into these, but
  there may be no upper limit to their combination into superagents, allowing
  concepts of any aggregate size.
  This ground state set must be finite, but might be infinitely extensible. Limits
  on the size of conceptual agent clusters would likely be economic in nature.
\begin{definition}[Continuity of concepts]
  An overlapping collection of clusters, composed from base
  agents, and linked by association, 
  bridges clusters into a single continuous cluster.
\end{definition}
From these simple speculative proposals, grounded in graph theoretic
parsimony, we arrive at a natural picture of concepts.  Concept space
is continuous, in the sense that it contains the overlap of open
collections of clustered basis agents, but concepts themselves are more or less discretely
identifiable, with the caveat that it might be difficult to decide exactly where
the boundaries between them lie in general.
\begin{example}[Cognitive linguistics]
  In the field of cognitive linguistics, concepts are treated as
  `semantic units', (i.e. signal agents in PT parlance) that exist
  independently of language\cite{langacker1}. Although the concepts of
  cognitive linguistics are not well developed in a formal sense (even
  by their own criteria), they seem to be broadly compatible with a
  promise theory view in which concepts act as spanning sets for words
  in contexts.  This is the view I take here, based on experiences
  with systems of agents in CFEngine\cite{burgessC1}, etc.  Langacker
  is vague on what constitutes the spanning sets, but indicates that
  the following general containers are involved in classifying
  concepts: space, size, shape, orientation, function (container,
  force, motion, role), material nature or composition, and also
  economic issues like cost.  These regions are called `spaces', and
  have a close resemblance to the notion of
  namespaces\cite{spacetime2} as containers of concepts, analogous to
  directories of files on a computer.  In Langacker's view.  metaphors
  like in the associative overlap between conceptual
  classifications\footnote{Langacker riles against the notion that
    concepts must be discrete structures, presumably in protest
    against the discrete mutually exclusive treelike structures of generative
    linguistics\cite{generativesemantics}, suggesting instead that
    concepts might overlap; however, there is no contradiction between
    discreteness of superagency and the possibility of overlap.} .

  There is a superficial similarity in structure between Cognitive
  Grammar\cite{langacker1} and the spacetime view taken in this note
  series. It is plausible that this spacetime view could provide a
  more rigorous notion of conceptualization that is compatible with
  it.
\end{example}

\subsubsection{Discreteness of concepts}\label{zzz}

Some properties of concept agent representations:
\begin{itemize}
\item The discreteness of concepts does not imply discreteness of
  their semantics, and vice versa.

\item Concepts need not be mutually exclusive in their degrees of
  freedom (orthogonal with respect to a basis set, as described in
  paper I): just as vectors may be partially independent without
  actually being totally orthogonal (e.g. East and North-East are
  linearly independent, but not orthogonal). So we can think of concepts
  as tensors in the promise matroids of knowledge representation space.
\end{itemize}
\begin{example}[Non-orthogonal concepts]
The concept of blue is independent of the concepts of colour or
turquoise, but it is not completely orthogonal or unrelated to them.
\end{example}

\begin{example}[Artificial learning networks]
  Artificial Neural Networks (ANN) can be used to learn concepts from
  a curated input stream. Due to the nature of ANNs, concepts are
  not represented by any single node, but rather
  as virtual knowledge space agents that are distributed
  throughout links between aggregated basis agents in a
  distributed and overlapping way. This contrasts with a semantic
  network, e.g. Topic Maps (TM), where concepts are pre-curated
  without learning.  In ANN, concepts are decentralized in their
  representation in order to promote fuzzy matching (as a non-linear
  function of a sum over weighted bases from multiple network layers,
  with $i,j$ running schematically over layers and members); in TM concepts are
  completely centralized in single graph nodes (with matroid basis
  vector $\hat e_i$ and $i$ running over the graph nodes) and have no
  representation that can be matched to data: \beq
  \text{concept} &\stackrel{\text{ANN}}{\simeq}& V\left(\sum_{ij} w_{ij}\hat e_{ij}\right)\\
  \text{concept} &\stackrel{\text{TM}}{\simeq}& \hat e_i.  \eeq
Thus TM has simplicity and efficiency of representation, while ANN
has generality and a basis in sensory learning.
\end{example}

Real world objects are composed of discrete elementary properties
combined into compounds (like the table of atomic elements, or the
19th century preoccupation with taxonomies), while the conceptual
models need not be limited to that. An appeal to topology allows us to
formulate concepts as patches, which (together) form a covering of a
space of basis parts.  This shows that, even without such a
fundamental `table of elemental concepts', it is possible for a finite
number of agencies to encode properties to span an effectively
infinite realm of possible multiset combinations.

\subsubsection{Role of context in discretization}
Contextualization of concepts further allow single concepts to be represented
as a set of agents, distributed over the combinatoric contexts.
The encoding of concepts in the neuronal space of the human brain is one of the
mysterious but highly relevant examples of encoding of concepts.
\begin{example}
  Recent studies of brain activity using MRI\cite{semspc} to scan the
  brains of listeners during English language readings showed how
  brain activations were grouped into to broadly cohesive cortical
  regions\cite{semspc}, with similarities even across different
  subjects.  The authors posited that English words were mapped by
  brains into broad clustered meanings that they labelled with the
  following categories: visual, tactile, outdoors, place, time,
  social, mental, person, violence, body part, number.  In current
  parlance, these could be also called spacetime, state, self, senses,
  relationships, and danger. The status of these studies is currently
  under scrutiny, at the time of writing, due to the discovery of
  software bugs going back several years, that may be showing false
  positive results.
\end{example}

The aggregation of observables in this example suggests how apparent
`types' or categories of knowledge emerge, which could, in turn, be
identified with tokenized context states (see later discussion around
figure \ref{contexts}). Context-delineated aspects of a concept
representation are `roles' in the promise theoretic sense.  Since
these are stable under repetition, by the assumption that context is
stable, they must be shaped, relative to a network spanning set, by
the same semantic discriminator that fashions context, working in a
predictable manner.

A closer look at context is thus warranted, based on the foregoing
remarks, and a rephrasing of the definition of context may also be posited:
\begin{definition}[Context (version 2)]
  The accumulation of state, over the recent history of an observer, such as to
form the basis for the semantics of long term observations.
\end{definition}
This rough description will be the model of context for explaining how
tokenized names themselves become concepts.  From this starting point,
we can obtain trains of thought, patterns of usage, and so on, just by
linking agents together.  Because context linkage plays the same role
as matroid representation, context effectively becomes a part of the
semantic coordinate system for observational memory.

\subsubsection{How are concepts acquired?}\label{necmin}

Concepts are memory agents (and clusters of agents, linked by
association) in knowledge space; that is, they are interior agents,
within the interior of an observer horizon.  The promises they can
make are about being associated with one another in different ways,
their strength of association, freshness, etc (note the difference
between a promise and an association). A concept may be promised
internally (i.e. constructed) as a mirror of a promised relationship
in the exterior `objective' world of the observer; alternatively, it
may be promised entirely on the basis of interior patterns, by
introspective thinking about tokenized patterns and their perceived
similarities.
\begin{figure}[ht]
\begin{center}
\includegraphics[width=6.5cm]{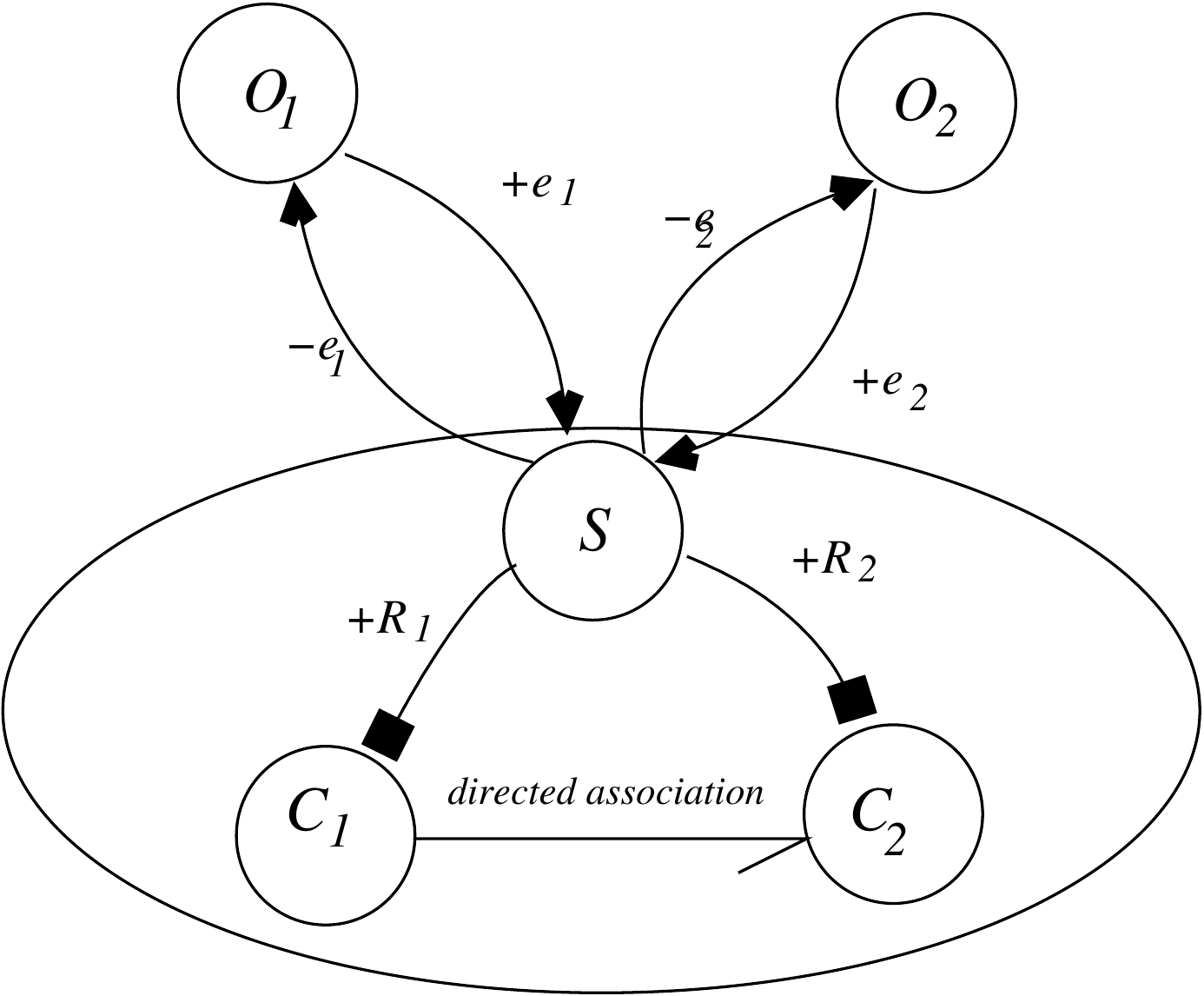}~~~~
\includegraphics[width=6.5cm]{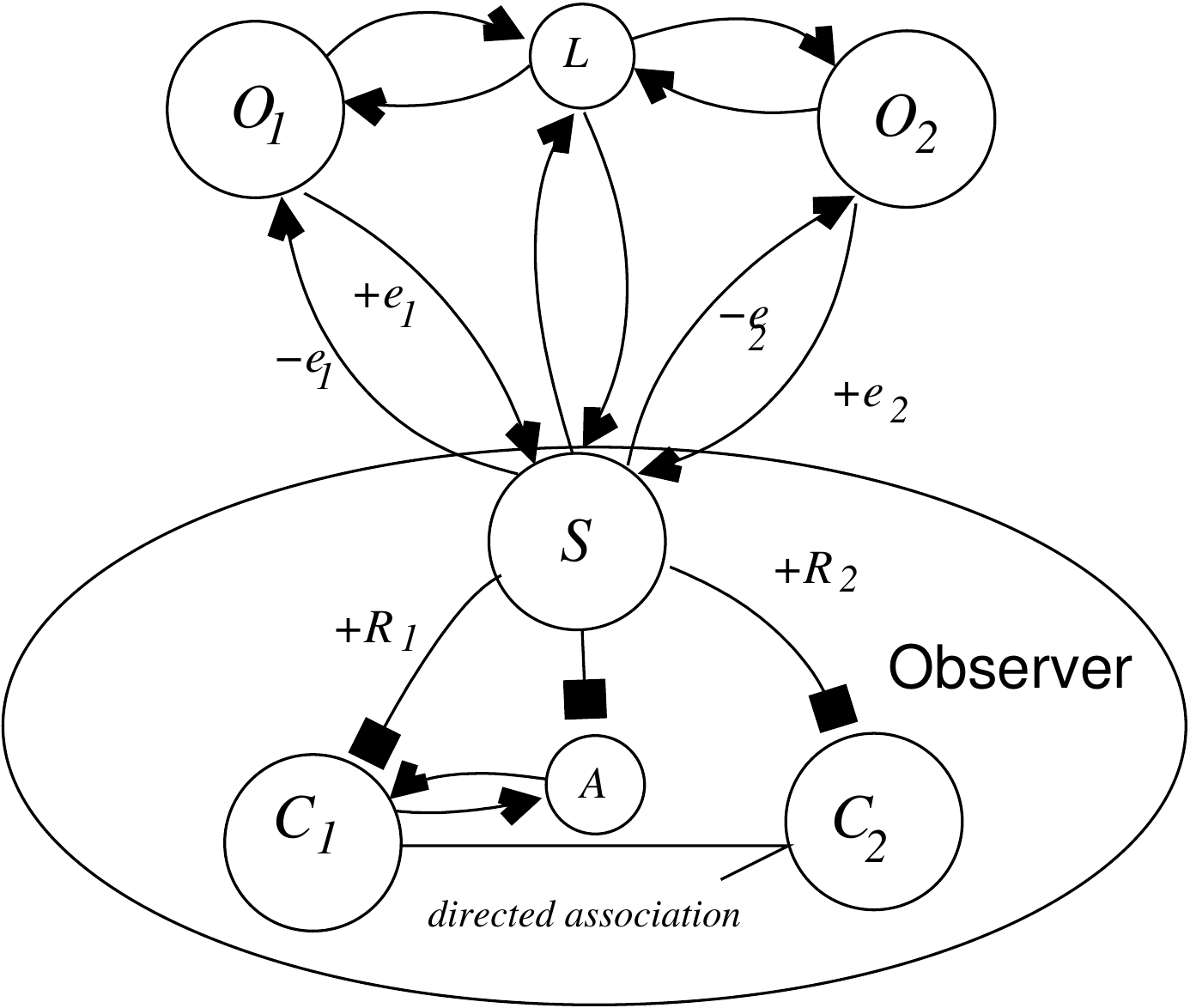}
\caption{\small Associations are formed in the mind of an observer.
  on the left, the association is formed without external impetus.  On
  the right, the observation of a link (perhaps through an
  intermediary $L$) is observed and represented as an association.
  Associations may or may not reflect actual structure of the real
  world: some we call abstract, imaginary, or conceptual.\label{association}}
\end{center}
\end{figure}

For an imprint to be memorized, a {\sc thing} $O_i$, in the objective exterior, promises its visible
`essence' $e_i$ to an observer's eyes or sensors $S_i$, allowing the observer to sample it
(see the left hand view in figure \ref{association}): 
\beq
O_1 &\promise{+e_1}& S\\
S &\promise{-e_1}& O_1\label{c1} 
\eeq 
If the promises are kept, this sampling occurs on a timescale $T_\text{exterior}$.  The
agent $E$ imposes a representation into the memory container whose image is
some agency $C_1$:
\beq
S   &\imposition{+R_1|e_1}& C_1\\
C_1 &\promise{-R_1}& S\label{c2} 
\eeq 
If kept, the latter exchange happens a timescale $T_\text{interior}$.  For
this to work, one must have 
\beq 
T_\text{interior} \le T_\text{exterior}, 
\eeq 
i.e. a mind must work quickly relative to what it hopes to observe.

The importance of these timescales is particularly significant when
sensor discriminators are chained together, in order to modulate one
anothers' capabilities. In that case, the timescales modulate and
throttle one another. Chaining is a technique attempted as a way of
using neural networks for non-trivial reasoning (see example
\ref{reasonnet}).

\begin{example}[Social networks as concepts]
  The formation of social networks has the same basic structure as the
  knowledge networks alluded to above. A social network may be
  considered a smart space, and indeed an intrinsic knowledge
  representation.  Links between members of a social network lead to
  clustering of cliques and regions.  Recommendations by one member of
  a group get passed on to others until a group also represents an
  interest in that topic, by epidemic transmission\cite{roles,vulnerability}.

  Members who are not interested in the mentioned topic will fall away
  or start new groups. Thus overlapping memory locations cluster
  around topics to which they feel an affinity, forming concepts.
  Their affinity depends on a context in the wider world that brings
  about the topics in question. Parts of the concept cluster may come
  at the topic from different angles, triggered by different atoms of
  context during a concurrent activation.
\end{example}

\subsubsection{Concept feedback, or introspection about {\sc things} through things}

Once learnt, the stability of concepts and their address locations can
be improved by feedback from the inner representation into the context
assessment stage of the sampling sensory apparatus
$S$, which allows recognition and discrimination of internal memories
along side new inputs.  This serves both a calibrating, self-stabilizing role,
and it also suggests the possibility of deliberate 
calibration extension, by matching and linking concepts relative to
similarly patterned concepts, when inputs are subdued.
This introspective processing of things, without the anchor of sensory reality, seems like
the likely origin of what we consider to be new ideas.
If existing memories shape new versions of concepts, then conceptual
reinforcement may be cyclic: 
\beq
O_1 &\promise{+e}& S\\
S   &\promise{-e}& O\\
S   &\imposition{+R_\text{new}|e_1,R_\text{previous}}& C_1\\
C_1 &\promise{-R_\text{new}}& S \\
C_1 &\promise{-R_\text{previous}}& S 
\eeq 
This has precisely the form of a Bayesian learning update process for
non-empty $e_i$. For empty $e_i$, it represents thinking or introspection,
while input is suppressed\footnote{Dreaming and introspection are not exactly
the same, as we know from brain research. During dreaming, not only
are sensory inputs suppressed, but long wavelength waves stimulate the brain
at a coarse scale, in some kind of annealing fashion.}.
\begin{figure}[ht]
\begin{center}
\includegraphics[width=5.5cm]{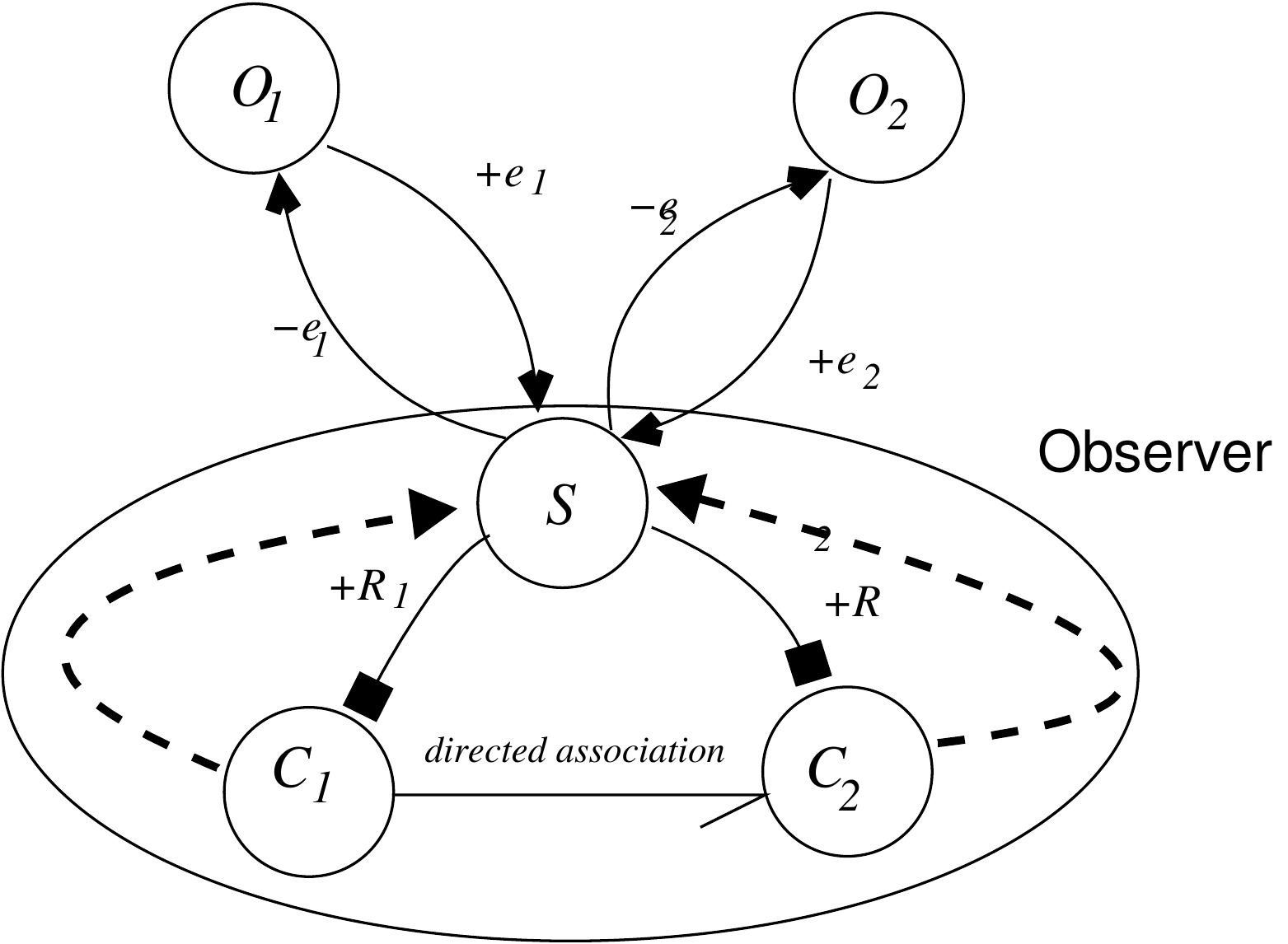}
\caption{\small Once learnt, the stability of concepts and their address locations can be improved by
feedback from the inner representation into the sampler $E$,
allowing recognition and discrimination.\label{associationfeedback}}
\end{center}
\end{figure}
For now, I will leave this possibility open ended. However, there is some
evidence that this mechanism is active in human brains\cite{hawkins}, as in figure \ref{awareness}.
\begin{definition}[Introspection]
  When an observer agent feeds back concepts from long term learning,
  back into its sensor and context stream as though they were new
  input.
\end{definition}

\subsection{Associations defined}

Associations are the links between the material agents of conceptual
knowledge space. Let's consider how a promise viewpoint points towards
the origins of associations as a subject of the promises made by these agents.

\subsubsection{Semantic addressing}

How one enters or stimulates activity in a knowledge network at a
suitable location, for thinking about a particular concept, is a
problem where context and association must overlap. Knowing both the
symbols and the exact coordinate address of every concept in memory
would be potentially cumbersome. Indices and directories can be used
to short-cut the linking of semantics to locations, but there is also
the possibility that memory could be addressed by its own tokenized
content.  This is the idea behind semantic addressing, as used in
databases and other key-value stores. The underlying mechanisms of
databases make use of a indices and directories to simulate semantic
addressing; however, other technologies, based on hashing, or
functional symbolic discrimination, do not need to use the
indirections of look-up tables.

\subsubsection{Concurrently active concepts}

\begin{figure}[ht]
\begin{center}
\includegraphics[width=10.5cm]{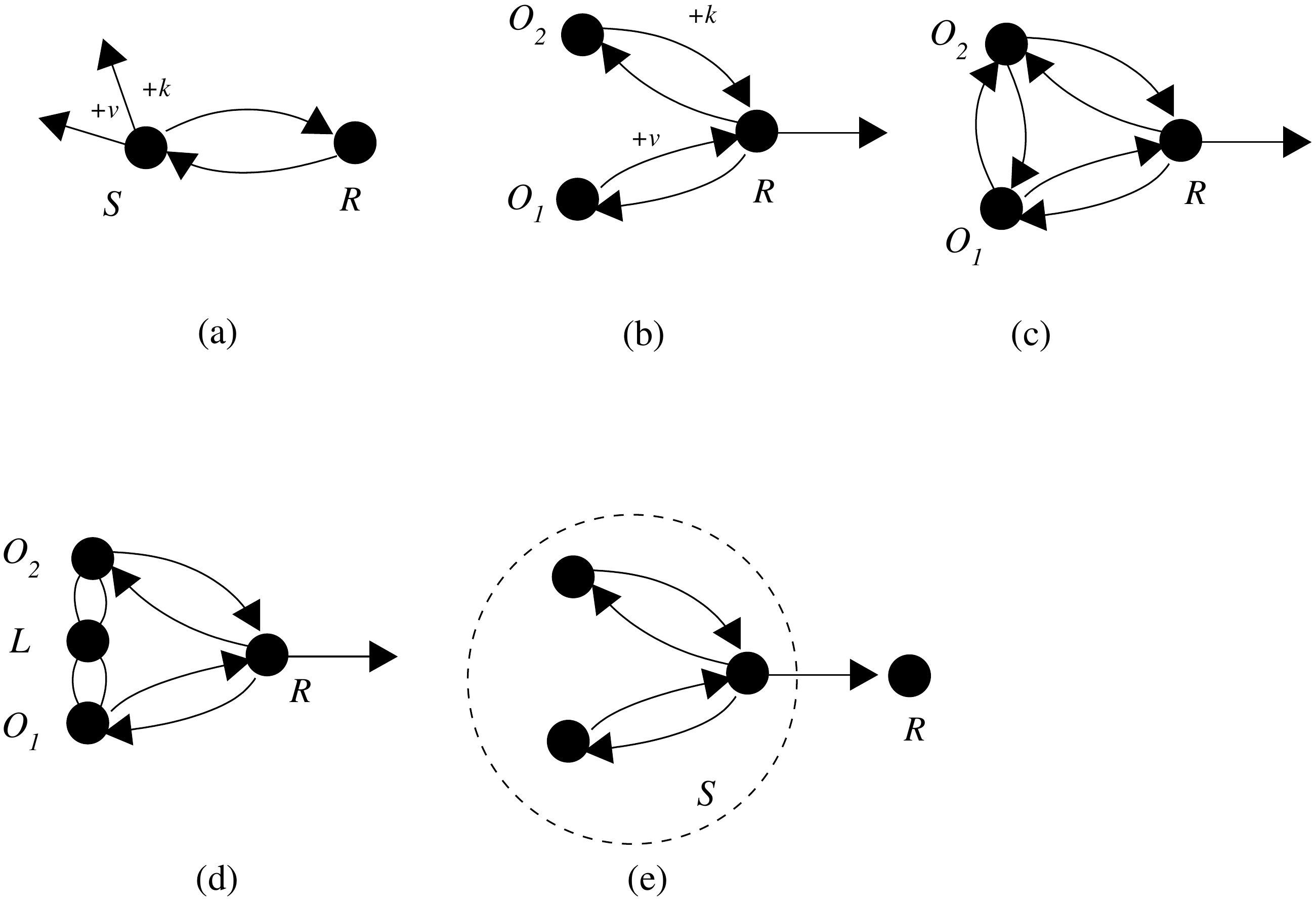}
\caption{\small The tensor forms of correlation, inferable as associations (see definitions in text).\label{association1}}
\end{center}
\end{figure}

Semantic addressing is related to conceptual {\em association}.  An
association is a pairing of concepts or data, such that one is a token
representation for the other, such as in a key-value pair. 
It involves a correlation of some kind. From
promise theory axioms, we know that a promise of correlation can only
be made locally by a single agent, as an independent
arbiter\cite{promisebook}.  This can take several forms (see figure
\ref{association1}) and explanations below.
Associations are semantic links between concepts. They originate from clusters
of promises.

\begin{definition}[Association]
  An association $\cal A$ is a labelled map between concepts. A simple
  diadic vector association $C_1$ has a relationship with $C_2$
  involves a link between two concepts, and a third concept which is
  name of link type/role itself.  \beq {\cal A}: \text{concept} \times
  \text{concept} \rightarrow \text{concept}.  \eeq We use the
  lower-half-arrow notation represent a directed association: \beq C
  \assoc{\text{named association}} C' \equiv C \promise{\cal A} C'.
  \eeq Unlike a simple promise, an association $a$ always has a
  conjugate or reciprocal interpretation $\overline a$.  An
  association implies the existence of a directed path between $C$ and
  $C'$.
\end{definition}
An association may be the subject of a cooperative promise between agents.
\begin{lemma}[Unlabelled concept-association graphs are non-directed]
The assumption of auto-reciprocal associations means that each association
provides bidirectional labelled connectivity in the knowledge graph
with a conjugate label.
\end{lemma}
The proof follows from the assumption that each association has a
mandatory conjugate or reciprocal meaning, and the existence of a path
between the concepts.  This means there is a bidirectional association
between any two concepts, but with conjugate labels.  \beq C \assoc{a}
C' \implies \left\lbrace
\begin{array}{c}
 C' \assoc{\overline a} C\\
 C \promise{+\adj} C'\\
 C' \promise{-adj} C
\end{array}\right.
\implies
\begin{array}{c}
 C \promise{-\adj} C'\\
 C' \promise{+adj} C
\end{array}.
\eeq
Hence there is a non-directed adjacency between $C$ and $C'$.
The relationship between associations and promises is complicated. An association
is a promise of virtual adjacency between concepts, which are superagent cluster;
however, the meanings of the concepts and associations depend on the promises
of the entire knowledge system. I'll leave it as an exercise to the reader to
imagine this this entire construction (see figure \ref{association1}).
\begin{example}[Associative notation]
\beq
T \,\assoc{\text{inside}}\, T',
\eeq
meaning $T$ is associated with $T'$, with the interpretation $T$ `is inside' $T'$.
\end{example}

\begin{definition}[Association structure]\label{assocstruct}
  A collection of assessments, which may be interpreted by an observer
  $R$ as a relative adjacency between two things or concepts, taking
  the form of a total function ${\cal A}: K \rightarrow V$, (written $k\assoc{\cal A}v$). It maps
  keys $K$ to values $V$, and is defined over some interior representation belonging to $R$.
\begin{enumerate}[label=(\alph*)]

\item {\bf Auto-calibrated association}. Scalar association by coincident promiser. A source agent $S$
promises both a key $+k$ and a value $+v$ to an unspecified agent.
Agent $R$ is in scope of these promises and assesses them
simultaneously:
\beq
S &\scopepromise{+k}{R}& \Unspec\\
S &\scopepromise{+v}{R}& \Unspec\\
R &\promise{{\cal A}: k\assoc{} v}& \Unspec
\eeq
\begin{example}[of definition \ref{assocstruct} (a)] These are dual or multiple roles, i.e.
scalar promises that are associated by virtue of originating from the same agent: e.g.
linguistic compounds like police-woman, 
bank-teller, food-truck, leather-jacket,
table-leg, car-factory, apartment-building, living-room etc. These are not merely linguistic compounds
but conceptual associations too.
\end{example}

\item {\bf Observed correlation}. Bi-scalar association, with key and value observed in separate agents.
\beq
O_1 &\promise{+v}& R\\
O_2 &\promise{+k}& R\\
R &\promise{-v}& O_1\\
R &\promise{-k}& O_2\\
R &\promise{{\cal A}: k\assoc{} v | k,v}& \Unspec
\eeq
$R$ assesses that $k$ and $v$ are simultaneously kept from different dependent sources
and associates them with the name $\cal A$.
\begin{example}[of definition \ref{assocstruct} (b)] The observation of
  correlations between qualities that have not been promised. Agents
  are authors, or police, seeing shapes in clouds, or faces on cars,
  i.e. the association by the observer, without reference to the outer
  reality.
\end{example}

\item {\bf Reported association}. Vector association, inferred by scalar promise from separate agents.
\beq
O_1 &\promise{+\cal A}& O_2\\
O_2 &\promise{-\cal A}& O_1\\
O_1 &\promise{+(O_1 \promise{+\cal A} O_2)}& R\\
O_2 &\promise{+(O_2 \promise{-\cal A} O_1)}& R\\
R &\promise{-(O_1 \promise{+\cal A} O_2)}& O_1\\
R &\promise{-(O_2 \promise{-\cal A} O_1)}& O_2\\
R &\promise{{\cal A}: O_1\assoc{} O_2 | (O_1 \promise{+\cal A} O_2),(O_2 \promise{-\cal A} O_1)}& \Unspec
\eeq
$R$ assesses the promises made by $O_1$ and $O_2$ as simultaneously kept, and
infers the association between these two agents under the name $\cal A$.
\begin{example}[of definition \ref{assocstruct} (c)]
  Observation of a connection, which the agents both promise to have,
  may be based on adjacency, causation, generalization, etc: e.g. a
  person on bike, device plugged in, Mark's dog, $A$ supplies $B$, $A$
  depends on $B$, $A$ and $B$ are partners, the agents `know' each other
  in the real world, the agents agree about subject $X$..  This is a
  formation of cluster of collaborative molecular roles from independent atomic
  roles.
\end{example}

\item {\bf Reported calibration}. Bi-vector association through intermediate agent.
Generalization of (c).
\beq
O_1 &\promise{+b_1}& L\\
O_2 &\promise{-b_2}& L\\
O_1 &\promise{+(O_1 \promise{b_1} L)}& R\\
O_2 &\promise{+(O_2 \promise{b_2} L)}& R\\
R &\promise{-(O_1 \promise{b_1} L)}& O_1\\
R &\promise{-(O_2 \promise{b_2} L)}& O_2\\
R &\promise{{\cal A}: O_1\assoc{L} O_2 | (O_1 \promise{+\cal A} O_2),(O_2 \promise{-\cal A} O_1)}& \Unspec\label{bbb}
\eeq
\begin{example}[of definition \ref{assocstruct} (d)]
  The calibration of roles, and innovative mixing.  These are promises that
  unite agents by a common (matroid) basis element: partners brought
  together by an interloper or matchmaker, the members of a club,
  agents authorized to act (e.g. keepers of the password), members of a country, all agents facing North.
\end{example}

\item {\bf Reported encapsulation}. Scaled superagent informs as a function via index/directory lookup table.
Generalization of (a).
\beq
S &\promise{+(k \assoc{\text{asserted}} v)}& R\\
R &\promise{-(k \assoc{\text{asserted}} v)}& S\\
R &\promise{k\assoc{\cal A} v}& \Unspec
\eeq
$R$ parrots the trusted association promised by superagent $S$, about its interior state.
\begin{example}[of definition \ref{assocstruct} (e)]
An association promised by a single encapsulating agent: e,g. I am a shopping mall, a city, a community, and these agents
are part of my makeup (here is is my internal directory).
\end{example}
\end{enumerate}
\end{definition}
The type of exterior promise involved in the simultaneous assessment
by $R$' need not be mentioned, as it plays no real role, as long as
the promise type matches across the concurrent assessment. Also, as usual, it is understood that the
promises, on which $R$ bases its assessments, may not always be kept.
Associations are, consequently, at least as uncertain as the
promises that lead to them.

In terms of physical representation, spacetime proximity (e.g. adjacency,
such as in a cellular automaton or an ionic chemical bond) is a more
robust or less uncertain promise of association than say remote
correlation by message exchange (e.g. pen pals or a covalent chemical
bond), because it does not rely on the promises of intermediate agents to glue
together the association. In either case, the associative bindings are
reinforced by frequent revisitation, i.e. learned familiarity,
increasing trust\cite{burgesstrust}.

\subsubsection{Tensor structure of association}

Associations are constraints on the spacetime structure of a semantic
space.  Associations are made of promises, but not vice versa.
As a constraint, an association involves spacetime symmetry
reduction, such as a regular ordering or clustering around
anchor points. Concepts-association graphs are a subset of promise graphs.
\begin{lemma}[Associations are tensors]
All tensor forms may be constructed by involving clusters of $N$ agents.
\end{lemma}
Referring to paper I, the proof follows trivially;
the rank of the tensor is less than or equal to the number
of agents involved in an association. 
In a mathematical tensor, each collection of indices represents a
subjective agent viewpoint, which could potentially be transformed into
another agent's perspective, by redefining agents. 
This is known as
covariance\cite{burgesscovariant}. It is intriguing to wonder if the
spacetime properties of memory allow covariance of tensor
representations to play a role in cognitive perspective too.  
Context may play the role of such a transformation, as it alters the
clustering structures of agents in a knowledge representation, 
and the promised associations made by the superagent clusters.
\begin{example}[Associative arrays]
In a computer, associative structures form directories or index maps
(see paper II). In order to link concepts semantically, there has to
be an underlying dynamical addressing of spacetime, which acts as
intermediary. Transforming the addresses, in a way that respects
spacetime structure, leads to a different perspective.
\end{example}
\begin{example}[Superagent directory lookup table]
A superagent directory is an associative array in which keys are mapped to promises.
Agent directories map interior agent names/addresses to their promises.
\end{example}

\begin{example}[Entity-Relation models]
  In Entity Relation (SQL) modelling, the graphical structure is built
from lookup tables, and addresses are hidden from users, replaced with
primary keys. Thus spacetime locations may be found only by searching,
  or using an index to map primary keys (see figure \ref{ERvsSL}).
  Underlying this user interface legerdemain is a ordinal addressing scheme
  for referring to the locations: just as a book index maps from
  keywords to page numbers, the ordinal page numbering is complicit in
  making the scheme predictable and efficient.  An index exploits the
  solid state structure of a space make location predictable.  There
  are then all kinds of structures with properties that optimize the
  time to retrieve the item.

\begin{figure}[ht]
\begin{center}
\includegraphics[width=8.5cm]{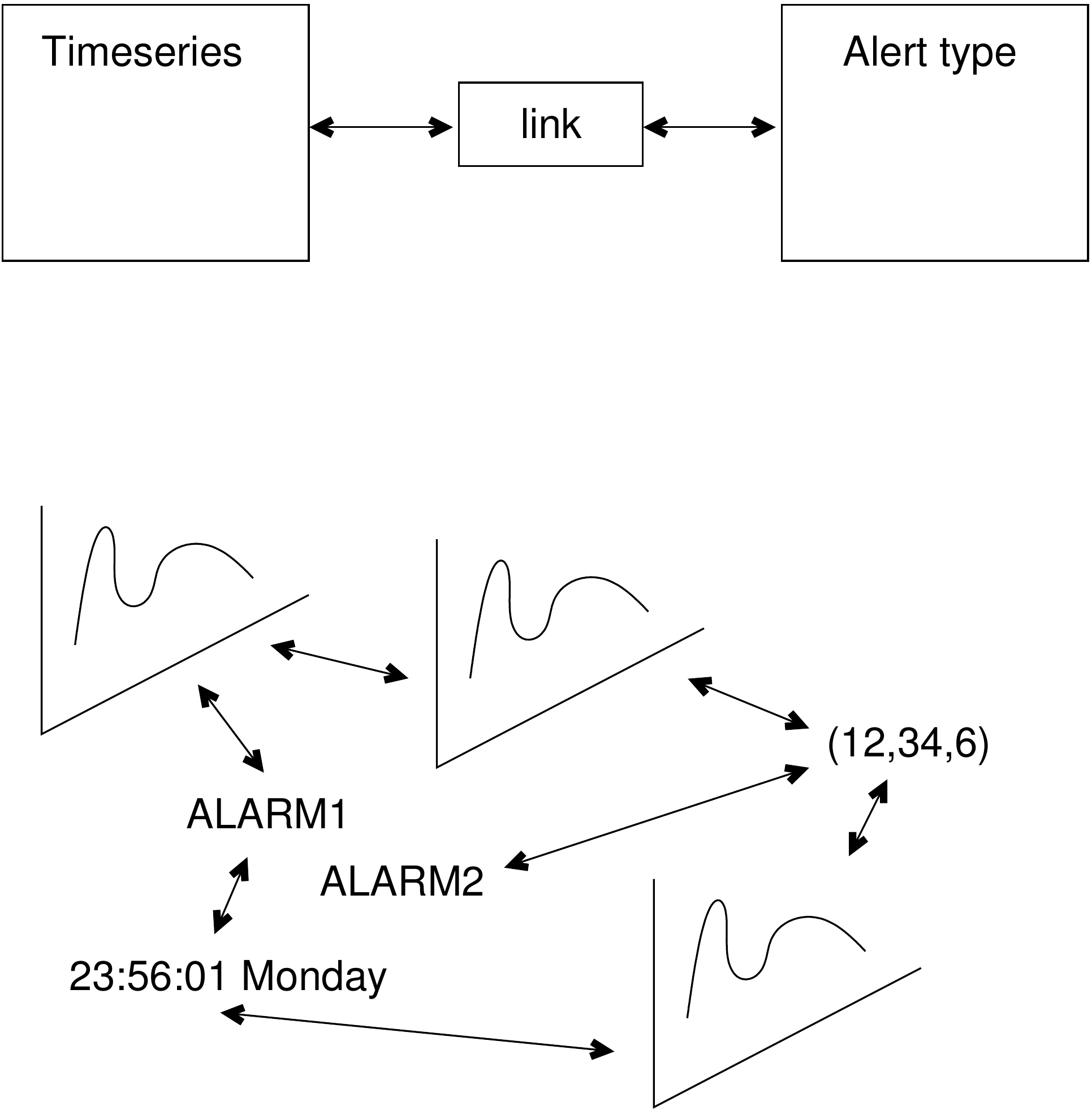}
\caption{\small Entity relation models in databases allow different
  types of data to be standardized as tabular templates and then
  linked together through data intermediaries forming a map, as long
  as each instance of a table they has a primary index, i.e. a coordinate system that can be
  indexed against.\label{ERvsSL}}
\end{center}
\end{figure}
\end{example}

\begin{example}[Lookup tables]
In data science, associative arrays are simple
data maps, mapping a domain into a codomain.
As associative data array is a simple mapping between sets. Consider a set of three colour
concept classes: $C = \{ {\rm red, green, blue} \}$ and a set of data values $D = \{ 000000, \ldots, FFFFFF\} $ corresponding to these
classes. We may form a mapping M between these concepts and ranges of discrete colour codes
in RGB space.
\beq
{\cal A}[{\rm red}] &=& \{ 010000, \ldots, FF0000 \}\\
{\cal A}[{\rm green}] &=& \{ 000100, \ldots, 00FF00 \}\\
{\cal A}[{\rm blue}] &=& \{ 000001, \ldots, 0000FF \}
\eeq
\end{example}

\subsubsection{Associations may connect any agent or scale}

An association is not limited to a particular kind of agent, as illustrated by these examples.
\begin{example}[Functions]
Functions association inputs with outputs, e.g. looking up RGB codes for colours:
\beq
Colour(c) \leftrightarrow Codes(d)
\eeq
Or as an associative functional map function $\cal A$, with $c\in C$ and $d\in D$:
\beq
c = {\cal A}(d).
\eeq
\end{example}

\begin{example}[Associating different schemas]
  What if we want to keep the different memory types distinct, in
  special kinds of memory, and still weave them together into an
  associative web? i.e. we don't try to transduce them all into words,
  or data values, but keep timeseries, words, image fragments, etc all
  separate but Entity Relation theory introduced for database
  modelling attempted to answer this question\cite{date1}.
\end{example}

\subsubsection{How are associations identified?}

Consider how two concepts can become associated, in a memory
representation.  Referring to equations (\ref{c1}) and (\ref{c2}), we
may add a single promise pair from $S$ to itself, which loops back, to
form the association.  By continuously keeping and assessing this
promise, the agent can learn the association, as a stable
representation in its knowledge space.
An association, intended by the sensor observer $S$ is a
promised scalar relationship about two sub-agents, inside its horizon:
\beq
S \promise{\pm C_1 \assoc{\text{some association}} C_2} S
\eeq
If the association builds on an actual observation of
objects $O_1$ and $O_2$, then that relationship would be communicated by the promises
of $e_1, e_2$, and then one could consider the association to be
a conditional promise:
\beq
S \promise{\pm C_1 \assoc{\text{some association}} C_2 | e_1, e_2} S\label{cas1}
\eeq
If the association is based on reasoning by the observer then
\beq
\text{\rm Observer} \promise{\pm C_1 \assoc{\text{some association}} C_2 | e_1, e_2} \text{\rm Observer} \label{cas2}.
\eeq
To associate concepts, we must have simultaneous triggering of concept
clusters. No other information exists to join two concepts together.
Since the concepts themselves are bootstrapped by the equation (\ref{cas1})
\begin{lemma}[Simultaneous coincident context]
The semantics of association are those of simultaneous activation of context.
\end{lemma}
From equations (\ref{cas1}) and (\ref{cas2}), there are two possible
forms of concurrent activation, which are the only concurrent events
that contain both concepts as information. Thus the only causal
possible relationships are these two routes.  Since exact simultaneity
is impossible, coarse graining of simultaneity is an essential
requirement to coarse graining together concepts into associated clusters.
This means there is a timescale for association formation
$T_\text{association} \ge T_\text{interior}$.

\subsubsection{Flooding, routing,  and association}

It is worth remarking that, in order to form concepts by association,
concepts may be addressed from inputs according to either the flooding
or routing addressing methods. Flooding (even within a restricted
region) is a divergent process, which is therefore expansive and leads
to possible generalization of concepts. Routing is convergent and
specific, allowing precision. These complementary `advanced' and
`retarded' propagation mechanisms are at work in spacetime networks
at all levels.

\subsubsection{Autonomy and association (a notational caution)}\label{caution}

\begin{figure}[ht]
\begin{center}
\includegraphics[width=10.5cm]{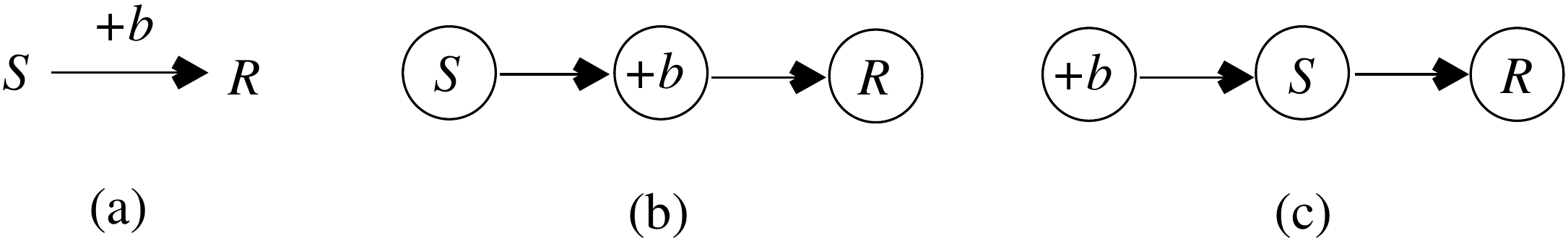}
\caption{\small A promise of a body $+b$ (a) can impose signal agents,
  or route the third party without it being emitted/absorbed. It is not correct
to imagine the body provider as an intermediary (b), rather it is a source
that must be relayed by the agent who makes the promise (c), because $b$
cannot promise on $S$'s behalf.\label{proxypromise}}
\end{center}
\end{figure}

Consider figure \ref{proxypromise}. If we represent composition,
within a knowledge space, as a promise (case (a)), a literal reading
of the notation 
\beq 
S \promise{+b} R; 
\eeq 
appears to claim that the associative label comes
between the two concepts it links (case (b)). If this were the case,
the intermediate agent would need complex semantics to observe both
$S$ and $R$ and act as a switch to relay the intention between the two
rather than impeding them. As an intermediary, it would be more
likely to impede or distort transmission than represent the other's
intent by proxy\cite{promisebook}.
Thus a connection would require the creation
of a new agent and a conditional promise to activate the connection.
Associations promised by autonomous agents, like cells or dumb
devices, do not have the internal capabilities to keep advanced
promises.
If $b$ is realized as an intermediate agent, then it blocks $S$'s
direct promise, rather than enabling it, and makes no guarantee of a
promise of its own to $R$. This is both a complication, expensive, and
uncertain.  A simpler representation, which observes the principle of
autonomy\cite{promisebook}, is shown in case (c), in which $b$ serves
as the source of the promise body, and $S$ promises to route it
directly to $R$. In this form, the source of the intention acts as a
basis (matroid) agent, which may be constant, and $S$ has a fixed
promise for all $b$ rather than requiring a specialized promise
for every $b$. Thus, this approach is economically more scalable. All the dynamics are
directly between the concept agents.
By the conditional promise law:
\beq
S \promise{+b} R  ~~~~~~~ \equiv > ~~~~~~~ 
\left\{
\begin{array}{ccc}
b &\promise{+\text{data}} &S\\
S &\promise{-\text{data}} &b\\
S &\promise{+\text{data}|\text{data}(b)} &R\\
R &\promise{-\text{data}} &S
\end{array}\right.  ~~~~~~~ \equiv > ~~~~~~~ 
\left\{
\begin{array}{ccc}
b &\promise{+\text{data}} &S\\
S &\promise{+\text{data}(b)} &R\\
R &\promise{-\text{data}} &S
\end{array}\right. 
\eeq

\begin{theorem}[No go for bipartite lattice]
  A programmable promise graph can not be implemented as a bipartite graph, by proxy
  (see figure \ref{proxypromise}(b)), i.e. $S \promise{+b} R$ cannot be
represented in the form $S\promise{x} B \promise{y} R$, for $b$, for any $x,y$.
\end{theorem}
From the discussion above, the proof follows from the fact that a
bipartite lattice would not propagate the intended promises without
additional promises to switch on the intermediate node, leading to a
recursive deficit in switching. This suggests that a bipartite lattice
is not a viable way to implement associations.  
In order to make a proxy representation, we would need
\beq
S &\promise{+x}& R,B\\
B &\promise{+b}& B,S,R\\
R &\promise{-x}& S,B,
\eeq
where $x$ is to be determined. In order to transmit intent $+b$
to intermediate agent $B$, $+x$ would need to be of the form
\beq
E &\promise{\pdef(b)}& B\\
B &\promise{-\pdef(b)}& S,R\\
B &\promise{+b|\pdef(b)}& R\\
S &\promise{?}& R
\eeq
for some external control source agent $E$.
The latter is needed to promise intent from $S$ to $R$, but there is no $?$ that
can promise the intent to follow $B$ from $S$. If $S=E$, there would need to be
a direct channel from $S$ to $R$, which violates the bipartite structure.
Section \ref{embellishedsec} illustrates a workaround for this,
however, using intermediate switching agents for more complex linkage;
the result is not a direct association, but an implicit association by
reasoning.  The routing construction does not violate the principles
of autonomy, and can lead to a simple direct association:
\begin{lemma}[Promises may be routed]
A promise graph in space can implemented as a conditional routing switch (figure \ref{proxypromise}(c)).
\end{lemma}
This follows from the conditional promise law\cite{promisebook}, in which
a promise from the basis agent for the association label acts as the switch.
\beq
B &\promise{+ \pdef(b)}& S\\
S &\promise{+b|\pdef(b)}& R\\
S &\promise{-\pdef(b)}& B,R.
\eeq

\subsubsection{Quasi-transitive or propagating associations}

Intent propagates between agents if there is what we might call a {\em
  quasi-transitivity} as a result of promises made and
kept\cite{faults}. In a knowledge network, this may translate into
concept associations which propagate if there are associations that
have certain spacetime transformation properties.  Associations can correlate causal
influence over two dimensions (scale and location). Let $C^{(n)}$ be conceptual
(super)agents at an aggregate scale, labelled by $n$:
\begin{itemize}
\item {\bf Vertical scaling associations} (North-South) which aggregate or decompose
concepts by membership, i.e. scale resolution:
\beq
C_i^{(n+1)}  \assoc{+\rm generalizes} C_j^{(n)}
\eeq
\begin{figure}[ht]
\begin{center}
\includegraphics[width=10.5cm]{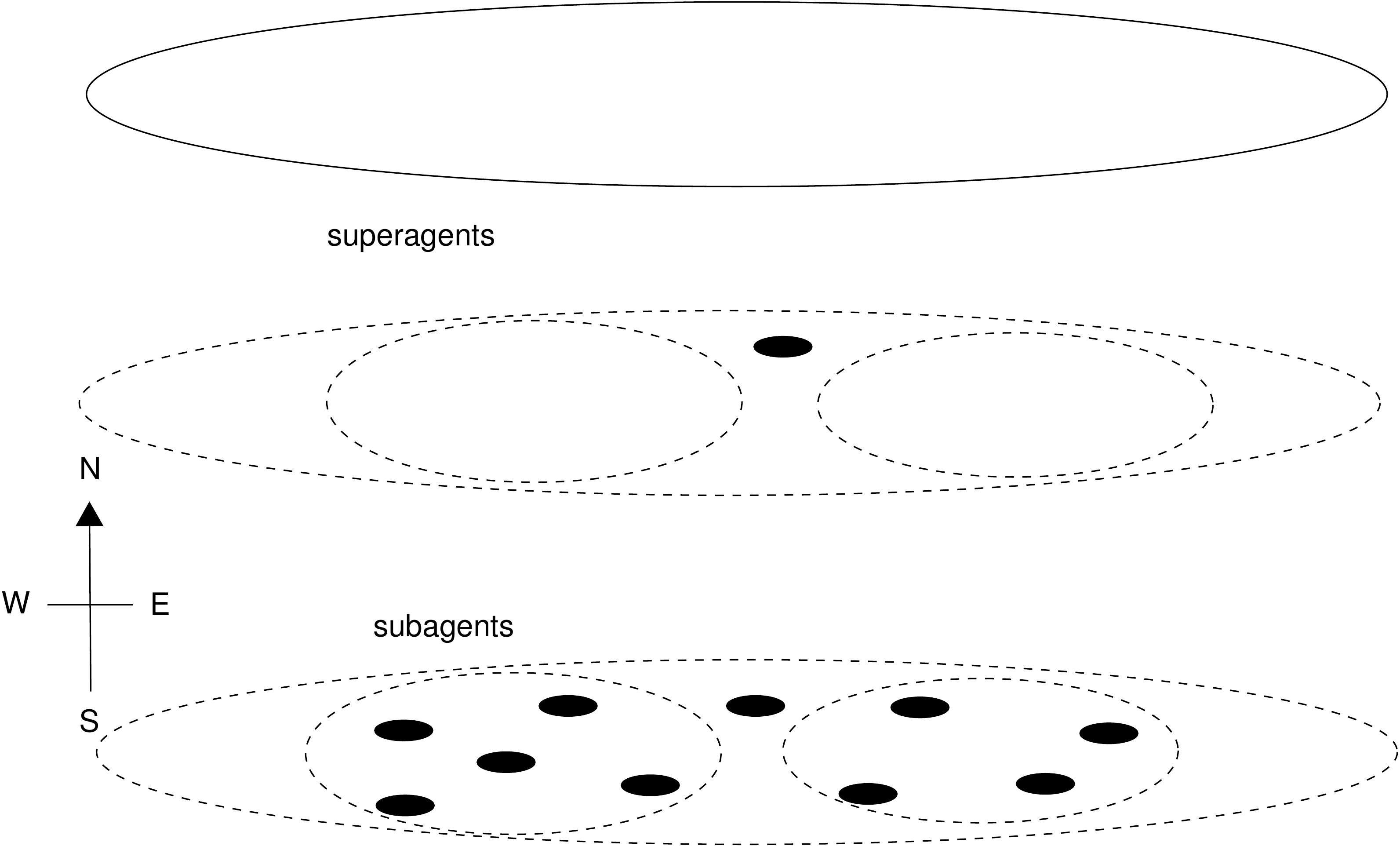}
\caption{\small The common use of compass directions to refer to 
a scaling hierarchy of increasing generalization, i.e. increased coarse graining,
while other associative properties happen in the planar scaling layers. This language of N,S,E,W is
common in the networking industry.
All the irreducible associations refer to constraints on
  the scaling and the topology of the space of agents.\label{scales}}
\end{center}
\end{figure}
These associations break scaling symmetry by pinning concepts (see below),
or selecting a direction for scale.
\beq
C_\text{My dad} \assoc{\text{is bigger than}} C_\text{your dad}.
\eeq

\item {\bf Horizontal associations} (East-West):
within each superagent (see figure \ref{scales}):
\beq
C_i  \assoc{\text{similarity}} C_j
\eeq
or between superagents, e.g.
\beq
S_i  \assoc{\text{depends on}} S_j.
\eeq
\end{itemize}

\begin{itemize}
\item Vector associations that may result in a quasi-transitive
  inference, i.e suggest a propagation of intent over a potentially long range.
  These are related to spacetime properties (see table \ref{assoc}).

\item Vector associations that result in no transitive inference, i.e.
  suggest short range finite-dimension transitions rather than
  spatially extensive properties.  These are related to interior state
  machines (like hidden variables).
\end{itemize}

\subsubsection{The irreducible positive associations}\label{qtassoc}

The irreducible quasi-transitive structural associations are documented in table \ref{assoc}.
\begin{table}[ht]
\begin{center}
\begin{tabular}{|c|c|c|c|}
\hline
\sc ST Type & \sc Forward & \sc Reciprocal & \sc Spacetime structure\\
\hline
\hline
&is close to & is close to &\\
&approximates  & is equivalent to &\sc PROXIMITY\\
1&is connected to & is connected to&``near''\\
&is adjacent to & is adjacent to&   Symmetrizer \\
&is correlated with & is correlated with&\\
\hline
&\sc Forward & \sc Reciprocal & \sc Spacetime structure\\
\hline
&depends on & enables&\\
2&is caused by &causes & \sc GRADIENT/DIRECTION \\
&follows & precedes& ``follows''\\

\hline
&\sc Forward & \sc Reciprocal & \sc Spacetime structure\\

\hline
&contains & is a part of / occupies&\\
3&surrounds  & inside & \sc AGGREGATE / MEMBERSHIP \\
&generalizes & is an aspect of / exemplifies& ``contains''\\
\hline

\hline
&\sc Forward & \sc Reciprocal & \sc Spacetime structure\\
\hline
&has name or value &  is the value of property&\\
4&characterizes &  is a property of &\sc DISTINGUISHABILITY\\
&represents/expresses &  is represented/expressed by & ``expresses''\\
&promises && Asymmetrizer\\
\hline

\end{tabular}

\end{center}
\caption{\small The four irreducible association types are characterized by
their spacetime coincidence or adjacency. Note that, in promise theory, even relationships
like `is correlated with' are directed relationships: one may not assume that
the assessment of a mutual property is mutually assessed. Similarly the expression
of a property is a cooperative relationship.\label{assoc}}
\end{table}
The generally transitive or propagating properties fall into the
familiar irreducible classes (see table \ref{assoc}).
Let $A^{(n)}$ be agents in either a real or a knowledge space, which
may include $C^{(n)}$:
\begin{enumerate}

\item {\bf Proximity/Adjacency}: (space) mutual, classically
  non-directed relationships represent non-directional role symmetries
  that have not been broken.  e.g.  {\em overlaps with, shares with,
    associates with, joins, connects with, relates to, etc.}
\beq
{\cal A}_\text{similarity}: A^{(n)}_i \rightarrow A^{(n)}_j.
\eeq
For every ${\cal A}_\text{similarity}: A^{(n)}_i \rightarrow A^{(n)}_j$, there is also
a ${\cal A}_\text{similarity}: A^{(n)}_j \rightarrow A^{(n)}_i$.

\begin{lemma}[Similarity associations are symmetry preserving]
The class of similarity associations represents symmetry preserving
transformations, mapping one concept into an equivalent one
that may be reversed by inverting the direction
\end{lemma}
If $s$ is a similarity association $C_1 \assoc{s} C_2$, then $C_2 \assoc{s} C_1$,
because $A \promise{C_1 \assoc{s} C_2}A \implies A \promise{C_2 \assoc{s} C_1}A$.

\item {\bf Sequence/Dependency}: (time) associations that describe causal pathways
  in a system, whereby one agent depends on the prior keeping of a
  promise by another, or builds on the existence of a prior concept.
  e.g. {\em depends on, reinforces, contributes to, asserts, documents,
  annotates, expresses, elaborates, extends, involves.} 
\beq
{\cal A}_\text{causal}: A^{(n)}_i \rightarrow A^{(n)}_j.
\eeq

\begin{lemma}[Causal associations break equivalence symmetry]
Causal associations break the directional symmetry of mutual semantics.
\end{lemma}
Selecting a precedence between two concepts imposes a local direction
on the map, thus selecting a preferred direction or concept, and
breaking the `reflection' symmetry or equilibrium between concepts.

Causal association may be chained together to represent world lines,
because any change in an association measures time, and selects a preferred direction.

\item {\bf Scaling/containment}: (boundary enclosure) describes a transformation
  from one scale to another, in which multiple concepts at the smaller
  scale contribute to the larger scale.  e.g. the relationship between
  a collective agent and a member agent: {\em generalizes,
    exemplifies, abstracts, models, etc}.

Let (super)agents at scale $n$ be represented by $A^{(n)}$ (see paper II), then
\beq
{\cal A}_\text{coarse}: A^{(n)} \rightarrow A^{(n+1)}.
\eeq
\begin{lemma}[Coarse grain associations break universal scaling]
  Coarse graining associations pin semantics at a particular semantic
  scale, breaking renormalization scaling symmetry.
\end{lemma}
By naming a specific agent or superagent, e.g.  
\beq 
C_\text{animal} \assoc{\text{generalizes}} C_\text{dog}.  
\eeq 
we break a generic scale-free relationship.

  The formation of superstructure by associative clustering leads to
  generalizations and new types.  Generalization of instances into
  conceptual hubs allows self-organized model learning, and increases
the scope for fuzzy matching.

\begin{lemma}[Smoothing]
If we take a vector field and coarse grain the space so that the
points defining the vector are effectively coincident at the new
scale, then we average away the gradients.  This is the process of
smoothing and noise reduction. It transforms similarity states into
a single thing with representative semantics, i.e. a common concept.
\end{lemma}

\item {\bf Promise/Expression}: (property) if we associate agents, while
  retaining the special names and characters, it is like the
  formation of molecules. This breaks the symmetry of roles between
  the component agents in a cooperative association, e.g.
{\em represents, is a name for, is the doorway to, etc}.
\beq
{\cal A}_\text{representation}: A^{(n)}_i \rightarrow A^{(m)}_j.
\eeq

\begin{lemma}[Representation breaks role symmetry]
Representational associations break the symmetry of mutual semantics,
selecting one part to act as a proxy labelling the other.
\end{lemma}

\end{enumerate}

\begin{lemma}[Transitivity no go]
Transitivity of associations cannot be promised for arbitrary associations.
\end{lemma}
The proof follows from the basic axiom that no agent may make a
promise on behalf of another. Concepts are (super)agents, and no agent can promise a
property on behalf of another, i.e. that an association will be 
enforced beyond itself.  
\begin{example}[Transitivity is not assumed]
If $A$ promises it is
connected to $B$, and $B$ that it is connected to $C$, we simply don't
know whether $A$ is connected to $C$, either directly or through $B$.
it might be true that there is a connected path from $A$ to $C$. This
is not a certainty, because $B$ might not allow transmission of
anything between the two. The fact that agents are chained together in
relationships is no guarantee of their.
\end{example}

\begin{example}[Vertical association limits (hierarchy of generalization)]
Vertical associations, which take us from exemplary {\sc thing}s  to a generalized
class or concept, could be stacked
to any depth, in principle. In practice, however, human faculty for
stacking concepts is quite limited: we can typically keep only a
handful of ideas in mind at any one
time\cite{certainty,depth1,dunbar2}.  Moreover, in any practical
system, the apparatus by which associations are made and classified is
finite, and will have to be understood by a human with limited
faculties, thus it is not implausible that these vertical links would
be finite in number, and perhaps specialized with distinct semantics
that reflect the transformation staging in learning process (recall the
stages in the parsing example \ref{parsing}).
For example, a limited set of stages could lead to the following interpretations:
\begin{enumerate}
\item Species in a taxonomy.
\item Things in a taxonomy.
\item Concept hierarchy.
\item Sensor data hierarchy.
\end{enumerate}
Going up the list corresponds to `is generalized by', and going down
the list corresponds to either `is exemplified by' or `is composed
of'\footnote{It is interesting to note that the brain cortex has
neuron column structures that can form hierarchy up to depth 6, and
to speculate about whether this is related\cite{hawkins}.}.
\end{example}

\subsubsection{Non-transitive semantics in associations}\label{ntassoc}

Not all associations make even quasi-transitive sense. Consider human
relationships, like father, daughter, etc.
If we write
\beq
\text{Fabian} \assoc{\text{is father to}} \text{Simon} \assoc{\text{is father to}} \text{Dawn},
\eeq
it is by no means true that Fabian is father to Dawn. The father
relationship is pinned at a specific short range: it cannot form a
state of long range order. It is, in fact, is a short-range form of
 association, and a causal precedence
association. Interestingly, it does not propagate, because the concept
of father is limited. This now has the direct implication of that
we can state as a general law: that loss of long range symmetry is
directly associated with knowledge encoding, as stated formally in the final
conclusions.
\begin{figure}[ht]
\begin{center}
\includegraphics[width=6.5cm]{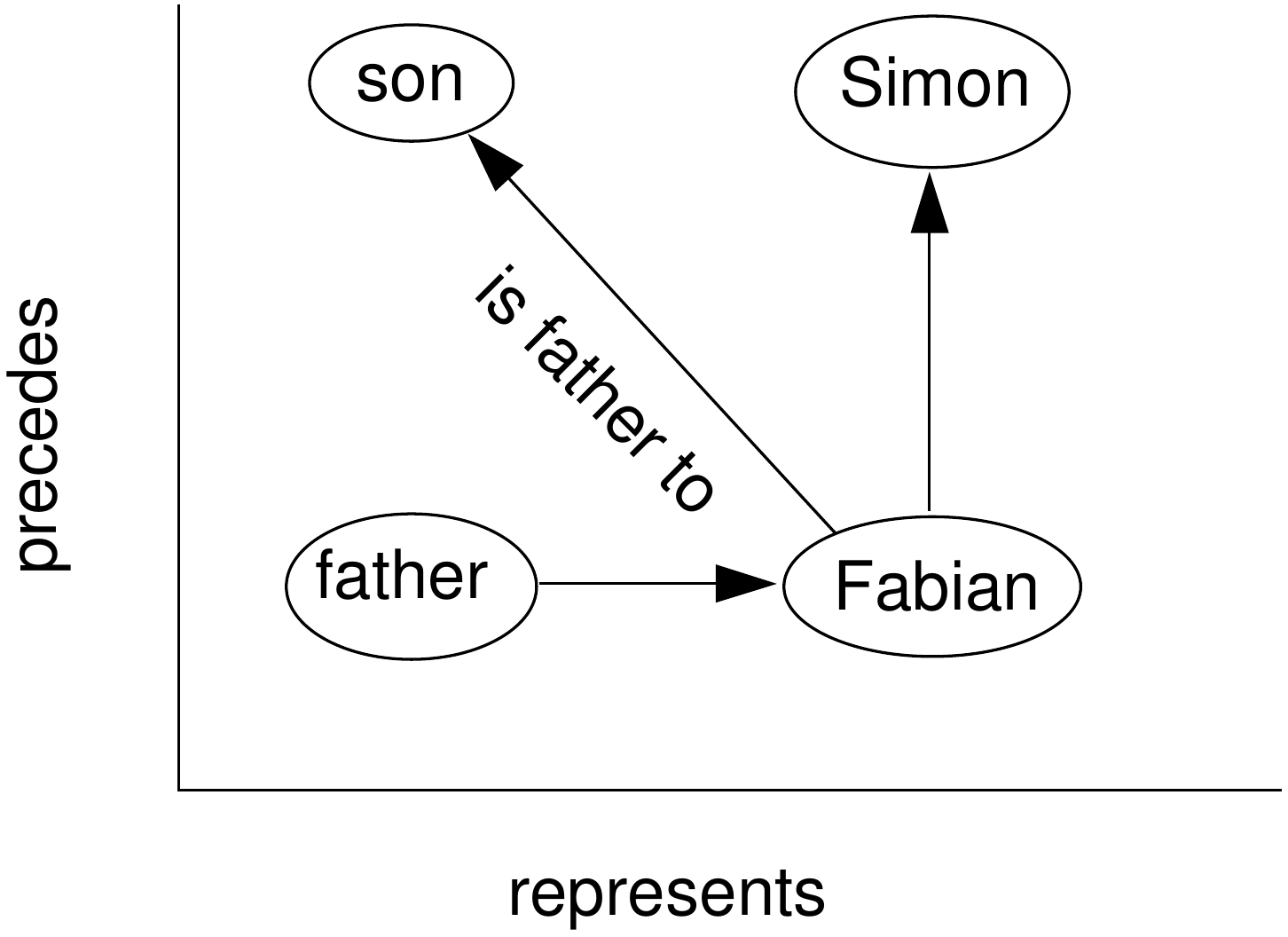}
\caption{\small Constrained association by something like vector addition of
irreducible types.\label{father}}
\end{center}
\end{figure}

Imbuing relationships with labels that constrain the semantics locally also
constrains the propagation of intent. This is significant: it means 
that a plain graph theory cannot represent hierarchical semantics, and we
need to understand specifically how concepts, embedded in the associations,
further constrain the spacelike clustering of concepts. If we decompose
`is father to' into a causal/precedence association and a representation
association (see figure \ref{father}, and recall section \ref{caution}):
\beq
X \assoc{\text{father of}} Y = \left( \text{father}\; \rassoc{\text{represents}}\right)  X \assoc{\text{precedes}} Y\\
\assoc{\text{father of}} \simeq \rassoc{\text{represents}} \oplus \assoc{\text{precedes}}.
\eeq

What this shows is that our understanding of concepts is more
complicated than our understanding of spacetime; it involves higher
tensor forms. If the father concept were just any old agent, and `is
father to' were just a generic `source of', then the sequence 
\beq
\text{Fabian} \assoc{\text{is source of}} \text{Simon} \assoc{\text{is
    source of}} \text{Dawn}, 
\eeq 
now makes quasi-transitive sense.  Why the father concept does not
propagate does not have an obvious spacetime explanation, but may be
addressed from the viewpoint of reasoning (see section
\ref{reasoning}). Semantics are sometimes pinned to particular
spacetime scales, and cannot be generalized beyond them. Thus, the
emergence of semantics may be associated with spacetime symmetry
breaking.

\subsubsection{Embellishing associations for deeper inference}\label{embellishedsec}

To represent higher tensor forms, where associations are more complex
than a simple link between two concepts, we may consider embellished
associations. These are constructed by introducing an intermediate
concept and treating the association as an inference (see section
\ref{reasoning}).  Although the direction of a narrative still follows
a direction from a promiser concept to a promisee, other concepts may
embellish the descriptions of the associations, as labels. Thus vectors associations
generalize to tensors, with multiple concepts clustered into the
association itself.

\begin{example}[Rolling context]
Context plays a key role in the iterative association of concepts (see figure \ref{embellished}):
\begin{figure}[ht]
\begin{center}
\includegraphics[width=10.5cm]{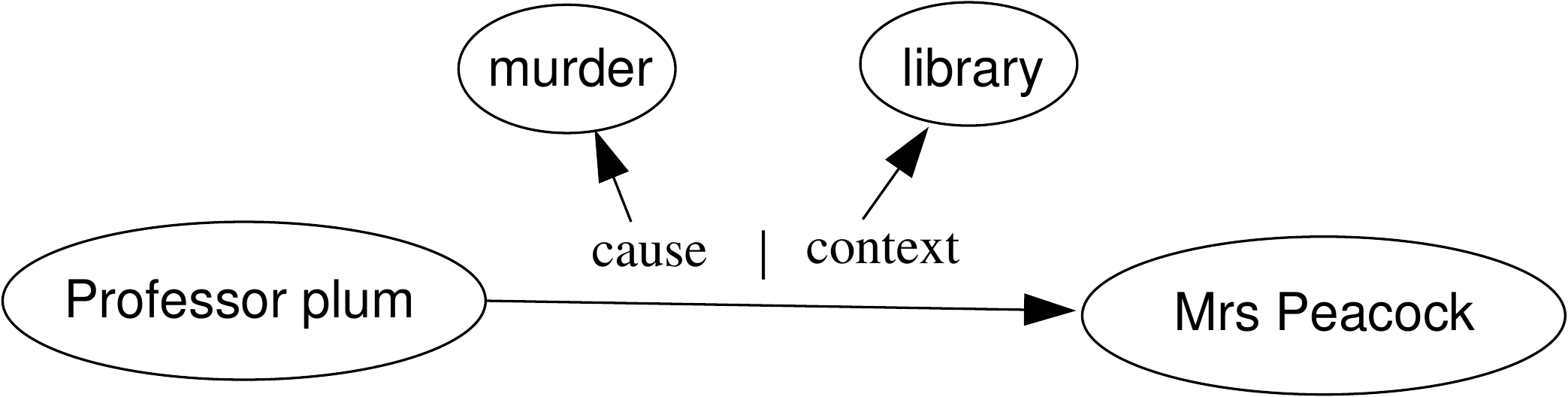}
\caption{\small Embellished association is a general
tensor.\label{embellished}}
\end{center}
\end{figure}

\begin{center}
\begin{tabular}{|l|l|l|l|}
\hline
From-Concept & Association & Context & To-Concept\\
\hline
\hline
Professor plum & murders with a knife & in the library & Mrs Peacock\\
\hline
\end{tabular}
\end{center}

In the next iteration, the context would accumulate the foregoing concepts
until they age beyond relevance: 

\begin{center}
\begin{tabular}{|l|l|l|l|}
\hline
From-Concept & Association & Context & To-Concept\\
\hline
\hline
Police & arrive & the murder of professor plumb & in the library\\
       &        & in the library by Mrs Peacock.&\\
\hline
\end{tabular}
\end{center}
\end{example}

What kind of spacetime structure accounts for this graph (see figure \ref{embellished})?
It is a tensor of rank less than or equal to four:
\beq
{\cal A}_{c_1c_3c_4c_2}
\eeq
\begin{figure}[ht]
\begin{center}
\includegraphics[width=10.5cm]{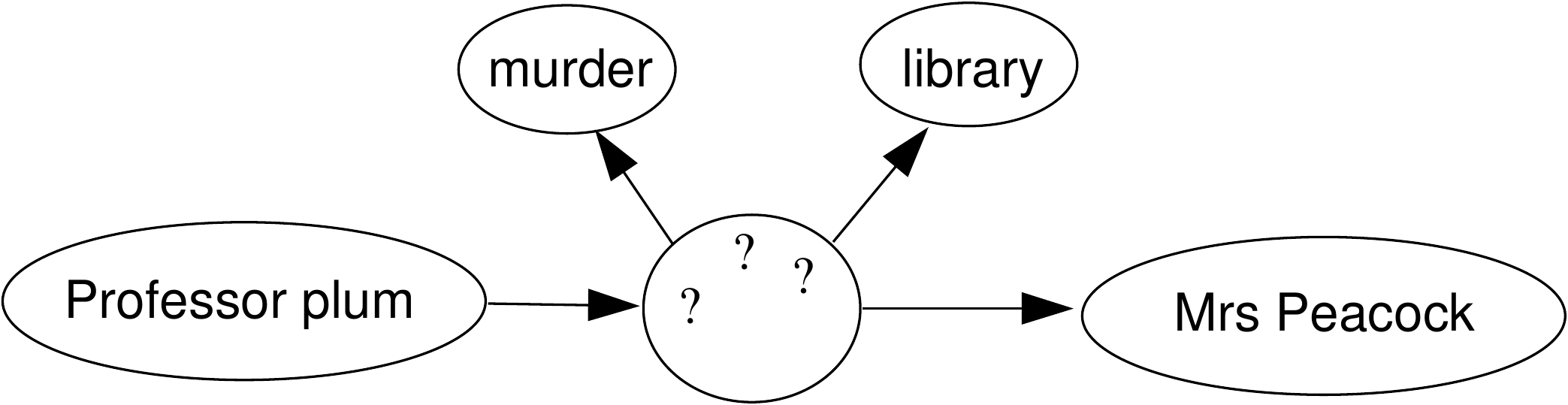}
\caption{\small Embellished association is an intermediate routing hub or switch
that allows simultaneous activation of the concepts. The hub plays a key role in
knowing such complex associations.\label{embellished2}}
\end{center}
\end{figure}
The only natural object with this structure, which does would not
violate the `no go' theorem for bi-partite causaton, is the context
aggregator itself, which would build the embellishments as context.
This is not yet well understood, at the time of writing, but should
become clearer with some model building\footnote{Experiments are
  underway in the `percolibrium' project.}.  The conditional
activation decided by the hub could be based on a summation threshold
value or it could be a binary logic. In a computer system, binary
decisions are assumed.  In biochemical processes, they seem
unlikely\footnote{Since we cannot extended the number of these
  simultaneous connections indefinitely, it suggests that there is a
  limit to how many simultaneous concepts can be processed in one go,
  unless they are parallelized, meaning that a Turing machine trying
  to simulate the knowledge process would take longer to understand
  complex associations than simple ones, proportional to the number of
  concepts involved.}.

\subsection{Context as a conditional switch for associative memory}

A knowledge system needs to be able to map associations to contexts,
in order to differentiate and nuance usage, so a semantic space must be able
to handle context-specific adjacencies, both to distinguish and
coordinatize multiple meanings.  To realize this, associative memory
must be switchable.  A switch is a device that exploits boundary
conditions in order to link or unlink spatial regions. We can imagine
context as acting analogously to the bias current of a transistor:
\beq
{\rm Collector} &\promise{\rm +current|bias}& {\rm Emitter}\\
{\rm Concept} &\promise{\rm +related|context}& {\rm
  Exemplar}\label{transistor} \eeq Context thus becomes a modulator
for permanent conceptual and associative memory, in accordance with
the conditional assistance law\cite{promisebook}.  From
(\ref{transistor}), and the rules of promise theory, we see that the
promise to deliver context information must be kept by a distinct
agent in order to assist in the keeping of a conditional promise.  As
a model for this, we return to figure \ref{awareness}, consider how
this makes sense\footnote{Any superagent capable of assembling a
  working context can use it as a coded memory address in a routing
  fabric analogous to a Banyan switch\cite{banyan}.}.

\subsubsection{Conditional association and structure of context}

Conditionality breaks translation symmetry explicitly, by introducing
boundary conditions which modulate the concept space creating
inhomogeneity, and it is associated with switching; thus, now we see a
direct link between switching, memory, and knowledge representation at
a fundamental spacetime level.  Consider associations of the form:
\beq C \assoc{\text{association}|\text{conditional context}} C'.  \eeq
The associational link is now considered only to be a link if the
condition is part of the context of the observer.
\begin{example}
In tabular form:
\begin{center}
\begin{tabular}{|l|l|l|l|}
\hline
From-Concept & Association & To-Concept& Context \\
\hline
\hline
Hot & characterizes & Spain& weather \\
Hot & characterizes & Chilli& food \\
Hot & characterizes & Sarah& sexual attraction \\
\hline
\end{tabular}
\end{center}
\end{example}
We see that concepts may have many interpretations that are labelled
by context.  In ontology, e.g. in topic maps, associations and
concepts are classified into types; this approach was criticized, as
it leads to static, impositional data models that easily form contradictions to
invalidate themselves\cite{topicmaps,burgesskm}.  An approach using
conditionals, where the current context belongs to a set of token
characteristics, addresses the deficiencies of a type system, though
it leads to more complex graphs\footnote{The ideas is similar to the
  demotion of static schemas in database technology and their
  replacement with schema-less data formats.}.

\subsubsection{Constructing embellished or contextualized associations}

\begin{figure}[ht]
\begin{center}
\includegraphics[width=8.5cm]{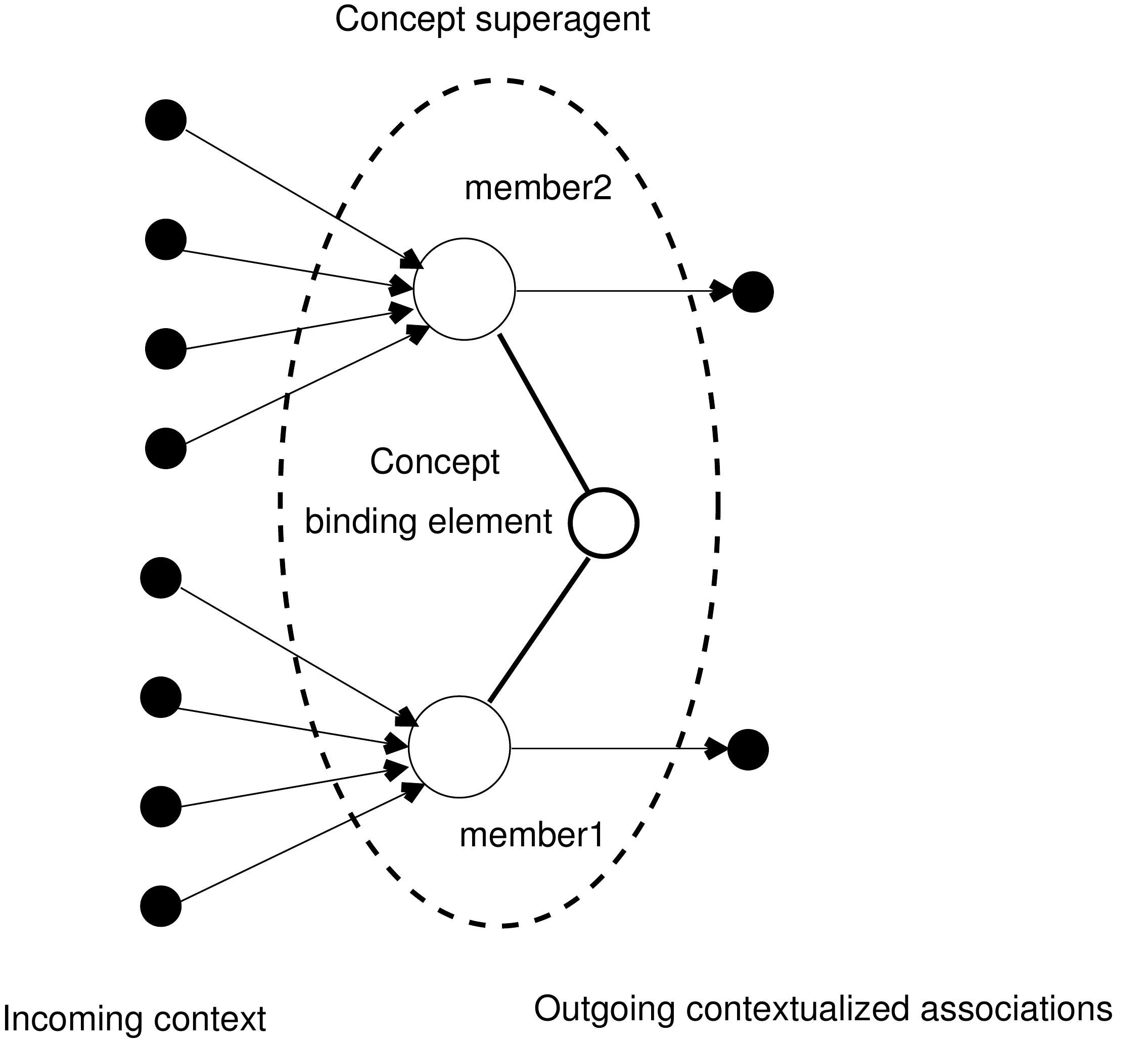}
\caption{\small A concept forms as a cluster of several
aggregating switches, joined through a binding basis agent. \label{binding}}
\end{center}
\end{figure}

The need to represent the kind of contextualized associations as
described in the previous section, has structural implications for
concepts. 
\begin{figure}[ht]
\begin{center}
\includegraphics[width=12.5cm]{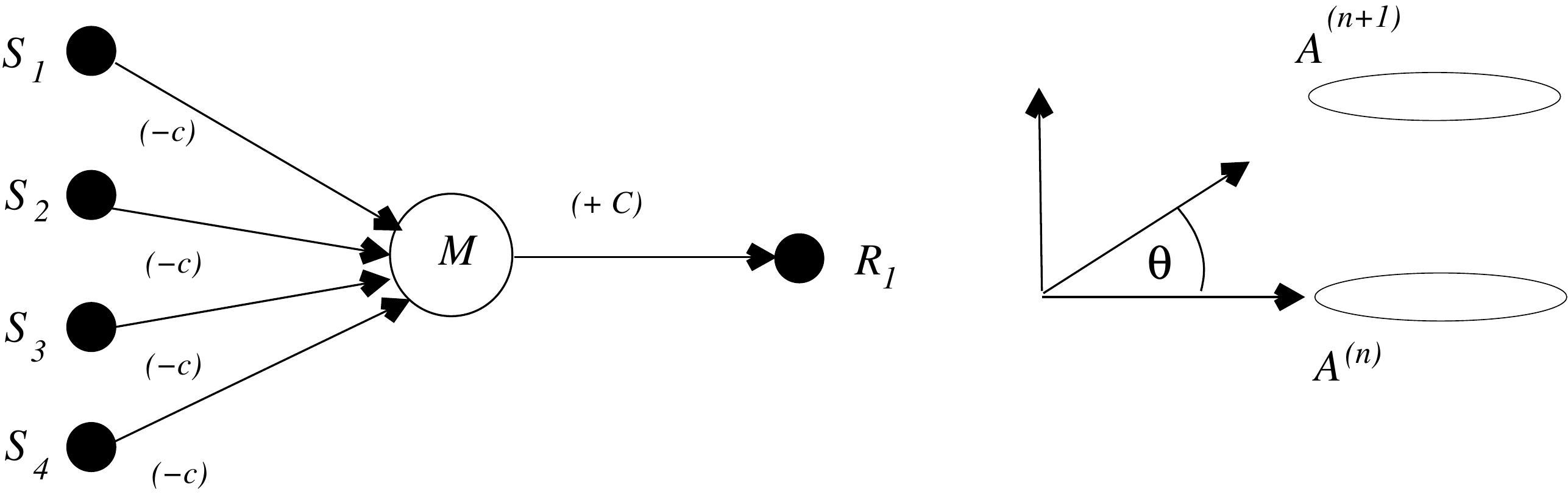}
\caption{\small An aggregating agent, whose promise to $R_1$ depends
  on the promises from a number of source agents is an ambiguous
  semantic situation. An aggregation of equivalent sources is
  essentially a membership relationship in which the agent $M$ plays
  the same role as a generalization, mapping from level $A^{(n)} \rightarrow A^{(n)}$
or from level $A^{(n)} \rightarrow A^{(n+1)}$. \label{switching2}}
\end{center}
\end{figure}
If we assume the existence of aggregating agents, such as the one
shown in figure \ref{switching2}, then we see that the inputs to an
agent from different contextual triggers begin to dominate the links
to the concept superagent, shown by the dotted line in figure
\ref{binding}.  For each outgoing association from the single concept
to other concepts, conditionalized by context, there must be an
incoming bundle of promises to accept context inputs. This suggests a
potentially large asymmetry between inputs and outputs that places a
burden on a single agent. A hypotheical resolution would be to build a
concept bundle from interior subagents, say one for each association.
A concept with, say, two outgoing associations (shown by the dotted
line in figure \ref{binding}), would then have two such basic
aggregating elements. This generalizes to more.  This, in turn,
suggests that the concept forms as a cluster of several such
aggregating switches, linked through a binding basis agent (see figure
\ref{binding}), which demarks the single concept as a discrete entity.

For each association, promised by the concept agent, there are likely
to be several context triggers, arising from different stages of the
sensory inputs chain. Given that agents have limited internal
capabilities, there is also likely to be a limit to the number of
contextual inputs that can condition an associative memory link.

\subsubsection{Causation and aggregation are not completely orthogonal}

The irreducible associations are not without some overlap or
ambiguity.  Consider the case of an aggregating agent, like the one
discussed in figure \ref{switching2}. This is a structure that is key
to representing conditional promises, and thus a contextualized
associations of the form:
\beq
M &\promise{+C | c_1,c_2,c_3,c_4}& R_1\\
M &\promise{-c_1,c_2,c_3,c_4}& R_1
\eeq
The triggering context $-\{c_1,c_2,c_3,c_4\}$ is aggregated by $M$ in order to
activate the promise $+C$ of an association between $M$ and $R_1$.
Another interpretation might place $M$ in the role of a generalization
of the concepts $S_1,S_2,S_3,S_4$. The ambiguity arises because a generalizing aggregate
concept may be considered to be caused (in some sense) by the members that
belong to it. However, aggregation usually presupposes that the $S_i$ are symmetrical
in their semantics.
\begin{itemize}
\item {\bf (Symmetry of contexts)} Each instance leads to (has a causal influence on) the existence of a
generalization.
\item {\bf (Asymmetry of contexts)} A dependency or contributing ingredient is not generalized by the
combination of contributors.  
\end{itemize}
Thus causation is a more primitive association, in some sense.  The
main difference is that generalization requires the ingredients to be
symmetrical in their semantics. The members of a group need some
unifying property (such as a basis set or matroid element). In
practice, a generalization is a membership in a superagent $M$ whose
representative basis is the superagent itself. For source agents that
are symmetrical in their semantics, one could imagine a transformation
from one type of association to another, formed as a superposition of
semantics, represented by the angle $\theta$ in figure
\ref{switching2}. This transformation does not invalidate the
irreducibility of the coarse graining association. Indeed, it is
natural that causal influence would be able to traverse concepts
belonging to different levels of aggregation. Introspective association
would tend to level the hierarchy of concepts.

Finally, it is worth noting that the law of dual or complementary
promises\cite{promisebook} indicates how $\pm$ can be exchanged,
turning what is context into what is concept, or vice versa.  Thus, in
a reasoning situation, the roles of context and representative
concepts can easily be thrown into reverse.

\subsubsection{Information argument for separation of context}\label{contextsec}

To be more precise about the multiple stages of conceptualization, and
their timescales, from quick the identification context tokens to
long-term associative clusters (in different kinds of memory
representation), we consider an argument based on intrinsic
information, as measured by the Shannon entropy.  The Shannon entropy
is a measure of information
efficiency\cite{shannon1,burgessbook2}. Both short term context and
long term memory contain information, in the form of
internal links and concepts. We don't need to now specifically how
information is encoded, but we can assume that it is quantifiable, through a
symbolic alphabet, as an informational entropy.
\begin{definition}[Shannon entropy of a memory configuration]
  Let ${\cal A}$ be some associative configuration, formed from an
  alphabet of links between concepts, in a graph ${\cal
    A}_{ij}$ of all possible concept agents $C$ and superagents $S$,
  where $i=1,\ldots,C+S$, and let $p({\cal A})$ be the probability of
  links being active.  We may define the information of this
  configuration to be: \beq S({\cal A}) = -\sum_{i < j=1}^{C+S} \;
  p_{ij} \log p_{ij}.  \eeq
\end{definition}
This intrinsic information may be used to compare the amount of
information in short-term context memory with the amount of
information in long term memory.\footnote{Physicists and computer
  scientists define information differently, in a subtle way.
  Information in physics is about uniqueness of an experiment within
  an ensemble of experiments, so it measures the uniformity of
  statistical patterns across many episodes (external time).
  Information in Shannon's theory of communication is about uniformity
  of states within a single episode or sample, thus it measures the
  uniformity of patterns within a single experiment (over internal
  time). This is why information scientists would say that information
  is entropy, and why physicists say that is it the opposite of
  entropy!}  An agent or collection of agents responsible for context
classification takes observations about the state of the world, at
each clock tick of the knowledge system, and labels a set of states,
which are used to predicate new associations between pairs of
concepts. Let's define these collections of agents using
superagents at a large scale:
\begin{definition}[Adaptive knowledge system components]
\beq
{\cal S} &=& \text{Sensor input agent graph}\\
{\cal C} &=& \text{Context agent graph}\\
{\cal A} &=& \text{Knowledge/Concept association agent graph (sum of all persistent memories)}
\eeq
\end{definition}
We can use information to make a strong statement:
\begin{hype}[The information resolution of context must be less than that for knowledge]
  A context-switched knowledge system, capable of distinguishing
  non-trivial contexts, with introspective feedback, divides into two
  independent agencies $\cal C$ and $\cal A$, representing short-term
  context and long-term associative memory, where $S({\cal C}) \ll
  S({\cal A})$ and the timescales for acquisition $T({\cal C}) \ll
  T({\cal A})$. This is necessary to have differentiated classes of
  associated knowledge determined by context.
\end{hype}
The proof of this lies in the following considerations:
\begin{itemize}
\item Suppose that the context memory were the same agency as the long
  term memory plus the new sensory input, i.e. ${\cal C} = {\cal A}$.
  The feedback from $\cal A$ onto itself is now indistinct from the
  ordinary associations, and thus the use of total knowledge as a
  switch for knowledge context is an automorphism with no class
  distinctions. This memory must be homogeneous, and therefore does
  not allow for persistent context distinctions. In other words, it
  cannot remember distinct associations as conceptual entities,
  because ${\cal C} = {\cal A}$.  This means we can rule out
  $\cal C=A$ in a contextual associative system, and there must be
(at least) two agencies.  A further
  consequence of this case is that the timescale for context
  evaluation must be the same as the timescale for knowledge
  acquisition and retrieval. Since long term knowledge requires
several iterations of sampling over a context to stabilize, with introspection
involving both ${\cal C}$ and ${\cal A}$, $nT({\cal C}) < T({\cal A})$, for some number of
iterations $n > 1$.

\item Next, suppose instead that $S({\cal C}) \not= S({\cal A)}$. The
  average amount of independent information in each context is then of
  the order $S({\cal A}) / S({\cal C})$ in each context, allowing
  contexts to be inhomogeneously encoded scenarios within the total
  amount of long term knowledge. In order to contain a non-trivial
  amount of context specific knowledge, we therefore need $S({\cal C})
  \ll S({\cal A)}$, implying that whatever agency keeps context, its
  capacity for concept resolution must be low compared to long term
  memory. If this condition were not met, the contexts would merge
  into one another, and lose distinction, eliminating smart
  behaviour\footnote{This may well be an effect that happens as the
    memory of a smart observer reaches its full capacity, or a critical percolation
threshold between contexts is reached.}.  With a lower
  information content, context can be addressed and operated at a
  separate and faster timescale than long term memory processes,
  leading to fast short term memory, and slower learning of knowledge
  $T({\cal C}) \ll T({\cal A})$.
\end{itemize}

\begin{corollary}[Short term context is faster than long term association]
If $\cal C$ is a distinct agency that is a conditional prerequisite
for switching $\cal A$, then the timescale $T({\cal C})$ for a transition is ${\cal C}$
must be less than or equal to the timescale $T({\cal A})$ for a transition
of ${\cal A}$'s configuration.
\end{corollary}
Context is an agent's assessment of the present, or its `situation
awareness'; this may include an interpreted `emotional state' for the
agent; these structural entities can all be derived from basic
spacetime considerations, in a representation independent
way\footnote{There has been some speculation about the nature of brain
  consciousness as a state of matter, that makes some connection with
  spacetime notions too\cite{tononi,tegmark}. That work goes far
  beyond the scope of the present work however.  The extent to which
  the present structural concerns are represented in brains remains to
  be seen, though there does not seem to be any obvious contradiction
  as far as I can tell.}.

\begin{example}[CFEngine's model]
  The CFEngine software\cite{burgessC1} makes use of a
  strongly similar system, defining classes based on environment
  discovery, and using hints about what the user is thinking, e.g.
  {\tt -Dclass}. This replaces a more familiar form of reasoning by
  first order branching logic. Sensory inputs may be extended
  indefinitely to discriminate very specific states.  Thus $CC$
  represents a kind of short term memory for what the operational
  system of machines and human operators is collectively thinking about,
  or focused on.
\end{example}

\begin{example}[Immunology model] Context is provided by antigen
  presenting cells.  Long term memory is provided by the populations
  of epitopes on antibodies in lymph nodes and in circulation as
  memory cells\cite{lisa98283,perelson1}.
\end{example}

\begin{example}[Genetics and epigenetics] 
Genetic encoding operates at two levels: there is long term memory encoded in
gene sequences on chromosomes. Then there is an environment of protein
networks that activate genes at specific moments. These encode short term
context, within an organism as a learning system.
\end{example}

\begin{example}[Do neural networks have fast and slow memory?]
  In an artificial neural network, layers of neurons are connected
  into a high density feed-forward network, in which propagation
  occurs at a single speed. This fast propagation acts as a context
  encoding for conditioning the long term memory tokenization.  Once
  trained, this means that a neural network has only a single
  timescale.  However, when the network is learning, it feeds sensory
  data forward as context, to be adjusted by back-propagation over a
  longer training timescale. This is the separation of timescales for
  ANN training.  Once the network is trained, it operates purely as an
  advanced multi-dimensional sensor at high speed, similar to the way
  an eye can process visual signals quickly because of millions of
  years of evolutionary learning.  Similar to the eye, once training
  has stopped, the ANN cannot learn anything new, without a separate
  slow process of adjustment, thus ANNs do not contradict these
  considerations.
\end{example}

The implication of conditional promises in context switching reaffirms
the basic separation of learning into two timescales: a short-term
evaluation or classification of the current `state of mind' in a
thinking process, and a long-term linkage between concepts, which can
be paged in by the short term modulation.  The relative information
also supports the notion that knowledge would converge, by feedback,
to a tokenized representation, separate from detailed sensory input.
The question remains, however, as to whether context has to be a
completely separate memory, or simply a distinct agent that transforms
long term memory into context?  From the foregoing considerations, it
seems that $\cal C$ would be a coarse graining of both $\cal A$ and
sensory input $\cal S$?

\begin{figure}[ht]
\begin{center}
\includegraphics[width=10.5cm]{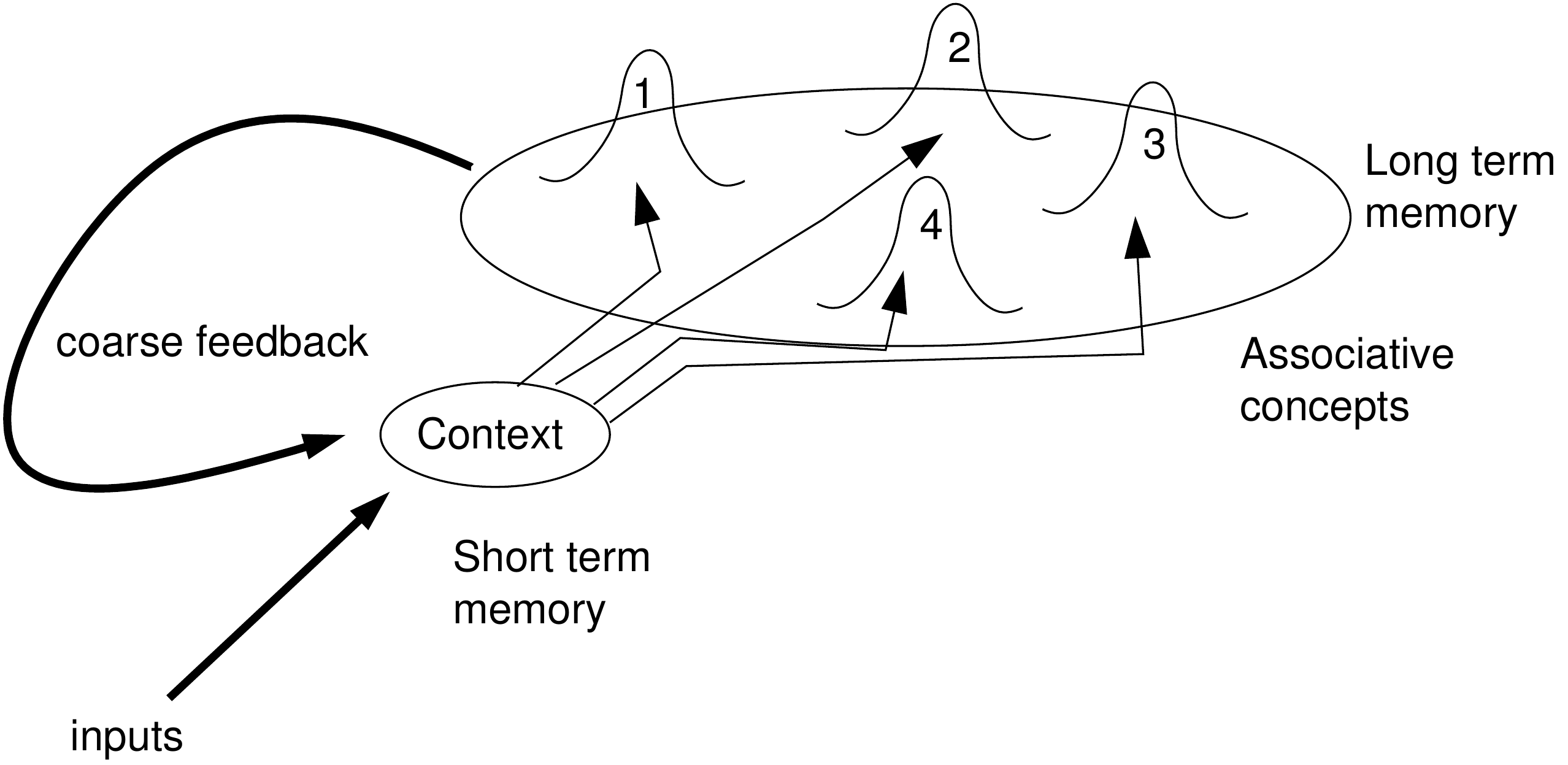}
\caption{\small Context discrimination leads to different versions of
  a concept, according to sensory input and coarse feedback. Context
  assessment needs to be fast and disposable, i.e. short term.  The
  formation of associations, or contextual relationships to other
  memories and concepts, can be slower, to achieve higher certainty.
  The integration of experiences into a network of associated concepts
  could perhaps be interpreted as a form of quality control.  Once
  concepts have been learnt, over long times, they too can be used to
  build more robust contexts, feeding back into the perception of new
  states. \label{contexts}}
\end{center}
\end{figure}

In an introspective system, a component of context could derive from
fed-back long-term associative knowledge, along side the
interpretation of direct observations (see figure \ref{contexts}). However, it also
seems likely that context has a role closer to observation than to
accumulated knowledge.  The question of how knowledge is bootstrapped
provides some direction here.  A memory that has no initial concepts
still has to be able to generate context according to this
architecture. This suggests that context is closer to sensory
interpretation than to long term memory. That hypothesis is further
backed up by the argument that contextualization has to happen at
approximately the same timescale as observation; thus, even with
feedback from long term concepts, context is the imprint of the
sensory stream onto specialized knowledge.
Thus, promise theory makes a simple prediction:
\begin{hype}[Separate agent for context adaptation]
  Context needs a distinct and internally coordinated agent, of
  unspecified structure, that promises to quickly summarize the
  current state of the the external sensory input and recent history,
  for use in selecting the relevance of context-dependent
  relationships between the representations above. The agent may
  promise only limited capabilities, meaning that context has to focus
  on particular parts of the sensory input in practice.
\end{hype}
Because the only mechanism for coupling of concepts is their
simultaneous activation, and activation is assumed to be modulated by
context, the prediction is that conceptual memory is built on a
hierarchy of timescales, not on a hierarchy of a priori semantics.  In
a knowledge space, associational adjacency is conditionality on a
context.  It is thus entirely reasonable to expect to find different
kinds of memory for different purposes, supporting different
recognition mechanisms and timescales just as we make different kinds
of data store in computing.

\begin{example}[Scientific method as a learning system]
The scientific method is a learning system that acts at the level of a society.
\begin{itemize}
\item $\cal S$: the sensors are the scientists who study the world. 
\item $\cal C$: the context is the papers and the theoretical and experimental results.
\item $\cal A$: the long term knowledge is the cultural memory and literature.
\end{itemize}
At the scale of a society, the observation horizon is formed by a
changing population of scientists, who produce noisy inputs, because
they are constantly in flux, like the disposable cells in a single
organism. Since scientists come and go much more quickly than the
literature $\cal A$, the results could appear unstable. However, there
is a stability mechanism for $\cal S$ arising by social interaction.
This leads to norms and cultural inertia, which may drift slowly over time,
but offers a more evolutionary timescale that allows $\cal A$ to
stabilize.
\end{example}

\subsubsection{Context atoms: coarse grained compression to awareness of `now'}

To remain viable, as a fast switching discriminator, a context
classifying agent must assemble from input and long term memory and
compress or tokenize state in `real time'\footnote{Realtime is
  self-defining here, as the perception of the observer is only as
  fast as it can evaluate context. Evolutionarily, it can only adapt
  to its environment and be `smart', if its internal clock is at least
  as fast as its environment.}. The coding of irreducible
associations, discussed in section \ref{qtassoc}, enable long term
concepts to aggregate hierarchically (though not mutually exclusively)
into superagent generalizations.  The names or identities of these
generalizations would be suitable compressible coarse grains for use
in context too. In a similar way, input sensors could perform a
partial coarse graining automatically as part of their function.  This
is how one may dissect deeper specialized knowledge into multiple
categories, in real time, without them becoming so many as to merge into one
another, i.e. $S({\cal C}) \ll S({\cal A})$, from section \ref{contextsec}.  
We know, from spacetime physics, that structural stability
relies on the ability to separate scales by assuring weak coupling or
full decoupling between contexts\footnote{Some contexts might be
  triggered by all concepts, affecting all parts of memory equally.
  This might happen when a concept is not clearly related to an
  existing concept in the current context. Then context together with
  feedback from long term concepts cannot narrow a focus to a
  specialized area, and thus might tend to map to all or no areas.
  Such unspecific stimuli might be what animals experience as
  emotional sensations.}.

Samples are themselves signal agents, which encode data, and are associated
with the spacetime locations of the sampling points. Thus, given the assumption that the variety
and size of the external environment greatly exceeds that of the observer, samples
need a notion of context in order to support adaptation. They
cannot easily be attributed meaning without the annotation of context.

What might context atoms look like in practice? The table below
shows some examples that lend themselves to specific kinds of sensors:
\begin{center}
\begin{tabular}{|l|l|}
\hline
Coarse grained sensory context atom & sensor type\\
\hline
\hline
day, time, location           & spacetime\\
light, dark, red, green, blue & sight\\
hot, cold, free, constrained  & touch\\
salt,sweet, sour, savory, etc     & taste\\
\hline
\hline
open, closed & bridge\\
blocked, open, snowed under & road\\
available, unavailable & shop supplies\\
parking, parked, shunt, cruising & transport\\
inside, outside & sense of self ``me''\\
\hline
\end{tabular}
\end{center}
In spacetime terms, context is associated with coarse boundary
information about an observed scenario. Knowledge is accumulated
gradually by observation with continuous introspection. This is
integrated relative to an already accumulated notion of `self'.
Similarly, in the sensory data collection, one always associates the
concept of ``here and now'' with a sensed context, calibrated by the
long-term accumulated experience.  The evolving semantics of ``here
and now'', developing over time, are what leads to higher semantics of
associations, i.e. that which goes beyond `concurrent activation'. 

It seems likely, given the sheer volume of sensory inputs, and
possible feedbacks, that only a partial prioritized context would be
used to encode and retreive memories\footnote{We humans tend to focus
  one particular sense at a time, when we are reasoning; e.g. we
  concentate on sounds instead of sight, or touch instead of looks,
  taking these one at a time in order to `concentrate'.}. Thus, one
would expect a reasoning engine with limited processing capacity to
`miss' in comprehending certain concepts, if by chance a partial
context led reasoning astray. Concepts associated with coarse grained
context may well only converge into coherent clusters over many
averaged experiences and long times. Thus the alphabet of context one
could access, at any given moment, might be much smaller than the
actual potential resolution of a sensory input system.

\begin{example}[Some human context concepts]
The separation of human senses lead us to distinguish spatial concepts from other variations in the
environment. The elements of context may include measures that we do not consider to be spatial, but which
fit easily into a concept of semantic spacetime:
\begin{center}
\begin{tabular}{|l|l|}
\hline
Sense & Dimension\\
\hline
Sight &time, space, colour.\\
Sound &time, space, pitch, timbre.\\
Touch &time, space, composition, temperature, pressure, humidity.\\
Smell &time, composition.\\
Taste &time, composition.\\
\hline
\end{tabular}
\end{center}
\end{example}

\begin{example}
  A sequence of sampled music can be labelled at a variety of scales.
  The sampling of a piece of music is digitized (or packetized) into a
  sequence of standardized, temporally ordered samples that capture
  the dynamics of the sound. The semantics of these samples are that
  they promise a representation of music according to an agreed
  encoding convention that may be used by any promisee to reproduce
  the intended meaning of the performance. Each packet sample
may be labelled with:
\begin{itemize}
\item The time and location of recording.
\item The relative times (i.e. order) of the discrete sound samples, from the perspective of the observing agent.
\item The data values of the sampled sound representation.
\end{itemize}
To reproduce the music, we don't need to know the spacetime location
of the original sampling process, but we do need to know the relative
spacetime arrangement of the samples, so as to retransmit them faithfully
over the scale of the total sample, during playback. This suggests a minimal
data representation of the form
\begin{quote}
  {\tt (ordinal sequence label, data value)}\\
  i.e. (time stamp, digitized sound frequency, amplitude)
\end{quote}
\end{example}

\subsubsection{Spacetime template patterns for efficient recognition and recall}

Optimizing the use of space and time to to encode specialized
patterns, suggests that certain patterns or `fabric' topologies for
memory might assist in the identification and discrimination of
sensory forms.  Directional patterns could benefit from directional
sensors; visual sensors that measure relative positions would benefit
from a coordinate grid representation, etc. This is one more way in which
the structural properties of knowledge are affected by the structural
properties of spacetime: they are constrained by them. The earlier
discussion of grid cells, place cells, directional cells in biology
are an example of this (see section \ref{topofit}).

\begin{example}[Efficient fabrics for mapping semantic context into long term memory]
The Sparse Distributed Representations
(SDR)\cite{sdm1,sdm2,hawkins, hawkins2} provide a plausible (even
likely) mechanism that could evolve in biological systems.  The
connection between running timelines and context are best developed in
studies of linguistics\cite{feldman1,langacker1}, and the notion of a
state of mind in contemporary artificial knowledge systems need not be
extrapolated too far to explain the whole of free will and
consciousness\cite{dennet1}, but it is clear that there are common
characteristics that open an avenue of plausible investigation.
\end{example}
Directed networks, including Bayesian networks and neural
networks\cite{duda1,bayes} are a natural area of study, as they form
generic discriminating structures. As branching, rather than
converging processes, they may not be efficient, but they can be
generic, and they automatically route connections.  Routing structures
are an attractive model for knowledge addressing, because a flooding
of a bus would be inefficient and would block the system with a strong
coupling to a single timescale $T(\cal A)$ (see section
\ref{learnassociation}), with inputs constrained to the width of the
context bus.

\subsubsection{Stability of contextual knowledge and weight maintenance (sleep functions?)}

It seems plausible that `state of mind' may be considered a fair
synonym for context: i.e.  a collection of contemporaneous or
coincidental signals and concepts associated by simultaneous or
concurrent inputs.  The number of sensory inputs feeding into that
state thus plays a role in determining whether a rich enough context
can be found to refining and tokenize concepts, and provide robust
conceptual representation in long term memory; similarly in
determining the number of independent degrees of freedom by which
sensations can be felt or imagined by the observer\footnote{It seems
  plausible that emotional sensation is no more than a particular
  state of mind.  In an organism with myriad sensors, an emotional
  sensation could be complex, an approximate facsimile of that entire
  complex sensory apparatus. For a system with few sensors, an emotion
  would be little more than a feeling of hot and cold. In humans, emotions
can feel intense, presumably for this reason.}.

\begin{lemma}[Concepts are distributed]
Different aspects of a `thing' may end up being encoded in quite
different parts of a semantic space (e.g. a brain). Localization of
concepts is not guaranteed.
\end{lemma}
If multiple sensory inputs bring simultaneous activity into data
summarization, different impressions may not all converge to a single
localized region to form a coherent representation as a `thing', and
an observer equipped with a knowledge bank will be thrown into
confusion. Thus, a knowledge representation needs to be able to handle
multi-dimensional parallel assessments of context as an addressing
scheme for concept storage.
\begin{figure}[ht]
\begin{center}
\includegraphics[width=10.5cm]{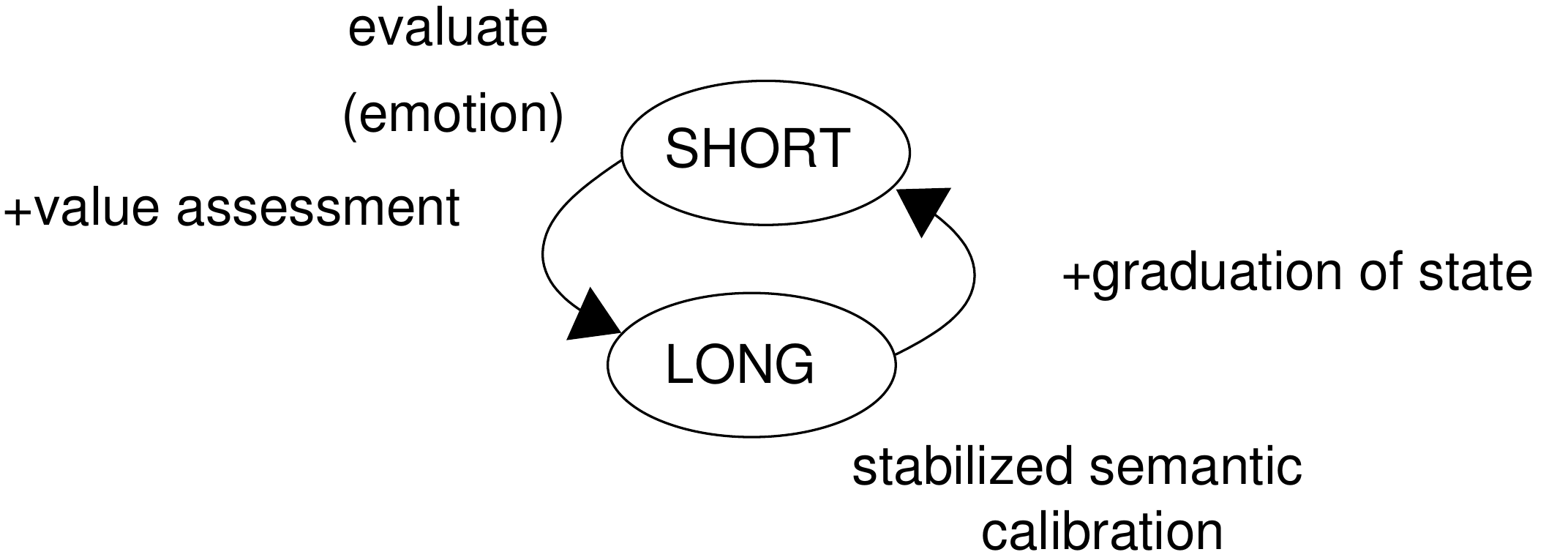}
\caption{\small Self-calibration, where previously learned patterns are used as a measuring stick for
new observations, leading to a convergent result. If there is no convergence, we remain
confused about measurements.\label{selfcalibration}}
\end{center}
\end{figure}

Because context acts either like a multi-switch, or as a network of
signal concentrations that form a basis for
routing like in ANN, the routing mesh needed to coordinatize concepts
has the structure of a directed graph, from the simplest signal wire
to a very complicated routing fabric.
Two dynamical processes may be at work in associative knowledge
representation: iterative learning (cyclic repetition, as in figure
\ref{selfcalibration})\cite{burgessDSOM2002,burgessC14}, as in section
\ref{learning}, and directed flow through discriminating fabrics. The
latter have been studied in \cite{graphpaper}.
What is known, from such studies of the underlying graph dynamics, is that
all of these structures and processes may be unstable to perturbations.
\begin{lemma}[Instability of associations]
Knowledge may be unstable to perturbations.
\end{lemma}
\begin{figure}[ht]
\begin{center}
\includegraphics[width=10.5cm]{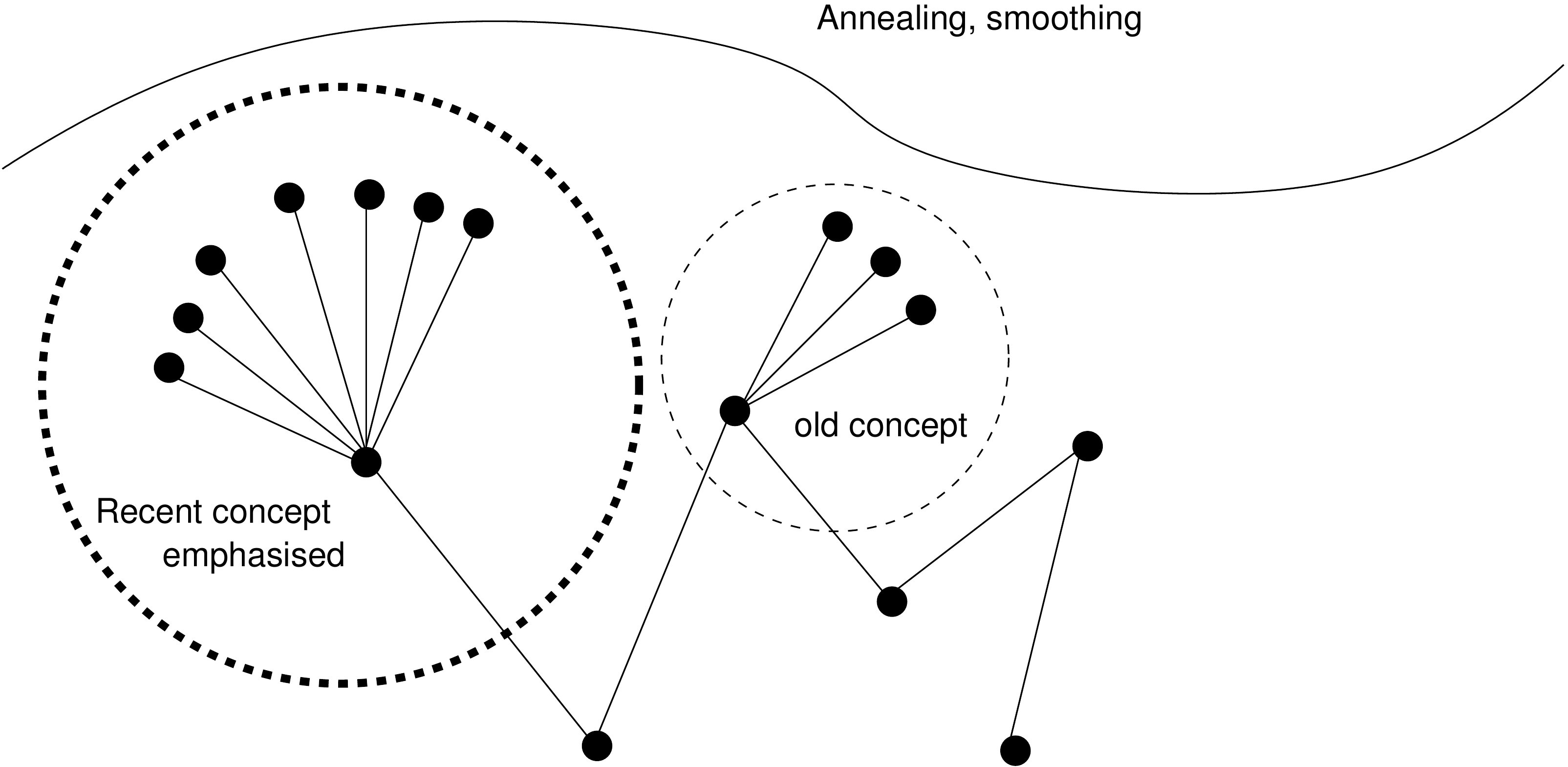}
\caption{\small De-emphasizing recent memory to stabilize the
  associative graph, by locally diffusing importance scores
with post-processing (like stimulated coherence in physics).\label{smoothing}}
\end{center}
\end{figure}
Instability is combatted by regular maintenance, at a spacetime scale
that matches potential perturbations. This is the essence of Shannon's
error correction theorem, and the maintenance
theorem\cite{burgesstheory}.  In sensory learning and context
awareness, it makes sense for a knowledge system to bias inputs in favour of
what is happening right now, rather than paying too much attention to
the past. However, if taken to the extreme, focus on too short a
timescale would lead to ignoring long term experience in favour of ad
hoc spikes of activity in disparate locations.  Agents that
experienced many different contexts over long times, but only a few on
short times, would be constantly destabilized by a fickle accumulation
of new unbalanced inputs. In order for memories not to become too
biased in favour of the present, some form of annealing or smoothing
of the memories may be needed to compensate for these biases (see
figure \ref{smoothing})\footnote{It is fascinating to speculate as to
  whether this is the (or one of the) function(s) of dreaming in
  animals.}.
\begin{hype}[Annealing of knowledge]
  A form of low frequency (long wavelength) process that is de-focused
  relative to the spatial scale of a concept representation (so as not to cause damage to
  individual memories), is needed to smooth out associative link
  strengths in a dynamically learning knowledge representation, to
  apply a fair weighting policy on memories.
\end{hype}
The need for such a process makes complete sense from a spacetime
viewpoint, but semantically it seems speculative. It is only by
unifying both viewpoints that it becomes a very plausible
stabilizing maintenance mechanism for a learning system, analogous to time series smoothing\footnote{In
  the field of cybernetics, the idea of building a brain as a
  controller has been discussed for nearly a century, in different
  forms. Ashby, in particular, discussed the notion of designing a
  brain based on the continuum feedback principles of his
  day\cite{ashby2,ashby1}. In his work, he notes the separation of
  reflex responses and centralized control feedback.  He also
  identifies the reliance of learning on a selection criterion that
  leads principally to stability of knowledge. This suggests a
  separation of scales balancing dynamical convergence with
  modulation, very like the discrete version here.}.

\subsubsection{Sensory perception: from context to reasoning}

We associate intelligence with adaptation based on {\em reasoning}.
It is therefore instructive to think about how tokenized context can lead to
knowledge representations that support reasoning.  These forms
should emerge, rather than being imprinted by design. This topic
is a bridge to the next section about automated reasoning and
inference from knowledge representations.

There are two obvious reasons for wanting to study the emergence of
reasoning structures.  The first is pragmatic: the less supervision a
learning process requires to think about context, the more impartial
and valuable it is. The second is more related to understanding how
our arbitrary semantic spacetimes (including brains) may actually
work: no one is obviously born with a large hardwired conceptual
ontology, so we have to explain how ontologies emerge through
learning\cite{burgesskm}.

The semantics of spacetime meet the semantics of observational input
in the concept of streams. Streams are ordered samples of data that
form the basis for data acquisition, repetition, and thence stable
knowledge (figure \ref{modulation}).  A key question, in the formation
of practical knowledge representations, is: how can we automatically annotate the
a collection of data, to bring about an automatic emergent semantic model,
i.e.  not design an information model, but a model of associative meaning?
We return to this practical subject more fully in section \ref{annotation}.
The process has to involve a series of dynamical transformations, 
which correspond to spacetime structures. Schematically: 
\beq
\text{Sensing} \rightarrow \text{Agentization(basis)} \rightarrow
\text{State} \rightarrow \text{Association graph},
\eeq 
i.e.
\beq 
\text{Arrival process}
\promise{\text{semantic coordinatization}} \text{Matroid basis}
\rightarrow \text{High level graphs}.
\eeq 
New impulses are recorded and recognized, then they are consistently
routed to certain `empty' memory agent locations. These take on the
role of proxies, in order to represent the incoming signals and
eventual join them up into concepts\footnote{This process is sometimes
  called {\em reification} in computer science, but this terminology
  is closely associated with the Resource Description
  Framework\cite{rdf} for the semantic web\cite{robertson1}, and
  related technologies\cite{rizzi02complexity}, so I shall avoid using
  it.}.  Encoding of concepts and associations into a physical medium
is a spacetime process: it involves coordinatized spatial location of
agents for pattern encoding, and sequential as well as cyclic time
patterns for coding and decoding with repetition for iterative
confirmation.  It is a process by which scaled knowledge `symbols' are
constructed from pre-adapted sensors\footnote{A semantic `symbol' may
  be represented by any convenient clusters of agents, they needn't be
  represented as a non-overlapping alphabet of agents, though there
  will always be a transformation to such a picture, in principle.},
attending to a separation of timescales. Without such a separation, a
confusion of semantics would result.

Through the separation into context and associative concepts, we can
now explain how a set of coarse characteristics, named or labelled
quickly, may lead to the formation of a switched network, connecting
or isolating concepts only under specific circumstances, stimulated
from the environment and from introspective feedback. A bus of context
reaches into long term memory, acting much like a transistor bias, or
network routing fabric, but with semantic addressing forming the
interface to coordinate tuple addresses. Any spatial structure with
this quick approximate response to its surroundings can thus store
contextual memories.

In the next section, we consider how this leads to associatively
connected routes through a knowledge space. If one starts at some
context relevant location, and traces neighbouring concepts by
association, focused pathways and divergent trees (explosions from a
starting point) can spread out, activating related concepts in order.
This may be perceived as reasoning, explanation, or brainstorming,
depending on how focused the result seems to be. How we get from
associations to storylines and what we consider to be rational logic
is the subject of the next section.

\begin{example}[Named Data Networking (again)]
  The Named Data Networking example, used earlier\cite{ndn}, has the
  structure of a routing network, based on semantic addressing rather
  than on explicit spacetime coordinates, but does this make it also a
  representation of concepts, i.e. a knowledge representation (see figure \ref{ndn2})?
\begin{figure}[ht]
\begin{center}
\includegraphics[width=6.5cm]{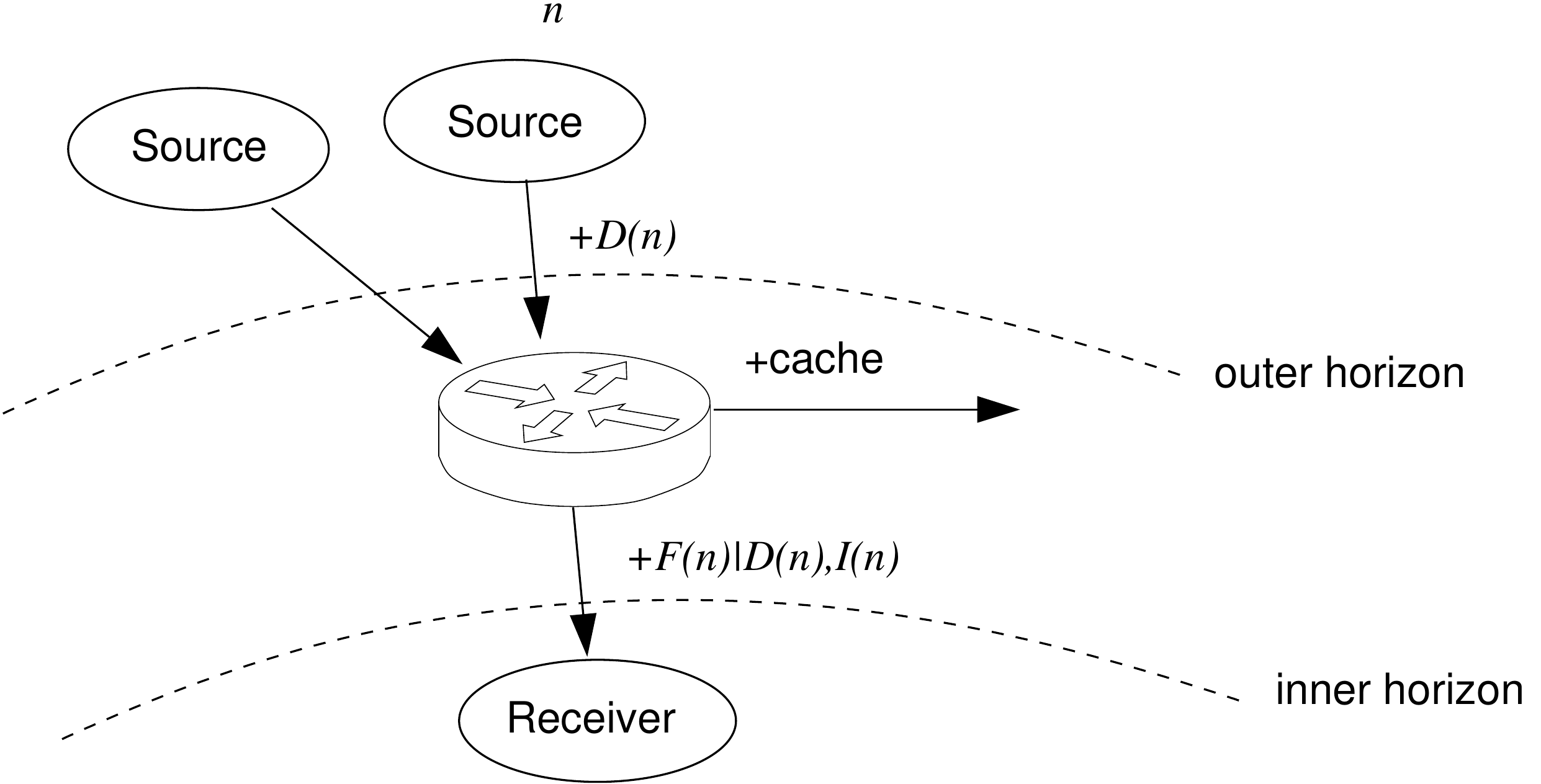}
\caption{\small Is Named Data Networking (NDN) a knowledge representation supporting 
concepts and associations?\label{ndn2}}
\end{center}
\end{figure}
A knowledge representation lies inside a single observer.  Based on
the figure from example \ref{ndnex}, we may draw two horizons that
represent where we consider the observer to be, in alternative
interpretations of NDN as a knowledge space. We can say immediately
that NDN forms a fabric that switches data based on specialized
sensors.  

The observer sensors are the sources of data that transmit data by
name. It is most natural to choose the outer horizon as the boundary
of the observer.  Thus the sum of all receivers is the `mind of the
observer', in this proposed mapping, i.e. `interest space' is where
concepts could be encoded.

If the router keeps transaction logs, then it has a kind of memory of
simultaneous activation of receivers that showed common interest in a
piece of data at the same moment. However, there is no exterior
correlation between data files, as there is no spacetime model
relating files and data streams to a single model, so the a meaning of
such a correlation would be dubious.  Because all data have to have a
unique name in NDN, the `sensors' (or data sources) are unlike the
sensors that provide data streams in organisms or knowledge systems,
as the named data never change: they are frozen in time.  However, we
could instead define sensors that point to aggregations of similar
things, e.g. different versions of a document, or chunks of a video
stream. The aggregation of these can be understood as the naming of
superagents in the observer's exterior.

Can we map concepts and associations to NDN structures?  A concept
would have to be mapped to a resting place for related data in the
space of receivers. The clustering of receivers with similar receipts
could be interpreted as a concept, however these are not linked except
by common routing (like coins sorted into categories by a machine).
An association would be a link between two concepts. There are no
structures that join the receivers together except the routers.
Correlations between the interests of autonomous receivers are the
most tangible form of semantic link. Unlike a true knowledge space,
there is no way to promise persistent associations. 
Data interest may be indirectly associated through the routing hub's
ephemeral promise to forward inputs, but these links are neither
persistent nor.  

In conclusion, NDN does not form a knowledge
representation, even though it is a semantic routing fabric.  It's
function is not be form a memory of the world, but to disseminate data
in a fluid gaseous state. Chance correlations could be observed
between receivers, analogous to concepts, but the lack of context or
permanent association prevents them from iterative associative
learning.
\end{example}

\subsection{Spacetime representation of scaled concepts}

Let's consider how invariant and tokenized representations emerge from
scaled agent clusters and irreducible association.

\subsubsection{Examples of irreducible positves}

To see how the irreducible types can lead to plausible encodings of
concepts, through spacetime associations, it is helpful to look at
examples.  By reducing to the most elementary forms, we end up with a
kind of pidgin language, but this seems like a good start that mirrors
how language begins. Let's characterize the irreducible associations
generically, without trying to nuance the expressions, to see if we
can make ideas comprehensible by reduction:
\begin{center}
\begin{tabular}{c|l|l|l}
Type & Verb & Prepositional association & Spacetime interpretation\\
\hline
1.& approaches & near & similarity/distance\\
2.& follows & before/after & causal/ordered\\
3.& contains & inside/around & scale/aggregation \\
4.& expresses & instead of &boundary structure (name)\\
\end{tabular}
\end{center}
Let's try to combine these into more complex concepts, br reducing
them to triplets of the form: \beq (\text{concept},\text{irreducible
  association},\text{concept}) \eeq These examples may not be the only
possible representations (if they were, they might be fragile), but
they are hopefully plausible and maximally simple, avoiding higher
tensor forms in the interest of low cost simplicity.
\begin{example}[Scaled conceptual cluster reductions for inference]
Consider the following linguistic representations of common
concepts:
\begin{enumerate}
\item {\em ``She is sitting on the roof''}

This is a description of a static atemporal state. We could argue that these capture the situation:
\begin{quote}
(she, near, roof)\\
(she, is contained by, roof)
\end{quote}
However, they cannot easily be distinguished from other similar events.
Two different versions irreducible concepts may apply to this
situation. Either is approximately true, and the superposition of them,
in provides more specificity. Being inside a roof might sound odd, but we
have no trouble thinking of someone being inside the boundary of a roof, as
a two dimensional space that marks a three dimensional volume. Our visual
apparatus is used to projecting images and concepts into two dimensional frames,
which has the effect of blurring these distinctions `metaphorically'.

A recursive approach offers a more precise representation than this
simple direct representation, and illustrates how grammatical recursion helps
to sort out the represenation of events.
\begin{quote}
(she, expresses, sitting on roof)\\
(sitting on roof, expresses the role, sitting)\\
(sitting on roof, expresses the attribute, on roof)
\end{quote}

\item {\em ``She climbs onto the roof''}

This is a description of a temporal process, leading to a future state.
\begin{quote}
(she, expresses, on roof)\\
(she, expresses, climbing onto)
(climbing onto, expresses role, movement towards)
\end{quote}
The superposition of these two association paths captures the idea.
Alternatively, a interpretation that includes causal transition may be
included:
\begin{quote}
(she, leads to, on roof)
(she, causes, on roof)
\end{quote}

\item {\em ``Acid burns stuff''}

This is a description of a temporal process.
\begin{quote}
(acid, leads to, burning)\\
(acid, expresses, burning)\\
(acid, leads to, burned stuff)
\end{quote}
The challenge here is in linking the concept of burning (a verb) to a
pair of things: acid and stuff. The use of adjective clusters `burned
stuff' as an independent agent is one solution. This has the form of a
superagent, linked by context.  One solution is the
use of aggregate superagent clusters to employ recursion:
\begin{quote}
((acid, near ,stuff), expresses, burning)\\
((acid, near ,stuff), leads to, burning)
\end{quote}

\item {\em ``Alice looks good''}

This is description of state.
\begin{quote}
(Alice, near, good)\\
(Alice, expresses,good looks)
\end{quote}

\item {\em ``The dog who loves adventure''}

This is an explication of state, with an additional twist.
\begin{quote}
(Dog, near, adventure)\\
(Dog, expresses,love of adventure)\\
(Love of adventure, has role, love)\\
(Love of adventure, has attribute, adventure)
\end{quote}
What is new here is that it introduces emotion into the expression.
The use of `love' may add weight or priority to this property above
others. It suggests a dimension of representation that is otherwise
missing in the simple spacetime focused discussion here.

\item {\em ``People must obey the law''}

This is an explication of {\em desired} state, which introduces
a temporal aspect pointing to the future.
\begin{quote}
(People, near, law)\\
(People, express, obey the law)\\
(Obey the law, has role, obey)\\
(Obey the law, has atribute, the law)\\
(Obey the law, depends on, the law)
\end{quote}
The notion of force, expressed by `must' is not a spacetime concept, but could
be captured by the final causal annotation.
It refers to a desire, which has an emotional component, and also
a future (temporal) component. This is an intention, a goal, or a desired
state.
\end{enumerate}
\end{example}
These formulations may not seem linguistically familiar, but
they are logically fair approximations to what we mean by more habitual
expressions. It is possible that there may
be value in distinguishing refinements of these wordings as independent
associative link types, e.g. refinements of the form
\begin{quote}
X `leads to' Y from X `affects' Y \\
X `expresses' Y from A `has property' Y 
\end{quote}
may be necessary distinctions or merely scaled refinements, in the
long run, in order to separate causal ordering from the notion of
generalized influence, but parsimony suggests otherwise at this stage
of understanding. What is important and striking is how few concept types are
needed to represent most conceptual scenarios, and how these shadow spacetime relationships.
More explanation of these cases, with examples, is given in \cite{cognitive}.

\subsubsection{Noun and verb concepts}

Some concepts suggest an understanding of time relative to contextual `now'.
\beq
\text{Future}\; \assoc{follows}\;\text{Context/Now} \;\assoc{follows}\; \text{Past} 
\eeq
Concepts may be both {\em things} or {\em processes} (nouns or verbs,
in linguistic terms). These have distinct semantics.  
Although, ultimately, everything must be encodable as spacetime
patterns, elemental spacetime concepts are not enough to represent the
full spectrum of meanings we would expect of knowledge
representations. However, they do take us much further than has
previously been realized.  

The results of the previous reduction exercise look remarkably close
in form and idea to the programme of cognitive grammar by
Langacker\cite{langacker1}.  Emotional priorities are another
dimension that needs serious consideration, even if these are
indirectly encoded through spacetime structures too.

\subsubsection{Irreducible positves and negatives (encoding NOT)}

The well known aphorism that `absence of evidence is not evidence of
absence' is a a basic property reflected in promise theory in the
form: the absence of a promise is not the promise of an absence. The
same is not true of complementary sets, or the Boolean algebra used in
information technology, and leads to a sometimes parodically
deterministic belief that absence of information is simply not even
taken as an option.

The human interpretation of `not' we seek is not a notion of set complementarity, but rather
a complementary intent. We could express this by:
\beq
\text{human NOT} \assoc{\text{expresses NOT}} \text{body set complement}.\label{prnot}
\eeq
This is easy to see. If we divide the world into
that which is carrots and that which is not carrots, then promising to
not like carrots is not the same as promising to like the complement
of carrots (everything else). Thus, when we say `$X$ is NOT a
generalization of $Y'$, we do not mean that $X$ is a generalization of
the complement of $Y$ (NOT $Y$); rather, we intend to add a specific
note to disregard the assertion that $X$ is a generalization of $Y$
from consideration.

The question arises as to how we represent this in a knowledge network.  Promise
theory guides us towards the form of (\ref{prnot}), by defining the a
promise of NOT $b \equiv \neg b$ to be the deliberate intent to not
keep the promise of $b$.  This guides us to represent NOT on a par
with the positives, i.e. for each positive irreducible association,
there is a mirror association that expresses the negative.
\begin{lemma}[Negative association]
For each promisable association in the irredicuble types, there is
a corresponding association type formed from the 
negation of the association promise that represents a denial of an association.
\beq
C_1 &\assoc{\cal A}& C_2 \\
\forall {\cal A} \exists {\cal A'}: {\cal A'} = \neg {\cal A}\\
C_1' &\assoc{\neg \cal A}& C_2'
\eeq
\end{lemma}
The proof of this follows from definition of negative promise\cite{promisebook}.
Let the promise of an association be $b = C_1 \assoc{\cal A} C_2$, e.g. from (\ref{bbb}),
so $\neg b = \neg (C_1 \assoc{\cal A} C_2)$ implies the intent not to associate, i.e. $\neg b = (C_1 \assoc{\neg \cal A} C_2)$.
So this suggests the existence of four more (effectively independent) types of
irreducible association for the negation of intent:
\begin{itemize}
\item Does not generalize,
\item Does not lead to,
\item Does not express,
\item Is not close to.
\end{itemize}
More explanation of these cases is given in \cite{cognitive}.

\subsubsection{Emotional origin to logical negation?}

There is a certain emphatic (even emotional) appeal to NOT statements,
such as we mean them humanly. This is not captured by the
dispassionate binary nature of logical complementarity. For example:
\begin{itemize}
\item I am {\em not} your friend. 
\item $X$ is {\em not} next to $Y$.
\item $X$ is {\em not} represented by $Y$.
\end{itemize}
It is interesting to speculate whether the origin of this reasoning
attribute lies more in an emotional reaction than in a direct
programming of embedded logic.  For logic to result from an emotional
outrage would be ironic to many. We may only properly understand the
nuances of reasoning once emotional weights can be incorporated into storylines.

Negations are very specific targeted promises, whose subject is a denial rather than
a logical complement. It is the binary nature of Boolean representation that
muddles these ideas together in modern information technology.
If we represent such associations for each of the irreducible types,
then the question arises what happens if nodes make both promises?
It is interesting to speculate whether the origin of this reasoning
Clearly, it makes sense for opposites to coexist in different contexts\footnote{Contexts
play the role of different Kripke or Everett universes.}, for instance:
\beq
\text{Alone} &\assoc{\text{generalizes} | \text{company}}& \text{isolated}\\
\text{Alone} &\assoc{\text{does NOT generalize} | \text{electricity}}& \text{isolated}
\eeq
These two coexist in different usage patterns, with unambiguous
semantics. However, the semantics of the following seem problematic:
\beq
\text{Alone} &\assoc{\text{generalizes} | \text{company}}& \text{isolated}\\
\text{Alone} &\assoc{\text{does NOT generalize} | \text{company}}& \text{isolated}.
\eeq
These are mutually exclusive, in a logical sense. However, in a
non-deterministic system mith Byzantine causation, it could be
difficult to prevent the formation of simulatenous links, so it is
important to understand the effect of such contradictory promises.
Could this mean indecision? Indecision would be more likely (and
cheaply) be represented by the absence of a relationship altogether.
The response to this `entangled' Schr\"odinger's catlike state
might be equally spurious, sometimes exciting one possibility and sometimes
the opposite. This suggests that reasoning may itself be stochastic in nature
in a sufficiently complex system, and that stability of reasoning emerges
from larger scale aggregation rather microconnectivity.

\subsubsection{Invariant representations and its relationship to tokenization}

The issue of how a knowledge representation can cope with all the
angles, perspectives, and distortions inherent in pattern recognition,
in multiple dimensions, has led to the idea that brains store imagery
and other sensory experiences in `invariant representations' rather
than as libraries of rasterized screen shots.  An invariant
representation implies the decoupling of a token representation from
acquired data patterns. 

\begin{figure}[ht]
\begin{center}
\includegraphics[width=10cm]{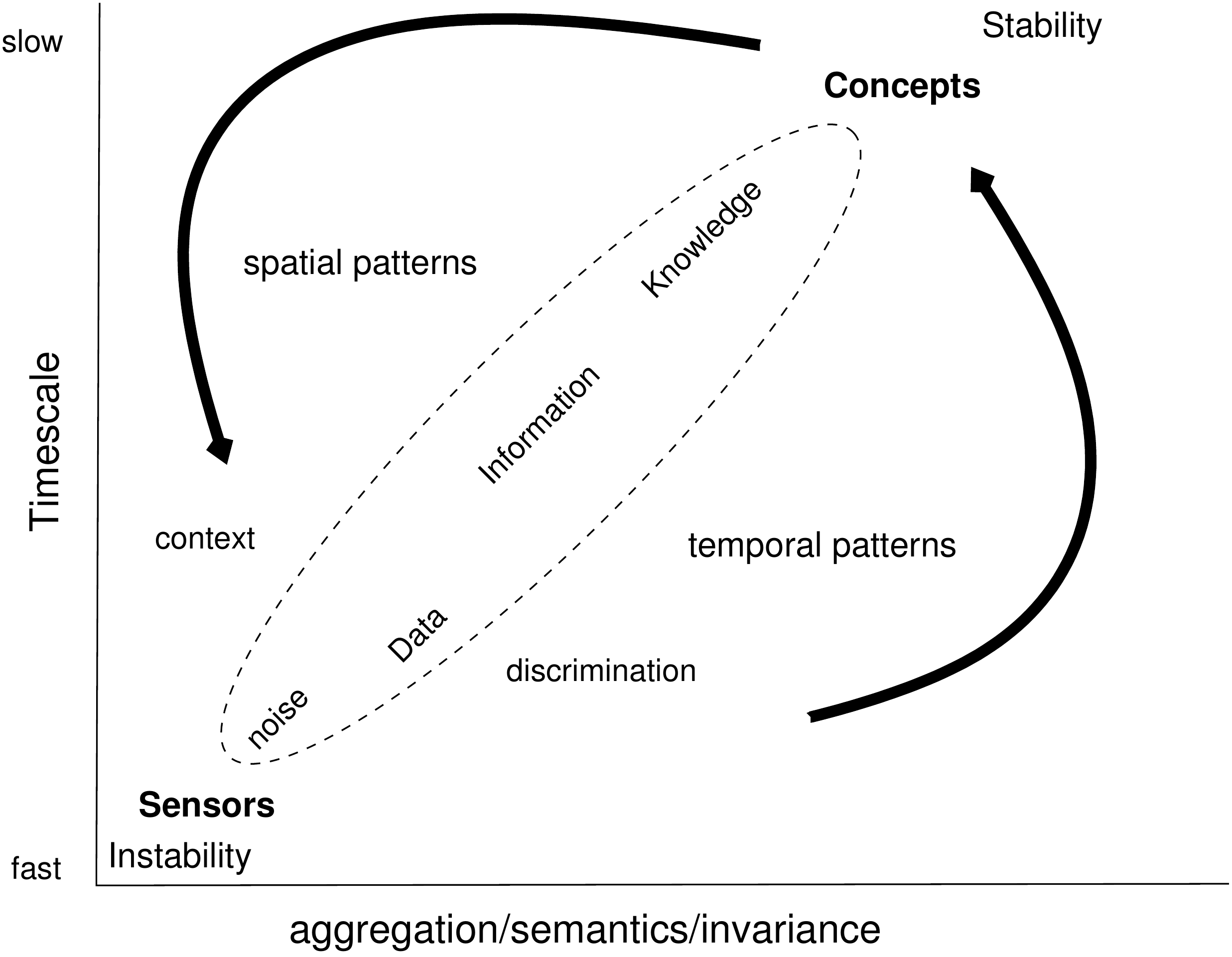}
\caption{\small Scale/invariance diagram template.\label{DIKW}}
\end{center}
\end{figure}

In mathematical physics, invariances and covariant transformations
base invariance around the existence of symmetry within a single
configuration space. For a semantic spacetime, tokenization (i.e.  the
association of a single point with an invariant meaning) is the
simplest invariant form.  Invariance is thus immediately supported by
the tokenization discussed earlier, and the invariance of semantics is
closely associated with spacetime structural stability under accretion
(see figure \ref{DIKW}).  The extreme end of this idea is that a
single neuron might actually encode a person or {\sc thing}; and,
while this might sound extreme, it has received some empirical
support\cite{invariantrep1,invariantrep2}, suggesting that there is
indeed some kind of matroid basis element representation going on in a
brain (see also the discussion by Hawkins \cite{hawkins}).

What remains to be understood in a learning representation is how
sensory inputs accrete into generalizations under the irreducible
associations, and these superagent structures become associated with
concepts. Understanding this process in a fully emergent system like
gene networks or a brain is a tough problem, but not an impossible one
(see, for example, the techniques of bioinformatics\cite{durbin1}).
For the annotation of expert input, in a human-artificial intelligent
cyborg approach, one can simply insert agents amongst a mixture of
partially learnt and manually inserted concepts and associations.
This hybrid approach becomes possible (even practical) now that we have
a simple spacetime representation with only eight irreducible 
relations. The main challenge lies in linking the data patterns to the
concepts seamlessly.

\begin{example}[Aggregation to invariants]
Figure \ref{DIKWexample2} illustrates an example of a
scale/invariance diagram, in which sensors recognize a plate by
matching sensory inputs to learned data patterns and associated
attributes expressed within a knowledge representation of a plate.
Initially, the focused sensory experience of a plate is too small to
single out a plate from other possibilities (e.g. flying saucer or
frisbee. However, once the context of being in a kitchen is established, the
concept of kitchen feeds back into context by `introspection' singling
out the plate, by context labelling, as the likely match.
\begin{figure}[ht]
\begin{center}
\includegraphics[width=11.5cm]{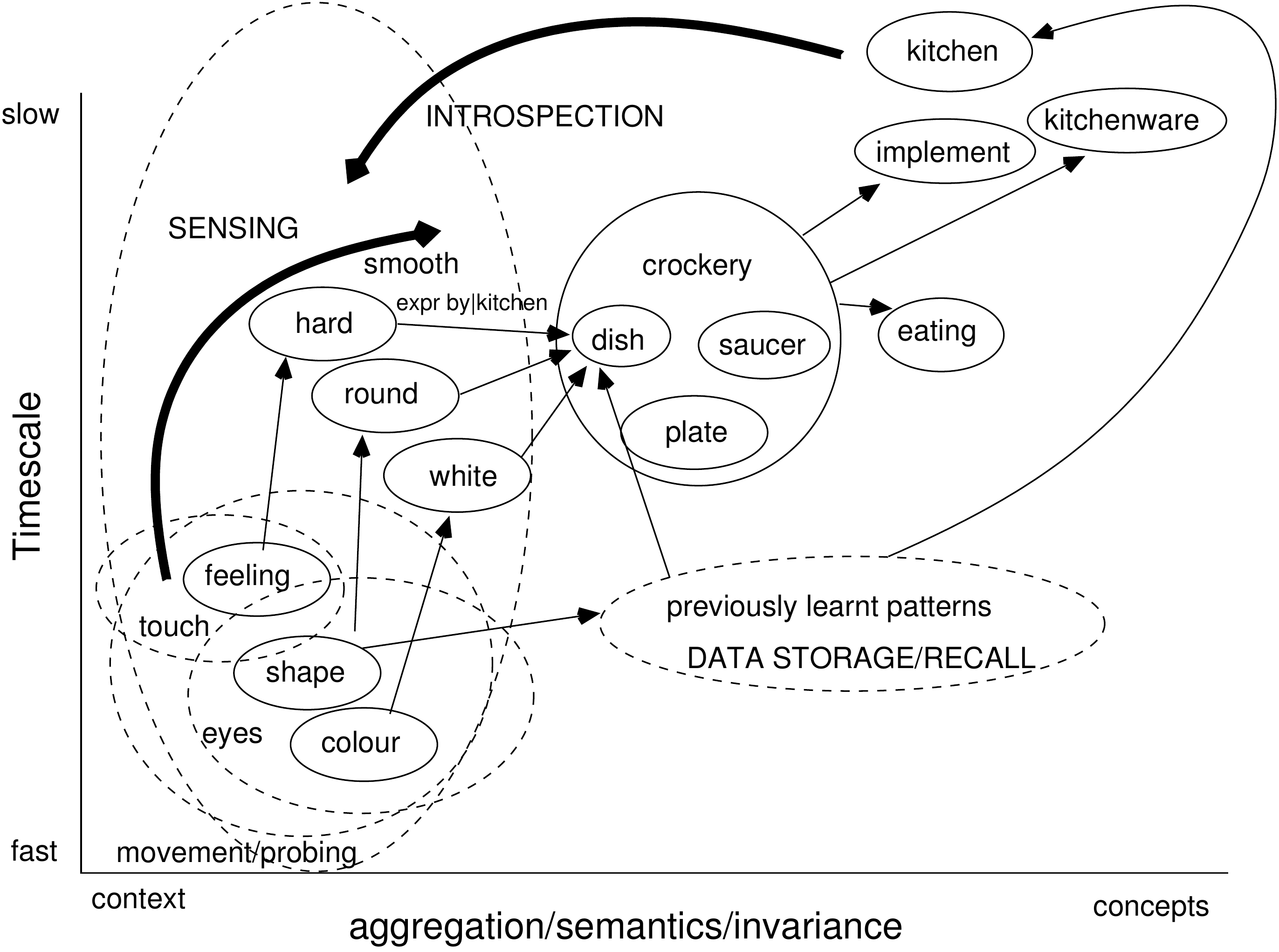}
\caption{\small Scale/invariance diagram example for recognition of a plate
in the context of a kitchen.\label{DIKWexample2}}
\end{center}
\end{figure}
\end{example}
The goal for stable invariant representations must be towards increasing
spatial aggregation and increasing (slower) timescale. Pattern similarities
that could lead to association into clusters include:
\begin{itemize}
\item Digitization into similar symbols alphabets $q$.
\item Same name or property expressed.
\item Similar changes in sensory inputs $dq$ (rising/falling).
\item Correlation and coactivation (timescales on $dq$).
\item Same function or purpose of a thing verb/noun representation.
\item Matching the same pattern (epitope matching)\cite{perelson1}.
\item Common activation context (origin path).
\item Dividing up and recombining at random (random recombination).
\item Spatial search paths (sequence).
\item Stable time-sequences and spatial aggregations.
\end{itemize}
Appendix \ref{appb} shows a further example based on more extensive
system monitoring sensors.

\section{Inference, logic, and reasoning in semantic spaces}\label{reasoning}

The ability to associate a sequence of concepts into a narrative path
is what we call reasoning. Reasoning takes many
forms\cite{boole1,boole2,firstorder,modallogic,suneel1,gridlogic2,vonneumann2},
and depends intrinsically on scale, because concepts scale like any
other kind of agent.  A narrative argument at one scale may make no
sense at another.  Since Boole's work\cite{boole1,boole2}, Hilbert's
challenge\cite{hilbert,hilbertack}, and the arrival of digital
computers, Boolean algebra and first order logic, have come to
dominate our thoughts about what is meant by reasoning. It has become
a mathematized form of arithmetic, for the specific purpose of making
decisions. However, these are crude forms of reasoning compared to the
processes by which humans reason.  To approach human forms of
reasoning, we need to think more carefully about how concepts and
associations are woven into narratives.

\subsection{Reasoning from learning}

Several common forms of reasoning express pathways that connect concepts. Classic steps
in reasoning include the following:
\begin{itemize}
\item {\bf Induction}: generalization based on specific instances (concepts from exemplars).
These are represented by the coarse graining associations between concepts (figure \ref{scales}):
\beq
C_\text{dog} \assoc{\text{is generalized by}} C_\text{animal}.
\eeq
Underlying this are spacetime promises of the form of superagency (see paper II):
\beq
A_\text{dog} \promise{\text{inside}} S_\text{animals}.
\eeq
The process of induction is an $N:1$ mapping from exemplars to generalizations.

\item {\bf Deduction}: inferring exemplars from identification of types.
These are represented by compositional associations:
\beq
C_\text{dog} \assoc{\text{has property}} C_\text{waggly tail}.
\eeq
In a promise model, the superagent representing `dog' makes a number of
exterior promises, such as `waggly tail'.
The process of induction is an $1:N$ mapping from generalizations to exemplars.

\item {\bf Abduction}: Inferring a theory from an observation
  (learning or model building\footnote{This is the approach used in
    Bayesian learning, or machine learning.}).  These are represented
  by compositional associations: \beq C_\text{waggly tail}
  \assoc{\text{characterizes}} C_\text{dog}.  \eeq These are
  retrofitted inferences of identity, analogous to the matching of
  receptors\footnote{Machine learning is all about abduction.
    Abduction can be applied to any structure. Induction goes further
    by discretizing into qualitative attributes and concepts.}.
The process of induction is an $N:1$ mapping from exemplars to stories or models of explanation.

\item {\bf Causation}: Inferring the prerequisites for the occurrence
  of a concept\footnote{This is simply a definition of causation, as
    used pragmatically in the physical sciences. It assumes `intention
from within', but does not imply any notions of free
    will or the many other complications that arise in a more general philosophy.}.  These are represented by
  causal associations, \beq
  C_\text{Lights} &\assoc{\text{depends on}}& C_\text{electricity}.\\
  \eeq which, in turn, arise from underlying conditional promises:
  \beq
  A_\text{light bulb} &\promise{+\text{shine}|\text{power}}& A_\text{observer}.\\
  A_\text{power socket} &\promise{+\text{power}}& A_\text{light bulb}.
  \eeq
It is important to remember, when reasoning, that effects may have multiple
causal predecessors.

\item {\bf Free association / lateral thinking}: the exploration of
  concepts that are closely related to another, by some definition.
  These are represented by similarity or proximity associations, and
  spacetime adjacency promises.  \beq C_\text{dog} \assoc{\text{is
      similar to}} C_\text{wolf}.  \eeq This kind of reasoning is
  saying: if you are thinking about `dog', then what about `wolf' as a
  similar case. In other words, do you need more examples of your
  concept, without actually going to a new level of
  generalization\footnote{It is like Amazon's referral engine: users
who bought $X$ also bought $Y$.}.
\end{itemize}

\subsection{Associative model of reasoning}

\begin{figure}[ht]
\begin{center}
\includegraphics[width=10.5cm]{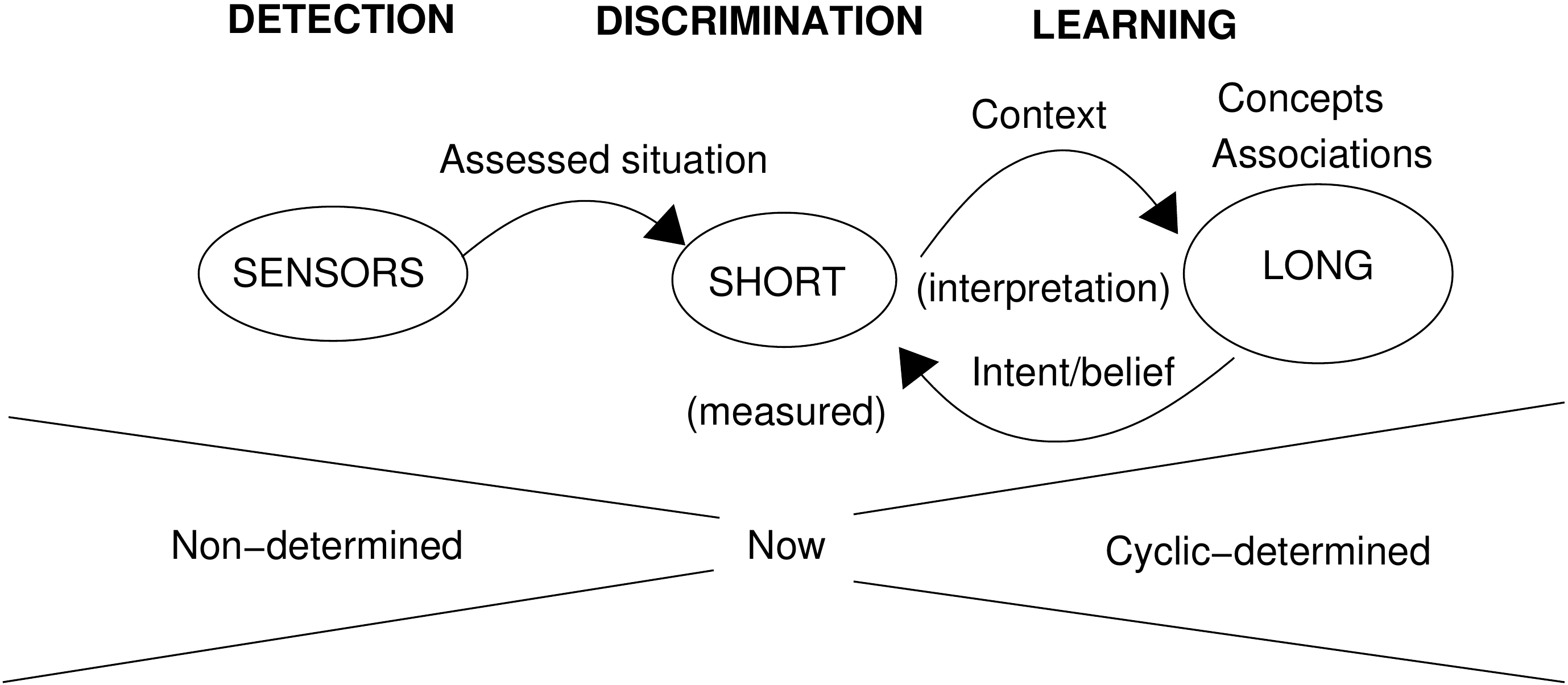}
\caption{\small Short term memory caches the state of `now' from sensor results, but the semantics are pre-supposed. Long term memory
builds up the semantics slowly over time, so that it can be used to interpret sensors as context.
Modulation of slowly varying long term memory by quickly varying short-term memory
allows us to switch large amounts of information into play (that otherwise takes a long time to assimilate)
quickly, depending on context.
\label{modulation}}
\end{center}
\end{figure}

We return to the model of a knowledge system as the interaction
between a single test user and its exterior world, via sensors that
channel their observations into a collection of internal agencies to
represent knowledge in a concept space (see figure \ref{modulation}).
Recall that the model is about separating timescales into short and
long term recall.  Sensors collect data and transform it into short
term context. Context already possesses the semantics of its sensors,
and collectively stimulates concepts by some unspecified form of
approximate addressing. Meanwhile, the activation of concepts can
trigger semantically close concepts which feed back into context as
memories and mix with the input stream to trigger even more related
concepts. Thus, there is a cumulative exploration of the space of
concepts, based on a `state of awareness'\footnote{This is how
  CFEngine policy engine, and knowledge map was constructed, with sensors evaluating to
  short term classes that trigger the long term encoded `policy'.}.
In order to be able to use these techniques, sensory inputs first have
to trigger a concept or observe a {\sc thing}, which in turn triggers an
association to a property, a question, or even a lateral guess. 

\begin{figure}[ht]
\begin{center}
\includegraphics[width=10.5cm]{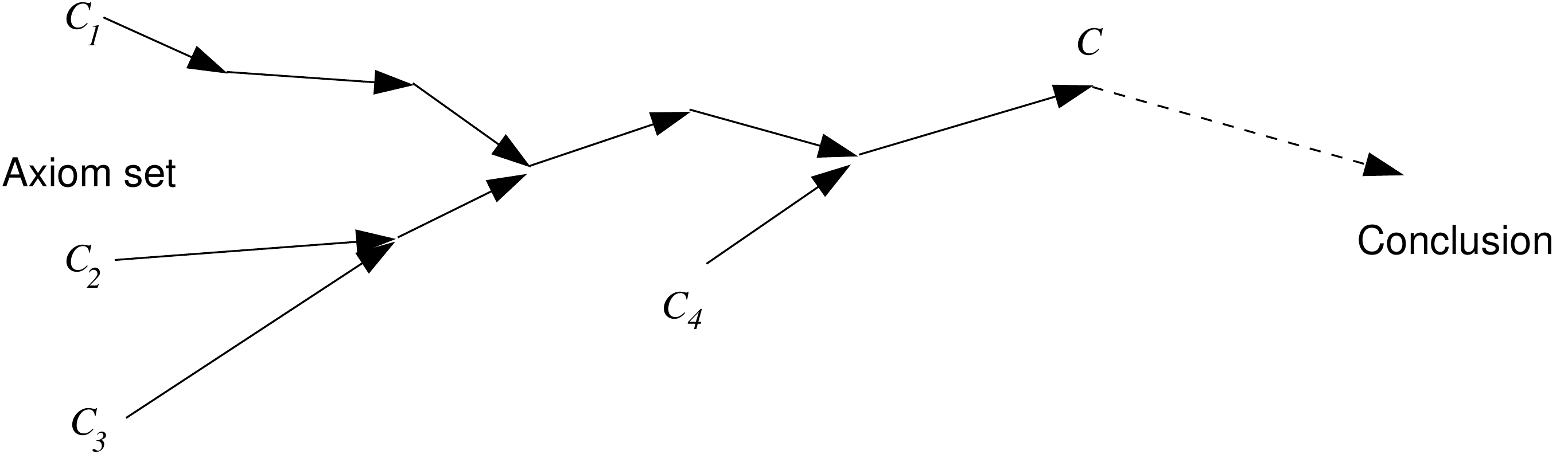}
\caption{\small A story is a directed graph that converges towards a
  single thread at the subject of interest (i.e. it may fan out in the
  past).\label{stories}}
\end{center}
\end{figure}

A linearized sequence of inferences may be accumulated into the idea
of a story or narrative \cite{stories,burgesskm}). A narrative is
closer to what we think of as everyday human reasoning than is
generally credited in the literature.  This approach
is based on the idea that much of what humans consider to be `smart' is based on
storytelling, a tradition that goes back to the earliest forms
of communication.  A story is a converging network of associations
that passes through a number of specific concepts of interest,
including the subject or topic of the story.  By tracing a selected
path of associations linking the subjects forwards or backwards, we
can can divine both predictions of consequences and possible related
causes about the subject. Associations are based on the causal
and structural spacetime associations represented by the irreducible
types\footnote{Since the first work about promises and
  knowledge\cite{knowledgemace2007,burgessaims2009}, I have still to
  find any related work on the importance of stories to human
  thought.}, so we already know that they encode spacetime relationships
and causation.
The concept of stories is thus of great importance to understanding the
role of knowledge and inference in smart behaviours (see figure
\ref{stories}).
\begin{definition}[Story or narrative about $C$]
  A directed acyclic graph that passes through $C$, and whose
preceding association is of a single irreducible type. The story
begins with a set of axiom concepts and ends with a set of conclusion concepts.
\end{definition}
There may be several associations passing through $C$, but those of
different types represent different stories.
All forms of reasoning start from an initial concept or set of concepts, converging
towards a single storyline. 

\begin{definition}[Equivalent narratives/stories]
If two sequences of associations have the same axiom set and final conclusions,
then they may be considered equivalent alternative tellings of the same story.
\end{definition}
Since stories belong in the space of concepts, they exist
independently of the learning process. A story may have been learned
wholesale, or inferred from independent sensory input events, or even
imagined by introspection.  The concepts spanned by a story may have
been learned at any time prior to the formation of the story. 
\begin{example}[Can we short-circuit learning?]
  Is a story, learned wholesale, e.g. from a book, or around a
  campfire, a proper substitute for a sensory learning experience?
  The amount of information within a story, based on a
  context-concept-association representation, is significantly less
  than the information that was involved in the original learning
  process, through sensory recognition and repetition.  Suppose,
  instead, that the story was not experienced directly, but merely
  passed on as a linguistic rendition, e.g. by reading a book: now,
  the information in the tokens of the story may be transmitted
  exactly, from the book to the reader, and provide the same
  information to different recipients. It might seem that identical
  knowledge could be passed on.  What may not be the same, however, is
  how the concepts used in the story link up in the network of the
  receiver prior to the telling, thus extending beyond the telling of
  the story itself.  This may differ between the storyteller and any
  given story receiver.  The degree of similarity in the final state
  depends on the extent to which the observer already has all the
  dependent concepts in place to being with. Since the story modifies
  the network of concepts in the listener, the initial state before
  the story and the final state after may not correlate at all well
  between storyteller and story receiver. A full sensory learning
  experience will undoubtedly lead to more associations being formed
  around the storyline in memory, but this depends on both the initial
  state and the unique sensory experience of the receiver. In short,
  it is impossible to say what effect either the storytelling or
  sensory learning might have on different observers.
\end{example}

\begin{example}[Stability of chained neural networks]\label{reasonnet}
  Neural networks (ANN) are pattern discriminators, which approximate
  functional mappings, and learn by supervised training.  They have no
  other rational reasoning capabilities, so any attempt to solve an
  extended reasoning problem with ANN has to be solved in multiple
  stages, by turning a series of tailored discriminators into a series of
  transformation functions.  One approach is to chain together
  multiple pattern discriminators in a directional
  hierarchy\cite{deeplearning3,hawkins}, where early layers transmute
  sensory input into a reduced dimensional (tokenized) form, and later
  layers identify combinatorics of these reduced forms. This is a
  machine learning representation of the concept learning schema
  described in this section.

If independent neural nets are strongly coupled, in such an array, then
they are effectively a single tightly coupled network, with each
network $N_a$'s hidden layers trained independently by data sets $D_a$, but coupled as a
single structure.  If there is no separate training, they are a single
neural network, capable of discriminating only a single input to
output association.
Each layer must
  be trained independently, else the combined network is just a
  deeper single pattern discriminator.
  In such a system, the stability of a reasoned outcome builds on the
  stability of all the layers. A later layer that depends on a
  previous layer, inputs the modulations of the earlier layer and
  outputs a contextualized result to a later layer. This output is
  used in two ways: during operation and during training.
\begin{figure}[ht]
\begin{center}
\includegraphics[width=8.5cm]{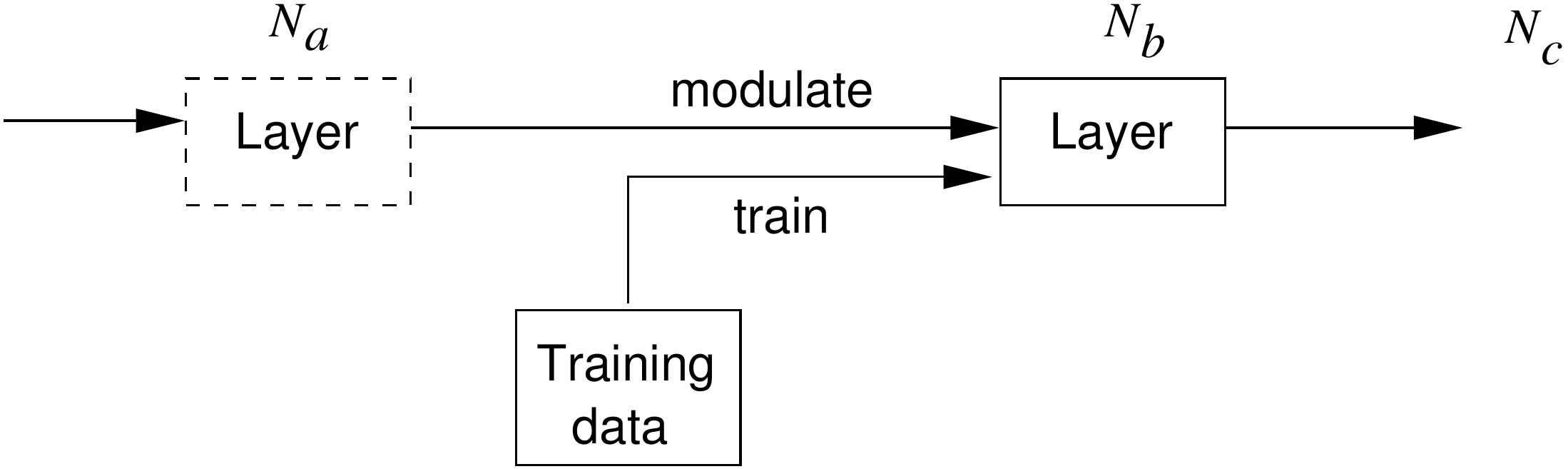}
\caption{\small Each layer in a serial network may be altered by operational modulations
or by retraining. Earlier layers promise to modulate later layers,
and each layer promises to retrain from datasets, on different timescales.\label{trainsw}}
\end{center}
\end{figure}
Referring to section \ref{tscond}, we can describe the timescales for deterministic operation.
\begin{itemize}
\item  $T(m_b) \equiv T(N_b \promise{+m_b(d_a) | m_a,d_b} N_c)$ the time it takes a layer to operate, transmuting input from layer $a$ to layer $b$, where $b>a$.
The promise is a conditional one, depending on both the modulation of the previous layer $m_a$ and the training data of the present layer $d_b$.
\item $T(D_b \promise{+\Delta d_b} N_b)$ the time in between changes of training data $\Delta d_a$ for the layer from world sources.
\item $T(d_b) \equiv T(N_b \promise{-\Delta d_b} D_b)$ the time to accept and process changes of training data $\Delta d_a$ by the layer $N_a$.
\end{itemize}
The switching will be semantically and dynamically stable (quasi-deterministic) iff
\beq
T(d_b) \gg T(m_a),~~~~ b > a\label{mdm}
\eeq
as described in section \ref{tscond}.
The world must not change substantially during the training of the
entire reasoning process, else the result will be invalid before a
conclusion can reached, and the work done in training will not be
reusable, i.e. it cannot keep up its evolutionary fitness for purpose.
This is an evolutionary constraint: the more primitive senses should
be basically fixed compared to the plastic later stages. This is
certainly true in biology, where basic senses are encoded genetically,
and brain learning of what there is to see is nurtured.

The training of later layers must depend on the training of the earlier layers,
thus the earlier layers
must behave consistently over the time it takes to train the later
ones. This means that the networks should not change significantly, or
\beq
T(D_a) \gg T(D_b) ~~~~~ a < b
\eeq
i.e. later stages may be more plastic than earlier ones.

However, in order to modulate later layers successfully, during
training, (later layers learn from repeated invocations of earlier
layers), the operation time of an earlier layer must be significantly
shorter than the training time of the later layers, else learning
cannot be repeated a sufficient number of times:
\beq
T(m_a) \ll T(d_a),
\eeq
which is consistent with (\ref{mdm}).
If these inequalities are not met, the training of one layer will be
rendered unstable by a change in an earlier layer. 

\end{example}

\subsection{Couch's rules for inference}

Alva Couch made the observation that the irreducible positive associations lead
to connected stories of decreasing certainty with increasing
length\cite{stories}.  The essence of these rules was based on the irreducible
spacetime structures described earlier, and the uncertainty from
semantics is introduced semantically rather than quantitatively (as,
for example, in statistics).  From the irreducible associations, the following
examples illustrate the idea:
\begin{enumerate}
\item If there is a route from $A$ to $B$, and from  $B$ to $C$, then there might be a route from $A$ to $C$.
\item If $A$ is inside $B$, and $B$ is inside $C$, then $A$ might be considered inside $C$.
\item If $A$ depends on $B$ and $B$ depends on $C$, then $A$ might depend on $C$.
\end{enumerate}
The appearance of `might' signals an increase in
uncertainty\footnote{The actual rulesets were much more specific and
  detailed than these, based on a particular model implementation, but
  this captures the principle.}.  There is no guarantee of the
conclusion, because the semantics of the agents need not be uniform;
nevertheless, the outcome is possible or even plausible.  For
instance, we might throw our hands up at 2 and say that the second of
these must allow us to say that $A$ is inside $C$.  However, what if
we open up $C$ then we find only $B$ because $B$ is opaque? How could
we know that we are supposed to open $B$ too?  And when are we allowed
to stop opening boxes? It is a perfectly valid semantic point of view
that $C$ is not really inside $A$, because being inside suggests being
able to see or touch the boundary.  Whenever autonomous agents becomes
intermediaries for a process, one cannot be certain that they will
behave uniformly\footnote{One way to be assured of homogeneity is for
  them to promise homogeneity.}.
With these rules in mind, the process of inference over arbitrary
associations lies in the ability:
\begin{itemize}
\item To interpret associations as aliases for the irreducible association types, so that
we can always reduce chains of arbitrary associations to chains of the irreducible types.
\item To form triangle reductions that condense multiple concepts
or associations into fewer based on the irreducible associations\cite{stories}. 
\end{itemize}
Inference, based on the irreducible associations, is a powerful tool for
working in tokenized form. Simple search algorithms may  be used to generate
stories. While not all these stories emerge immediately in well-formed
human readable language, the essence of a skeletal story does emerge from
simple promises or rules about parsing the association graph.
This is an indication of the potential simplicity of reasoning, but at the same time
the complexity of linguistic expression.

\subsection{Stories and narrative reasoning}\label{storysec}

The importance of stories, as defined, is much deeper than a cultural
memory of tales told around a campfire.  Stories form patterns of
functional relationships in spacetime: in other words, they are the
map between algorithm, reasoning, and machinery. We use them for
reasoning by following pathways until we find concepts that match the
criteria of a question. Stories lie at the heart of processes,
algorithms, and also what we humans mean when we speak of {\em
  explanation} and {\em understanding}.

\begin{definition}[Understanding or explanation of $C$]
The ability for an agent tell a story that converges on $C$.
\end{definition}
\begin{definition}[Depth of understanding of concept $C$]
  The network distance, or number of associations between concept
  nodes that precede the concept $C$ in an explanatory story.
\end{definition}
People differ in what standard of depth they consider to be sufficient
to understand something. Yet these simple definitions capture the
basic essence of what is commonly meant by humans concerning
imagination and understanding.
\begin{example}[When do we really understand something?]
  Alice may say that she understands atoms, by telling the following
  story: atoms are made up of protons, neutrons, and electrons.
  Later, she may change her mind and add another level of predecessor:
  atoms are made up of protons, neutrons, and electrons, and protons
  and neutrons are made up of quarks. The number of levels of
  explanation in terms of associations to other things may continue
  indefinitely (turtles all the way down).
\end{example}
During the parsing of a story, the {\em context} of a knowledge
agent's mind is modified by the successive details of the story: new
context is refreshed as the stages of the story are revealed. This
context may condition the associations to particular concepts, and
feedback into the stream. Thus the accumulation of context helps to
build a case for an `answer' or `conclusion' to the story.

\begin{lemma}[Story limits]
If a concept $C$ has at least one association, there is a story.
There is no limit to the possible length of a story.
\end{lemma}

\subsection{Story types, classified by their promises and spacetime associations}

We may classify stories into different types, according to the type
of irreducible linkage they use.

\begin{example}[Episodic history]
A sequential timeline story (history) such as:
\begin{enumerate}
\item John got up
\item John went to work
\item John had lunch
\item John came home,
\item etc
\end{enumerate}
may be represented at a story formed from time markers joined by precedence associations,
each of which is attached to a concept associated with the event. This is a simple, regular
molecular chain:
\begin{figure}[ht]
\begin{center}
\includegraphics[width=10.5cm]{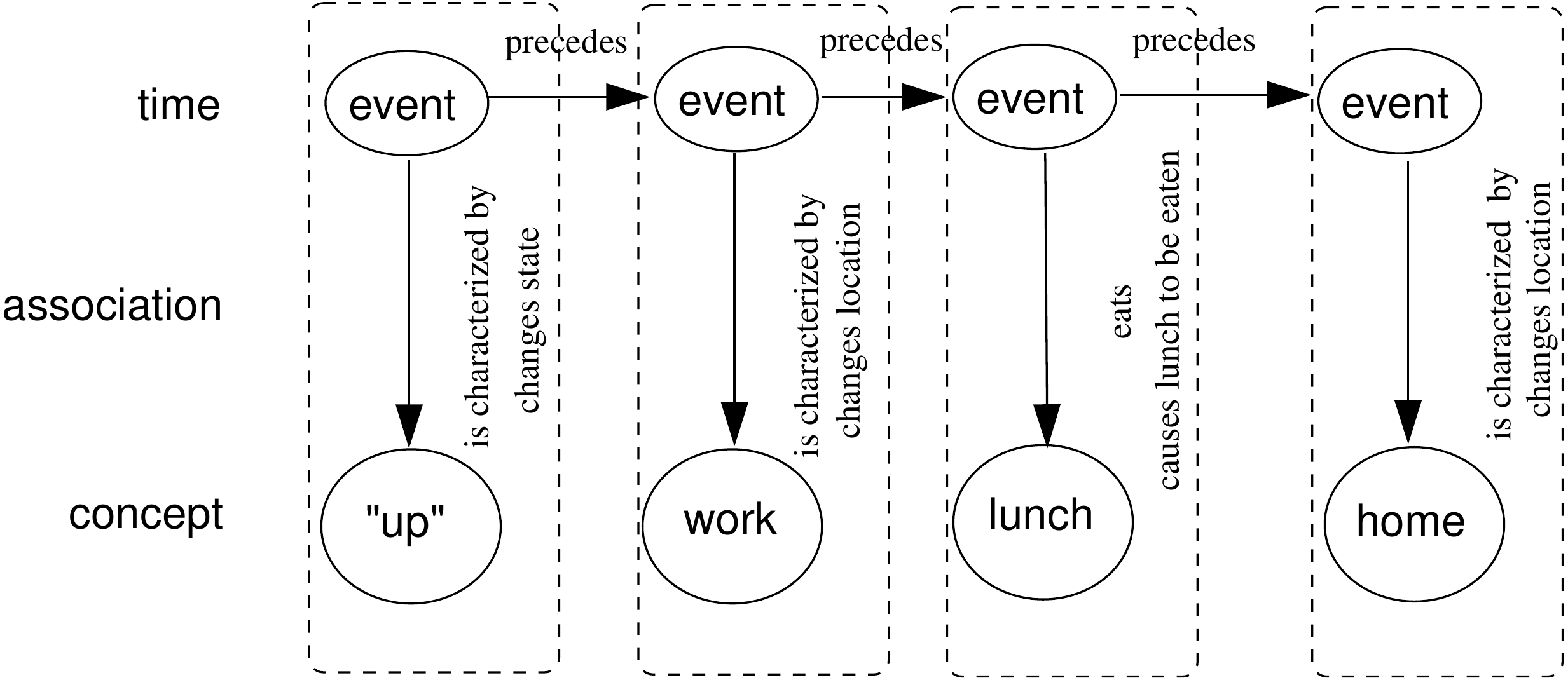}
\caption{\small A timeline history: ordered molecules of timestamps and associated
happenings.\label{history}}
\end{center}
\end{figure}
First we notice that the story is made up of promises, not associations,
and that each promise precedes the next.
\beq
\pi_1: \text{John} &\promise{\text{got up}}& \text{observer}\\
\pi_2: \text{John} &\promise{\text{went to work}}&  \text{observer}\\
\pi_3: \text{John} &\promise{\text{ate lunch}}&  \text{observer}\\
\pi_4: \text{John} &\promise{\text{went home}}&  \text{observer}
\eeq
The first is a change of state, i.e. it is a scalar promise. The latter three contain
both verbs (links) and nouns (agents), i.e. they are vector promises.
To turn these into associations, we separate out the concepts. In the first case,
this involves introducing a `state' agent, as in the example for `blue' in figure \ref{calibration}.
The timestamps act like road signs guiding the story forward and preserving the
order. They need not be numbered, provided they preserve their solid state order.
The result is a spacetime crystalline phase in the form of a chain.
Without forming these molecular pairs, the concepts have no ability to
become associated only for this one example. It would be wrong to
permanently associate up, work, lunch, and home with a precedence
relationship: what we want to express is what happened in the context
of this episode.
\end{example}
\begin{example}[Conceptual story]
Look at the knowledge space or concept network in figure \ref{story1}.
This shows the concepts in the mind of an observer ``me''. 
By exploring at random, with a god's eye view of the mind of ``me'', we can
infer that ``me'' works for the police, and that there has been a theft at
the village bakery, somewhere in Oxford.
\begin{figure}[ht]
\begin{center}
\includegraphics[width=12.5cm]{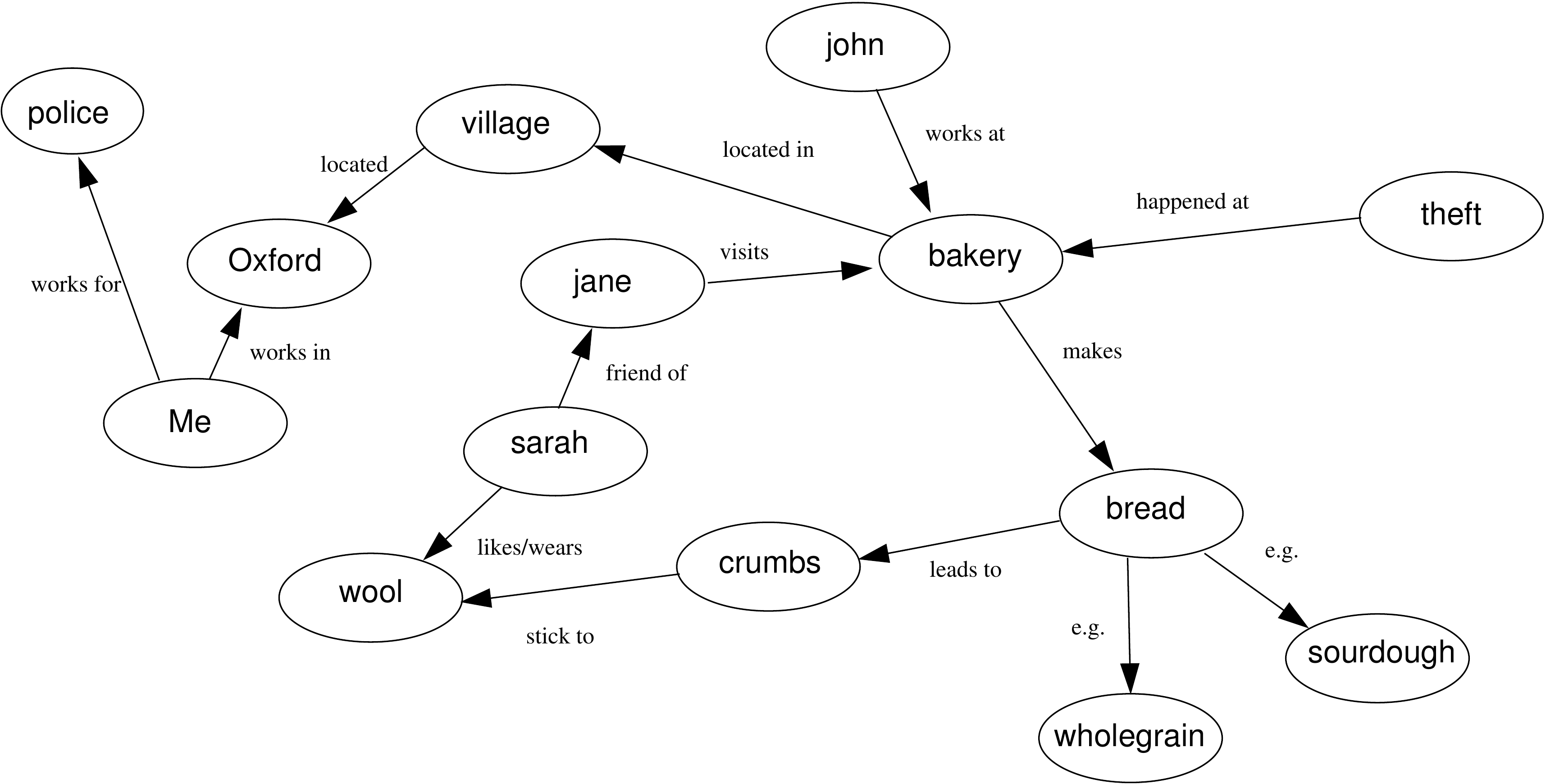}
\caption{\small Knowledge space belonging to ``me''.\label{story1}}
\end{center}
\end{figure}
The extent to which this picture is available to the individual agents
is not important, because we know that knowledge and concepts lie in the
distributed network of concepts, not in any viewpoint located inside it
(there is no neuron that knows what we are thinking: ideas are something that
emerge at a larger scale).

By sensing the exterior world, police person ``me'' observes that Jane
has crumbs on her sweater. The concepts of Jane and crumbs and sweater
are thus activated simultaneously as the current short term situation
context. By following causal links, we find a specific story about Jane, from ``my'' perspective is
shown overlaid on figure \ref{story2}.
\begin{quote}
crumbs $\assoc{\text{stick to}}$ wool $\assoc{\text{is liked by}}$ Sarah $\assoc{\text{friend of}}$ Jane $\assoc{\text{visits}}$ bakery $\assoc{\text{occurrence of}}$ theft.
\end{quote}
\begin{figure}[ht]
\begin{center}
\includegraphics[width=10.5cm]{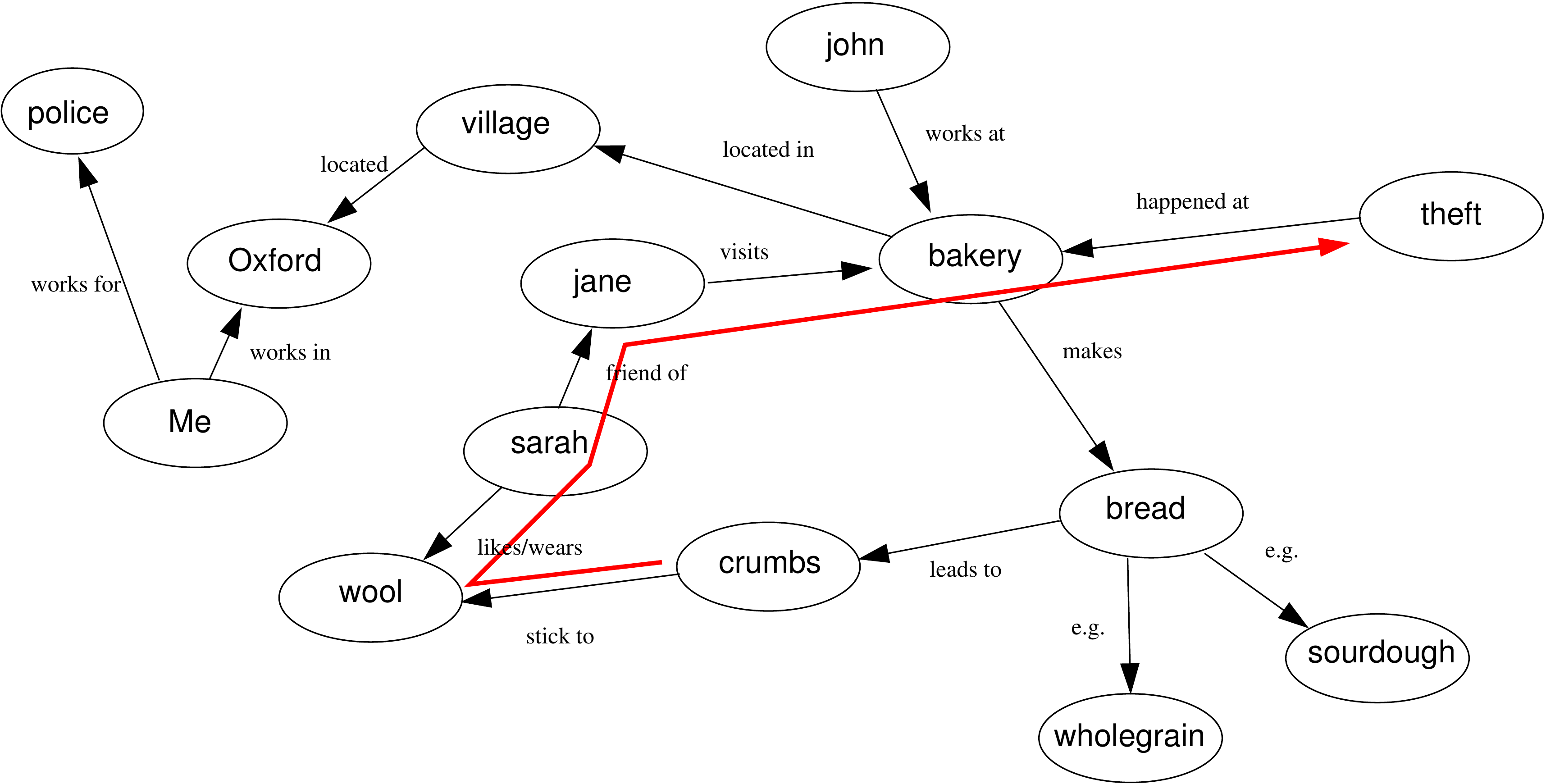}
\caption{\small A story about Jane.\label{story2}}
\end{center}
\end{figure}

\begin{figure}[ht]
\begin{center}
\includegraphics[width=6.5cm]{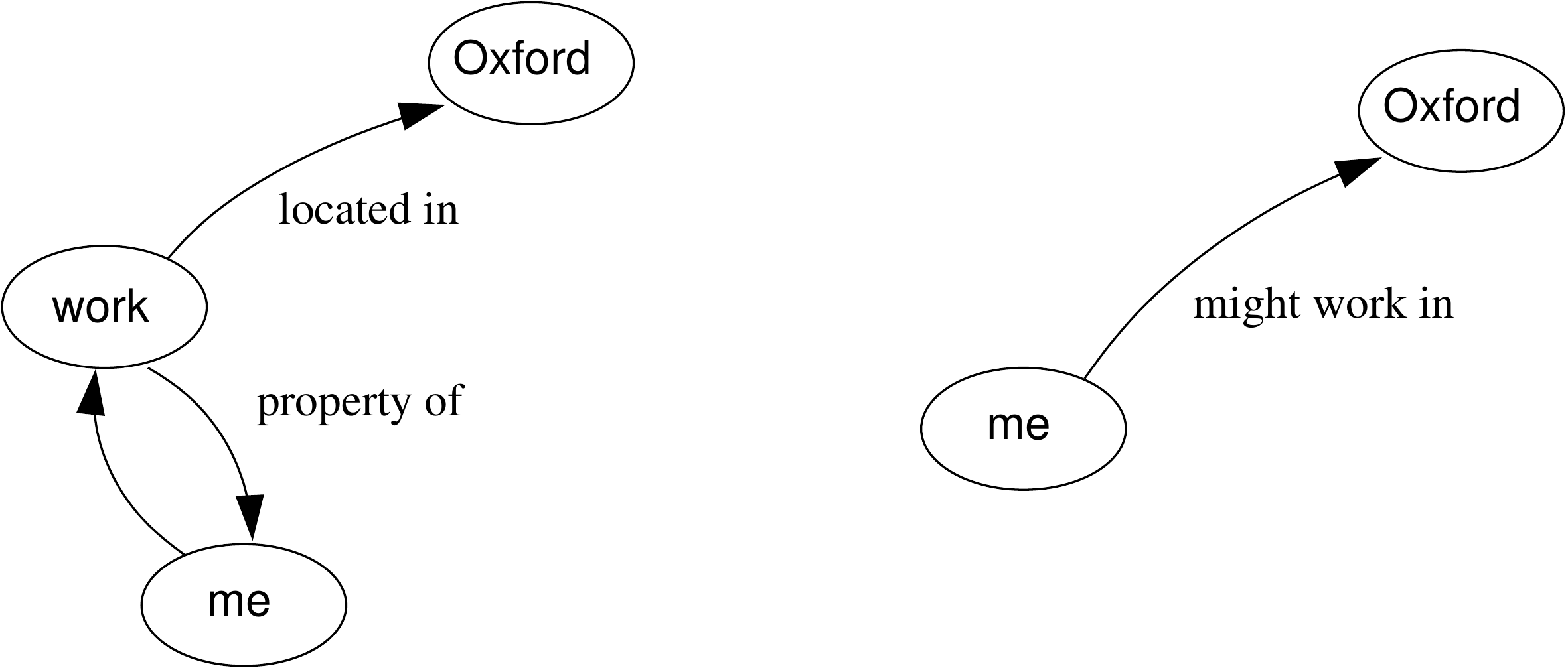}
\caption{\small Quasi-transitive inference altering associations.\label{story3}}
\end{center}
\end{figure}

\begin{figure}[ht]
\begin{center}
\includegraphics[width=6.5cm]{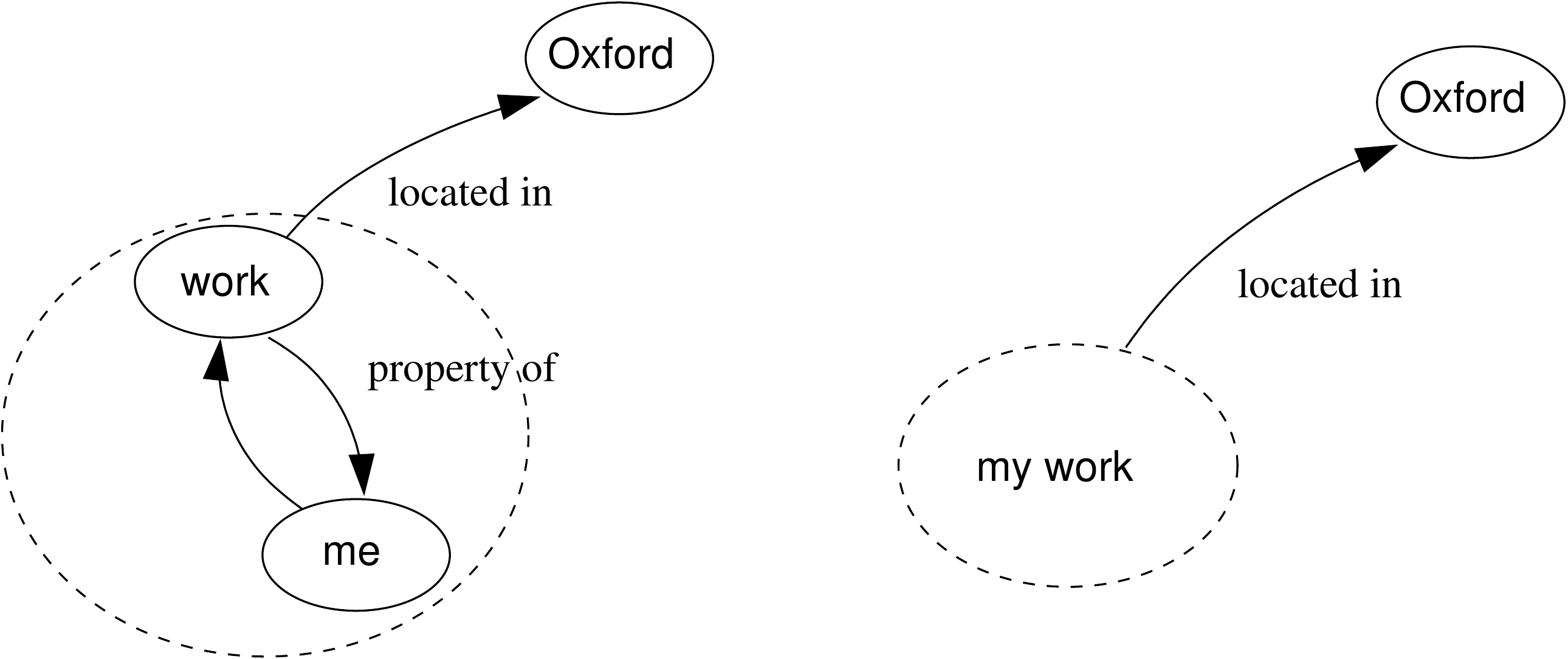}
\caption{\small Another way for the scaling of agency to represent composite concepts.\label{story4}}
\end{center}
\end{figure}

\end{example}
In this example, we see how the scaling of agency and inference of quasi-transitive
association, mimicking basic spacetime concepts, allow us to anthropomorphize inanimate
things, and attribute intentional language to simple inanimate {\sc things}.
This is a natural extension of the overlap between human and non-human concepts in an
associating network.

\begin{figure}[ht]
\begin{center}
\includegraphics[width=8.5cm]{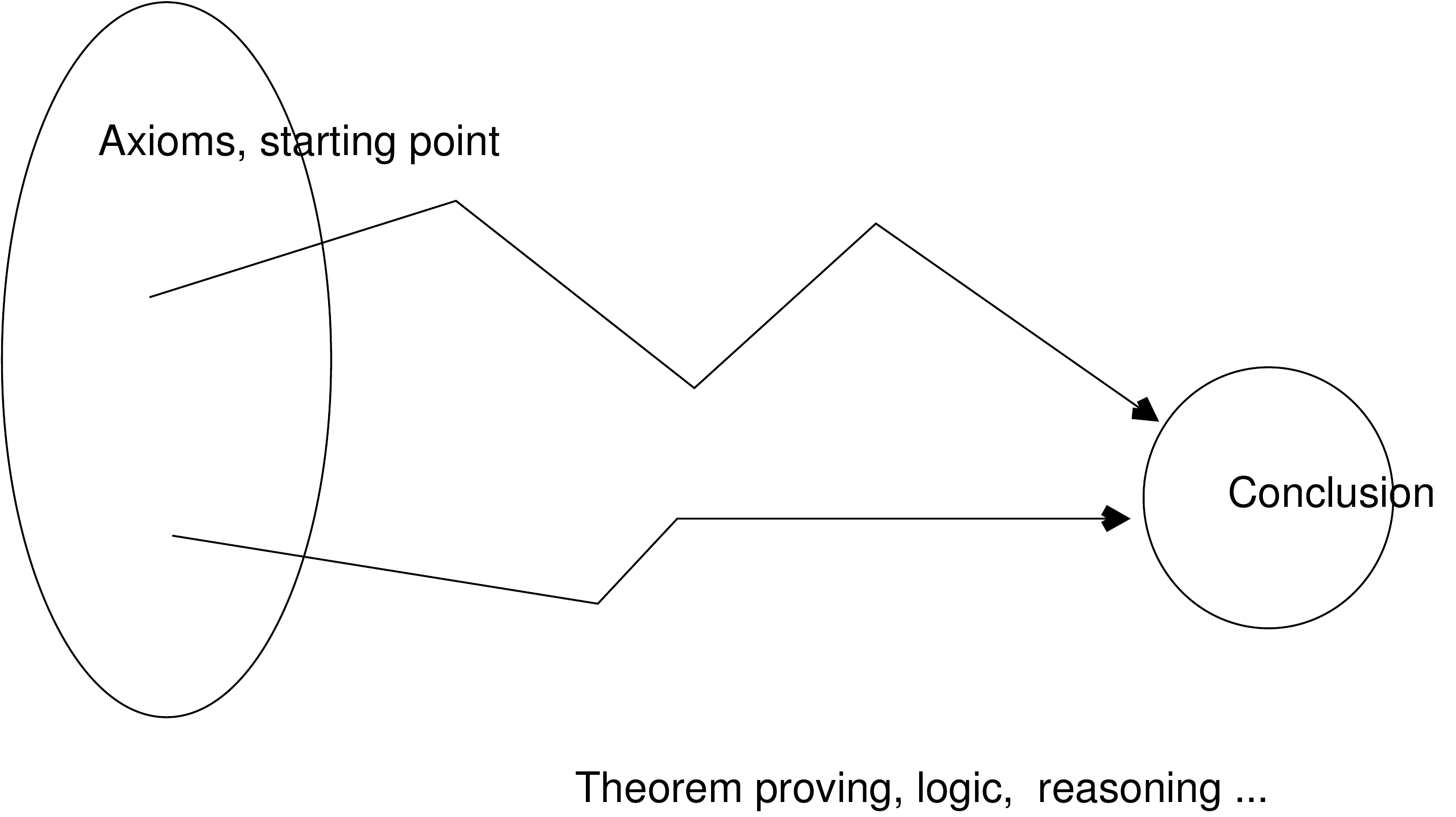}
\caption{\small Logic is not a succession of inevitable mandatory steps, but rather
a sequence of axioms promised and conclusions accepted on the trust of the
axioms.\label{theorem}}
\end{center}
\end{figure}

If stories are as important as they seem to be, our understanding of
reasoning in computer science has to change (see figure \ref{theorem}).

\subsection{Inferences from algebra of association: type symmetry transformations}

Another class of stories are those that allow combinatoric inference.
If we return to the example of family roles, we see that some
associations do not require an inverse to be promised: they simply
follow as facts and require no additional documentation, as both
directions are properties of the `mind' of an observer.  However, {\em
  promises} have a different logic, and do need reciprocal
confirmation, because they originate from independent agents that may
not agree.

\begin{itemize}
\item {\bf Promise logic}:
Suppose exterior agents make promises:
\beq
\textsc{JOHN} &\promise{+father\;of}& \textsc{JANE}\\
\textsc{JANE} &\promise{+daughter\;of}& \textsc{JOHN}\\
\textsc{JOHN} &\promise{-daughter\;of}& \textsc{JANE}
\eeq
John promises that he is Jane's father, but Jane does not confirm.
This could mean that he is lying, and an observer would not be able to
trust the promise without some corroboration.  However, Jane promises
that she is John's daughter and John confirms, so an implicit
confirmation is in fact given.  Because the relationship between
father and daughter is an independent linguistic fact, this is
sufficient to conclude that the missing promise is given\footnote{It
  is still possible that they are both lying, but there are stronger
  grounds to trust the promises now. Also, an adopted daughter may
lead to further semantic quarrels about the reciprocity of the relationship,
but this serves to illustrate the principle, and I'll leave the remainder
as an exercise for the reader.}. Thus, we may infer:
\beq
\textsc{JANE} \promise{+daughter\;of} \textsc{JOHN}
\text{  implies  }  \textsc{JANE} \promise{-father\;of} \textsc{JOHN}.
\eeq
These two promises are not independent.
Similarly, daughter, son, and child  are not independent vectors:
\beq
\{ +male , -daughter\} \rightarrow father
\eeq
Perhaps we need to use conditionals to vectorize
\beq
+daughter &=& +child | female\\
+son &=& +child | male\\
\eeq
Note that these are conditional promise relationships, not associations.

\item {\bf Association logic}: Unlike the case for promises,
  knowledge, i.e. that which has been learnt, has already been trusted
  and inferences may be predicted, even if they haven't been observed.
  This is because associations are all inside the mind of a single
  observer.  The representation might be formed of independent agents,
  but we can assume that the knowledge is likely consistent.  This is
  how ontology arises.  So, assuming that the family concepts are
  known to the observer, we can make some inferences about named reducible
associations that would not necessarily follow for promises. This is because
we are assuming that all the promises may be taken as given.

\begin{figure}[ht]
\begin{center}
\includegraphics[width=10.5cm]{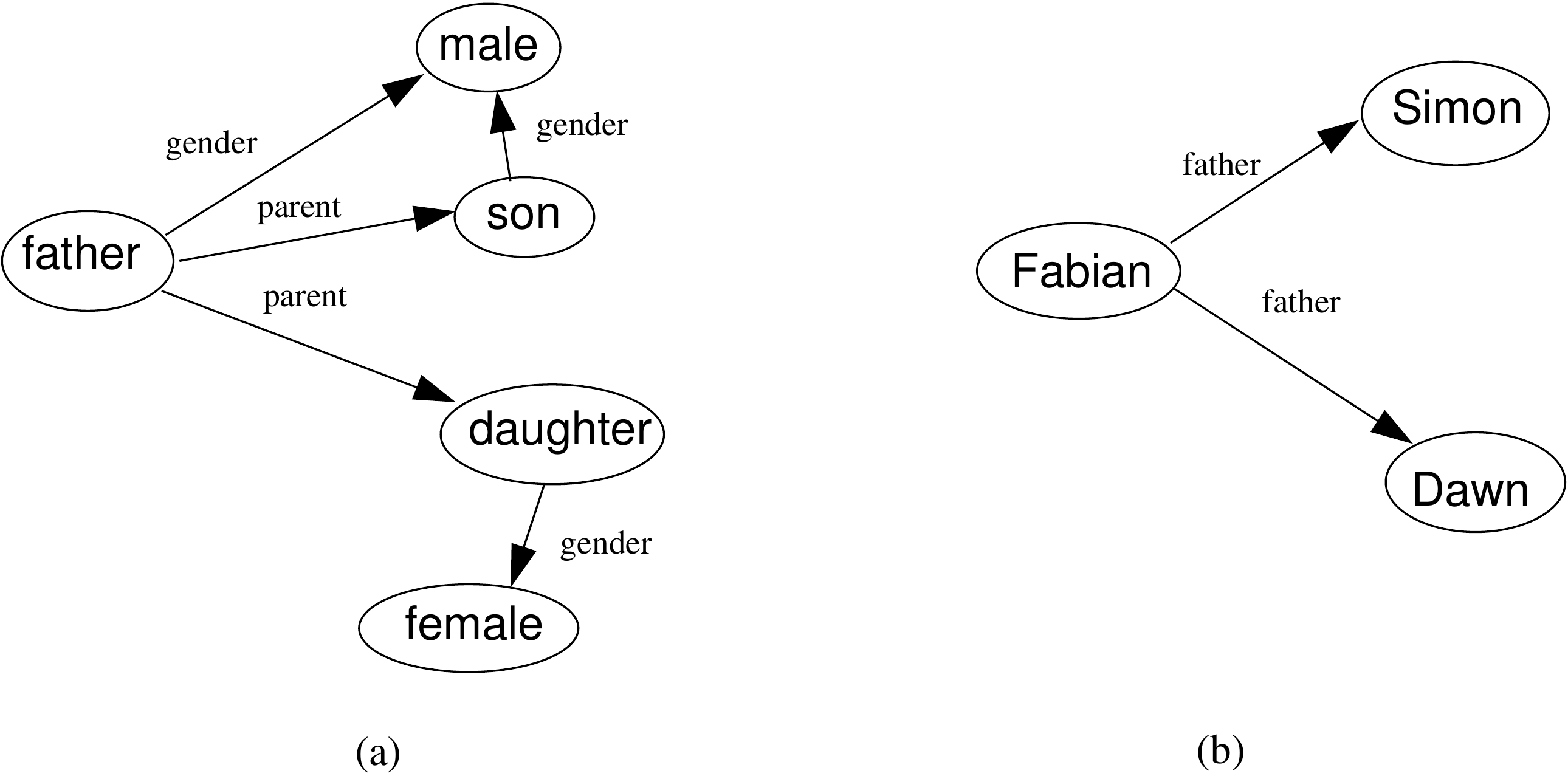}
\caption{\small Different models of the same associative information, one using
names, the other using roles. The named version looks simpler, but the
associations conceal a new concept: father, so they must be reducible.\label{fatherchild}}
\end{center}
\end{figure}
As noted in section \ref{ntassoc}, we can break complex associations
down into irreducible parts\footnote{This is a hypothesis,
  demonstrated in a few cases. It remains to be proven as a theorem.}.
Let's return to the father-child example (see figure \ref{fatherchild}).
The basic associations act as transformations (functional vectors)
that enable immediate inferences to be made:
\beq
\text{Father} \assoc{precedes} \{ son, daughter, child, \ldots \}
\eeq
There is another association to gender, so it is possible to relate
the inverse to the context of the gender. 
\begin{itemize}
\item For gender neutral, the inverse of father is child.
\item For gender male, the inverse of father is son.
\item For gender female, the inverse of father is daughter.
\end{itemize}
We can proceed in this way to deconstruct the complex association
`is father to' as `male' and `precedent', except male is a role
association like `father', so it makes little difference which
we choose. What is useful in terms of spacetime inference is that
the causal precedence is represented, as this can be used
for potentially transitive reasoning.

\item {\bf Combining vertical and horizontal association}

Deduction:
\beq
\text{general concept} &\assoc{\text{likes}}& \text{fish}\\
\text{general concept} &\assoc{\text{has exemplar}}& \text{particular example}\\
\textbf{ infer: }~~~~ \text{particular example}&\assoc{\text{likes}}& \text{fish}.
\eeq
The reverse inference cannot be made:
\beq
\text{particular example}&\assoc{\text{likes}}& \text{fish}\\
\text{particular example} &\assoc{\text{is generalized by}}& \text{general concept}\\
\textbf{ infer: }~~~~ \text{general concept} &\assoc{\text{likes}}& ????
\eeq

Induction:
\beq
\textsc{thing} &\promise{\text{moves}}& \Unspec\\
\text{thing} &\assoc{\text{is characterized by}}& \text{motion}\\
\text{motion} &\assoc{\text{is affected by}}& \text{force}\\
\textbf{ infer: }~~~~ \text{thing}  &\assoc{\text{is affected by}}& \text{force}
\eeq

\end{itemize}

\subsubsection{Questions and answers: contextual stories}

\begin{figure}[ht]
\begin{center}
\includegraphics[width=10.5cm]{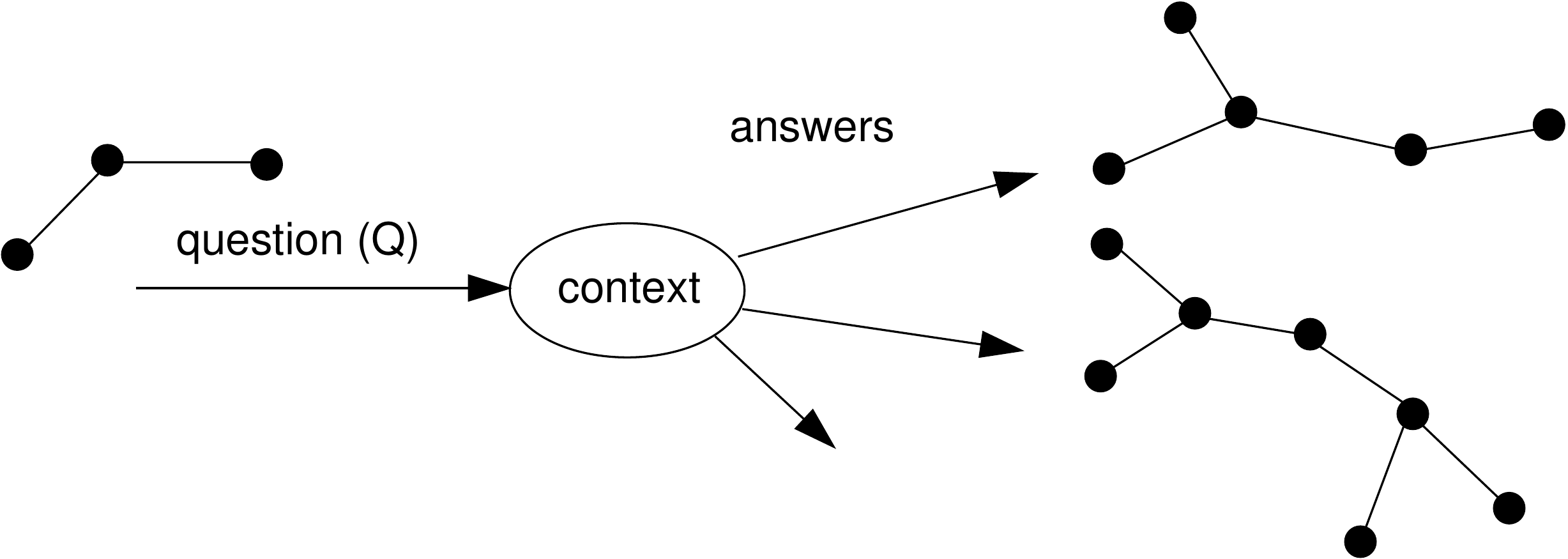}
\caption{\small Questions matched to answers for each context of the question is
a 1:N process.\label{qna}}
\end{center}
\end{figure}

A facility that we associate with smart behaviour is the ability to
answer questions.  Answering questions is similar to the process of
deduction. Learning answers to questions is like contextualizing
input. The reverse is like contextual recall (see figure \ref{qna}). 
\begin{definition}[Question]
  A proposition about a subgraph of concepts and associations (called
  the query set $Q$), or a prompt to tell a story involving the
  mentioned concepts and associations.
\end{definition}
\begin{definition}[Answer]
  A response to a question, proposition, or situation, defining a
  query subgraph $Q$ of concepts and associations, i.e. the telling of
  a story about the query set, in which the query set plays the role
  of the subject of the story. 
\end{definition}
Answers may be given in terms of generalizations and stories about
concepts stated in the question.  It is not necessarily true that the
query set maps to the axiom set of story.  These would be learned by
induction or abduction over time.  When we answer a question, we are
asked to reply with a story matching the context of the question.
Question answering is thus a form of (or at least related to)
deduction.

\subsubsection{The storytelling hypothesis}

The central importance of storytelling to reasoning and problem
solving suggests that no knowledge system can be considered
self-adapating without this capability.  So, it seems natural
to highlight this by mirroring the context
adaptation hypothesis, we can now add a promise that must be given in
order for a knowledge representation to be capable of knowledge-based
reasoning.
\begin{hype}[Separate agent for story inference and navigation]
  In order to extract knowledge introspectively from a knowledge
  representation, there must exist an independent mechanism or agency
  capable of searching and following paths surrounding a concept and
  feeding back the narrative.
\end{hype}
It is fascinating to speculate on the relationship between this
narrative engine and the phenomenon of consciousness in sufficiently
complex knowledge systems.

\subsection{Semantic commentary of structures}\label{annotation}

In spite of the successes of machine learning, and pattern
recognition, humans excel at learning, and each individual carries
vast reservoirs of knowledge and experience. Some level of
understanding can be passed on without extensive experiential
training, through books, stories, or by encoded as expert systems, in
order to be built on by learning or compounded by other multiple
sources.  In the modern approach to so-called artificial intelligence,
much effort is expended to accomplish pattern recognition by expensive
supervised training.  This need not be pursued to the exclusion of the
fruits of human knowledge.  Given a data set, from
some kind of sensor, it may still be helpful for experts to annotate
the data, based on their broader contextual understanding of what is
intended and what is accomplished.  This can be done by annotating
keys and values to explain the semantic significance, at a level that
can be used for reasoning. This is a form of documentation, and may be
applied to any data process to enhance the interpretation of sensory
data.

\begin{example}[Semantic annotation of learning networks]
  Unsupervised learning, with no initial boundary semantics, have no
  way of developing semantic concepts over time. Without external
  guidance, a network that gathers data may see patterns cluster into
  agencies that resemble concepts, but the names of those concepts are
  unknown. Learning systems need grounding in the human realm to give
  meanings to their discoveries that we can understand.  The quick and
  simple way to assign meaning to an artificially learned network of
  concepts is to have human curators name and comment on the
  structures.  This is exactly what happens in the training of human
  brains from birth.

There are two forms of semantic grounding that allow animals to learn
the same concepts: i) is the evolutionary hardcoded information that
has been groomed by selection, and ii) is the reservoir of social
norms and writings that stabilizes behaviours and passes on names in a
long term memory from generation to generation. Mass culture plays a
key role in the development of knowledge. If we continue to give
society tools to communicate in private channels, closed rooms,
cliques and tribes, it seems likely that society would lose its
ability to pass on certain kinds of knowledge by equilibration with a
mass norm.
\end{example}

\begin{example}[Semantic commentary]
Consider a fragment of programming pseudo-code:
\begin{verbatim}
string animal_of_the_day = "vegan lamb replacement";
string text = "Mary had a little " . animal_of_the_day;
integer result = count_words(text);
\end{verbatim}
Using the core association types, we could annotate:

\begin{itemize}
\item \verb+animal_of_the_day+ is a member of \verb+string+.
\item \verb+text+ is a member of \verb+string+.
\item \verb+result+ is a member of \verb+integer+.
\item \verb+count_words+ is a member of \verb+functions+.
\item \verb+count_words+ is a member of \verb+integer+.
\item \verb+result+ depends on \verb+text+.
\item \verb+result+ depends on \verb+count_words+.
\item \verb+text+ depends on \verb+animal_of_the_day+.
\end{itemize}
Hence we may infer that result depends on \verb+animal_of_the_day+ through the following story:
\begin{enumerate}
\item \verb+result+ depends on \verb+text+ and \verb+count_words()+.
\item \verb+text+ depends on \verb+animal_of_the_day+.
\item By inference rules: \verb+text+ may depend on \verb+animal_of_the_day+.
\end{enumerate}
We don't know this for certain, because we don't know whether \verb+count_words()+
actually promises to use \verb+text+. However, we can say that it is possible.
\end{example}
Using this approach, one can generate semantic networks, as 
concept graphs, which are more efficient to construct and update than
techniques that only employ probabilistic networks for machine learning.
The coordinatization of documents allows a practical annotation mechanism
to be implemented as a technology (see the example in appendix A).

\subsection{Summary}

Setting aside the many deep technical issues in realizing concept networks, the basic processes of
knowledge representation are represented schematically in figure
\ref{awareness}.
\begin{itemize}
\item Knowledge is acquired by learning from sensors and by thinking
  (introspection): an iterative process of associating ideas to
  understand and react to the world.
\item Concepts and associations form a tokenized model, analogous to language, of what is sensed
and understood.
\item Concepts and associations form a semantic space, with four irreducible association types.
\item Concepts grow in significance the more often they are revisited.
\item The iterative generation of context from combining sensing and introspection drives
an observer's clock sense of time.

\end{itemize}
These ideas can now be applied to a multitude of representations, to
help understand the process of adaptation, and harness technologies
for improving the value of learning to organisms or materials at all scales.


\section{Smart spaces}

The concept of a `smart space' may refer to any kind of structure in
space and time that mimics or otherwise represents knowledge-oriented
behaviours.  This may be at any scale, e.g. it could include new kinds of material, buildings, homes,
vehicles, towns, etc.  Thus it readily encompasses systems we do not normally
associate with intelligence.
In the human realm, the aim of `smart spaces'
is to enhance the experience of people living in or around them, or
bring a positive benefit to society. In an industrial capacity, the
aim might be to allow specialized functional spaces (like factories,
farms, etc) to become more self-sufficient in operation and adapting
to contextual challenges. In this latter view, smart spaces are a form
of cybernetic organism, or what is currently referred to as artificial
intelligence.
Smart spaces may be addressed from three perspectives:
\begin{itemize}
\item {\bf Smart environment}: where space records knowledge about its
  exterior interactions with occupants and environment, and may be
  equipped to react and interact more intelligently. If the occupants
  are humans, then this is a form of ergonomics, i.e. a fitness for purpose
  strategy.

\item {\bf Smart occupant}: where occupants of a space are equipped to
  interact more intelligently with the space they occupy. If the
  occupants are humans, then this is a cyborg strategy.

\item {\bf Smart organism}: as a single system in which some agents
are mobile and others are fixed, but all work cooperatively, over some bounded
region.
\end{itemize}
Smart need not imply the direct insertion of information technology,
as is sometimes assumed in technology circles. It might simply 
refer to the ability to build valuable stories between initially unrelated
concepts within the space, including connecting people, functions, and businesses
together to adapt to new challenges. From the foregoing
discussion, it should be clear that smart behaviours can emerge in all
kinds of representations, without necessarily trying to bend systems to human ways
of thinking. The immune system remains a powerful example of how an
assembly of `dumb' agents exhibit smart adaptive behaviour.

Several interpretations are possible.
Should we consider occupants (human or otherwise) to be part of a
process (as in a factory), or as guests (as in a hotel), or even as
intruders (as in an immune system), or as symbiotic partners in the
space (as in a shopping mall).  For example, we could think of a smart
home as providing a service to its residents (like a hotel), or as an autonomous
smart building embedded within an external region acting cooperatively
to share resources and minimize waste.  Systems are often strongly
interactive, thus boundaries may have fluid meanings, and clean cut
roles could be difficult to decide {\em a priori}. A smart space might
balance all these viewpoints simultaneously.

\subsection{Interactive spaces}

In studying semantic spaces, we have a choice:
\begin{itemize}
\item Consider how to make tools that enhance the abilities of humans.

\item Consider a structure (e.g. a city, a house, a material etc) not as a
external thing to be learned about in the mind of an observer, but as
representation of a mind itself, looking onto something else or
perhaps introspecting about itself.
\end{itemize}
From the perspective of the latter, we see adjacencies not as
distances but as associations, places as concepts, and the environment
plus its visitors as sensory stimuli.  What might this space now be
thinking about? If an imaginary deity were to look down on the mind of
a city, would it appear to be smart?  If so, at what scale? Would it
appear as a single organism, or as a swarm of interacting
organisms\cite{swarm,kazadi1,kazadithesis}?  Who is it, in reality,
who judges these behaviours, and on what basis?

We may thus pose the question: what is the relationship between smart
spaces and semantic spaces?
Why should we care about such questions? One answer is that our brains
are such a smart spaces that we want to understand.  Another is to ask
whether we could improve our interactions, to better utilize knowledge,
by the formation of new structures that go beyond the scale of an
individual human: teams, costumes, rooms, communities, cities, etc. 

\begin{example}[Where is the memory of a community?]
In the past we have assumed that the memory of a community was the
memories of the people within it, and that their personal
relationships formed the adjacencies to pass on stories representing
concepts to others. Later, we built non-human structures to remember:
imagery, art, monuments, tombs, statues, writing, books, libraries,
architecture, institutions, cultural habits, and so on.
\end{example}

\begin{example}[Social networks as smart associations]
  A social network represents an association between parts of a space
  of possibilities, mediated by humans. It forms a part of a knowledge
  representation.  Links between members of a social network lead to
  clustering of interests, i.e.  concepts.  Recommendations by one
  member of a group get passed on to others until the group represents
  an interest in that topic, while at the same time expanding the
  concept to overlap with other issues.  Social networks lead to
  mixing of concepts, and humans act as message carriers in the
  organism of a space/city/etc.
  Parts of the concept cluster may come at the topic from different
  angles, triggered by different atoms of context during a concurrent
  activation.
\end{example}

\subsection{What does smart mean?}

Smart is an assessment made by an exterior observer. It is based
prejudicially on our understanding of human capabilities and
ingenuity: we define our own idea of smart by our ability to use
resources as tools, to turn learning into practical and useful
behaviours, to innovate by combining resources into new functions, to save
time and effort or optimize processes. We believe that people who `see
more' are smarter.  From the perspectives of scale and relevance,
there are two pragmatic criteria that suggest `smart':
\begin{itemize}
\item When we perceive an action as helpful in a given context (fit for our present purpose).

\item When the outcome arrives `just in time' to be useful in that
  same context (anyone can come up with something too late).  
\end{itemize}
A smart space is about facilitating interaction that leads to
economics benefits, including advancement, innovation and happiness.
Emotional aspects of systems may not be discounted as long
humans make the assessments.

In the preceding sections, we've looked at how spacetime itself can
encode the basic elements of knowledge and reasoning. This may or may
not be sufficient to justify calling a space smart.  An independent
assessment of `smart' is a matter of perspective: each individual
observer decides whether they experience something as smart or not,
relative to their own expectations.  Finally, given our ability
to enhance systems using Information Technology (IT), could we make a space
even smarter?
We might posit that a space could be
considered smart if it could answer yes to a number of questions that
include the following:
\begin{enumerate}
\item {\em Does the space have some interpretation of sensory input, i.e. does it interact some a world beyond it?}

  Any space that is bounded, and which interacts with an external
  agency has the capability to exchange information. Where do we draw
  the line? Identifying the agents on the perimeter, what
  promises do they make, and how do these reveal the world as a sensory
  experience? Are there new agents and sensors that
  we can equip the space with the extend these capabilities?

\item {\em Does the space have a capacity to learn and encode concept and associations?}

  Any space, which can represent different internal states, possesses
  this capability, but does it have associative and temporal memory,
  it is plastic in its contextual adaptation? Can it mix and combine
  ideas into new ones (innovation), and is it fast enough to apply its
  learning in interactions with real time changes in its exterior
  environment? Can we extend its capabilities with artificial means?

\item {\em Do the normal dynamical processes of the space tell stories about itself?}

  Is there causal linkage between remembered events, forming a
  meaningful narrative?  Can a space reason independently, through its
  own processes? How could this reasoning benefit occupants or users?
  How we answer that question probably depends on our expectations.
  There are very simple examples of reasoning that we call algorithms:
\begin{example}[Mechanical reasoning]
  The stopcock in a water closet (the switch that fills up the water closet tank
  and stops filling when it is full) is a reasoning system. It has a
  sensor, which is the water level. The sensor forms a sense of
  context: full or not full, and transduces this into a switched on a
  water valve. Thus it understands two concepts: full (represented by
  the switch being open and the valve closed, which usually implies
  that the tank is filled with liquid) and not full (represented by
  the switch being close and the valve being open, for any other
  condition).
\end{example}

\item {\em Does the structure of spacetime alter the way new sensory inputs are perceived?}

  Do pre-existing concepts feed back into the sensory stream, acting
  as anchors for new ideas or external interactions to strengthen and
  expand upon?

\end{enumerate}
Smart appears to be something to do with a plasticity to
challenges, and innovating quickly around them, with approximate
goals. In promise theory language, we could define smart like this:

\begin{definition}[Smart promise or function]
We perceive an action as helpful in a given context (fit for our
present purpose).  The action arrives `just in time' to be useful in
that same context (anyone can come up with something too late).
\end{definition}

\subsection{Promise view}

Given that we can relate promises to knowledge representations,
from a promise theory viewpoint, we may begin by asking the usual questions
about the cases of interest, defining inside and outside of what we take
to be the semantic space, marked by its horizon.
\begin{enumerate}

\item {\em What or who are the operational agents inside and outside the semantic space?}

  In an arbitrary space, we can find agents, with diverse capabilities,
  at all scales.  Atoms, molecules, bacteria, viruses, people, tools,
  vehicles, machines, building, regions, malls, warehouses, shops,
  streets, shop signs, road signs, public structures, monuments,
  bridges, districts, parks, cities, countries, planets, etc.

  How are these agents ordered and addressed?

\begin{itemize}
\item Should we distinguish between the semantic space and its occupants at all, or simply treat
a system as self-contained, with some agents mobile and others fixed? If so, the occupants would
be formally outside the space in our model (even though the occupants of a smart house are within it
from a normal spacetime perspective).
\item What processes happen inside the space?
\item How does the space sense and interact with across its boundary or with its occupants, however they are defined?
\item Who or what is doing the thinking in the system, to judge `smart'?
\end{itemize}

\item {\em What promises do the agents make to one another, in what context?}

  Proximity, bonding, services, or how may we interpret these as
  knowledge of associations, by one of the four irreducible types or
  any possible alias of those.

\item {\em What phase(s) are the agents in?}  

  Gases, bacteria, animals, people, and vehicles are usually in a
  gaseous phase, as they move around and environment that is fixed
  with solid bindings.  There are anchored places, forming concepts,
  with freely flowing messengers to associate them.

\item {\em What are the important scales for their interactions?}

  For a knowledge representation, scales are pinned by the scale of
  sensory apparatus. A smart space interacts with its exterior through
  such sensors. 

  \begin{example}
    A smart material may feel pressure on a surface; a smart city may
    experience trade and migration on a timescale limited by surface
    area of transport hubs; a smart house is built for 1-10 people who
    have a fairly predictable size and speed. Dependencies (like
    transportation and interface surface) condition what happens
    inside a space on what happens outside it.
\end{example}
\end{enumerate}
Based on the principles of promise theory, using only agents and promises,
one may now begin to map behaviours encoded in different spacetime media
to challenges faced by a space formed from a collection of such agents.
Inside the observer horizon, we can build the notions of concepts and associations
on top of these. From this,
if there is a plausible similarity between such behaviours and the behaviours
we understand as smart in organisms, then it is fair to say that  we have
understood the meaning of an analogous smart space.

\subsection{Benefits of a knowledge representation in real space}

\subsubsection{Memory}

The ability for a functional space to remember something depends on
the shaping of structures within its observer horizon.  Any persistent
spatial configurations and spacetime processes (orbits, loops,
interactions) may be used to represent an addressable memory inside
the horizon, and this, in turn, can be used for learning. The
speed and spatial capacity of memory are determined by all the spacetime
scales of the agents.  Memory could be short term and long term,
coarse or refined. Realizing a knowledge representation in real space,
is a process of iterative learning, by direct analogy with learning by
an individual, resulting in:
\begin{itemize}
\item {\bf Concepts}: formed from any agents or patterned clusters of
agents.  
\item {\bf Associations}: links, by adjacency, or by realtime
processes, e.g. transportation or messenger.
\end{itemize}

\begin{example}[City memory 1]
  In a city, short term memory is found in the locations of mobile
  agents like people, cars, taxis, work crews, or in the mutable
  states of agents, e.g. the state of traffic lights, the number of
  cars in a parking lot, or the stock levels in a shop. The wear and
  tear on roads and public infrastructure recalls its usage.  The
  memory of sunshine could be the worn-away grass from walking or
  playing in the park.  This imprint might be associative, as it joins
  locations the start to the end of the track.  On longer timescales,
  buildings records cultural history.  The locations of parks and
  monuments.  In buildings, allusions to Greek styles or familiar
  forms (doorways, windows, etc) root function in historical memory
  associations.
\end{example}
\begin{example}[City memory 2]
  A cloud of pollution is a spacetime trace that recalls a process or
  event that caused it. If the process continues, it will be
  replenished and the trace will be learnt by becoming persistent,
  e.g. smog.  If it is ephemeral, it will be quickly forgotten.  Long
  term memories may survive as  chemical traces in concrete or trees.
Other external sensory events may leave memories over time, e.g. an earthquake
may result in a change in building codes, and stronger buildings.
This would be the memory imprint of the concept of an earthquake.
\end{example}
Stored memories must also be recalled in order to be of value. In
other words, recognition has to be mutual. 
\begin{example}[Vestigial structures]
  Greek columns on the face of a building recall Greek culture or
  symbolize grandeur, but the allusion is meaningless unless it has
  resonance with observers' and their shared ability to recognize the
  allusion. Similarly, the presence of a phone box, or disused railway
  tracks, or wartime bunkers.

We may also grow accustomed to the Greek facade and relearn its
significance without any direct transmission of intent.  These are
memories encoded in structures, without a doubt, but they do not make a city
smarter unless they assist other processes in the space in overcoming
challenges.
\end{example}
It may be agreements, rather than promises alone that lead to memory.
In every learning relationship, there is a mutual interaction leading to a local
Nash equilibrium\cite{nash1,certainty}.  This is why bidirectional
iterative relationships are the key to knowledge.

\subsubsection{Sensor types and dimensionality of context}

On the outside of an observer horizon, space may extend into dimensions other than
the obvious spatial dispersion of locations joined up by vector promises; at each
scale, there may be interior scalar promises, at embedded locations
that encapsulate their own semantics; these can always be transformed
into a representation which uses vector links to link them to an other
agent. Human sensory perception probes not only spacetime but
additional dimensions of description, like heat and light, etc.  Tables \ref{sensortab} and
\ref{sensortab2} shows a few sensors that could be interpreted as
inputs to a smart civic region.
\begin{table}[ht]
\begin{center}
\begin{tabular}{|l|l|}
\hline
Coarse grained sensory context atom & sensor type\\
\hline
\hline
day, time, location           & spacetime\\
light, dark, red, green, blue & sight\\
hot, cold, free, constrained  & touch\\
salt,sweet, sour, savory, etc     & taste\\
\hline
\hline
open, closed & building door\\
open, closed & building window\\
open, closed, partial & building window blinds\\
\hline
\hline
open, closed & city bridge\\
blocked, open, snowed under & city road\\
available, unavailable & city shop supplies\\
parking, parked, shunt, cruising & city transport\\
\hline
\hline
inside, outside & sense of self interior/exterior\\
incoming, outgoing & external transport\\
supply, demand & external trade\\
energy/money usage & external trade/accounting balance\\
\hline
\hline
Person by face & spacetime pattern\\
Vehicle by number or appearance & spacetime pattern\\
Pollution cloud & spacetime pattern\\
Flooding & spacetime pattern\\
Weather & spacetime pattern\\
\hline
\end{tabular}
\end{center}
\caption{A few sensors that could feed into a large scale civic smart spaces.\label{sensortab}}
\end{table}

\begin{figure}[ht]
\begin{center}
\includegraphics[width=12.5cm]{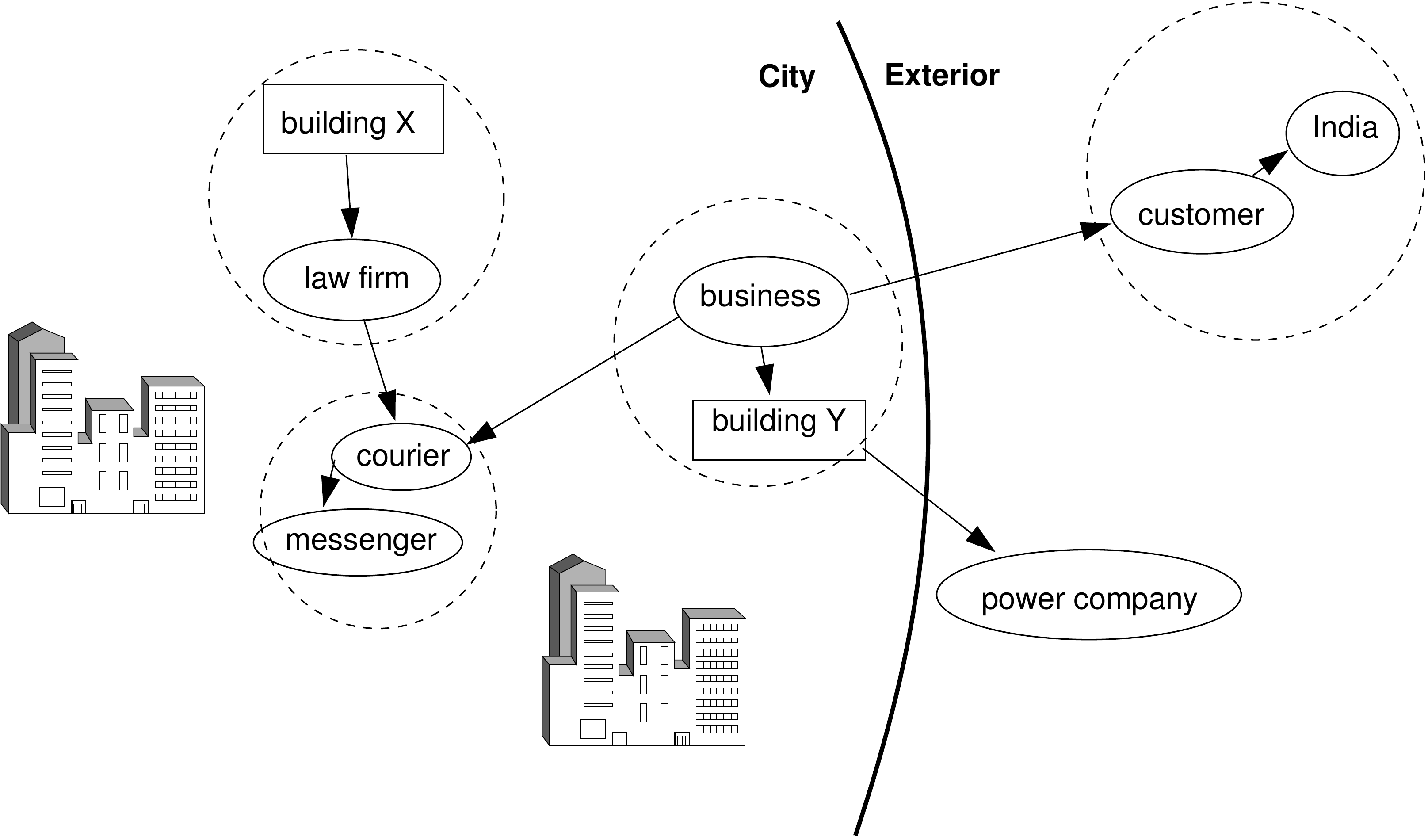}
\caption{\small A city as a knowledge space. A business interacting with a remote
customer acts as a sensor viewing the exterior world.\label{storycity}}
\end{center}
\end{figure}

\begin{table}[ht]
\begin{center}
\begin{tabular}{|l|l|}
\hline
{\bf Sensor} & Spacetime scale / usage\\
\hline
\hline
Smart phone information & \\
\hline
Lidar & Laser ranging, 100m\\
\hline
Radar & 1000m\\
\hline
Microwave & Astronomical and atmospheric\\
\hline
Gyroscope & Orientation\\
\hline
Microphone & Audio analysis\\
\hline
Camera & 10m \\
\hline
Edge detection & Computer vision, object classification\\
\hline
GPS & positional tracking\\
\hline
Seismic & used in oil and gas, geological sink hole discovery\\
\hline
Sonar & deep sea\\
\hline
(f)MRI & brain and body\\
\hline
Neural net pattern recognition & Trainable\\
\hline
CANbus & \\
\hline
\end{tabular}
\end{center}
\caption{A few sensors that could feed into generalized smart spaces.\label{sensortab2}}
\end{table}
Hierarchies of encapsulation have always been an important way of
separating scales, and decoupling concerns. It is this process of
aggregation (rather than fragmentation) that enables economical
scaling.  Current information technology has access to a growing array
of sensors, at all scales, to measure increasingly specific
semantically distinguishable patterns. In addition to simple physical
sensors for scalar promises like heat, humidity, light, etc, agents
who interface to the exterior world can also act as sensors for the
interior (see figure \ref{storycity}), and, of course, there are
artificial information channels, including sophisticated neural
networks that can be trained to recognize almost any complicated
pattern, representable as ordered collections of agents, with tensor
structure, that promise a wide range of characteristics.  What we need
to ask, in order to design appropriate future 
technologies, to exploit information, is how should we organize the
spatial interactions in such a way as to make desired outcomes and processes
efficient.

\subsubsection{Spatial reasoning with concept and association structures}

If a space is smart, in the sense of representing concepts and
associations about its external interactions, then what do the
internal structures of a space say about those external structures,
and at what scale? The sensory scales in table \ref{sensortab} show
examples.  While each space has a maximum spatial scale, limited by
its size, we could extend probes almost indefinitely at smaller
scales by equipping a space with measuring devices, or at larger scales, through
satellite observation, to create something analogous to 
highly extensive nervous systems\footnote{The huge
  extent of human sensory nerve bundles is no doubt responsible for
  the powerful and complex sensations we experience, both by direct
  stimulation, and via the coarser emotional states that are based on
  them. In simpler machines, the analogy of emotional sensations would
  be limited in expression by the palette of sensory inputs it has to
  draw on to represent sensation.}.

Do the conceptual features of the space enable it to think for itself,
or merely serve up data for human consumption? In the words of the
analogy, do they have an active functional significance or merely a
passive archival significance? Do they trigger new ideas and
associations (processes) by virtue of being there (see the Tsunami
example \ref{tsunamiex}, where the concepts became processes)?. This is
what is meant by {\em introspection}.

\begin{example}[Can a smart city think?]
  Could a city really possess a memory analogous to a human, or a
  computer? Memory is everywhere: from the patterns of worn grass,
recalling a hot summer in the park, to the architecture of the buildings
that recall historical and cultural periods and artefacts.
Could it behave like a simple organism? I believe this is also
  plausible.  First, a computer memory is not like human memory: it is
  a linear transcription of data, with an encoding into language, or
  video, or some other kind of media formatting. Computer memory is
  based on a model of the sensors and retrievers. When we record or
  play video, there is a model of a wide-screen television involved in
  the coding of data; there is no interpretation of the data. That
  kind of memory may exist in humans too (at least for short term
  memory) but beyond that, there is a long term summary of experience that
  transmutes raw feeds from sensors into an approximate spacetime
  invariant model, using a lot of summarization, interpretation, and
  conceptualization. 
  Crucially, it is possible to mimic some of the behaviours of human
  cognition to imagine `smart' behaviour by imagining a series of
  transformations from raw sensor data to generalized concepts and
  associations between them.
\end{example}
\begin{figure}[ht]
\begin{center}
\includegraphics[width=12.5cm]{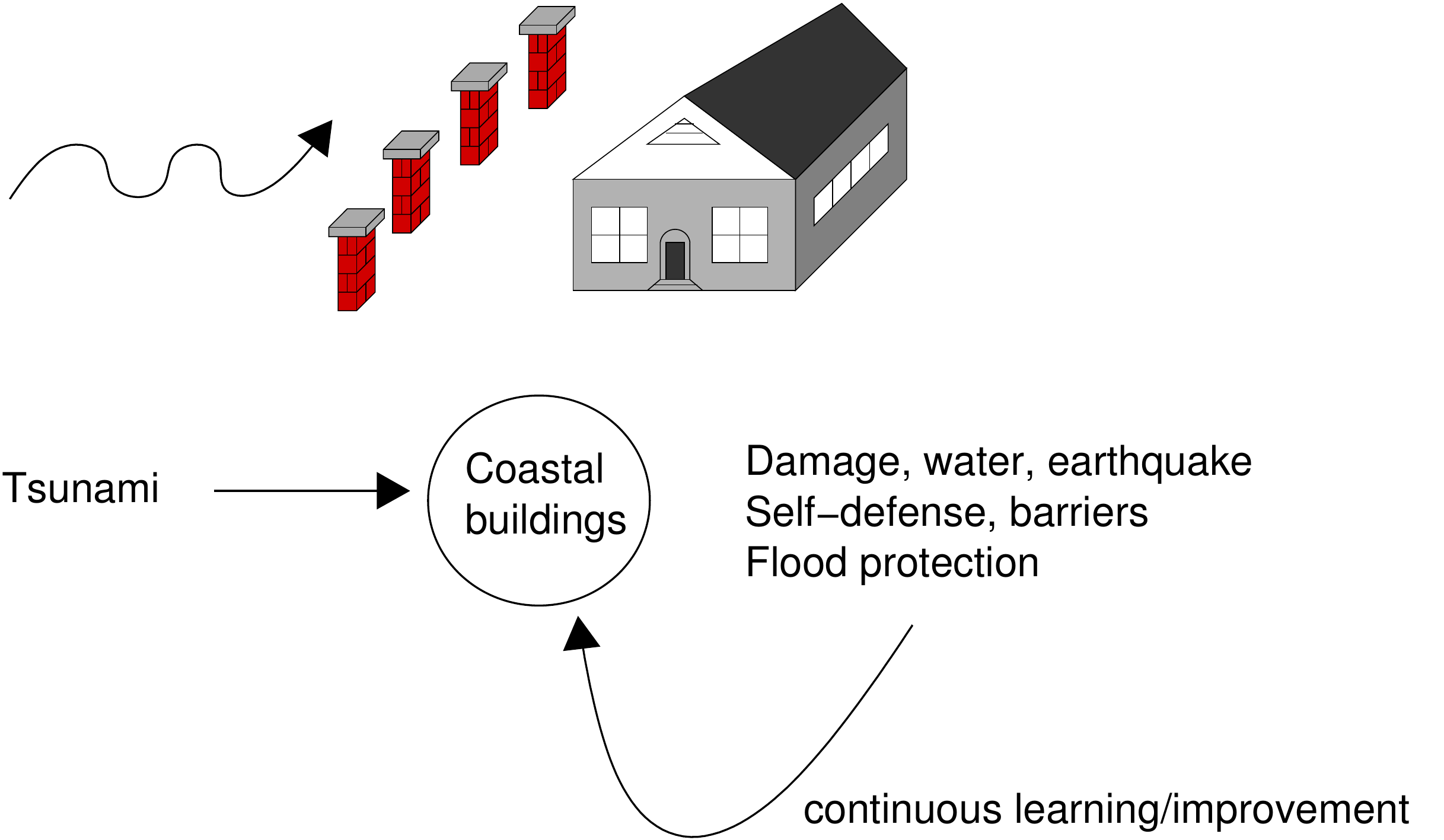}
\caption{\small Learning about Tsunami in a smart city: flood defenses
are the first line of memory that accumulates.\label{tsunami}}
\end{center}
\end{figure}
\begin{example}[A cognitively responsive smart city?]\label{tsunamiex}
Imagine a city by the sea, which is vulnerable to storms and Tsunami.
To image a cognitively smart city, we may ignore the humans living in
it, and just think in terms of abstract processes (see figure
\ref{tsunami}).  How would the city remember a Tsunami? The sensors
would be the coastal structures, which would be flooded or destroyed.
Over time, the response to learning a bout Tsunami would be the
building of flood defenses. This is cumulative learning loop, and
results in a memory imprint: the flood defenses. As these are being
built, the patterns are learnt and summarized, recorded in books and
documents.  The events are symbolized in movies, etc. Associations
form between Tsunami and related concepts: an alarm system to trigger
flood defenses and alarms on earthquakes is built. This is the
physical encoding of an association between the concept of earthquake
and Tsunami. So now there are specialized sensors capable of
transmuting raw data into learning, over time, through the
construction of new hardware in the city. Memory, concepts and
associations have been encoded in the physical structures of the city.

The test of a memory lies in the recall. To be somewhat like a human
memory, recall should be triggered by the occurrence of an earthquake,
or the imagination of an earthquake. So now consider what happens:
when an earthquake happens, it triggers an autonomic Tsunami warning,
and flood defenses activate.  This signal changes the behaviour of the
city and the processes of recall are activated. If something like an
earthquake occurs, then it might trigger the alarm too (a false alarm,
to be sure, but nevertheless a semantically related event that
triggers the memory of a Tsunami). Finally, the city could simply be
dreaming, with no inputs to the earthquake or tsunami detectors: if
something accidentally triggers the alarm, or if all the alarms are
deliberately tested by sending a planned theta wave of activations
across the city, then the defenses are once again activated as if in
rehearsal, and the memory is recalled.
The autonomic response of the city configuration is the key to
triggering a memory based on a change of configuration.
\end{example}

\begin{example}[Memory at different scales (encapsulated specialists)]
  If concepts are not represented directly by the civil engineering of
  a space, e.g.  buildings, parks, transportation, and bridges that
  encodes information about the outside, then is there documentation
  in processes, books and institutional habits?  Would it be a smart
  city that paid attention to the wear and tear of usage in the park
  lawns, and altered its behaviour to build paths along these trails?
\end{example}
The irreducible association types map to the basic processes in
spacetime:
\begin{center}
\begin{tabular}{|c|c|}
\hline
Irreducible Association & Semantic function in smart space\\
\hline
Causal association & Sequential spacetime processes, data storage\\
Containment & Conceptual generalization and accumulation\\
Molecular composition & Compound roles, identities and names\\
Proximity & Lateral association and approximation\\
\hline
\end{tabular}
\end{center}

\begin{example}
  The factories supply an external region around them, and response to
  their needs so the internal state of the industrial region mirrors
  the sensed supply or demand.
\end{example}

Associations can be dynamical
Stories that play out as processes
\begin{itemize}
\item Agents and processes associated by compatibility of promises
\item Ability to align for a common purpose
\item Mixing concepts to bring about innovation
\item Identification of similarities alternatives (functional thesaurus)
\item Speed of recognition: awareness of situation.
\end{itemize}

\subsubsection{Temporal reasoning with pathways and stories}

Processes and connected pathways that join concepts via
associative connections, within a smart space, lead to stories,
as described in section \ref{storysec}. Stories enable
semantic predictions, by relating compressed tokenized concepts.
\begin{example}
  The arrival of a load of fish or fruit to a town market stimulates
  processes that activate the town and mirror the external process of
  harvest.
\end{example}
Here are some examples:
\begin{example}[Episodic story]
An example story about a single agent, represented in city processes:
\begin{enumerate}
\item A person eats breakfast, 
\item then drives to the mall, 
\item then goes shopping,
\item then eats lunch, 
\item then sees a movie.
\end{enumerate}
An episodic story describing the world-line of an agent within the
space.  This represents a kind of process, but one of limited
significance to the whole city or its outside environment. The story
describes the process of a part in a machine, because it is about the
orbits of only a single agent. What we could say is that the single agent
introspectively binds together the concepts of breakfast, mall, shopping, lunch and movie
as a single concept, whose significance remains to be decided.
\end{example}

\begin{example}[An associative story]
  Associative stories can reveal dependencies and be useful for
  diagnostics, see figure \ref{storycity}. For example, a law firm in
  the city fails to perform an important transaction because:
\begin{enumerate}
\item The law firm is located in building $X$
\item The building is connected by courier to a business client
\item The business client is located in nearby building $Y$,
\item Building $X$ depends on power from an exterior company. 
\item A power failure in building $Y$ prevents the business client from receiving word from a customer in India.
\end{enumerate}
These associations are causal in nature, and lead to direct
predictable consequences\cite{stories} within the city, e.g. perhaps loss
of business for the city. Still, the processes of the city do not
represent the world beyond.
\end{example}
\begin{figure}[ht]
\begin{center}
\includegraphics[width=8.5cm]{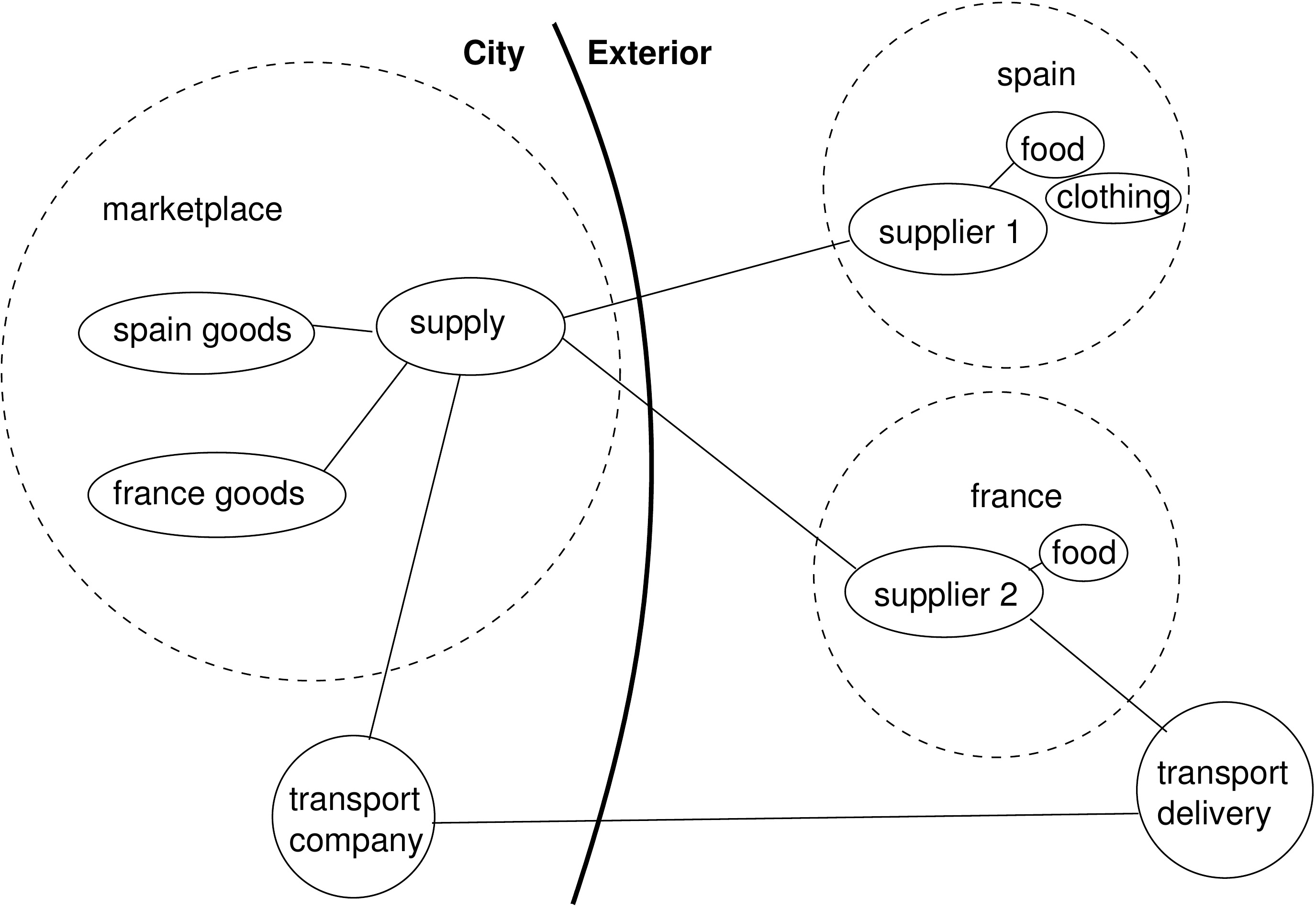}
\caption{\small City structures are a facsimile of the world outside, thus encoding
concepts of the countries beyond in a knowledge representation using a marketplace.\label{storycity2}}
\end{center}
\end{figure}

\begin{example}[Exterior agent stories]
Consider a view of a city in which the internal structures are direct representations 
of the outside world (see figure \ref{storycity2}).
\begin{enumerate}
\item Many suppliers deliver to a supply chain hub in the city. The supply chain
hub encodes the tokenized concept of supplier/supply.
\item The suppliers are variously from France of Spain, depending on context.
Simultaneous activation means that suppliers from both France and Spain become variations
of the same concept, but also that France and Spain get represented as separate
entities in the city: a French food market and a Spanish food market.
\item The generalization of these is the concept of a market.
\item The city observes that deliveries depend on the transportation
  of goods by lorry.  The shipping company has its own support depot
  and business office in the city, mirroring the external activities.
  Thus the memory concept of transportation is encoded by the local
  office.  This is associated with the supply hub, by mutual
  activation, when the lorry arrives at the hub, with the logo of the
transport company.
\end{enumerate}
\end{example}

\begin{example}[An interior agent story]
An example story represented in city processes:
\begin{enumerate}
\item A person eats breakfast, 
\item breakfast contained coffee
\item coffee was purchased from coffeehouse
\item coffeehouse belongs to a franchise
\item the franchise is owned by the coffee group.
\end{enumerate}
This is a pure pathway in a semantic graph. The story significance of
the story is in the eye of the city, e.g. it might be of benefit in
diagnosing a connection involving the person concerned to a particular
batch of coffee, as a way into an abduction process. For instance, the
coffee might have made some other people ill; it would therefore be
further associated with a story about visits to a hospital for medical
attention, etc.
\end{example}

\begin{example}[A process story]
  The arrival of snow from outside a city is a sensory experience that
  could trigger drifting and road blockage (see figure \ref{snow}). That is associated with
  snow clearing (or perhaps later flooding). Snow forms the context
  for these associated concepts and the city may recall and even enact
  several stories.
\begin{figure}[ht]
\begin{center}
\includegraphics[width=5.5cm]{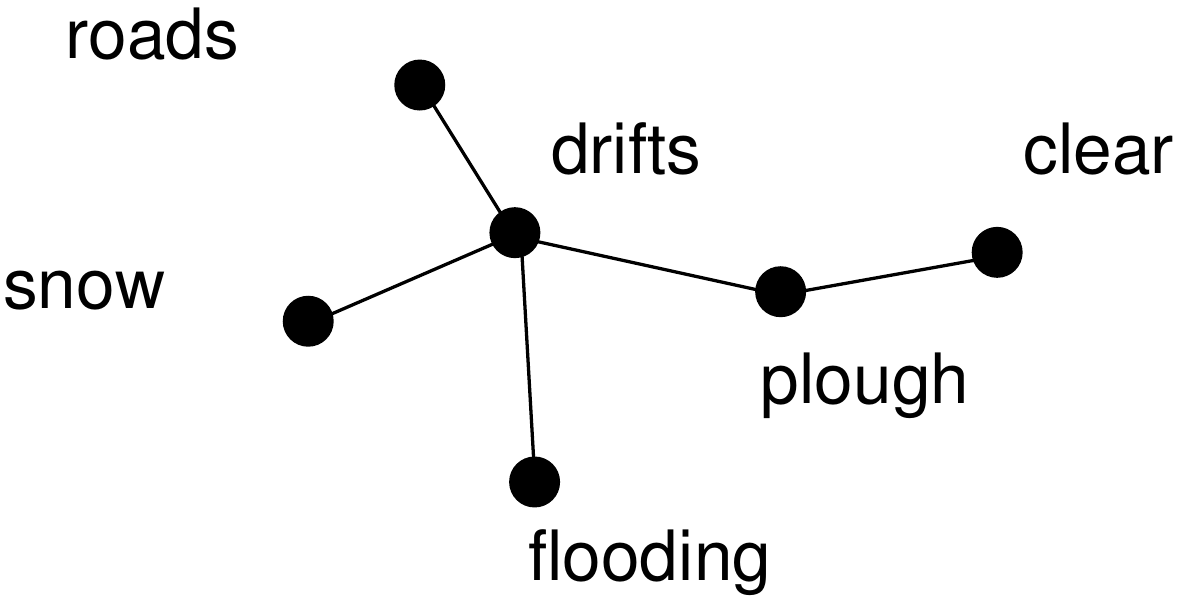}
\caption{\small Sensory activations and concepts for a city associated, leading 
to several possible stories.\label{snow}}
\end{center}
\end{figure}
The story may be extrapolated further by association and
generalization to make predictions like occurrences of flooding or
hypothermia, etc.  These predictions can have actionable consequences:
perhaps roads, bridges, or airports are closed.
\end{example}
These examples show how the processes of agents, in a sense, tell
their own stories, within the contexts of a city that faces an outside
world.  People and vehicles, as mobile agents, act as messengers,
binding city concepts together, or acting as sensory stimuli.  As they
move, leakage and random motion around the conceptual focal points
leads to an effective blurring and association of ideas.

Irrespective of whether one buys into to notion of a smart space
without information technology, it is clear to see, from these
examples, how information technologies could be used to bring
plausible depth to this story, whether for the benefit of an abstract
space (region, city, building, etc), or for the people who inhabit it.
By designing appropriate sensory probes (detecting specifically
interesting semantics, rather than just raw physical observables), 
one gives immediate relational meaning to context and long term memory.

\subsubsection{Locality, centralization, and economies of scale}

Not all smart spaces are equal.  Some may have more knowledge than
others, but we can recognize, measure, and perhaps improve smartness
by understanding how it occurs.  Concepts form localized clusters,
represented as a promise graph, via all of the agents and their
interactions. In semantic knowledge space, concepts may be localized
as single concept agents. This does not imply that the physical
agents, which represent such concepts, have to be localized in a
single location. Culturally, there is a tendency to centralize and
focus on clear locations: cognitively, we prefer low entropy
structures to represent isolated concepts, because they are easily
identifiable; this may lead to practical bottlenecks in accessing the
knowledge or location.

Contextualization has the effect of dissecting a single concept into
several smaller parts.  Collective centralization of any shared
service or function is an expression of identity: it allows us to
visit, recognize, and learn to know things, as if they were themselves
agents, with their own personality\footnote{This is how humans `know' each
  other and know things: we form a relationships with them as if they
  were characters in a story, hence our tendency to anthropomorphize
  concepts.}.  Thus, humans can use the same mechanisms that we evolved,
to learn and know other humans, for learning and knowing other abstract
things.

\begin{example}[Where is a city's knowledge?]
  Does knowledge lie in a city, or in the people who live in the city?
  The answer we discern from this discussion is that it lies in both.
  The interactions between people and things are part of processes
  that lead to new concepts and associations. As automation takes on
  more of the jobs that drive a city, more of the knowledge will be
  encoded in processes that belong either to the city, or to a
  placeless repository of information.
\end{example}

Easy access to functional services that we depend on suggests
localization.  Close proximity promotes speedy access, but context
switching can also provide fast routing to a conditional part.
Connection, or adjacency, may be arranged in two ways: either by
topological and geographical proximity, or the availability of
messengers to act as go betweens.  Specific concepts or functions may
be available to us through either one of these mechanisms.
Communication and transport play a central role in enabling processes
to associate agents.

\begin{example}[Cities or communities?]
Traditionally we think of cities and geographical communities, but
the development of modern communications has rendered this a less useful
definition. When a city resident moves beyond its geographical domain,
does he/she leave it behind, or carry it with them?
In other words should we say:
\beq
\text{space/city} &=& \sum \; \text{agents ?}= \sum \; \text{locations ?}
\eeq
Many processes, like taxation and personal health issues, are connected
to our place of residence. What happens when we are no longer within
our home geo-political domain? How distributed can a representation of
a process be? Could we transform a static adjacency into a mobile utility?
It is interesting to speculate on how smart adaptations could evolve
the city's sense of awareness to cope with this.
\end{example}

\begin{example}[Partitioning: rooms and districts]
Limiting concept scope helps us to grasp concepts with less processing.
We make buildings for specific purposes: shops, houses, museums, offices, etc;
we also make districts around specific economies of scale: shopping malls,
financial districts, red light districts, etc.
As we embed a concept in a larger framework, i.e. as a
region of space gets bigger, central points become harder to visit due
to congestion of pathways, unless there are direct routes. Scale
therefore plays a deciding role in identifying concepts and making use of
knowledge. The Dunbar limits place bounds on how large human work groups
can become while remaining effective.
\end{example}

\begin{example}[Implicit learning in a space]\label{sunworship}
  On a street, in jazz city, during the context of daytime, the sun
  shines on the north side of the street, and south side is in shadow
  all day long.  Over time, sun worshipers gravitate to the sunny side
  of the street, while night birds flock to the shady side. Thus the
  street remembers the concept of sun, and associates shops, clubs,
  and services with the happy sun people on the north side, by
  concurrent activation.
\end{example}

\begin{example}[Economies of scale in smart cities]
  Recent work on what makes cities work and grow identified the
  economics of infrastructure at a key part of the
  scaling\cite{cities,bettencourt2}.  What smart knowledge base could
  assist in facilitating economies of scale in a city?

  Economies of scale relate to specific goods or services. To scale
  these by introducing more agents involved in the keeping of
  promises, some measure of coordination is required. A calibrated
  source concept provides the semantic basis for making a quantitative
  promise.  The association is semantic but the physics lies in the
  real world representation.

\end{example}

\subsubsection{Central controller versus brains}

For many, the intuition of what `smart' means suggests a reasoning
brain', or some central guiding intelligence. We imitate some of the
the symptoms of human intelligence by using algorithms and processes,
with the help of machinery that is mechanical or computational, as if
trying to place a homunculus within as the {\em deus ex machina}. This
form of mimicry uses spatiotemporal memory to emulate behaviour through
processes rather than through static structures.  Our
attachment to brains is clearly a liability in automating processes.
Human brains are quite over-qualified for many of the tasks to which we
apply ourselves: far simpler mechanisms can easily cope with the state
spaces and constraints of isolated processes in, say, manufacturing or
quasi-stable repetition (e.g. autopilots).  If smart implies quick and
timesaving, then devices much simpler than ourselves may be called
smart.

Turing's halting theorem tells us that we can simulate any computable
result with a brain with a central Turing machine eventually, but this
is not to say that a brain is a Turing machine, works anything like
one, or that such a simulation would work at even close to the same
timescale\footnote{Ironically, once we've finished constructing
  automata to mimic smart behaviour, the proxies which enact
  programmed patterns often seem very dumb to us, because they are
  frozen into behaviours that are little more sophisticated than the
  opening and closing of a flower in response to sunlight.}. Of
intelligence, we expect agents that adapt {\em continuously} and {\em
  cumulatively} to their context, or have awareness of sufficient
local state. The emergent behaviour of an immune system exhibits smart
behaviour, repairing us when we are sick and mostly sleeping when we
are well, yet it is the most decentralized of systems in our bodies\cite{lisa98283}.

\begin{example}[IBM's True North `cognitive computing' architecture]
  In 2014, IBM introduced an extensible hardware programmable chip
  architecture called True North\cite{truenorth}, with a parallel
  cloud simulator `compass'.  The modular design and inbuilt routing
  capability allows users to build massive deep layered neural networks
  with a quasi-deterministic event driven input scheme based on data
  oriented signal agents called `spikes'. As a programmable fabric,
  this architecture enables massive learning capability, with
  connectivity approaching the scale of a human brain, and at low
  power. So far demonstrations have been limited to quite simple,
  traditionally recognizable tasks. Crucially, this architecture shows
  how non-von Neumann architecture computational fabrics may be built
  as relatively dumb (but programmable) semantics spacetimes at a low
  level.
\end{example}

\begin{example}[Smart organization]\label{sorg}
  Many firms adopt hierarchical forms of organization.  Large firms
  may have up to 10 or more layers of management, arranged in a
  hierarchical tree. Can a management hierarchy act as a trained
  neural network, i.e. a learning space that has its own knowledge
  representation?

  The structure shown in figure \ref{firm} is drawn deliberately to
  suggest a layered neural network structure, with the `top' of the
  company to the left. Customer demand is the sensory apparatus of the
  organization, which enters through top level management and sales
  (the customer facing part of the organization). The inputs get
  handed down through product managers and developers to final
  outputs, which represent the products or services of the company.
  These, in turn, feedback back around to the inputs allowing the
  company to `think' as a continuous iteration of products and
  services\footnote{In modern parlance, this process is called
    Continuous Delivery\cite{humble}.} This has the superficial
  structure of a semantic learning network, in which concepts are
  represented both by agents and clusters of agents. Unlike what one
  would normally consider to be a neural network, the agents are
  humans (or perhaps processes), which are themselves already quite
  complex networks, so the learning capabilities are quite different
  to an artificial neural network structure, made from dumb agents.

  The assumption here is that there are tight horizontal connections,
  and no vertical connections in the layers.  In a real organization,
  the layers also collaborate as teams, not as individual nodes. This
  may help or hinder the learning. Since, in a human organization, one
  assumes that the nodes are not dumb threshold devices, but contain
  significant memory each, there is useful equilibration and
  redundancy by having intra-layer connections. What we must have is
  stronger interlayer connections to forward learning and connect
  inputs to outputs.

  The hierarchical structure represents a directional routing
  structure, or an aggregator of information, if the agents promise to
  faithfully forward information up or down the hierarchy.  In some
  companies the elements of the layers do not map to all the elements
  in the next layer. There are branches for `separation of concerns',
  sometimes called a silos, in which each branch attempts to represent
  a different concept. From a semantic space viewpoint, this is a
  mistake, as it tries to hardwire concepts rather than learning them.
  Concepts like `customer type', `trends', `competitors', etc, may be
  distributed throughout the layers in a neural network (without
  necessarily being named explicitly). In a branching hierarchy, the
  associations between layers are encoded only through their `direct
  reports' or managers, leading to association pattern (b) in figure
  \ref{association1}, which is the observation of a coincidental
  correlation if a branch manager imposes coordination on the
  sub-managers, and may be reported calibration (d) if the
  sub-managers promise an agreement with their mutual manager
\begin{figure}[ht]
\begin{center}
\includegraphics[width=13.5cm]{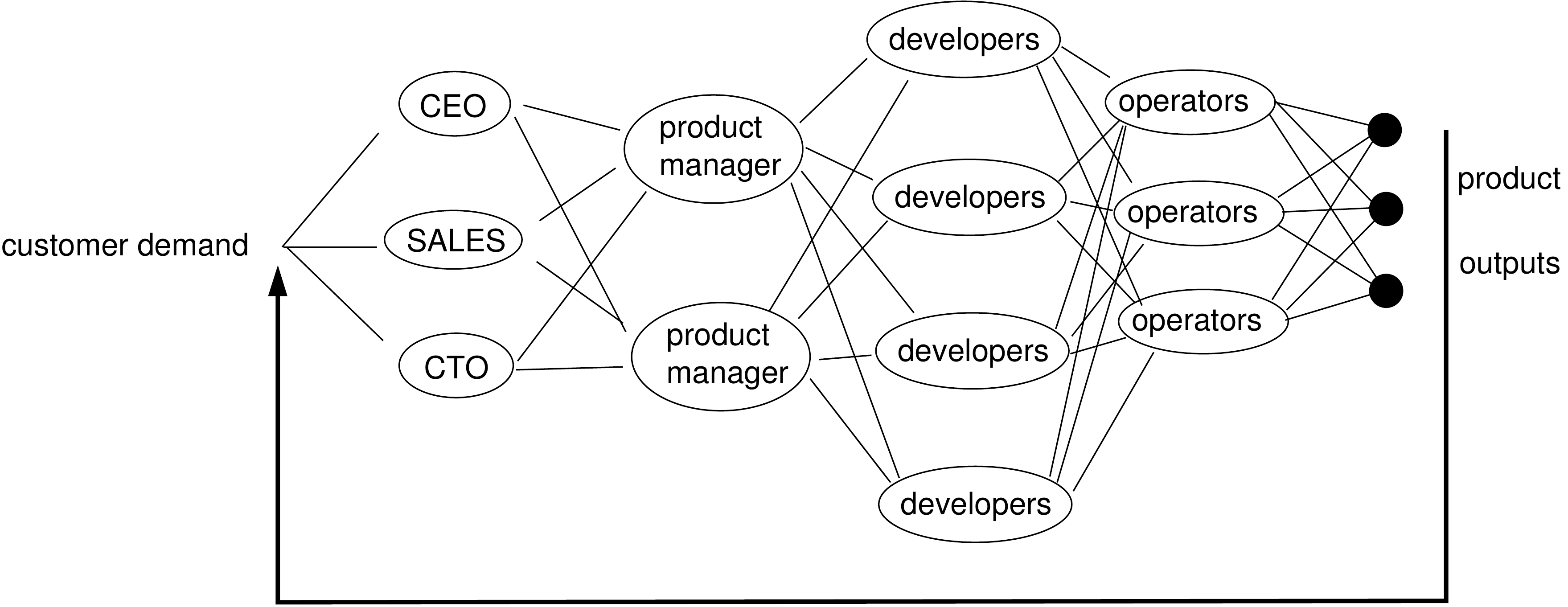}
\caption{\small Is a hierarchical organization like a learning
  semantic space?  The structure is potentially similar if there are
  strong links between layers, without siloing.\label{firm}}
\end{center}
\end{figure}
What is missing from a silo topology, or purely branching network
(which is a classic computer science class hierarchy or spanning
tree), is the ability to learn and freely associate ideas. All ideas
are related to a single concept representation, which is reified as
the top manager (CEO).
\end{example}

\begin{example}[CRISPR gene editing]
  Clustered regularly inter-spaced short palindromic repeats (CRISPR)
  is a gene editing technique for scanning and locating specific codon
  sequences in the genome. With centralized intent at a large scale,
  and distributed cells containing centralized DNA at a small scale,
  decentralized molecular agents attached to immune cells can edit the
  genomes of diseased cells, in a semantically specific way\cite{crispr}.
\end{example}

\subsubsection{Innovation in smart spaces}

\begin{figure}[ht]
\begin{center}
\includegraphics[width=5.5cm]{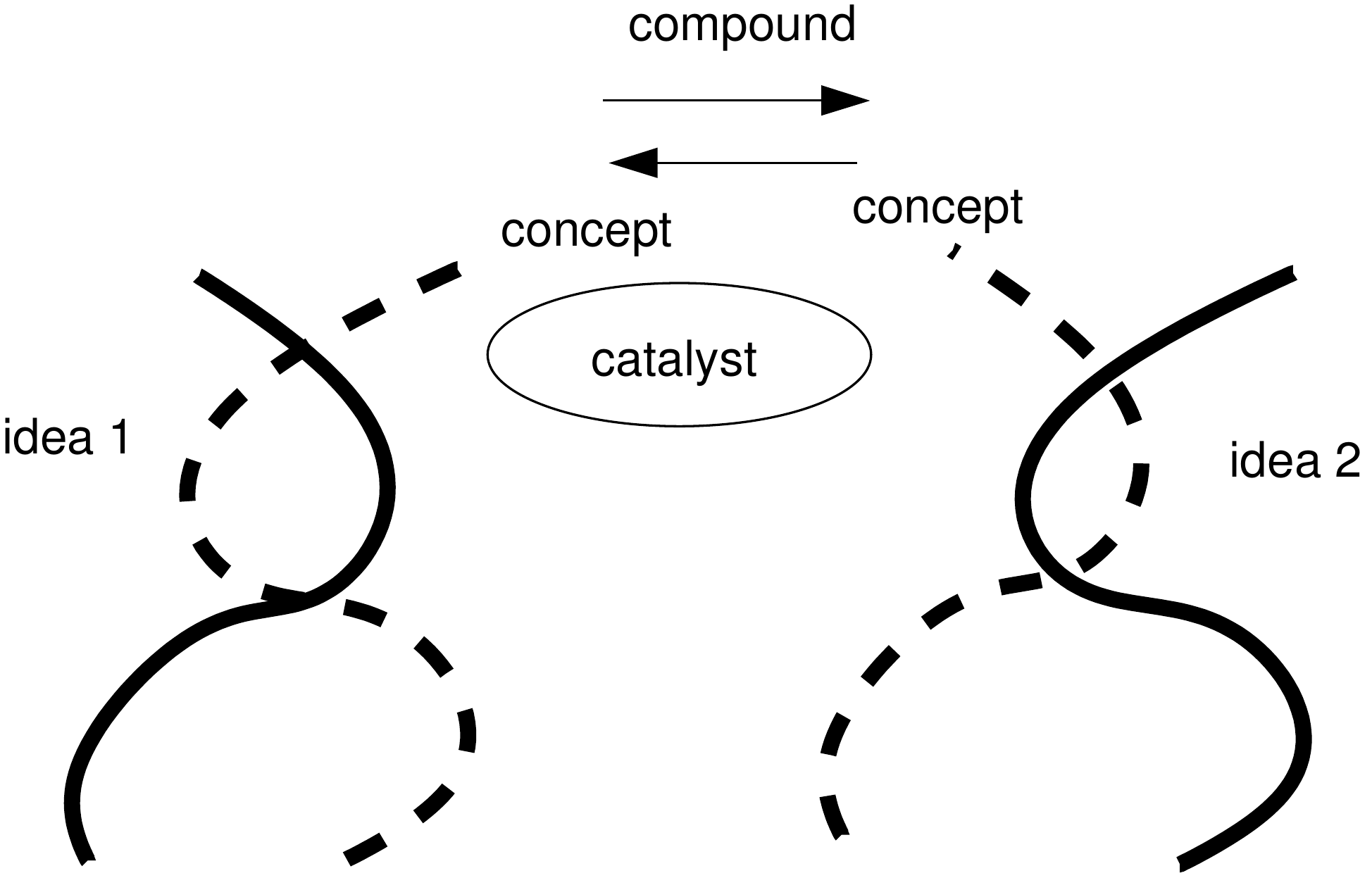}
\caption{\small Innovation happens by mixing ideas.\label{catalyst}}
\end{center}
\end{figure}

The processes of innovation haven been studied from a promise theory
perspective elsewhere\cite{innovation}. The key elements of innovation
are the random mixing of concepts, followed by outcome gestation and
selection. Mixing of ideas can be a quick process, catalyzed by
accessible localization (e.g. centralization); the latter is a slower
process, in which something is built from the new plan, functional
parts align with a recombined mixture of promised concepts, and it is
then subjected to a selection process (testing), in order to judge its
fitness for purpose (see figure \ref{catalyst})\footnote{This is
  functionally equivalent to the mixing of DNA to form a zygote.}.

In a smart space, the agents that define the space innovate by
interacting with their exterior to learn new ideas. This might be the
occupants of the space interacting with it, or the environment beyond
it.  To facilitate this interaction, catalysts can be provided to
match-make the association of compatible concepts.  The efficiency of
this process is accelerated by a large contact surface, so larger
spaces could be smarter than smaller ones. Matchmakers and go-betweens
thus have roles to play in making a city or space smart. This kind of
`thinking' is what a knowledge space does when it focuses more on
acquired memories and attempts to combine them into new ones.

\begin{itemize}
\item Spacetime mixing of agents is an important part of
  recombination.  Mixing is most effective in a gas phase, provided
  agents are brought together to form combinations, but could also be
  mediated `pollenated' in a solid phase by messengers.

\item Outcome selection requires an oriented notion of a goal, or
  promised outcome and a judging agent, or simply failure.
\end{itemize}
Could an inanimate space (at any scale) help humans or organizations work together,
just by virtue of how it bounds, facilitates, and shapes their interactions?

Independent researcher Meredith Belbin has studied human collaboration
through teamwork, and has identified nine abilities or roles (kinds of
promise) to be played in a team collaboration (regardless of how many
people there are in the team)\cite{belbin}. In Belbin's view, human
dynamics require roles for: plant (a creative ideas person who solves
problems), shaper (with the drive and courage to overcome obstacles),
specialist, implementer (a practical thinker), a resource investigator
(who knows where to find help), a coordinator (an arbitrator), a
monitor (who judges progress, team worker (someone concerned with the
team's inter-personal relationships), and finisher (someone analytical
who does quality assurance).  Belbin does not claim that all these
roles must be taken on by different agents (people in this case), but
that these functional roles define the promises that need to be kept
for an effective teamwork. So, should we expect to find these promised
roles in smart spaces too?  In terms of spacetime processes, these
incorporate learning or memory (specialist), mixing (plant,
coordinator, finisher).  the limitation of Belbin's model lies in
assuming human involvement, and human psychology, the following of a
plan, and a notion of progress towards a goal.

\begin{example}[Revised simple view of a smart company]
  In example \ref{sorg}, we represented an organization in the form of
  a neural network, deliberately, in order to contrast a simple flow
  discrimination network with an organizational hierarchy. Now let's
  make a more realistic sketch. A company has an outside world, with
  users, customers, competitors, which it interacts with through
  sensor-actuators, represented roughly by sales and support.  These
  sensors forward their observations to the internal representations
  of the company: the people who remember the relationships, including
  developers and operations people, i.e. the memory agents (see figure \ref{devops}).

  Unlike the previous example of the flow network, where agents were
  imagined as driving a company from a helm, this model is more
  fatalistic: inputs are passed to both developer and operations
  agents to be interpreted. Neither is above the other in a
  preferential hierarchy.  Data from the outside world about successes
  and failures are applicable to both equally, and they will also
  share associated impulses with one another.

The agents in a company are not dumb switches, they are already
significantly established networks of knowledge themselves.  The
training of the nodes over time corresponds to their schooling and
experience in doing their jobs, as well as adapting and modernizing
individually.  A hybrid structure like this cannot be understood in
precisely the same terms as an atomic knowledge representation: it is a hierarchy
of superagents, each of which is a learning knowledge representation.
The question is whether the organization as a cooperative whole may be considered
a learning knowledge space.
\begin{figure}[ht]
\begin{center}
\includegraphics[width=9.5cm]{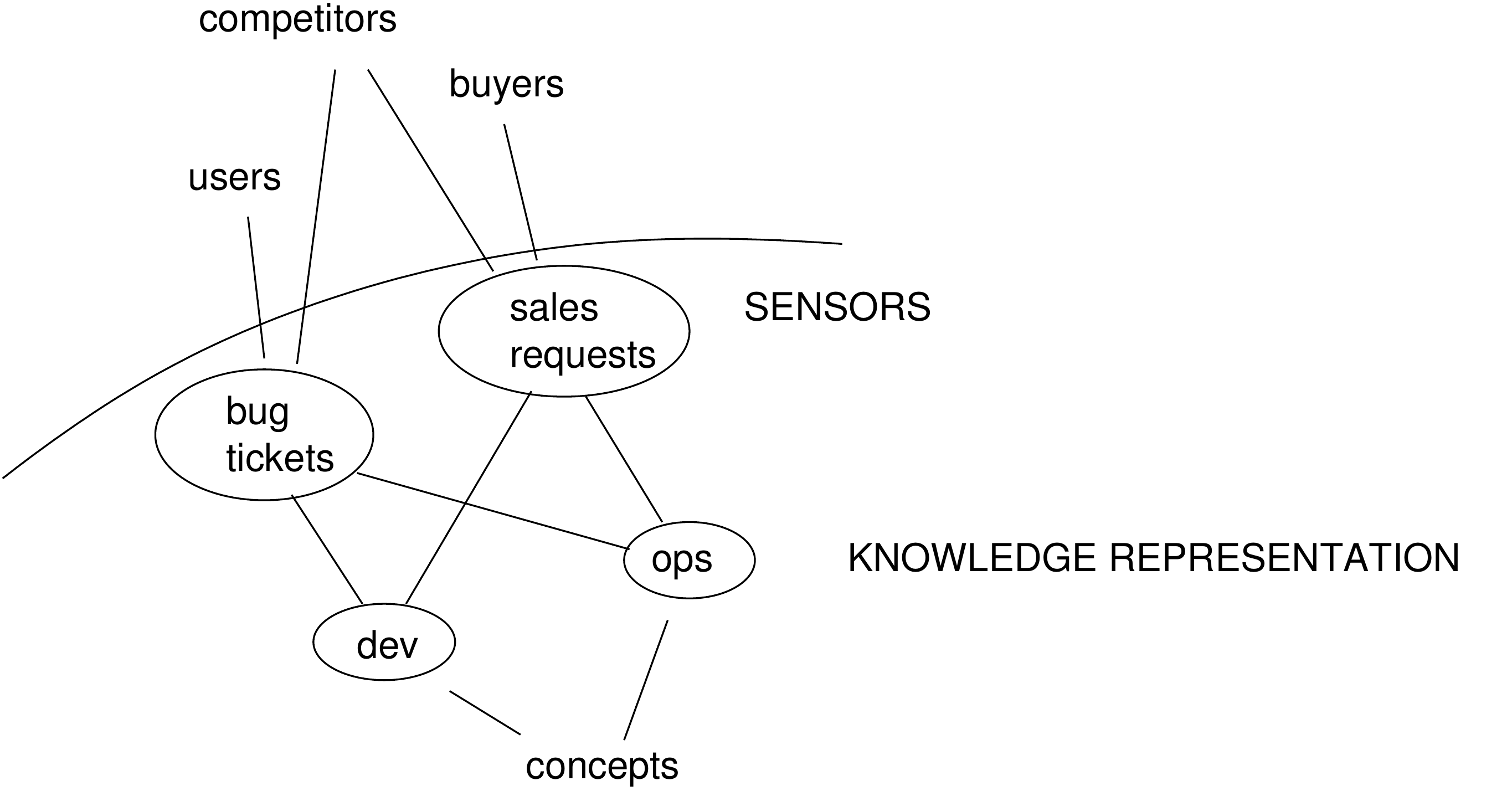}
\caption{\small A company as a knowledge represent ion.\label{devops}}
\end{center}
\end{figure}
The stability of the organization depends on it being able to handle
the perturbations from the sensors (fix bugs and operational issues,
i.e.  keep its business promises) much faster than the time it takes
to train the internal nodes in their basic functions of interpreting
the sensors.  However, there is a second rate of learning, which is
the operational knowledge, learnt `on the job', which is formed in
real time from context activations. This fits with the model of knowledge
representation presented in section \ref{knowledgerep}, but it does not
fit well with a neural network discriminator structure, since the latter
is mainly about recognition of known patterns.

Even these simple considerations seem to indicate that a company,
which interacts with the world, cannot be steered by a single helm, in
a quasi-deterministic hierarchy, else it will be unable to represent
the operational knowledge it acquires through interactions. A flatter
organization, where all knowledge workers are closer to their customers
leads to a more reliable knowledge retention.
\end{example}

\subsubsection{Auto-reasoning and searching for answers}

The ability to construct a coherent narrative from a collection of
parts is what we call reasoning. This could be addressed in a tokenized (story)
representation or in an implicit learned network representation such as example \ref{sunworship}.
The aim of a smart space would be to improve the likelihood that a
space, when faced with a challenge or question, would automatically
evolve towards a state representing a reasoned outcome. This need not
guarantee the correctness of an answer, only the convergence towards
the best answer it could find. For example, in the case of the Tsunami
challenge (example \ref{tsunami}), the responses of the city unleashed
a chain of processes that automatically protected the city. This
exhibited a primitive form of autonomous reasoning. Reasoning that happens
on a larger scale would not be clearly visible to the occupants of a city,
and may take place on a much slower timescale than they normally observe.
Thus, a smart city might not be apparently smart to human concerns.

The use of information technology to enhance spaces, using special
sensors and training, is clearly a popular idea today.  In recent
years, it has also been popular to answer questions by trawling
`big data', i.e. unspecific sources of sensory information collected
from a wide range of sources, in hope of discovering interest concepts
and relationships by introspection, whence to pursue reasoning based on these.
Finding answers is a form of pattern recognition, in semantic patterns
(stories and narratives) built from data patterns that reveal
interpretations (concepts).

Are we making the best use of knowledge? People, machines, spaces, and
communities all learn, but not all do much to propagate or facilitate
the application of their expertise.  The smartest part of searching
lies in setting up the boundary conditions, i.e. in {\em posing the
  right question}: what pattern are we looking for, and at what scale?
Scale plays a special role in reasoning.  If we search for a structure
that operates at a scale much larger than our sensory experience (like
a city), it is entirely possible that behaviours might emerge that we
would be entirely oblivious to, because the benefits are not directly
apparent at the scale of our own sensory apparatus.  Generalized
clustering techniques, like Principal Component Analysis\cite{duda1,kyrre3}
can find conspicuous patterns, but cannot attach significance to them.
It is the latter process which is the most expensive long term
approach.  Intelligent assistants might identify relationships that we
missed with our limited human faculties, but human expertise can offer
helpful input without having to go through expensive training, using a
semantic network or topic map approach.

\section{Summary}

\begin{quote}
\rm Parsifal: Ich schreite kaum, doch w\"ahn' ich mich schon weit.\\
\em (I scarcely move, yet already it seems I have travelled far.)\\
\rm Gurnemanz: Du sieh'st, mein Sohn, zum Raum wird hier die Zeit.\\ 	
\em (You see, my son, here time becomes space.)\\
\rm -- R. Wagner, Parsefal
\end{quote}

In these notes, I have tried to sketch out how to apply spacetime
concepts to practical questions about the semantics of data, and how
data become knowledge. The aim is not to examine a particular approach
to knowledge representation or learning, or to describe and test a
particular technology, but to rather expose the general structural
aspects, using a minimum of assumptions based on Promise Theory.  This
includes the emergence of what we know as concepts and how they
network into larger knowledge representations. 

This effort began, motivated by observations concerning IT
infrastructure\cite{burgessC9,burgessDSOM2002,burgessaims2009,stories},
and this later carried over into the possibilities for future
large-scale infrastructure. It has since assumed much wider scope.
Throughout the notes, we have been homing in on the conclusion that
knowledge is represented principally by concepts encoded through
patterns that manipulate the various aspects of spacetime symmetries.
This suggests that information and knowledge representation are
fundamentally connected to the structure of semantic spacetime, and
few other concepts are really needed (see section \ref{ntassoc}). This
simple insight takes one remarkably far, and qualifies as a basic law:
{\em `All representations of knowledge have semantic content 
 proportional to the loss of spacetime symmetry.}

\begin{figure}[ht]
\begin{center}
\includegraphics[width=12.5cm]{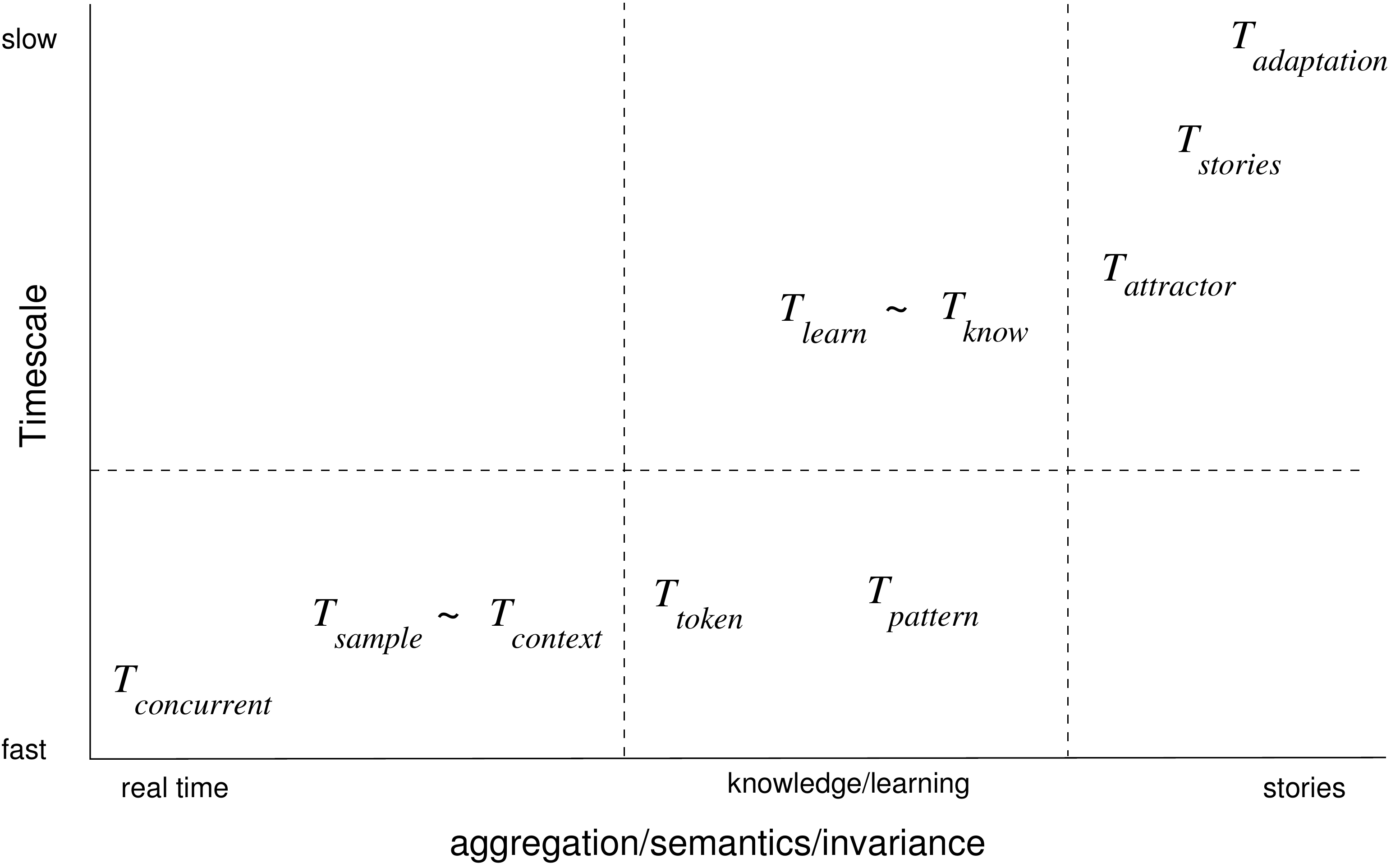}
\caption{\small Timescales overlaid onto a scale/invariance diagram.\label{timescales}}
\end{center}
\end{figure}

The detailed mechanisms for encoding of concepts, associations, and
semantics can be identified with spacetime scaling, directed
adjacency, and boundary interactions. A key requisite is that there be
memory, with predictable and repeatable addressability over
distinguishable spacetime locations. Addressability is needed for
semantic continuity: recalling and updating knowledge. Although a
database is a semantic space, knowledge is not a database, but a
continuous process, and a clean separation of timescales is essential
for the dynamical stability of that process (see figure
\ref{timescales}).

A number of hypotheses emerged from basic promise theoretic considerations during this study:
\begin{enumerate}
\item The weights or significances of patterns are in inverse relation to intrinsic information of their representation.
\item Usable knowledge requires a minimum level of dynamical stabilty.
\item Context must be evaluated faster and promises less information than the derived knowledge it modulates.
\item Context evaluation suggests coordination by an independent agent.
\item Knowledge maintenance, by partially annealing of learning weights, is needed for long-term stability.
\item Story inference and navigation suggest coordination by an independent agent.
\end{enumerate}
The representation of concepts and associative pathways, as spacetime
structures, builds on basic irreducible associations of spacetime:
\begin{itemize}
\item Nested containment (scaling).
\item Causal history (pre-requisite dependency).
\item Composition, i.e. cooperative refinement.
\item Similarity (proximity).
\end{itemize}
With the benefit of a semantic spacetime viewpoint, many apparently
complex semantic issues reduce to combinations of these spacetime
associations, and effectively a memory of what, where, and when.

Semantic tokenization is key to compressing cognition. We are most
familiar with this in the context of language representation.  In stable
situations, we can achieve this tokenization with high levels of
probability, but that is a best case scenario. The ability to learn
and crystallize concepts from `big' noisy data, is strongly
constrained by the capabilities of the memory used, and hence the
properties of the semantic space.
Tokenization solves three issues:
\begin{itemize}
\item Reduction of the dimensionality of sensory data (token formation).
\item Compression of sensory information to a symbolic form of low information (context).
\item Mapping variable inputs to an invariant representation (conceptualization).
\end{itemize}
This tokenization, which emerges as central to concepts, is not
intentionally a `linguistic' construction. It emerges from the
sequential nature of coarse grained input (both of which are just
spacetime concepts). However, I believe that the resemblence to
language is not spurious. Tokens or symbols emerge from concurrent
semantics of a sensory stream. The discreteness of symbols emerges
from the approximation of concurrency across multiple sensory threads.
Simultaneity of semantics is an implicit seed for binding sensory
properties into prototype concepts.  The result is that there is a
mutual affinity between this view of knowledge representation and
familiar models of linguistics (cognitive linguistics on the interior
representation and generative linguistics on the exterior).

Present day artificial intelligence research has most recently focused
on the advances in neural network for `deep
learning'\cite{deeplearning1,deeplearning2,deeplearning3}, where one
uses directional networks to learn patterns, such that weighted
connections encode pre-trained semantics. While fascinating, to be
sure, this approach may not be the cheapest route to technological
solutions: machine learning ignores several issues, like the problem
of explicit tokenization, and the opportunities to merge human and
machine knowledge, in order to focus on trying to engineer its
emergence.  It makes weak contact with existing human concepts, like
knowledge bases and libraries, and then only through arduous and
expensive training. There remains the possibility of improving on
this, by direct human input at the symbolic level (like words of
encouragement from a teacher), in order to imagine `cyborg' systems, at all scales,
which might make the best use of both humans and machinery.  The
spacetime approach described here may help to put exploit these pieces
and bring them together.

Although it would be impossible to bring rigour to a subject of this
scope, in a single series of notes, no matter how detailed, I hope
that there is sufficient material here to bring some clarity to a few
of the larger questions around artificial reasoning, and help
formulate more of the key questions theoretically. I find it
compelling that the most fundamental of spacetime notions leads to a
simple unified way of framing the subject across multiple disciplines.

This completes the initial overview of semantic spaces as
representations for organization and cognitive reasoning, and perhaps
opens a door to a more complete notion of artificial intelligence.
When all is done, the simple outcome of this work seems to be that our
most basic learning and abstraction capabilities have their origins,
plausibly, in the fundamental structures and processes of space and
time.

\section*{Acknowledgement}

I am grateful to Steve Pepper for getting me interested in many of the
topics covered within this paper, and also to Alva Couch with whom I
co-developed the notion of stories and inference between 2007-2008,
for a draft reading of the manuscript. The writing of this work has
been partly sponsored by Cisco Systems in 2016. None of these necessarily
endorse or agree with the presentation here.

\appendix
\section{Coordinatization of data streams}\label{appendix}

This shows a coordinatization of a multi-dimensional document, which
could be used to describe and document its interpretation, without
alteration, post hoc.
\small
\begin{verbatim}

{
  "kind": "Service",
  "apiVersion": "v1",
  "metadata": {
    "name": "myapp"
  },
  "spec": {
    "ports": [{
      "port": 8765,
      "targetPort": 9376
    }],
    "selector": {
      "app": "example"
    },
    "type": "LoadBalancer"
  }
}

\end{verbatim}
\normalsize

\small
\begin{verbatim}

Dimension/Region ( 1,  0,  0,  0,  0, )  -> 

Dimension/Region ( 2,  1{,  0,  0,  0, )  -> 
 - path/proper time location[0]("kind")
 - path/proper time location[1](:)
 - path/proper time location[2]("Service")
 - path/proper time location[3]("apiVersion")
 - path/proper time location[4](:)
 - path/proper time location[5]("v1")
 - path/proper time location[6]("metadata")
 - path/proper time location[7](:)

Dimension/Region ( 2,  2{,  1{,  0,  0, )  -> 
 - path/proper time location[0]("name")
 - path/proper time location[1](:)
 - path/proper time location[2]("myapp")

Dimension/Region ( 2,  3{,  0,  0,  0, )  -> 
 - path/proper time location[0]("spec")
 - path/proper time location[1](:)

Dimension/Region ( 2,  4{,  1{,  0,  0, )  -> 
 - path/proper time location[0]("ports")
 - path/proper time location[1](:)

Dimension/Region ( 2,  4{,  2{,  1[,  1{, )  -> 
 - path/proper time location[0]("port")
 - path/proper time location[1](:)
 - path/proper time location[2](8765)
 - path/proper time location[3]("targetPort")
 - path/proper time location[4](:)
 - path/proper time location[5](9376)

Dimension/Region ( 2,  4{,  3{,  0,  0, )  -> 
 - path/proper time location[0]("selector")
 - path/proper time location[1](:)

Dimension/Region ( 2,  4{,  4{,  1{,  0, )  -> 
 - path/proper time location[0]("app")
 - path/proper time location[1](:)
 - path/proper time location[2]("example")

Dimension/Region ( 2,  4{,  5{,  0,  0, )  -> 
 - path/proper time location[0]("type")
 - path/proper time location[1](:)
 - path/proper time location[2]("LoadBalancer")

Dimension/Region ( 2,  5{,  0,  0,  0, )  -> 

Dimension/Region ( 3,  0,  0,  0,  0, )  -> 

Dimension/Region ( 4,  0,  0,  0,  0, )  -> 
 - path/proper time location[0](@)

\end{verbatim}
\normalsize

A mixed format:
\small
\begin{verbatim}

{{{one { two, three } four <alpha> text ... </alpha> }

\section{section1}xx
Nothing happenings

\section{section2}

[windows1]

<greek>
<alpha>
AAA 
</alpha>
<beta>
BBB
</beta>
<gamma>
CCC
</gamma>
</greek>

[windows2]

xiug

\end{verbatim}
\normalsize

\small
\begin{verbatim}

Dimension/Region ( 1,  0,  0,  0,  0, )  -> 

Dimension/Region ( 2,  1{,  1{,  1{,  0, )  -> 
 - path/proper time location[0](one)

Dimension/Region ( 2,  1{,  1{,  2{,  1{, )  -> 
 - path/proper time location[0](two)
 - path/proper time location[1](three)

Dimension/Region ( 2,  1{,  1{,  3{,  0, )  -> 
 - path/proper time location[0](four)

Dimension/Region ( 2,  1{,  1{,  4{,  1<alpha>, )  -> 
 - path/proper time location[0](text)
 - path/proper time location[1](...)

Dimension/Region ( 2,  1{,  1{,  5{,  0, )  -> 

Dimension/Region ( 2,  1{,  2{,  0,  0, )  -> 
 - path/proper time location[0](\section)

Dimension/Region ( 2,  1{,  3{,  1{,  0, )  -> 
 - path/proper time location[0](section1)

Dimension/Region ( 2,  1{,  4{,  0,  0, )  -> 
 - path/proper time location[0](xx)
 - path/proper time location[1](Nothing)
 - path/proper time location[2](happenings)
 - path/proper time location[3](\section)

Dimension/Region ( 2,  1{,  5{,  1{,  0, )  -> 
 - path/proper time location[0](section2)

Dimension/Region ( 2,  1{,  6{,  0,  0, )  -> 

Dimension/Region ( 2,  1{,  7{,  1[,  0, )  -> 
 - path/proper time location[0](windows1)

Dimension/Region ( 2,  1{,  8{,  0,  0, )  -> 

Dimension/Region ( 2,  1{,  9{,  1<greek>,  0, )  -> 

Dimension/Region ( 2,  1{,  9{,  2<greek>,  1<alpha>, )  -> 
 - path/proper time location[0](AAA)

Dimension/Region ( 2,  1{,  9{,  3<greek>,  0, )  -> 

Dimension/Region ( 2,  1{,  9{,  4<greek>,  1<beta>, )  -> 
 - path/proper time location[0](BBB)

Dimension/Region ( 2,  1{,  9{,  5<greek>,  0, )  -> 

Dimension/Region ( 2,  1{,  9{,  6<greek>,  1<gamma>, )  -> 
 - path/proper time location[0](CCC)

Dimension/Region ( 2,  1{,  9{,  7<greek>,  0, )  -> 

Dimension/Region ( 2,  1{,  10{,  0,  0, )  -> 

Dimension/Region ( 2,  1{,  11{,  1[,  0, )  -> 
 - path/proper time location[0](windows2)

Dimension/Region ( 2,  1{,  12{,  0,  0, )  -> 
 - path/proper time location[0](xiug)

\end{verbatim}
\normalsize
Each of the addressable sections of this sensory data stream can now be annotated with a specific
meaning, either by representing as a pattern or as a specific address in the data `document'.

\section{Invariant representation by spacetime aggregation example}\label{appb}

\begin{figure}[ht]
\begin{center}
\includegraphics[width=16.5cm]{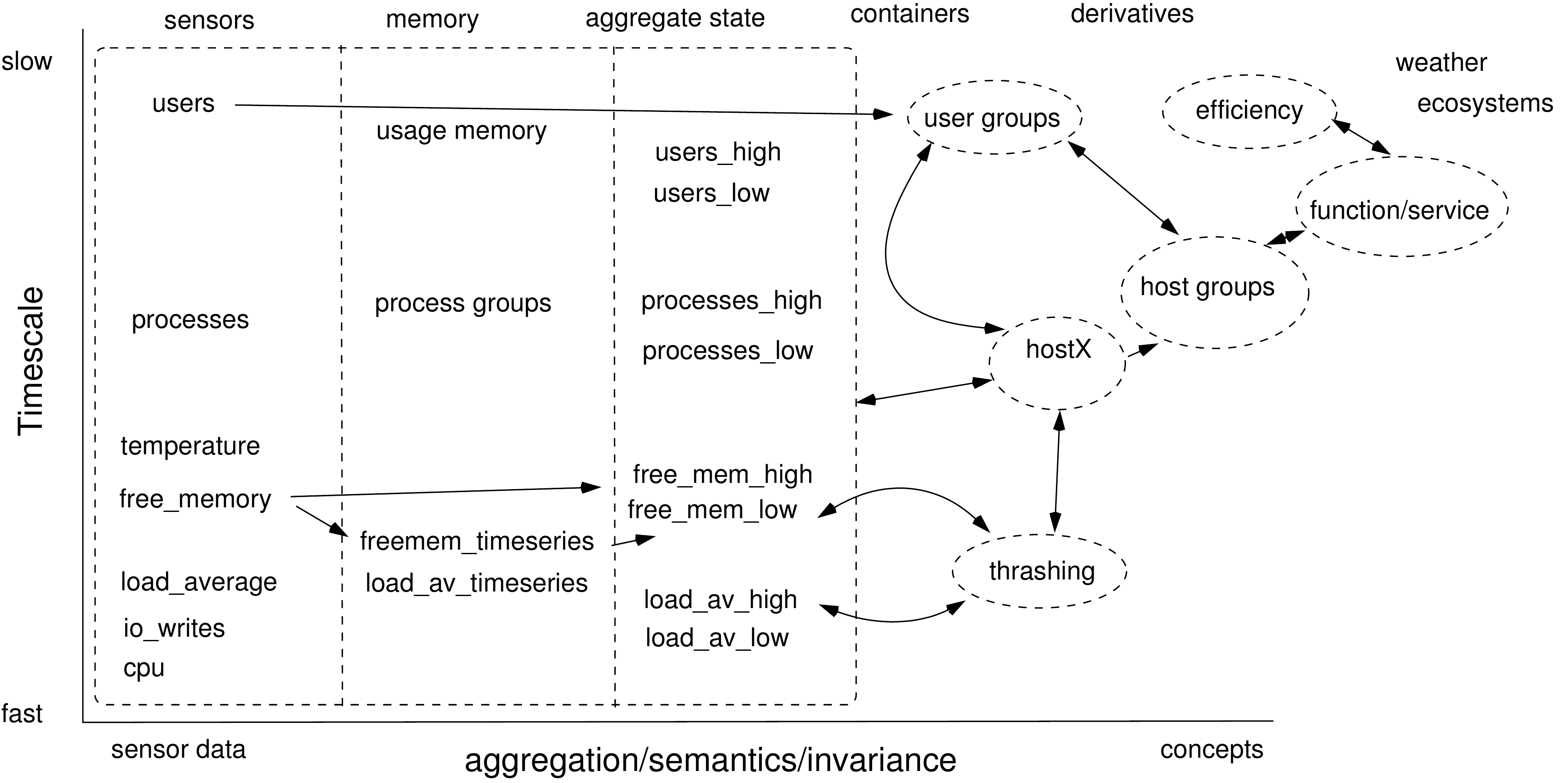}
\caption{\small Scale/invariance diagram example.\label{DIKWexample}}
\end{center}
\end{figure}

Figure \ref{DIKWexample} shows a scale/invariance diagram for a
hypothetical cloud system monitoring system. 
Data are collected from sensors, described at the bottom left of the
diagram, on individual computing nodes (where only one is shown). The
data are aggregated and memorized locally, as timeseries or state
caches in order to characterize a stable (and thus approximately
invariant) normal state.  Changes to what has been learned to be
stable normal state are classified as anomalous high or low
deviations, which are themselves invariant characterizations of a
changing state.  

At any given time, invariant characteristics may or may not be active,
depending on context. The tokenization of context must therefore be
complicit in the invariant representation. Invariance does not mean
something which is unable to change dynamically, but rather stability
of interpretation.  As we move to the right of the diagram, the
concepts become `larger' in scale (space and time) and thus more
stable semantically and dynamically. They accrete more support from
sensory data, derivative characteristics and even external annotation
(such as that provided by the technique shown in the previous
appendix).  Overlapping concepts, taken in clusters may represent
specific phenomena, e.g. thrashing be exhibited by a change in load
average, free memory, and several other characteristics dynamical
deviations.  The roles played by change, in delimiting regions of
stabilty, are likely key to finding semantics boundaries.  Each
independent grouping or clustering of agents can be given a new name
and represent a new concept, following the indentification of
autonomous agents in promise theory.

\bibliographystyle{unsrt}
\bibliography{spacetime}

\end{document}